\documentclass{article}
\pdfoutput=1
\usepackage{iclr2025_conference,times}
\usepackage{xspace}
\usepackage{mathpazo}

\usepackage{amsmath}
\usepackage{amsfonts}
\def\eqref#1{equation~\ref{#1}}

\usepackage{xcolor}
\usepackage{booktabs}
\usepackage{multirow}
\usepackage{array}
\usepackage{tabularx}
\usepackage{makecell}
\usepackage{graphicx}
\usepackage{tikz}
\usetikzlibrary{arrows.meta, backgrounds, calc, patterns}
\usepackage{pgfplots}
\usepgfplotslibrary{groupplots}
\pgfplotsset{compat=1.18}
\usepackage{sansmath}
\usepackage{colortbl}
\usepackage{subcaption}
\usepackage{fvextra}
\usepackage[T1]{fontenc}
\usepackage[export]{adjustbox}
\usepackage[most]{tcolorbox}
\usepackage{etoolbox}

\newtcolorbox[auto counter,number within=section]{prompt}[1][]{
    colframe=apertusSkyBlue,
    colback=apertusSkyBluePale!40,
    colupper=apertusAnthracite,
    colbacktitle=apertusSkyBlue,
    fonttitle=\small,
    coltitle=apertusAnthracite,
    fontupper=\footnotesize,
    boxsep=3pt,
    left=0pt,
    right=0pt,
    top=0pt,
    bottom=0pt,
    boxrule=1pt,
    breakable,
    #1,
}
\makeatletter
\def\tcb@cnt@promptautorefname{Prompt}
\makeatother
\newcommand{\promptsize}{\fontsize{8}{9}\selectfont}
\DefineVerbatimEnvironment{promptverbatim}{Verbatim}{breaklines,breakanywhere,fontsize=\promptsize}

\definecolor{apertusSkyBlue}{HTML}{7BBBD5}
\definecolor{apertusSkyBlueLight}{HTML}{AACFDC}
\definecolor{apertusSkyBluePale}{HTML}{BFD8E1}
\definecolor{apertusAnthracite}{HTML}{2E2F31}
\definecolor{epflRed}{HTML}{FF0000}
\definecolor{apertusWhite}{HTML}{FFFFFF}

\definecolor{c-navy}{rgb}{0,0.08,0.45}
\usepackage{hyperref}
\hypersetup{%
colorlinks=true,
linkcolor=c-navy,
citecolor=c-navy,
filecolor=c-navy,
urlcolor=c-navy,
}

\usepackage[section,above]{placeins}
\makeatletter
\AtBeginDocument{%
  \expandafter\renewcommand\expandafter\subsection\expandafter
    {\expandafter\@fb@secFB\subsection}%
}
\makeatother

\newcommand{\hflogo}[1]{\includegraphics[width=11pt, valign=c]{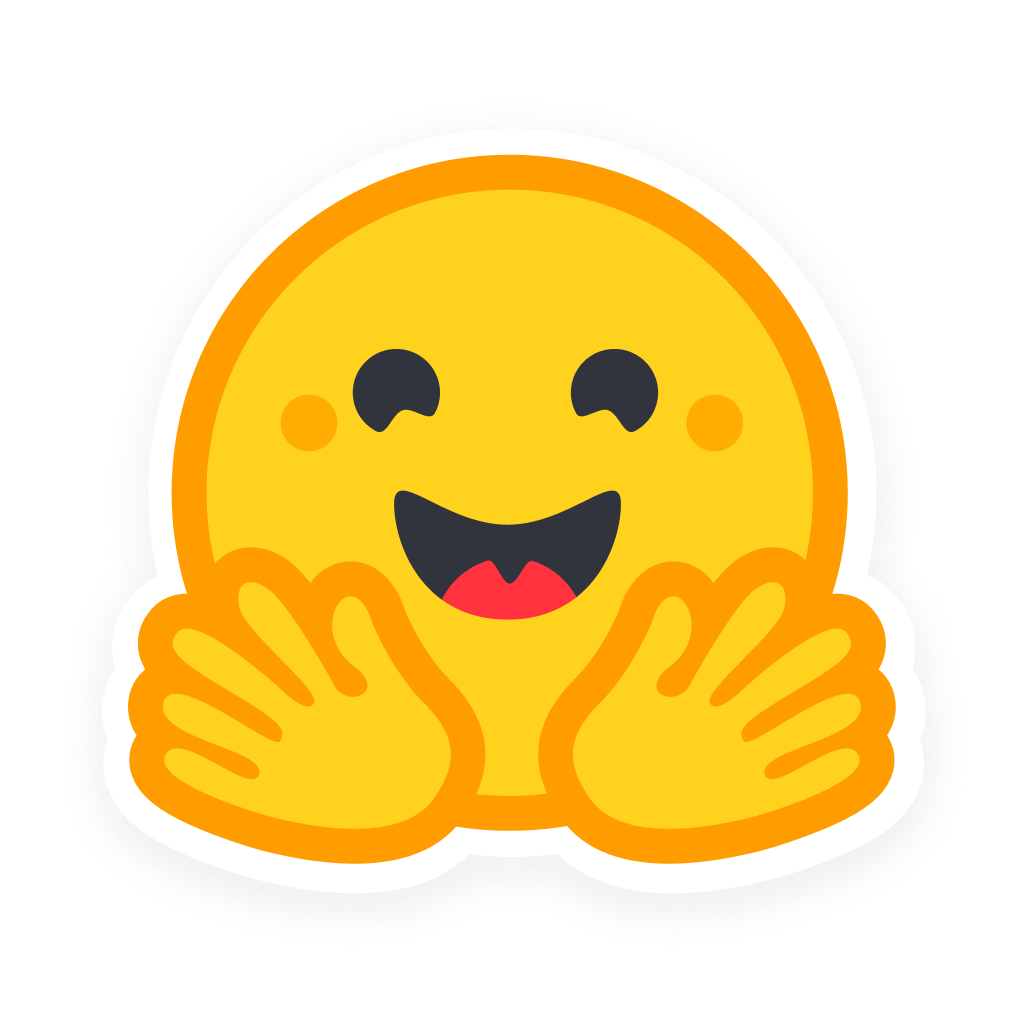}\nobreakspace\texttt{#1}}
\newcommand{\ghlogo}[1]{\includegraphics[width=11pt, valign=c]{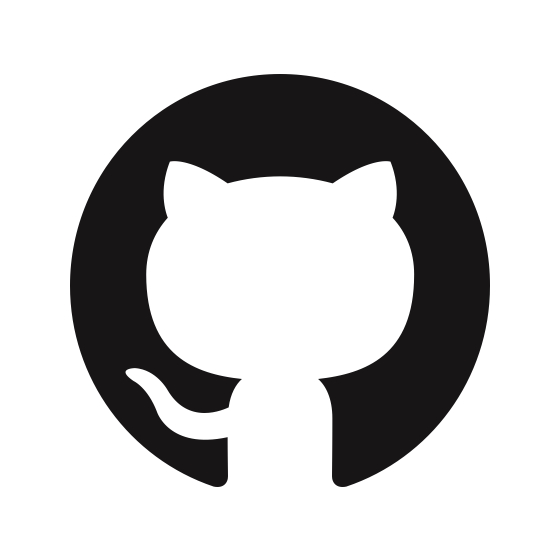}\nobreakspace\texttt{#1}}
\newcommand{\ghrepo}[1]{\texttt{\href{https://github.com/#1}{#1}}}
\newcommand{\hfrepo}[1]{\hflogo{\href{https://huggingface.co/#1}{#1}}}
\newcommand{\hfdata}[1]{\texttt{\href{https://huggingface.co/datasets/#1}{#1}}}
\newcommand{\ghrepoicon}[1]{\ghlogo{\href{https://github.com/#1}{#1}}}
\newcommand{\hfdataicon}[1]{\hflogo{\href{https://huggingface.co/datasets/#1}{#1}}}

\title{\apertustitle\\[1em]
From Open Text Models to Open Multimodal Reasoners\\[0.5em]
{\small \apertus{} v1.5 Technical Report}}

\input{cover/authors}

\newcolumntype{Y}{>{\raggedright\arraybackslash}X}
\newcolumntype{Z}[1]{>{\hsize=#1\hsize\raggedright\arraybackslash}X}
\newcolumntype{L}[1]{>{\raggedright\arraybackslash}p{#1}}
\newcolumntype{F}[1]{>{\columncolor{apertusWhite}[\tabcolsep][0pt]\hsize=#1\hsize\raggedright\arraybackslash}X}
\newcommand{\statuscell}[3]{\cellcolor{#1}\textcolor{#2}{#3}}
\newcommand{\statusdone}[1]{\statuscell{green!20}{green!50!black}{#1}}
\newcommand{\statuspartial}[1]{\statuscell{yellow!30}{orange!80!black}{#1}}
\newcommand{\statusnotdone}[1]{\statuscell{red!15}{red!70!black}{#1}}
\newcommand{\statusna}[1]{\statuscell{gray!20}{gray!70!black}{#1}}
\newcommand{\statusdonetext}[1]{\textcolor{green!50!black}{#1}}
\newcommand{\statuspartialtext}[1]{\textcolor{orange!80!black}{#1}}
\newcommand{\statusnotdonetext}[1]{\textcolor{red!70!black}{#1}}
\newcommand{\statusnatext}[1]{\textcolor{gray!70!black}{#1}}

\newcommand{\apertus}{Apertus\xspace}
\newcommand{\apertussmall}{Apertus-8B}
\newcommand{\apertusonefivesmall}{Apertus~1.5~8B}
\newcommand{\apertusonefivelarge}{Apertus~1.5~70B}
\newcommand{\apertustitle}{\resizebox{\linewidth}{!}{\includegraphics{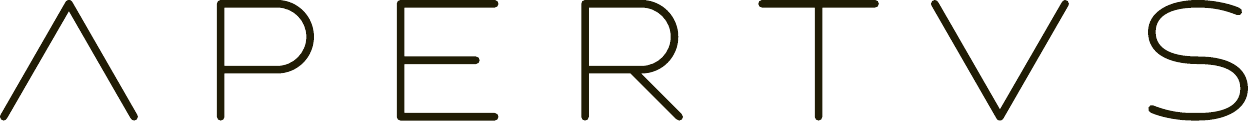}}}

\newcommand{\autorefnum}[1]{\hyperref[#1]{\getrefnumber{#1}}}

\newcommand{\citepsmall}[1]{{\scriptsize\citep{#1}}}

\newcommand{\effortrow}[2]{{\itshape #1: } #2\\[0.75em]}
\newcommand{\teamrow}[2]{{\bfseries #1:} #2\\[0.25em]}
\newcommand{\subeffortrow}[2]{\hspace*{1.5em}\begin{minipage}[t]{\dimexpr\linewidth-1.5em\relax}\raggedright{\itshape #1: } #2\end{minipage}\\[0.45em]}
\newcommand{\subeffortend}{\vspace{0.55em}}
\newcommand{\affiliationsrow}{\vspace{2.4em}{\bfseries Affiliations:} \afflist}

\begin{document}

\maketitle

\newpage

\begin{abstract}
Language models are becoming infrastructure, and infrastructure built in the open, by public institutions, on documented data, is the alternative to dependence on a few providers; Apertus shows this is viable at scale. \apertus{} 1.5 extends the fully open Apertus family from text to multimodal reasoning. Continued pretraining of the \apertus{} 1.0 models, on 4T additional tokens for the 8B and 2T for the 70B, adds image and audio understanding and extends the context window fourfold to 262,144~tokens; multimodality enters through the tokenizer, the language-model head remains restricted to text, and the architecture is unchanged. Supervised finetuning, long-context finetuning, reinforcement learning with verifiable rewards, and preference alignment complete the released models. Thinking is a flag of the chat template rather than a separate model, and lifts mathematical reasoning by up to 20~points on the 8B model. The released models improve over their predecessors on nearly every benchmark; among fully open models of its class, the 8B leads on knowledge, instruction following, and cultural benchmarks, and its instruction following surpasses all compared models, including open-weight ones. On image and audio understanding, a first for the family, the models are competitive with fully open multimodal models, strongest on document understanding and long-form and lower-resource speech recognition, while specialized vision and audio models remain ahead on perception and transcription of English and other high-resource languages. As in \apertus{} 1.0, we train only on openly available data, respect crawler opt-outs retroactively, and exclude restrictively licensed material. We release the model weights, data preparation and training recipes, and evaluation suites under permissive licenses; openness extends to values, with the \apertus{} charter guiding alignment stages published alongside the models.
\end{abstract}

\newpage
\begingroup
\linespread{0.97}\selectfont
\makeatletter
\patchcmd{\l@section}{1.0em}{0.8em}{}{\PackageError{main}{Could not tighten the table of contents}{}}
\makeatother
\tableofcontents
\endgroup

\clearpage
\section{Introduction}\label{sec:introduction}

The most capable large language models (LLMs) are accessible only through proprietary interfaces, and even the strongest openly released models typically publish their weights without the data, code, recipes, and values behind them. Fully open models are increasingly rare as the complexity and cost of LLM development have grown significantly, and they still trail their open-weight peers in capabilities. \apertus{} exists to close this gap. We understand full openness to extend across five axes: open weights, open data, open code, open recipes, and open values. Every claim in this report can therefore be traced to released artifacts, and the principles that shape the behavior of our models are published documents rather than internal policy.

\apertus{} is a collaborative effort led by ETH Z\"urich, EPFL, and CSCS, with contributors from many further institutions, to develop AI in an inclusive and trustworthy manner, and to provide a foundation on which developers build products and services. Pluralism is a founding goal: from the first release onward, \apertus{} pursues broad coverage of languages and cultures, and this focus continues in the present release. A second commitment is responsible development, guided by Swiss and European ethical and regulatory standards, while striving for performance and scale. LLMs are a central technology of our time, and we believe research institutions should engage with them directly. \apertus{} provides a scientific basis for research and development on open models by making its model weights, code, data, training recipe, and values available for study and reuse. This enables researchers and practitioners to examine the training process, deploy models independently, retain control over associated data flows, and develop the expertise needed to teach the next generation of engineers and researchers.

We released the first generation of models, \apertus{} 1.0, with 8B and 70B~parameters in September 2025, trained from scratch on 15T~tokens of openly documented text spanning over 1000~languages on the Alps supercomputer at CSCS~\citep{apertus2025apertusdemocratizingopencompliant}. Its data pipeline respected machine-readable opt-outs of domain owners, applying them retroactively to earlier crawls, removed personally identifiable information, and drew only from permissively licensed sources, and training suppressed verbatim memorization. The \apertus{} charter, published alongside the models, states the principles that guide their behavior, and the release included transparency documentation following the EU AI Act template, intermediate checkpoints, data preparation pipelines, and training code.

\begin{figure}[t]
\centering
\resizebox{\textwidth}{!}{%
\begin{tikzpicture}[
  every node/.style={text=apertusAnthracite},
  stage/.style={draw=apertusAnthracite, line width=0.45pt,
    fill=apertusSkyBlue, rounded corners=2pt,
    minimum width=2.55cm, minimum height=1.95cm, inner sep=0pt},
  stagetx/.style={stage, fill=apertusSkyBlueLight},
  title/.style={anchor=base, font=\fontsize{8.5}{10}\selectfont\bfseries,
    inner sep=0pt},
  detail/.style={anchor=north, align=center,
    font=\fontsize{7.5}{9.4}\selectfont, text width=2.38cm, inner sep=0pt},
  chip/.style={draw=apertusAnthracite, line width=0.45pt,
    fill=apertusWhite, rounded corners=2pt, align=center,
    font=\fontsize{8}{10}\selectfont, inner sep=4pt},
  final/.style={chip, draw=epflRed, line width=0.7pt,
    text width=2.15cm, minimum height=0.65cm},
  modtag/.style={fill=apertusAnthracite, text=apertusWhite,
    font=\fontsize{6.5}{7.5}\selectfont, rounded corners=2pt,
    minimum height=0.27cm, inner xsep=3pt, inner ysep=1pt,
    anchor=east},
  phase/.style={anchor=base west, font=\fontsize{9}{11}\selectfont\bfseries,
    inner sep=0pt},
  flow/.style={-{Latex[length=1.6mm, width=1.2mm]},
    draw=apertusAnthracite, line width=0.6pt},
]

\begin{scope}[on background layer]
  \fill[apertusSkyBluePale!35!apertusWhite, rounded corners=3pt]
    (2.08,0.62) rectangle (14.27,-2.17);
  \fill[apertusSkyBluePale!35!apertusWhite, rounded corners=3pt]
    (0.05,-2.88) rectangle (14.84,-5.67);
\end{scope}
\node[phase] at (2.25,0.25) {Continued pretraining};
\node[phase] at (0.22,-3.25) {Post-training};

\node[chip, text width=1.6cm] (a10) at (0.95,-0.975)
  {\textbf{\apertus{} 1.0}\\ 8B / 70B};

\foreach \id/\x/\heading in {
    s1/3.6/{1\enspace Unpaired},
    s2/6.65/{2\enspace Paired},
    s3/9.7/{3\enspace Cooldown},
    s4/12.75/{4\enspace Long context}} {
  \node[stage, minimum width=2.7cm] (\id) at (\x,-0.975) {};
  \node[title] at ([yshift=-0.45cm]\id.north) {\heading};
}
\node[detail, text width=2.5cm] at ([yshift=-0.9cm]s1.north)
  {Text, image, audio\\All-token targets};
\node[detail, text width=2.5cm] at ([yshift=-0.9cm]s2.north)
  {Image and audio\\paired with text\\Text-only targets};
\node[detail, text width=2.5cm] at ([yshift=-0.9cm]s3.north)
  {Higher-quality and\\instruction data};
\node[detail, text width=2.5cm] at ([yshift=-0.9cm]s4.north)
  {8k${\to}$32k${\to}$65k\\${\to}$131k${\to}$262k};

\foreach \id/\x/\style/\heading in {
    sft/1.495/stage/SFT,
    lcsft/4.465/stage/Long-context,
    rl/7.435/stage/RLVR,
    dpo/10.405/stagetx/DPO,
    sdpo/13.375/stagetx/SDPO} {
  \node[\style] (\id) at (\x,-4.475) {};
  \node[title] at ([yshift=-0.45cm]\id.north) {\heading};
}
\node[title] at ([yshift=-0.78cm]lcsft.north) {SFT};
\node[detail] at ([yshift=-0.93cm]sft.north) {16k context};
\node[detail] at ([yshift=-0.93cm]lcsft.north) {262k window};
\node[detail] at ([yshift=-0.93cm]rl.north)
  {Verifiable rewards\\Thinking on/off};
\node[detail] at ([yshift=-0.93cm]dpo.north)
  {Judge-scored pairs};
\node[detail] at ([yshift=-0.93cm]sdpo.north)
  {Charter alignment\\Self-distillation\\8B only};

\foreach \id in {s1,s2,s3,s4,sft,lcsft}
  \node[modtag] at ([xshift=-3pt]\id.north east) {T$\,\cdot\,$I$\,\cdot\,$A};
\node[modtag] at ([xshift=-3pt]rl.north east) {T$\,\cdot\,$I};
\foreach \id in {dpo,sdpo}
  \node[modtag] at ([xshift=-3pt]\id.north east) {T};

\node[anchor=east, font=\fontsize{8}{10}\selectfont\bfseries, inner sep=0pt]
  at (8.85,-6.55) {Released models};
\node[final] (m70) at (10.405,-6.55) {\textbf{\apertusonefivelarge}};
\node[final] (m8) at (13.375,-6.55) {\textbf{\apertusonefivesmall}};

\draw[flow] (a10.east) -- (s1.west);
\foreach \a/\b in {s1/s2,s2/s3,s3/s4,sft/lcsft,lcsft/rl,rl/dpo}
  \draw[flow] (\a.east) -- (\b.west);
\draw[flow] (s4.south) -- (12.75,-2.52) -- (-0.16,-2.52)
  |- (sft.west);
\draw[flow] (dpo.east) -- node[above=2pt, font=\fontsize{7}{8}\selectfont,
  inner sep=0pt] {8B} (sdpo.west);
\draw[flow] (sdpo.south) -- (m8.north);
\draw[flow] (dpo.south) -- node[left=3pt,
  font=\fontsize{7}{8}\selectfont, inner sep=0pt] {70B} (m70.north);

\node[anchor=west, font=\fontsize{7.5}{9}\selectfont, inner sep=0pt]
  at (0.22,-6.55)
  {Training modalities: \textbf{T} text, \textbf{I} image, \textbf{A} audio};
\end{tikzpicture}%
}%
\caption{The \apertus{} 1.5 training pipeline. Solid blue stages train on multimodal data (T$\,\cdot\,$I$\,\cdot\,$A, or T$\,\cdot\,$I for RLVR), light-blue stages on text only (T); red outlines mark the released models, which the 70B reaches without SDPO. DPO uses online pairs for the 8B model and offline and online pairs for the 70B model.}
\label{fig:pipeline_overview}
\end{figure}
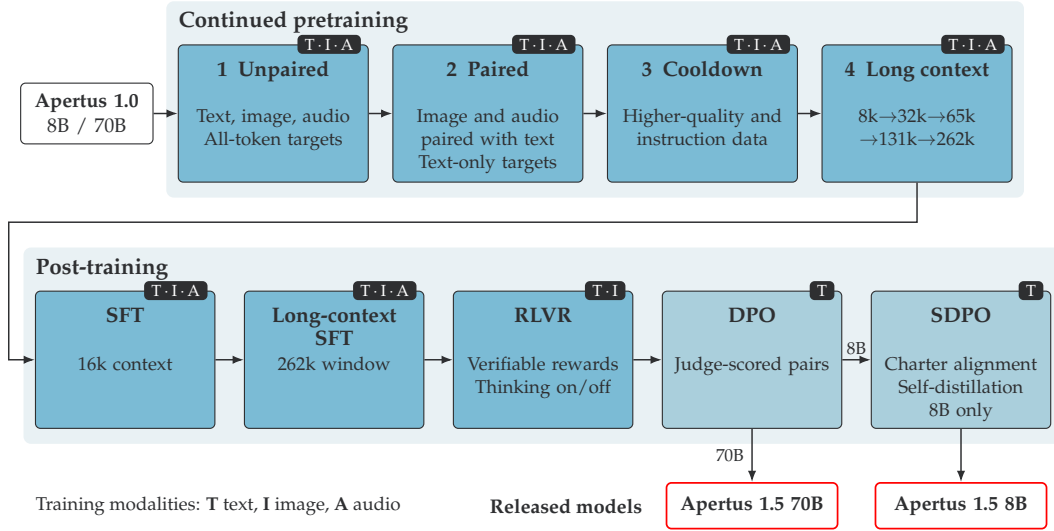

\apertusonefivesmall{} and \apertusonefivelarge{} build on this foundation through continued pretraining on a mix of text, image, and audio data, roughly 4T~tokens for the 8B model and 2T~tokens for the 70B model. Where \apertus{} 1.0 established that fully open training at scale is possible, \apertus{} 1.5 expands and improves the capabilities of the models. The models accept image and audio inputs in addition to text, support an optional thinking mode for harder reasoning problems, extend the context window fourfold to 262,144~tokens, follow instructions more reliably, and use tools more effectively. \autoref{fig:pipeline_overview} gives an overview of the complete training pipeline that produces both models, from the \apertus{} 1.0 checkpoints through continued pretraining and post-training to the released models. \autoref{fig:apertus_architecture} gives an overview of the shared model architecture and how it processes text, image, and audio inputs.

Our main contributions are:
\begin{itemize}
    \item \textbf{\apertusonefivesmall{} and \apertusonefivelarge{}.} Two fully open models that bring image and audio understanding, an optional thinking mode, a fourfold longer context window, and stronger instruction following and tool use to the \apertus{} family.
    \item \textbf{Multimodal continued pretraining.} A recipe that upgrades the \apertus{} 1.0 family in place: multimodality enters through the tokenizer, the language-model head remains restricted to text tokens, and the architecture is preserved.
    \item \textbf{Long-context extension.} The context window grows fourfold to 262,144~tokens, supported by dedicated long-context stages in both continued pretraining and post-training.
    \item \textbf{Our next generation of post-training.} Supervised finetuning (SFT), long-context SFT, reinforcement learning with verifiable rewards (RLVR), direct preference optimization (DPO), and charter alignment with self-distillation policy optimization (SDPO), delivering the optional thinking mode, more reliable instruction following, and more effective tool use.
    \item \textbf{Full release.} Weights, intermediate checkpoints, data preparation pipelines, training code, and evaluation harnesses, released under the Apache 2.0 license.
\end{itemize}

The remainder of this report follows the lifecycle of the models. \autoref{sec:multimodality} presents the multimodal tokenization, \autoref{sec:pretraining_recipe} the pretraining data and the continued-pretraining recipe, and \autoref{sec:post_training} the post-training pipeline. \autoref{sec:evaluation} reports evaluations across modalities and languages, \autoref{sec:infrastructure} describes the training infrastructure and its efficiency, \autoref{sec:governance} discusses the release and its governance, and \autoref{sec:conclusion} concludes.

\subsection{Released Models, Code, and Datasets}\label{sec:released_artifacts}

Full openness means every stage of this report can be retraced from released artifacts. We group the release into models, code, and datasets; all code and data carry permissive licenses, and the models are distributed under Apache~2.0.

\paragraph{Models.} The two final post-trained models are distributed on Hugging Face behind a gated download that records assent to the acceptable-use policy and reaches deployers with updates (\autoref{sec:license}); major intermediate checkpoints live as branches of these two repositories. The post-training stage checkpoints, one repository per stage under the same license and gating, let the lineage of \autoref{sec:post_training_lineage} be inspected checkpoint by checkpoint. Thinking is a request-time flag of the chat template rather than a separate model, even where inference providers expose it as a separate endpoint; usage instructions and supported inference stacks are in the model cards.

\hfrepo{swiss-ai/Apertus-v1.5-8B-base} \\
\hfrepo{swiss-ai/Apertus-v1.5-8B-SFT} \\
\hfrepo{swiss-ai/Apertus-v1.5-8B-RL} \\
\hfrepo{swiss-ai/Apertus-v1.5-8B-DPO} \\
\hfrepo{swiss-ai/Apertus-v1.5-8B} \\
\hfrepo{swiss-ai/Apertus-v1.5-70B-base} \\
\hfrepo{swiss-ai/Apertus-v1.5-70B-SFT} \\
\hfrepo{swiss-ai/Apertus-v1.5-70B-RL} \\
\hfrepo{swiss-ai/Apertus-v1.5-70B} 

\paragraph{Code.} We release the full training and evaluation stack: the Megatron-LM fork with the multimodal pretraining and SFT extensions, the multimodal tokenizer, the data ingestion inventory, the post-training framework, and the text and multimodal evaluation harnesses. Further special-purpose repositories, such as the tokenizer benchmarks and the synthetic long-context task generator, are linked in the sections where they are used.

\ghrepoicon{swiss-ai/Megatron-LM} \\
\ghrepoicon{swiss-ai/pretrain-code} \\
\ghrepoicon{swiss-ai/apertus-omni-tokenizer} \\
\ghrepoicon{swiss-ai/multimodal-data} \\
\ghrepoicon{swiss-ai/posttraining-data} \\
\ghrepoicon{swiss-ai/posttraining} \\
\ghrepoicon{swiss-ai/self-distillation} \\
\ghrepoicon{swiss-ai/lm-evaluation-harness} \\
\ghrepoicon{swiss-ai/MLLM-eval-suite} \\
\ghrepoicon{swiss-ai/HELMET} \\
\ghrepoicon{swiss-ai/evals-post-train} \\
\ghrepoicon{swiss-ai/model-launch} \\
\ghrepoicon{swiss-ai/verl-recipe} \\
\ghrepoicon{swiss-ai/verl}

\paragraph{Datasets.} Apertus is open data. All used training datasets are openly available for download and auditable via the list here \ghrepoicon{swiss-ai/apertus-data-transparency}, including reproducible dataset versions and corresponding licenses.
We also release the SFT mixture, the annotated completion pool, and the preference pairs built from it (\autoref{sec:preference_alignment}), as well as the datasets we curate ourselves. The respective sections detail the datasets used in each stage, including exact numbers and the applied preprocessing.

\hfdataicon{swiss-ai/Apertus-v1.5-SFT-mix} \\
\hfdataicon{swiss-ai/Apertus-v1.5-Preference-Pool} \\
\hfdataicon{swiss-ai/Apertus-v1.5-Preference-Data} \\
\hfdataicon{swiss-ai/nasa} \\
\hfdataicon{swiss-ai/owid} \\
\hfdataicon{swiss-ai/smithsonian} \\
\hfdataicon{swiss-ai/swisstopo}

\clearpage
\section{Multimodality}\label{sec:multimodality}

One core objective of \apertus{} 1.5 is to extend the models' capabilities to understanding and working with images and audio. We train both \apertusonefivesmall{} and \apertusonefivelarge{} to understand images and audio and respond in text. The multimodal effort spans continued pretraining, supervised finetuning, and long-context supervised finetuning (\autoref{fig:pipeline_overview}).

This chapter describes how the two new modalities enter the model. \autoref{sec:multimodal_representation} sets out the discrete-token representation and the reasons we prefer it to modality-specific encoders, while \autoref{sec:image_tokenization} and \autoref{sec:audio_tokenization} specify the two tokenizers with the vocabulary they contribute. \autoref{sec:multimodal_data} presents the staged multimodal training data used throughout the pipeline, whereas \autoref{sec:multimodal_infra} describes the tokenization pipelines and their integration into the training infrastructure together with the modifications of the pretraining and SFT codebase to enable multimodal training. Finally, \autoref{sec:multimodal_ablations} introduces ablations on preserving text capabilities while adding modalities during continued pretraining, and \autoref{sec:vision_dataset_tables} and \autoref{sec:audio_dataset_tables} record the data pool of each training stage. The following chapter, \autoref{sec:pretraining_recipe}, presents the continued-pretraining recipe.

\subsection{Unified Representation}\label{sec:multimodal_representation}

\begin{figure}[t]
\centering
\input{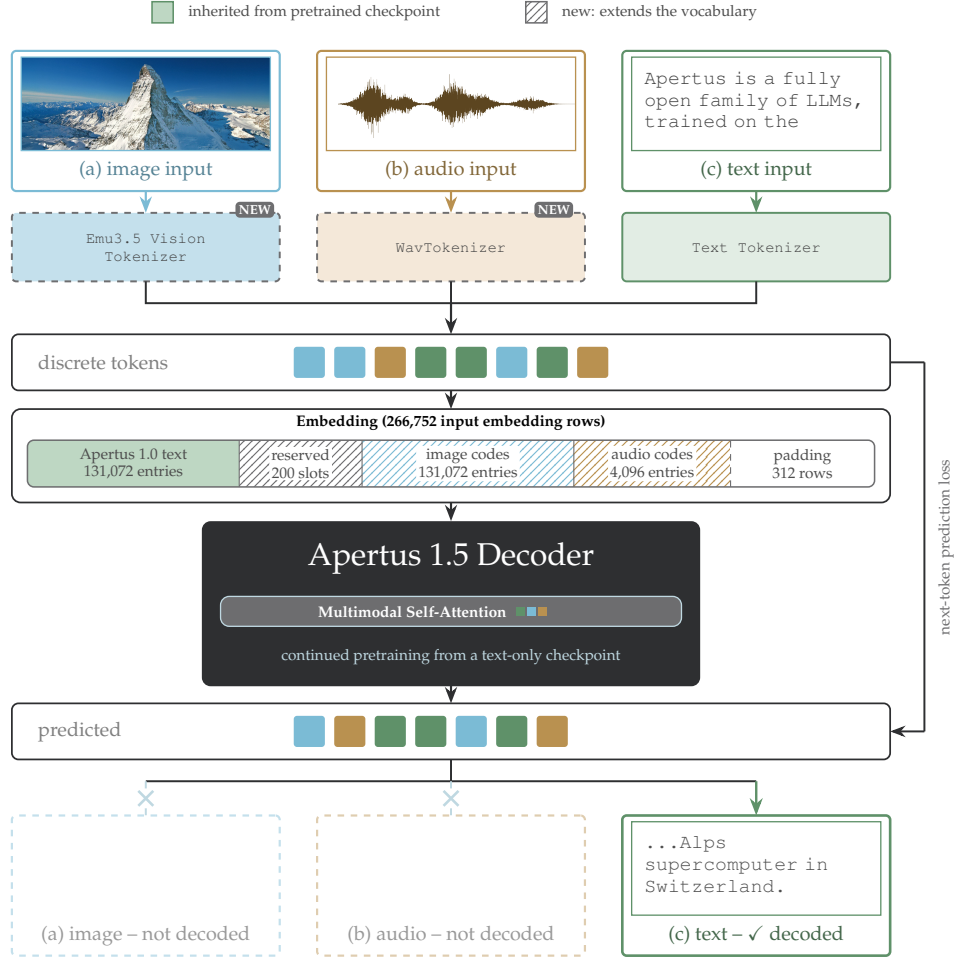}
\caption{The \apertus{} 1.5 architecture.}
\label{fig:apertus_architecture}
\end{figure}

We represent images and audio as discrete tokens, following unified models such as Chameleon and Emu3~\citep{chameleonteam2024chameleonmixedmodalearlyfusionfoundation, wang2024emu3nexttokenpredictionneed}. A shared vocabulary represents all modalities, and the language model applies a single next-token-prediction objective across them, as illustrated in \autoref{fig:apertus_architecture}.

A common alternative is to use modality-specific encoders whose features are projected into the language model, as in LLaVA~\citep{liu2023visualinstructiontuning}. More recent unified approaches instead avoid separate pretrained modality-specific encoders and process continuous inputs directly: SenseNova-U1~\citep{sensenova-si} operates directly on pixels and text in a unified end-to-end architecture, while the encoder-free Gemma~4 12B~\citep{gemmateam2026gemma4technicalreport} directly projects raw image patches and raw audio into the model's embedding space. These approaches unify multimodal computation within a single model, but represent non-text inputs continuously rather than through discrete modality tokens.

Our discrete representation allows the same next-token-prediction objective to be applied to images and audio, enabling the first stage of continued pretraining to draw on unpaired images and audio~\citep{irmiger-etal-2026-learning}. In this stage, the model learns the new modalities from image-only and audio-only examples alongside text, before learning their alignment with language from paired data in the subsequent stage (\autoref{fig:pipeline_overview}). Image and audio codes serve directly as prediction targets, preserving a common training objective across modalities. A controlled study shows that an image-only stage before paired training can improve image understanding on most benchmark groups~\citep{kreft2026representationlearning}; \autoref{sec:multimodal_ablations} summarizes its findings. Unpaired data also needs no captions, so it remains usable where captions are scarce or of low quality.

We convert training images and audio into discrete tokens once, during offline preprocessing (\autoref{sec:image_tokenization}, \autoref{sec:audio_tokenization}), just as we precompute text tokens. We reuse these tokenized datasets across subsequent training runs and experiments that use the same tokenizers and preprocessing. The language model consumes the precomputed tokens through its shared embedding layer; no vision or audio encoder runs in its forward pass. This separation lets us reuse the text-training infrastructure without the overhead and complexity of encoder forward passes during training, achieving throughput comparable to text-only training with only minor modality-specific modifications to the pretraining and SFT codebases (\autoref{sec:multimodal_infra}). A measured comparison substantiates the efficiency benefit: at an identical configuration (256~nodes, 8,192-token sequences), text-only ablation runs sustain 5,300 to 5,500~tokens per second per GPU, and the first multimodal production stage 5,474 (\autoref{sec:training_run_performance}).

\paragraph{Vocabulary Extension.} We extend the 131,072-entry vocabulary of \apertus{} 1.0 with 200 reserved multimodal slots, including structure markers, followed by 131,072~image codes and 4,096~audio codes. These contiguous blocks yield 266,440~vocabulary entries (\autoref{fig:apertus_architecture}). The released input embedding matrix contains 266,752~rows: its final 312~rows are padding, not additional vocabulary entries. The extended input vocabulary allows both models to consume every modality, while the released output head retains only the original text vocabulary (\autoref{sec:optimizer_recipe}).

\subsubsection{Images}\label{sec:image_tokenization}

We use the vision tokenizer of Emu3.5~\citep{cui2025emu35nativemultimodalmodels} to convert training images into discrete tokens during offline preprocessing. \citet{irmiger-etal-2026-learning} evaluate image tokenizers on reconstruction quality, on the average number of tokens per image, and on their support for arbitrary image shapes, and find that the tokenizers of the Emu family combine a favorable trade-off between reconstruction quality and token efficiency with native support for arbitrary shapes. \autoref{sec:multimodal_ablations} discusses the choice between the Emu3 and Emu3.5 tokenizers.\footnote{We provide code and benchmarks at \ghrepo{swiss-ai/benchmark-image-tokenizer}.}

The adopted tokenizer is trained solely with a pixel-reconstruction objective, rather than an explicit semantic or image-text alignment objective as used by vision encoders such as CLIP~\citep{pmlr-v139-radford21a}. We use its pretrained discrete codes directly as inputs and prediction targets for the language model.

The tokenizer maps every block of $16 \times 16$ pixels to one code from a codebook of 131,072~entries of dimension 256. An image of $512 \times 512$ pixels thus becomes a grid of $32 \times 32$ codes, or 1,024~tokens, and an image of arbitrary shape becomes a grid of its height and width each divided by 16. The codebook is trained with index backpropagation quantization, which keeps a codebook of this size in use where earlier vector quantization leaves most codes dead~\citep{shi2025scalableimagetokenizationindex}.

In the token sequence, an image opens with a start marker followed by its dimensions; its rows of codes are separated by end-of-row markers, and an end marker closes it. The visual codes extend the vocabulary directly after the text tokens of \apertus{} 1.0 and a block of 200 reserved slots, seven of which hold these structure markers.

\subsubsection{Audio}\label{sec:audio_tokenization}

We use the WavTokenizer codec~\citep{ji2025wavtokenizerefficientacousticdiscrete} to convert training audio into discrete tokens during offline preprocessing. The codec discretizes a waveform with a single quantizer over a codebook of 4,096~entries. It consumes mono audio at 24~kHz and emits one code per 600~samples, hence 40~tokens per second; a minute of audio costs 2,400~tokens. Inputs at other sampling rates are resampled. Because the codec was trained on peak-normalized audio, we normalize every input to a peak of $-3$~dBFS before tokenization.

Neural audio codecs such as EnCodec quantize each frame with a stack of residual codebooks~\citep{defossez2023highfidelityneuralaudio}, and their parallel code streams force an autoregressive model to interleave them~\citep{copet2023simplecontrollablemusicgeneration}. WavTokenizer emits one code per frame, so audio enters the model as a single flat sequence, exactly like text. This single code stream is the deciding criterion, as it leaves the training stack unchanged.\footnote{We provide code and benchmarks at \ghrepo{swiss-ai/benchmark-audio-tokenizer}.}

In the token sequence, an audio segment is enclosed by start and end markers, which live in the same reserved block of the vocabulary as their visual counterparts. The audio codes follow the visual codes in the extended vocabulary.

\subsection{Multimodal Training Data}\label{sec:multimodal_data}

Multimodal training data is central to this release. It teaches the models to use the image and audio token spaces introduced above. It also aligns both modalities with language and supports multimodal instruction following across continued pretraining and post-training. Across the three continued-pretraining stages, text-only data accounts for roughly 60\% of the \apertusonefivesmall{} training mix and 55\% of the \apertusonefivelarge{} mix, with vision and audio data accounting for the remaining 40\% and 45\%, respectively (\autoref{tab:token-budget}). We organize the discussion around the six data stages that serve these different roles, a design supported by the ablation of \autoref{sec:multimodal_ablations}. The six stages are the three continued-pretraining stages of \autoref{sec:optimizer_recipe}, whose third stage is the cooldown, followed by long-context pretraining (\autoref{sec:long_context}), supervised finetuning (\autoref{sec:sft}), and long-context supervised finetuning (\autoref{sec:long_context_sft}).

\subsubsection{Curation and Processing}\label{sec:multimodal_data_processing}

\paragraph{Acquisition and Normalization.}
The public repository \ghrepo{swiss-ai/multimodal-data} provides an inventory of vision and audio datasets, recording each source's pinned revision, download script, processing recipe, license filters, and training stages. The complementary \ghrepo{swiss-ai/apertus-data-transparency} repository records source versions and licenses across all modalities. We keep raw, processed, and tokenized layers separate to make the transformations between them explicit and repeatable. Download adapters handle repository snapshots, archives, object stores, public APIs, and URL-based image datasets. Processing decodes media, removes invalid records, checks minimum resolution, normalizes schemas, encodes media in standard formats, and converts records to Arrow or WebDataset. Dataset-specific adapters additionally handle large tiled geospatial images, multi-image conversations, spatial annotations, and interleaved documents. When the inventory lists no additional source-specific operation, we apply these shared checks without changing the publisher's filtering decisions.

\paragraph{Deduplication.}
For unpaired continued-pretraining images, we deduplicate samples with perceptual hashes and choose thresholds by domain. Most sources use exact equality of 64-bit hashes after reduction to $32 \times 32$ pixels. Satellite images use a smaller hash input to suppress repeated forest, water, and cloud tiles, while egocentric video frames allow a small Hamming distance to remove adjacent near-duplicates. These rules are inappropriate for text-rich images such as documents and web interfaces, where distinct pages can share a layout and produce the same hash. In early inspections of document and MINT-1T runs, perceptual hashing removed valid OCR examples. We therefore disabled it for later datasets centered on OCR and document understanding.

\paragraph{Caption Generation and Validation.}
For paired continued-pretraining data, we retain informative source captions and regenerate missing, repetitive, or low-information descriptions for selected sources with open-weight vision-language models. The generation pipelines reject empty outputs and apply source-specific output checks. For long captions of text-dense technical drawings, we additionally detect token loops, runaway number lists, and repeated endings caused by repeated labels and tabular schedules. We quarantine and regenerate flagged captions. These automated checks catch common generation failures, but synthetic captions can still contain factual errors. The public recaptioning pipelines and quality-control tools provide the shared generation and validation infrastructure.\footnote{Available at \ghrepo{swiss-ai/multimodal-data}: recaptioning pipelines under \texttt{recaption} and quality-control tools under \texttt{tools}.}

\paragraph{Processing at Scale.}
Preparing these corpora on shared supercomputing infrastructure introduces constraints beyond data quality: unavailable source URLs, storage and inode limits, scratch-retention policies, and network restrictions on bulk downloads. The raw MINT-1T snapshot alone occupies approximately 700~TB, so we compress it early and apply successive filters in place. URL-based datasets create a different problem: collecting the images generates crawl-like traffic across many domains, including destinations on security blocklists. Retries against denied endpoints put the Alps supercomputer's shared outbound IP at risk of blocking, with consequences for unrelated scientific workloads that depend on the same network access. Multimodal data preparation therefore requires coordination with the center's storage, network, and security teams to protect shared infrastructure while sustaining ingestion at scale. We describe the operational incidents and the layered proxy that enforces network safeguards in \autoref{sec:engineering_challenges}.

\subsubsection{Vision Training Data}\label{sec:vision_training_data}

Vision data enters all stages of pretraining and supervised finetuning. The tables in \autoref{sec:vision_dataset_tables} give the vision pool of each stage, with samples and input tokens per dataset, complementing the selection and preparation methodology in \autoref{sec:pretraining_data}.

Across all six stages, the vision pools for \apertusonefivesmall{} contain 821.6~million samples. Most belong to the unpaired and paired stages of continued pretraining (Stages~1 and~2). We report the share of vision data within each continued-pretraining stage's token budget separately in \autoref{tab:token-budget}.

Stages~1 and~2 together contain approximately 86\% of the vision samples across the six 8B stage pools. Their scale explains why the overall composition in \autoref{fig:mm_sample_composition} mostly reflects these early stages, despite the different emphasis of later pools.

\subsubsection{Vision Data Composition}\label{sec:sample_level_vision_composition}

\begin{figure}[!htbp]
    \centering
    \includegraphics[width=0.5\linewidth]{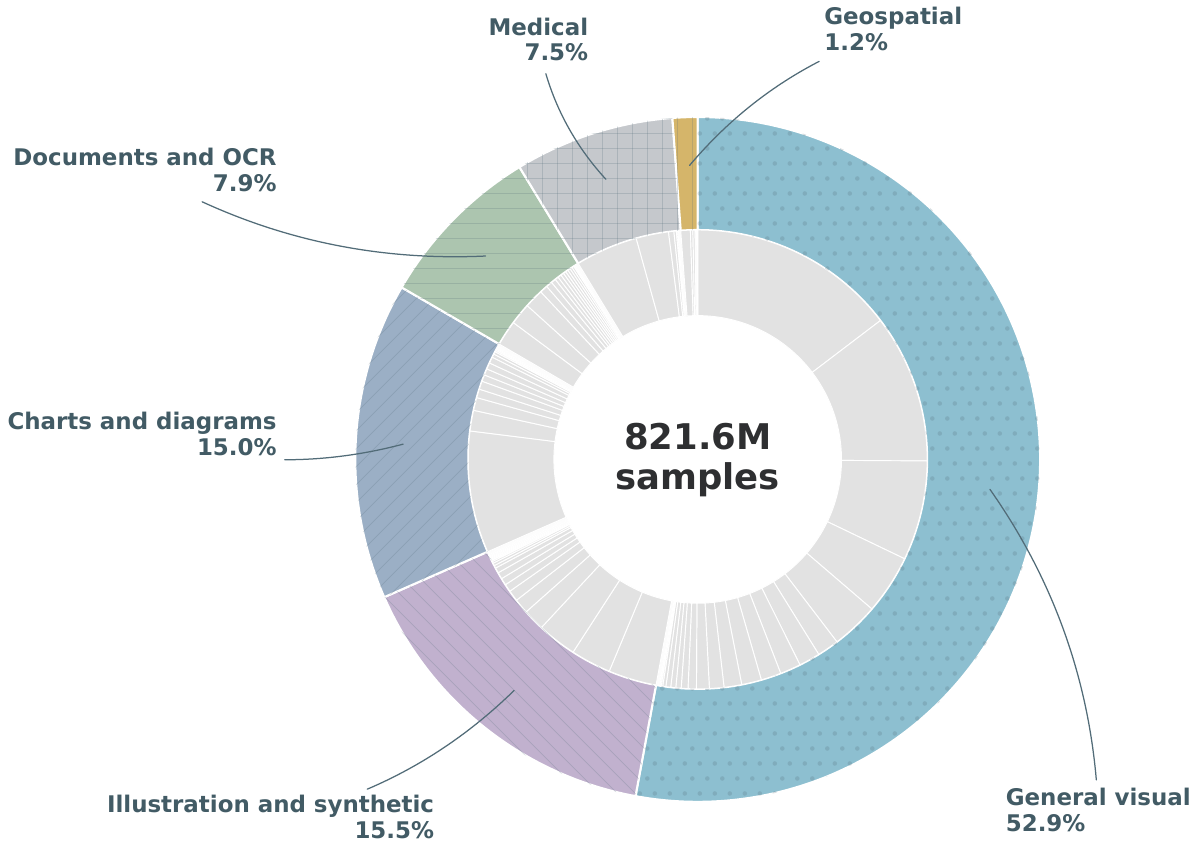}
    \caption{Composition of the 8B vision samples. The inner segments show the relative dataset sizes (largest first). A dataset can span multiple categories.}
    \label{fig:mm_sample_composition}
\end{figure}

We use \texttt{\href{https://huggingface.co/LiquidAI/LFM2.5-VL-450M}{LiquidAI/LFM2.5-VL-450M}} to assign each vision sample to one visual category in the composition figures.

General visual material dominates the overall vision pool (\autoref{fig:mm_sample_composition}). The inner ring shows that a few large datasets supply most of this category, while the other categories draw on a tail of smaller datasets alongside their larger sources. \autoref{app:image_exploration} refines these categories with a taxonomy of 61 fine-grained image types, applied to six large source datasets, and reports the prevalence of broad task signals across the mixture.

\begin{figure}[!ht]
    \centering
    \includegraphics[width=0.99\linewidth]{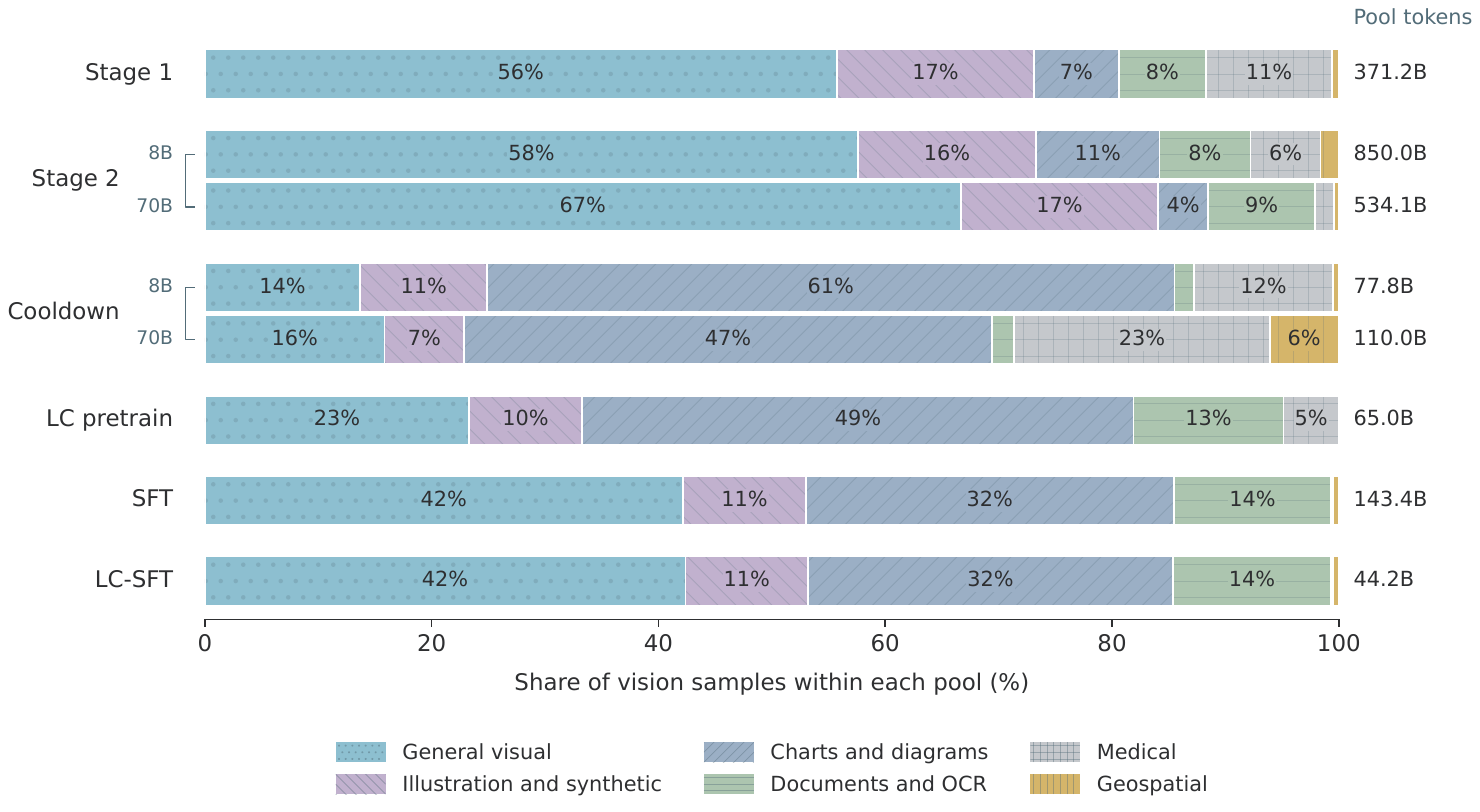}
    \caption{Vision data composition by stage, with pool sizes in input tokens on the right.}
    \label{fig:mm_stage_composition}
\end{figure}

Stage~2 has the largest vision pools, with approximately 850~billion input tokens for \apertusonefivesmall{} and 534~billion for \apertusonefivelarge{} (\autoref{fig:mm_stage_composition}). The composition shifts from general visual material in Stages~1 and~2 toward charts and diagrams in cooldown and long-context pretraining. Medical content is more prominent in the 70B cooldown pool. Geospatial content remains small overall.

\clearpage

\subsubsection{Audio Training Data}\label{sec:audio_training_data}

Audio data enters the same six stages. Stage~1 uses unpaired speech, music, and mixed audio, while the later stages concentrate on paired speech and text. Paired speech takes three forms: transcription pairs, translation pairs, and interleaved sequences that alternate the audio and transcripts of consecutive clips from one recording, with dedicated tokens marking each switch between speech and text. Interleaved sequences make up about a fifth of the Stage~2 and cooldown audio tokens and all of the long-context audio pool (\autoref{tab:audio_data_lc}). The tables in \autoref{sec:audio_dataset_tables} give the audio pool of each stage, with samples and input tokens per dataset.

\begin{figure}[!ht]
    \centering
    \includegraphics[width=\linewidth]{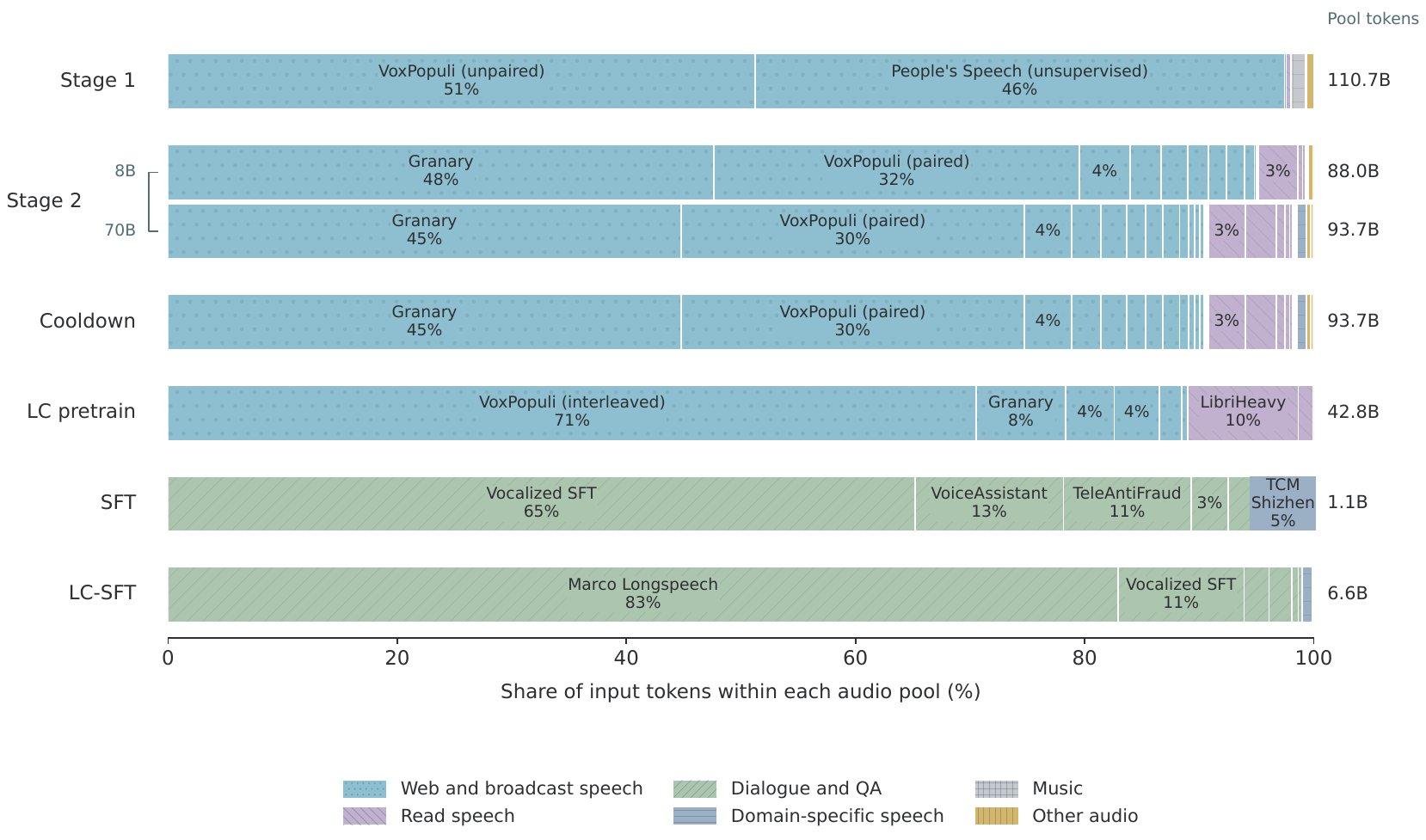}
    \caption{Audio data composition by stage, with pool sizes in input tokens on the right.}
    \label{fig:audio_stage_composition}
\end{figure}

Unlike vision, where we classify samples and weight categories by sample count, we group entire audio datasets into six categories by source and purpose. \autoref{fig:audio_stage_composition} shows their contributions by input tokens.

Web and broadcast speech dominate every pretraining pool. Read speech becomes more prominent in later pretraining stages. Music and other audio remain small components. The 70B Stage~2 and cooldown pools are identical. Dialogue and question answering dominate both SFT pools, with domain-specific speech (such as medical QA) contributing a smaller component.

\paragraph{Interleaved Speech and Text.} For the paired stages, we build interleaved sequences from transcribed speech corpora such as VoxPopuli, Multilingual LibriSpeech, and WenetSpeech. We group the transcribed clips of each recording into runs of consecutive clips. Within a run, a sequence alternates between modalities clip by clip: the audio of one clip is followed by a speech-to-text continuation token and the transcript of the next clip, which is followed by a text-to-speech continuation token and the audio of the clip after that, and every sequence ends with text. We build two copies of each run that start one clip apart, so that a clip that appears as audio in one copy appears as text in the other. Single-clip runs and the leftover clip of a run become transcription sequences that pair one clip's audio with its own transcript. Sequences of up to 8,192~tokens enter Stage~2, and longer ones enter the long-context extension. The builder is released in \ghrepo{swiss-ai/benchmark-audio-tokenizer}.

\paragraph{Speech SFT Data.} The auxiliary checkpoint used for capability filtering of text data (\autoref{sec:sft_capability_filtering}) and the production SFT runs draw from the same prepared audio pool of six speech datasets (\autoref{tab:audio_data_sft}). This pool contains speech inputs with text outputs and no music or environmental-sound supervision. Its largest component by sample count renders roughly 640,000~conversations from eight open chat corpora~\citep{lian2023SlimOrca, zhao2024WildChat, ding2023UltraChat, singh2024AyaDataset, durbin2023Airoboros, kopf2023OpenAssistant}, among them Capybara and Pure-Dove~\citep{daniele2023amplifyinstruct}, to speech with text-to-speech synthesis, so that the assistant behavior of the text mixture carries over to spoken instructions. For this component, we synthesize user turns with Qwen3-TTS~\citep{hu2026qwen3tts}, while retaining assistant responses as text. We align the generated speech to the corresponding source text using Qwen3-ForcedAligner~\citep{shi2026qwen3asr}. During dataset construction, we exclude conversations with unavailable audio and verify that each audio placeholder corresponds to an associated audio segment. The remainder combines a voice-assistant corpus of roughly 242,000~conversations~\citep{xie2024MiniOmni}, spoken question answering with both machine-read and human-read questions~\citep{wu2023HeySQuAD}, a fraud-detection dialogue set~\citep{ma2025TeleAntiFraud}, and a bilingual medical set that mixes Chinese dialogues with English audio question answering and speech recognition~\citep{chen2025shizhengpt}. We process each dataset separately: we decontaminate the vocalized conversations against the evaluation benchmarks, removing 2,153~conversations, and filter them to the languages supported by the synthesis pipeline, removing a further 6,141~conversations; we drop 4,306 identity-related rows from the voice-assistant corpus to protect the model's self-description; and we filter the medical set to its bimodal records (dropping 32,320 image-bearing records) and repair the missing audio placeholders.

\paragraph{Long-Context SFT Data.} The long-context SFT pool in \autoref{tab:audio_data_lc_sft} extends the ordinary SFT pool with Marco Longspeech and AudioMCQ GeminiCoT, as well as additional TeleAntiFraud and Vocalized SFT samples. Marco Longspeech is its largest component by token count. AudioMCQ GeminiCoT contributes only the speech partition of its source, with its reasoning traces placed in the thinking span, and enters the short-context bucket of both models' long-context SFT runs. The pool inventory gives the available data; \autoref{tab:long_context_sft_mix} reports the per-source counts used by each model.

\subsection{Tokenization and Training Pipelines}\label{sec:multimodal_infra}

Megatron-LM~\citep{shoeybi2020megatronlmtrainingmultibillionparameter} natively supports multimodality, with two limitations. First, it supports only continuous features: neither the multimodal models of Megatron Core nor the sample types of Megatron Energon carry discrete tokens~\citep{patel2025trainingvideofoundationmodels}. Second, it computes everything inside a training job: vision and audio encoders and a projection layer embed pixels and waveforms in the forward pass, and the supervised finetuning paths tokenize conversations and derive the loss mask from the chat template at load time. This adds encoder compute and media decoding to every step, and it leaves no entry point for pre-tokenized multimodal sequences, in pretraining or in supervised finetuning.

Following prior work on incorporating discrete multimodal representations into pretrained LLMs~\citep{ge2024making, zhan-etal-2024-anygpt}, we extend the base LLM by adding dedicated vocabulary ranges for vision and audio tokens. Each new token is assigned a corresponding embedding, allowing tokenized media to be consumed by the transformer in the same way as text tokens. The vocabulary is extended correspondingly, and the output projection is extended to produce logits for the newly introduced vision and audio tokens in addition to the original text vocabulary. The newly introduced parameters are randomly initialized, while all parameters inherited from the pretrained LLM are retained. Thus, the transformer itself requires no modality-specific computation: multimodal sequences can be processed with the standard autoregressive next-token prediction objective.

We therefore build two high-throughput tokenization pipelines from scratch, one for images and one for audio, that write Megatron indexed datasets directly. \autoref{sec:multimodal_data_processing} describes the upstream acquisition and processing that feed them. The pipelines convert each image or waveform into token ids from the corresponding added vocabulary range. This is the payoff of discrete tokens: a codec maps an image or a waveform to ids from a contiguous block of the vocabulary, so to the model a tokenized image is a text document in a language it has not seen before. The multimodal shards enter the training data mixture through the standard data blending and sampling of Megatron-LM, and stream through the unmodified pretraining data loader. Hence, continued pretraining can run on an almost vanilla Megatron-LM installation.

The necessary changes to Megatron-LM live in our fork (\ghrepo{swiss-ai/Megatron-LM}). For pretraining, we add options for masking multimodal tokens from prediction, which enables the text-only targets from the second training stage onward (\autoref{fig:pipeline_overview}), and for exempting the multimodal special tokens from Goldfish-loss dropping (\autoref{sec:optimizer_recipe}). For supervised finetuning, we add the missing entry point: a loader that reads pre-tokenized conversations from the same indexed format and derives the loss mask efficiently on the fly at the token level, masking user turns and tool responses without a separate preprocessing step or added overhead in the training step.

\subsubsection{Image Tokenization}

The pipeline, released in \ghrepo{swiss-ai/benchmark-image-tokenizer}, separates indexing, planning, and tokenization. A scan reads only tar headers and the first few kilobytes of each image, recording byte offsets and dimensions for hundreds of millions of samples in minutes, and never touches pixels. From the manifest a deterministic plan assigns every image to a GPU batch: images are grouped in windows of consecutive source rows, snapped to document boundaries, sorted by their exact post-resize shape, and packed to a token budget, so that every batch is shape-uniform and byte-adjacent, the ranks never communicate, and the measured rank imbalance stays below 2\% at 800~GPUs. Planning is NumPy-only and costs seconds, so re-planning for a different batch size or GPU count never requires a rescan. On each rank, loader threads decode, filter, and resize images while the GPU encodes the previous batch, and text tokenization and chat-template rendering run concurrently with the encoder, which keeps the encoder at roughly 92\% of wall time. Each rank writes its own Megatron indexed-dataset shards and checkpoints after every batch, so an interrupted job resumes from its last written batch, and a final step merges the shards; for SFT data, the merge can also write a copy with the reasoning spans removed. Two changes to the encoder itself (selecting codes by argmax rather than through a 131{,}072-way softmax and one-hot product, and pinning TF32 as an explicit precision contract), together with host-side overlap, roughly double throughput without changing a single emitted token, whereas bf16 would flip 7\% of codes and we therefore reject it.

\subsubsection{Audio Tokenization}

Audio tokenization, released in \ghrepo{swiss-ai/benchmark-audio-tokenizer}, combines the main systems challenges of text and image processing. Text inputs are variable-length, but tokenization is CPU-bound and scales across many cores. Image tokenization is GPU-bound, but resizing enables shape-homogeneous batches with high accelerator utilization. Audio is both GPU-bound and variable-length, making efficient batching difficult without excessive padding or underutilization. The pipeline first converts every source into Lhotse Shar shards~\citep{zelasko2021lhotse} and tokenizes any paired text during that conversion, so the GPU stage handles only audio. Each rank receives its own set of whole shards and never communicates with the others, and a dynamic bucketing sampler groups clips of similar duration into batches capped by total audio duration while CPU workers decode the next batch. Because batched encoding pads each clip to the longest in its batch and the last few codes before the padding are unstable, the pipeline drops the last five codes after padding.

\subsection{Ablations}\label{sec:multimodal_ablations}

Introducing new modalities into a pretrained language model can degrade its text capabilities, particularly during continued pretraining. \citet{lu2024deepseekvlrealworldvisionlanguageunderstanding} report catastrophic forgetting of language capabilities during joint vision-language pretraining and retain at least 70\% text data to mitigate it. Similarly, \citet{kimiteam2026kimik25} observe an immediate drop in text performance when vision data is introduced to a text-pretrained model, followed by only gradual recovery during subsequent training.

Preserving the language capabilities of the base model is therefore a primary constraint throughout the development of \apertus{}~1.5, alongside acquiring strong multimodal capabilities. This section studies the design choices of the continued pretraining that bear on this constraint; \autoref{sec:sft_ablations} turns to post-training, where multimodal data interacts with reasoning supervision.

Two design decisions of the continued pretraining rest on dedicated experiments: the choice of the image tokenizer, which trades reconstruction quality for token efficiency, and the two-stage recipe, which separates representation learning on unpaired images from modality alignment on paired data.

\paragraph{Image Tokenizer.} \citet{irmiger-etal-2026-learning} find that the tokenizers of the Emu family offer the best trade-off between reconstruction quality and token efficiency (\autoref{sec:image_tokenization}). Within that family, we choose the tokenizer of Emu3.5~\citep{cui2025emu35nativemultimodalmodels} over that of Emu3~\citep{wang2024emu3nexttokenpredictionneed} for its token efficiency. Emu3 encodes one code per $8 \times 8$ pixel patch and Emu3.5 one per $16 \times 16$ patch, so Emu3.5 represents the same image with a quarter of the tokens.

\begin{figure}[t]
\centering
\includegraphics[width=\linewidth]{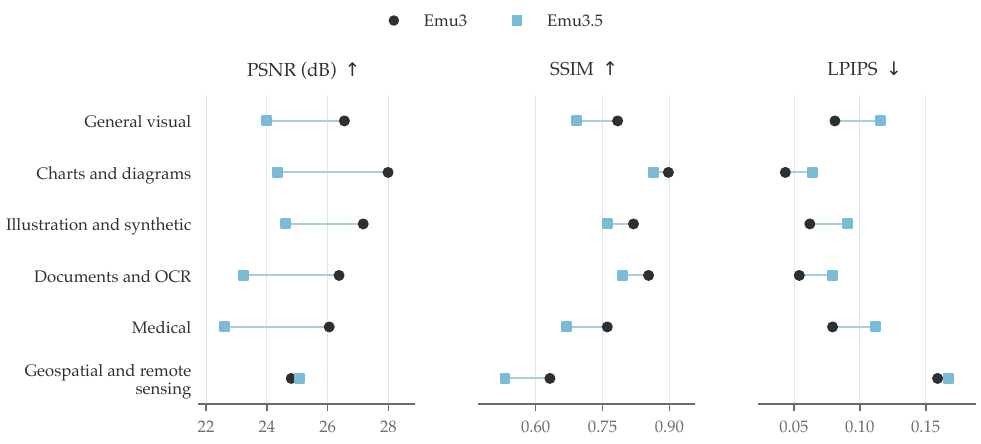}
\caption{Per-category reconstruction quality of the Emu3 and Emu3.5 tokenizers on the same 350 Stage~1 training images as \autoref{tab:emu_tokenizers}. Each marker is a category mean. Higher is better for PSNR and SSIM; lower is better for LPIPS. \autoref{tab:emu_tokenizers_categories} lists the values.}
\label{fig:emu_tokenizer_dumbbell}
\end{figure}

To quantify what this compression costs on our data, we benchmark both tokenizers on a stratified sample of the Stage~1 vision mixture, built on the sample-level categories of \autoref{sec:sample_level_vision_composition}. We draw 50~images per category with a fixed seed and report a representative subset of six categories, 350~image pairs, that spans natural photographs, synthetic imagery, structured graphics, text, and the two specialist domains. We omit the categories not listed and merge charts and diagrams into one category, as in the composition figures. We resize each image as we do for training \apertus{}, encode and decode it with each tokenizer, and score the reconstruction against the input with the peak signal-to-noise ratio (PSNR), the structural similarity index (SSIM), and the learned perceptual image patch similarity (LPIPS) (\autoref{tab:emu_tokenizers} and \autoref{fig:emu_tokenizer_dumbbell}).

Emu3.5 uses a quarter of the tokens of Emu3 in every category, at a cost of 2.7~dB PSNR, 0.07~SSIM, and 0.02~LPIPS averaged over the subset. The gap is consistent across categories: Emu3 reconstructs better on every metric except PSNR on geospatial imagery, where the two are almost level, and the largest PSNR losses fall on charts and diagrams, medical images, and documents, each between 3 and 4~dB. Medical images and documents are the two categories in which Emu3.5 reconstructs worst in PSNR, which confirms our qualitative observation that fine-grained detail, and text in particular, is its main weakness. We accept this trade: because image tokens typically dominate the token count of an image sample, the fixed token budget of each pretraining stage covers approximately four times as many images, and inference on image inputs is correspondingly cheaper. Although our models see more images during training, the coarser grid is the likely reason that OCR remains the capability group in which they trail other open models (\autoref{tab:image_eval}).

\begin{table}[!htbp]
\caption{Reconstruction quality and token cost of the Emu3 and Emu3.5 tokenizers, averaged over 350 Stage~1 training images from six representative categories of \autoref{sec:sample_level_vision_composition} and preprocessed as in training. Tokens are the mean per image, arrows give the better direction, and bold marks the better tokenizer; \autoref{tab:emu_tokenizers_categories} in \autoref{app:emu_tokenizers_categories} lists the values per category.}
\label{tab:emu_tokenizers}
\centering
\small
\begin{tabular}{@{}lrrrr@{}}
\toprule
\textbf{Tokenizer} & \textbf{Tokens} $\downarrow$ & \textbf{PSNR (dB)} $\uparrow$ & \textbf{SSIM} $\uparrow$ & \textbf{LPIPS} $\downarrow$ \\
\midrule
Emu3 & 8,372 & \textbf{26.70} & \textbf{0.806} & \textbf{0.075} \\
Emu3.5 & \textbf{2,080} & 24.04 & 0.740 & 0.099 \\
\bottomrule
\end{tabular}
\end{table}

\paragraph{Two-Stage Recipe.} A controlled study underpins the design of the first two pretraining stages of \autoref{sec:optimizer_recipe}~\citep{kreft2026representationlearning}. The study extends Llama 3.2 3B with the same Emu3.5 tokenizer and splits the topic-balanced LLaVA-OneVision 1.5 image-caption corpus~\citep{llava-onevision-1-5} in half: one half, stripped to standalone images, feeds a representation-learning stage that predicts image tokens, mirroring our Stage~1; the other half, kept as image-caption pairs, feeds a subsequent modality-alignment stage, mirroring our Stage~2. Both halves share one distribution, and every batch replays 10\% text-only data. Three findings carry over. First, giving image and text tokens equal loss weight during modality alignment is harmful: this setting scores far below all others on nearly every group of image-understanding benchmarks, showing that image-token prediction and modality alignment interfere when they share compute. Second, zero weight on image tokens during modality alignment is the strongest single setting for image understanding, best on the document-understanding, grounding, and hallucination benchmark groups, while variants that end the modality-alignment stage at zero weight stay competitive on the remaining groups; this backs our text-only Stage~2 loss targets. Third, the representation-learning stage pays off downstream: with the modality-alignment stage fixed, raising its budget from zero to roughly 57B image tokens lifts image understanding measured after modality alignment and SFT, for example GQA from 0.50 to 0.54, DocVQA from 0.67 to 0.71, and SEED-Bench from 0.61 to 0.65, with logarithmic gains that flatten at larger budgets; this backs our unpaired all-token Stage~1 and its bounded token budget. At the largest budgets, however, knowledge-heavy image benchmarks and text benchmarks erode, as both rely on the model's language capabilities; the deficits persist despite the replayed text and do not fully recover in the later stages, motivating the substantially larger text share of our mixtures (\autoref{tab:token-budget}).

\clearpage
\section{Continued Pretraining Recipe}\label{sec:pretraining_recipe}

\apertus{} 1.5 is obtained through continued pretraining of \apertus{} 1.0, with the goal of introducing multimodal capabilities, including image and audio, while preserving the text capabilities of \apertus{} 1.0. We start from the \apertus{} 1.0 8B and 70B checkpoints and proceed through four stages of continued pretraining (\autoref{fig:pipeline_overview}): unpaired multimodal training, paired multimodal training, and cooldown with higher-quality data, all at 8K sequence length, followed by long-context extension to support context lengths up to 262K.

This chapter gives an overview of the continued-pretraining data and recipe. \autoref{sec:pretraining_data} describes the revised data mixture including source selection, data processing, and provenance requirements. \autoref{sec:optimizer_recipe} specifies the optimizer, learning-rate schedule, prediction targets, and model-specific token budgets. \autoref{sec:long_context} describes the long-context extension and its data mixture. Finally, \autoref{sec:pretraining_retrospective} examines the training dynamics and lessons from the production runs.

\subsection{Pretraining Data}\label{sec:pretraining_data}

We continue training on a new mixture based on the \apertus{} 1.0 text mixture, refined with updated and additional text datasets and extended with image and audio data. We use only publicly available datasets and document their provenance and filtering. This subsection describes the data governance requirements shared across modalities, including compliance, provenance, and opt-out handling (\autoref{sec:data_compliance}), followed by the selection and refinement of the text mixture (\autoref{sec:data_selection}). We maintain the source versions and licenses for the full mixture in \ghrepo{swiss-ai/apertus-data-transparency} and the source-specific vision and audio processing recipes in \ghrepo{swiss-ai/multimodal-data}.

We introduce multimodality through the discrete-token early-fusion approach presented in \autoref{sec:multimodal_representation}: images and audio as discrete tokens that the model processes alongside text in a shared token space. These new token spaces motivate the staged recipe shown in \autoref{fig:pipeline_overview}. In Stage~1, we combine text with unpaired images and audio and train the model to predict tokens of all modalities. In Stage~2, we introduce paired image-text and audio-text sequences and restrict prediction targets to text, aligning the new modalities with language. The cooldown then emphasizes higher-quality and targeted sources, and the long-context extension adds longer and interleaved records. \autoref{sec:optimizer_recipe} specifies the model-specific token budgets, sampling ratios, and optimization schedule.

Details of the multimodal data recipe appear in \autoref{sec:multimodality}, which describes the multimodal design, the rationale for the staged training recipe (\autoref{sec:multimodal_ablations}), and data preparation (\autoref{sec:multimodal_data}). \autoref{sec:vision_training_data} and \autoref{sec:audio_training_data} describe the vision and audio data, respectively, while \autoref{sec:vision_dataset_tables} and \autoref{sec:audio_dataset_tables} list the prepared pools across stages. For multimodal sources, we apply the acquisition, filtering, deduplication, and caption validation procedures described in \autoref{sec:multimodal_data_processing}.

\subsubsection{Compliance and Provenance}\label{sec:data_compliance}

Every new modality is also required to satisfy the same requirements as the text data in \apertus{} 1.0~\citep{apertus2025apertusdemocratizingopencompliant}. Our review covers the upstream source, revision, declared license, constituent datasets, and sample-level provenance where available. For composite medical and remote-sensing collections, we select approved constituent sources or records rather than relying on the aggregate license. Non-commercial, research-only, and share-alike subsets are excluded, and the spreadsheet in \ghrepo{swiss-ai/apertus-data-transparency} remains the authoritative record of the final decisions. No paywalled or otherwise closed sources are used for training. The text pipeline additionally removes personally identifiable information and applies the multilingual toxicity filters of \apertus{} 1.0.

We apply machine-readable opt-outs (robots.txt files) following the same recipe as \apertus{} 1.0, and we honor robots exclusions retroactively: if a page owner opts out this year, previous crawls of the same URL are excluded as well. The opt-out filtering also applies to multimodal content when a record retains the page URL on which the content appeared. For example, MINT-1T HTML preserves original page URLs, so we apply the same domain exclusion list as the text pipeline. For images, a strict allow-list limited to domains represented in the FineWeb robots.txt database would discard roughly 96\% of MINT-1T, whereas the exclusion rule used by the text pipeline removes roughly 2\%. For datasets that retain only an image-hosting URL or the name of another dataset, we record that domain-level opt-out filtering is not technically available, rather than treating the hosting service as the content owner.

We materialize URL-based datasets by downloading original images from source URLs on the shared Alps supercomputer at CSCS, and therefore separately apply network safety filtering to protect the center's shared outbound network infrastructure. \autoref{sec:engineering_challenges} describes the proxy and blocklists. This safeguard is distinct from license review, opt-out handling, and image content review; it does not constitute a comprehensive scan for sexual or violent content. For image and audio content, we otherwise retain the filtering supplied by the dataset publishers.

\subsubsection{Text Data Selection}\label{sec:data_selection}

We refine the text mixture of \apertus{} 1.0 by replacing FineWeb2-HQ with FineWeb2-HQ+~\citep{turki2026crosslingual} and adding FinePDFs, Nemotron-CC-v2.1, SYNTH, FineTranslations, Nemotron-Pretraining-Code-v1, and legal-domain corpora. \autoref{tab:data_mixture} reports the per-stage sampling proportions of the resulting text mixture.

Every added source must be publicly available, pass license and provenance review, and improve on the \apertus{} 1.0 mixture in a small-scale ablation. The baseline experiment trains for 40,000~steps on FineWeb followed by a 10,000-step cooldown on FineWeb-Edu. A candidate source replaces 30\% of the cooldown data, and we evaluate it on the standard development suite.\footnote{The data ablation results are published at \url{https://api.wandb.ai/links/bmessmer/2l4dpwl1}.} For specialized sources, such as legal or medical data, we require an improvement on the corresponding domain evaluations without a material regression on the general suite. We also prefer sources that retain original URLs and other provenance fields needed for later filtering.

For multilingual data, we extend the FineWeb2-HQ dataset~\citep{messmer2025enhancingmultilingualllmpretraining} used for \apertus{} 1.0 to an improved variant, FineWeb2-HQ+~\citep{turki2026crosslingual}. The dataset retains the 10\% of FineWeb-2 documents per language rated highest by a multilingual quality classifier, covering the 100 highest-resource languages and a 10\% sample of the remaining languages (originally, FineWeb2-HQ's quality filter covered only 20~languages). The classifier recipe changes in two ways. First, we train a single classifier jointly on all languages from a pool of positive samples widened with multilingual instruction data (including MURI-IT, Tagengo, and WikiQA), with lower-resource languages upsampled by at most a factor of three so that high-resource languages do not dominate the decision boundary. Second, its negative samples are drawn from the 50th--75th percentile of a proxy quality classifier rather than uniformly from FineWeb-2, which sets the boundary against fluent but low-utility text instead of obvious noise~\citep{turki2026crosslingual}.

\subsection{Optimizer and Training Recipe}\label{sec:optimizer_recipe}

We continue pretraining both \apertus{} 1.5 models, the 8B and 70B, with the AdEMAMix optimizer~\citep{pagliardini2024ademamixoptimizerbetterfaster} from \apertus{} 1.0~\citep{apertus2025apertusdemocratizingopencompliant}, retaining the same momentum hyperparameters. We resume from the \apertus{} 1.0 checkpoints that have completed cooldown and use a warmup-stable-decay schedule: both models warm up from zero to their peak learning rates over 4.2B~tokens; the learning rate then stays at its peak through the first two stages and decays to 10\% of the peak with a $1-\text{sqrt}$ decay over the third. The 8B model keeps the peak learning rate of \apertus{} 1.0 8B, $1.1 \times 10^{-4}$. For the 70B model, we double the peak of \apertus{} 1.0 70B to $2 \times 10^{-5}$, which outperforms the original value in ablations, suggesting that the \apertus{} 1.0 70B peak learning rate was likely too small. To accommodate multimodal data, we double the sequence length from 4,096~tokens to 8,192~tokens. An image of $512 \times 512$ pixels is tokenized into 1,024~tokens and a minute of audio into 2,400~tokens (\autoref{sec:image_tokenization}, \autoref{sec:audio_tokenization}); multimodal sequences interleaved with text documents routinely exceed the previous 4,096-token limit, and the 8,192-token window ensures that entire documents fit within a single sequence. We halve the global batch size so that the number of tokens per batch remains the same as in \apertus{} 1.0, and reuse the same tensor and pipeline parallelism configurations to train \apertus{} 1.5. \autoref{tab:cpt_config} summarizes the resulting configuration of both models.

We apply weight decay to all trainable parameters, including normalization gains and the trainable parameters of the xIELU activation~\citep{huang2024xielu}, which we find helps reduce large activation and activation-gradient norms. We initialize the optimizer states from scratch, since the newly added input and output embeddings for image and audio tokens have no existing optimizer states. The corresponding embedding and output rows for the new tokens are randomly initialized, while the text rows keep their pretrained values; the fresh rows start at roughly a tenth of the norm of the trained text rows and adapt over the first pretraining stage. For the 70B model, we also evaluate splicing in the multimodal embeddings learned by the 8B model, but we observe that it destabilizes training, so we use random initialization instead. In the released checkpoints, we prune the output rows for image, audio, and reserved tokens from the language-model head, since the models never generate non-text tokens and the text logits are unchanged.

We keep the Goldfish loss of \apertus{} 1.0 with its parameters unchanged: it drops the prediction target of a random subset of positions, thereby preventing verbatim memorization of the training data and respecting copyright~\citep{hans2024likegoldfishdontmemorize, goldfish_calibration_thesis}. We extend this dropping to cover the predicted tokens of every modality; only special tokens, among them the markers that structure image and audio sequences, are exempt. Image and audio tokens are prediction targets in the first stage alone. The later stages predict text only, which prevents the model from generating non-text tokens. From the second stage, we likewise exclude from prediction the task-control markers that select between transcription and continuation: they condition the model but are never targets.

Before the long-context extension, continued pretraining proceeds in three stages: unpaired multimodal training, paired multimodal training, and cooldown with higher-quality data. The stages are summarized in \autoref{tab:cpt_stages} and \autoref{tab:token-budget}, and the text data mix in \autoref{tab:data_mixture}; the vision and audio datasets of every stage are listed in \autoref{sec:vision_dataset_tables} and \autoref{sec:audio_dataset_tables}. Each stage's dataset list is frozen before the stage begins: the deterministic sampling plan depends on the full list, and mid-stage changes would shift that plan, silently repeating or skipping documents. In the first stage we train on unpaired image, audio, and text data, with the loss computed over tokens of all modalities. In the second and third stages we add the respective parallel data (image-text and audio-transcript pairs) along with text-only data, and compute the loss over the text tokens only, since the models understand but never generate images and audio (\autoref{sec:multimodality}); an ablation further shows that text-only targets during paired training also yield the strongest image understanding (\autoref{sec:multimodal_ablations}). For the 8B, the Stage~2 text share rises to roughly two thirds, in part because multimodal data was still limited; the 70B, with its halved token budget, keeps a roughly even split between text and multimodal data, yet its text evaluations do not degrade. The third stage is the cooldown, in which we emphasize higher-quality data and include additional paired speech sources (\autoref{tab:audio_data_stage_two_cooldown}). The audio portion of the cooldown accounts for 95B~input tokens for \apertusonefivesmall{} and 92B for \apertusonefivelarge{} (\autoref{tab:token-budget}). These training counts include paired text and any repeated sampling; they do not measure unique hours of speech. The cooldown also includes a small share of instruction-formatted data (roughly 3B~tokens), for which the loss is restricted to assistant tokens. We train the 70B model on half the intended number of tokens to free resources for the next \apertus{} pretraining project, keeping the cooldown length unchanged and reducing the token budget mainly during Stage~2.

\begin{table}[!htbp]
\caption{Stages~1 and~2 fall in the warmup and stable phases of the LR schedule, while Stage~3 is the decay phase. Stage~1 is trained to predict tokens of all modalities. Stages~2 and~3 are trained to predict only text tokens.}
\label{tab:cpt_stages}
\centering
\small
\begin{tabularx}{\textwidth}{@{}cYl@{}}
\toprule
\textbf{Stage} & \textbf{Data} & \textbf{Prediction Targets} \\
\midrule
1 & Unpaired image, audio, and text data.  & All modalities \\
2 & Paired image-text, audio-transcript data, and text data  & Text only \\
3 & Higher-quality data, with some instruction-formatted data & Text only \\
\bottomrule
\end{tabularx}
\end{table}

\begin{table}[!htbp]
\caption{Token budget per modality across the three curriculum stages of \apertusonefivesmall{} and \apertusonefivelarge{}, that is, the mix each run trains on. Percentages are the within-stage share of tokens. For Stages~2 and~3, the Vision and Audio counts include their paired text (image captions and audio transcripts); the Text column is the separate text-only data.}
\label{tab:token-budget}
\centering
\small
\begin{tabular}{@{}l rrr r@{}}
\toprule
\textbf{Stage} & \textbf{Vision} & \textbf{Audio} & \textbf{Text} & \textbf{Total} \\
\midrule
\multicolumn{5}{@{}l}{\textit{\apertusonefivesmall{}}} \\
1 & 367B (45.6\%) & 110B (13.7\%) & 328B (40.7\%)   & 806B \\
2 & 852B (30.4\%) &  89B (3.2\%) & 1{,}861B (66.4\%) & 2{,}802B \\
3 &  78B (19.4\%) &  95B (23.4\%) & 230B (57.2\%)   & 403B \\
\addlinespace
\multicolumn{5}{@{}l}{\textit{\apertusonefivelarge{}}} \\
1 &  49B (12.2\%) &  52B (13.0\%) & 301B (74.8\%) & 403B \\
2 & 512B (42.4\%) &  90B (7.4\%) & 606B (50.1\%) & 1{,}208B \\
3 & 107B (26.6\%) &  92B (22.7\%) & 204B (50.7\%) & 403B \\
\bottomrule
\end{tabular}
\end{table}

\newpage

The full text data mixture for each stage is given in \autoref{tab:data_mixture}. Stages~1 and~2 share a single mixture dominated by high-quality web text (\hfdata{HuggingFaceTB/dclm-edu} and the top 10\% of \hfdata{epfml/FineWeb2-HQ-Plus}), supplemented by mathematics, code, translation, and legal corpora. The Stage~3 cooldown deliberately trades general web text for higher-quality and reasoning-heavy data: the share of general web text is cut substantially, while the combined code-and-mathematics allocation rises sharply, driven by larger allocations to mathematics (\hfdata{HuggingFaceTB/finemath}, \hfdata{IFM/MegaMath}, \hfdata{Infi-MM/InfiMM-WebMath-40B}) and code (\hfdata{HuggingFaceTB/stack-edu}, \hfdata{nvidia/Nemotron-Pretraining-Code-v1}). At the same time, the general legal corpora are dropped in favor of \hfdata{voilaj/swiss-caselaw}, the translation share rises (\hfdata{HuggingFaceFW/finetranslations}), and a small amount of \apertus{} 1.5 SFT data is mixed in.

\begin{table}[!htbp]

\setlength{\tabcolsep}{4pt}
\caption{Text data mixture per training stage, as a percentage of each stage's sampled text-only token budget.}
\label{tab:data_mixture}
\centering
\small
\begin{tabular}{@{}lrr@{}}
\toprule
\textbf{Name} & \textbf{Stage 1+2 (\%)} & \textbf{Stage 3 (\%)} \\
\midrule
\hfdata{HuggingFaceTB/dclm-edu} & 47.5 & 20.5 \\
\hfdata{HuggingFaceFW/finepdfs-edu} & 2.8 & 3.2 \\
\hfdata{nvidia/Nemotron-CC-v2.1} & 1.0 & 7.2 \\
\hfdata{PleIAs/SYNTH} & 0.7 & 0.0 \\
\hfdata{epfml/FineWeb2-HQ-Plus} (top 10\%) & 20.1 & 14.4 \\
\hfdata{HuggingFaceFW/finepdfs-edu} (multilingual only) & 8.5 & 3.6 \\
\hfdata{HuggingFaceFW/finetranslations} & 2.8 & 5.9 \\
\hfdata{HuggingFaceTB/finemath} & 0.9 & 6.3 \\
\hfdata{IFM/MegaMath} & 6.1 & 9.5 \\
\hfdata{Infi-MM/InfiMM-WebMath-40B} & 0.6 & 4.2 \\
\hfdata{HuggingFaceTB/stack-edu} & 2.1 & 6.2 \\
\hfdata{nvidia/Nemotron-Pretraining-Code-v1} & 5.3 & 15.1 \\
\hfdata{joelniklaus/legal-mc4} & 0.7 & 0.0 \\
\hfdata{joelniklaus/Multi\_Legal\_Pile} & 1.0 & 0.0 \\
\hfdata{voilaj/swiss-caselaw} & 0.0 & 2.7 \\
\apertus{} 1.5 SFT data & 0.0 & 1.2 \\
\bottomrule
\end{tabular}
\end{table}

\subsection{Long-Context Extension}\label{sec:long_context}
The long-context extension is the fourth and last stage of continued pretraining (\autoref{fig:pipeline_overview}), and follows the three stages of \autoref{sec:optimizer_recipe}. We extend the context window in four steps, from 8,192 to 32,768, 65,536, 131,072, and finally 262,144~tokens. Each step starts from the checkpoint produced by the previous one and reinitializes the optimizer states. Depending on the phase and model size, we train on 50B--63B~tokens per step.

We keep the learning rate fixed at the final value reached during cooldown. This provides a conservative continuation of training after cooldown, limiting changes to the model while adapting it to longer sequences. We also include synthetic long-context tasks to explicitly train capabilities that require using information over long distances. Whenever we double the sequence length, we halve the number of sequences per batch so that the total number of tokens per batch remains constant.

We also adjust the rotary position embeddings as the context window grows. The main runs initially keep the base frequency of 500,000 and scaling factor of 8 used in \apertus{} 1.0. By the end of the four extension stages, the final model uses a base frequency of 4M and a scaling factor of 32. Documents are packed using a best-fit-decreasing strategy~\citep{best-fit}. Cross-document attention masking under context parallelism also makes training substantially cheaper: training at 131,072~tokens costs roughly the same as training at 65,536~tokens did for \apertus{} 1.0.

The prepared data for each step keeps the same relative composition regardless of phase or model size: most tokens come from a subsample of the cooldown mixture, and the remainder splits into genuinely long documents and synthetic verifiable long-context tasks. Because each step trains on fewer tokens than this prepared pool, \autoref{app:long_context_mix} gives the exact, already-scaled per-dataset, per-step token counts and sources actually used in training. As throughout pretraining, we preprocess and filter every dataset before use. The long documents, split into English text, multilingual text, and multimodal image-text and audio corpora, are length-filtered subsets of similar pretraining sources used earlier in the recipe; we draw them proportionally to each source's own size so that no single source dominates a step. \autoref{tab:long_context_class_budget} gives the token budget and within-phase share of each class.

Each step targets a fixed image budget, but the natively length-filtered image pool cannot supply enough tokens at the 32k, 65k, and 131k phases, since fewer real documents in the pretraining corpora happen to span exactly those lengths; only the 262k phase has enough native long-context images on its own. We close the gap at the shorter phases by drawing additional images from short-context documents (1,000--16,384~tokens) so that every phase reaches the same image budget: native long-context images make up about 79\% of the 32k budget, 47\% of the 65k budget, and 34\% of the 131k budget, with the short-context top-up filling the rest. \autoref{tab:vision_data_lc} gives further per-dataset sample counts for the vision pool.

The remaining tokens per step are synthetic verifiable long-context tasks from \ghrepo{swiss-ai/LongContextQA}: we inject Q\&A turns into a long document and re-tokenize it as one training example. Common-word extraction (CWE)~\citep{olmo3techreport} and document-order reconstruction (DocOrder) together contribute most of these tokens: CWE asks how many times each of several sampled words occurs in the document, and DocOrder shuffles a document's sections and asks for their original order, each with 10 randomized question or answer formats so the model cannot rely on one fixed phrasing. Both tasks scale in difficulty with the packed length by varying how densely they quiz each document: the hard bucket draws shorter source documents and packs more items into them (10~questions for CWE, eight sections for DocOrder), so the model must track more items with less surrounding context; the easy bucket draws longer source documents and packs fewer items into them (five questions, four sections), testing the same skill with more context to spare. Mixing both buckets balances signal across the full range of packed lengths. Artificial needles~\citep{artificial-needles} add the rest, planting a retrievable fact among synthetic distractor listings. We add masking in the input sequence to avoid biasing the models toward incorrect or random long context in DocOrder and Artificial needles. The SFT token counts reported for these tasks count the whole packed example, including the injected document that the loss masking excludes; the number of assistant tokens actually optimized per sample is far smaller than these totals suggest.\footnote{Estimated from representative per-task answer lengths (CWE: about six tokens per turn; DocOrder: about 14~tokens for the single order-list answer; Artificial needles: about 25~tokens per turn) and the number of samples per phase: the actually optimized share of SFT tokens is well under 1\%, from about 0.3\% at 32k down to about 0.05\% at 262k.}

\begin{table}[!t]
\caption{Token budget per class across the four long-context extension phases. Percentages are the within-phase share of tokens beyond the Cooldown subsample of \autoref{tab:token-budget}.}
    \label{tab:long_context_class_budget}
\centering
\small
\begin{tabular}{@{}l rrrrr r@{}}
        \toprule
        \textbf{Phase} & \textbf{English} & \textbf{Multilingual} & \textbf{Image} & \textbf{Audio} & \textbf{SFT} & \textbf{Total} \\
        \midrule
        \multicolumn{7}{@{}l}{\textit{\apertusonefivesmall{}}} \\
        32k  & 6.07B (25.6\%) & 4.05B (17.1\%) & 6.66B (28.1\%) & 1.30B (5.5\%) & 5.60B (23.6\%) & 23.69B \\
        65k  & 5.63B (25.5\%) & 3.79B (17.2\%) & 6.25B (28.3\%) & 1.22B (5.5\%) & 5.17B (23.4\%) & 22.07B \\
        131k & 5.69B (25.9\%) & 3.80B (17.3\%) & 6.27B (28.6\%) & 1.23B (5.6\%) & 4.96B (22.6\%) & 21.95B \\
        262k & 5.69B (26.0\%) & 3.81B (17.4\%) & 6.29B (28.7\%) & 1.22B (5.6\%) & 4.86B (22.2\%) & 21.87B \\
        \addlinespace
        \multicolumn{7}{@{}l}{\textit{\apertusonefivelarge{}}} \\
        32k  & 5.81B (27.3\%) & 3.87B (18.2\%) & 4.97B (23.4\%) & 1.25B (5.9\%) & 5.35B (25.2\%) & 21.25B \\
        65k  & 5.77B (27.2\%) & 3.88B (18.3\%) & 5.00B (23.6\%) & 1.25B (5.9\%) & 5.29B (25.0\%) & 21.19B \\
        131k & 5.00B (27.7\%) & 3.35B (18.6\%) & 4.30B (23.8\%) & 1.07B (5.9\%) & 4.34B (24.0\%) & 18.06B \\
        262k & 4.99B (27.8\%) & 3.33B (18.5\%) & 4.31B (24.0\%) & 1.08B (6.0\%) & 4.26B (23.7\%) & 17.98B \\
        \bottomrule
    \end{tabular}
\end{table}

The long-context evaluations of the released models appear in \autoref{sec:long_context_eval}.

\subsection{Pretraining Run Retrospective}\label{sec:pretraining_retrospective}

\begin{figure}[t]
  \centering
  \begin{subfigure}{0.49\textwidth}
    \centering
    \includegraphics[width=\textwidth]{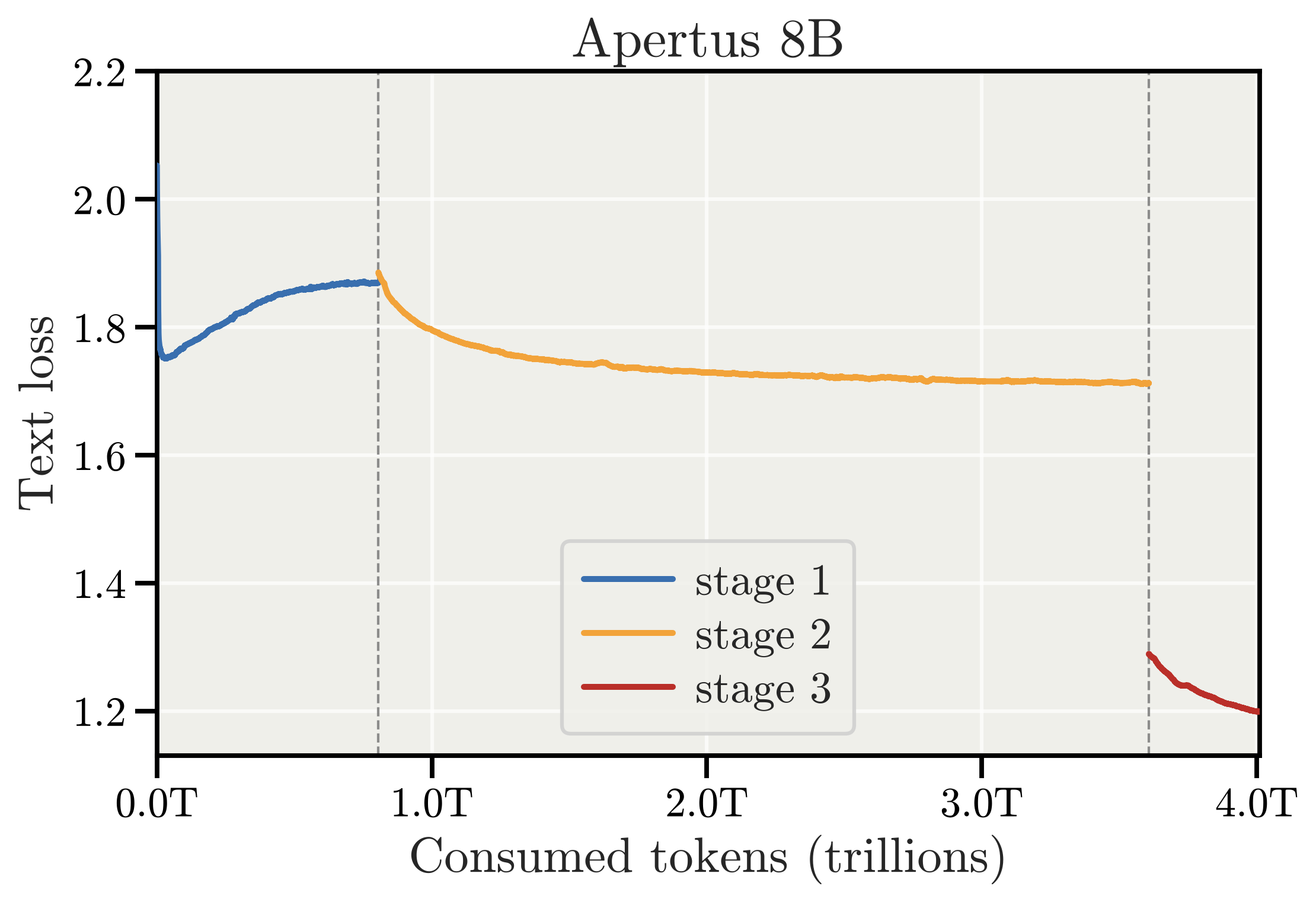}
    \caption{\apertusonefivesmall{}, text loss}
    \label{fig:apertus-8b-textloss}
  \end{subfigure}
  \hfill
  \begin{subfigure}{0.49\textwidth}
    \centering
    \includegraphics[width=\textwidth]{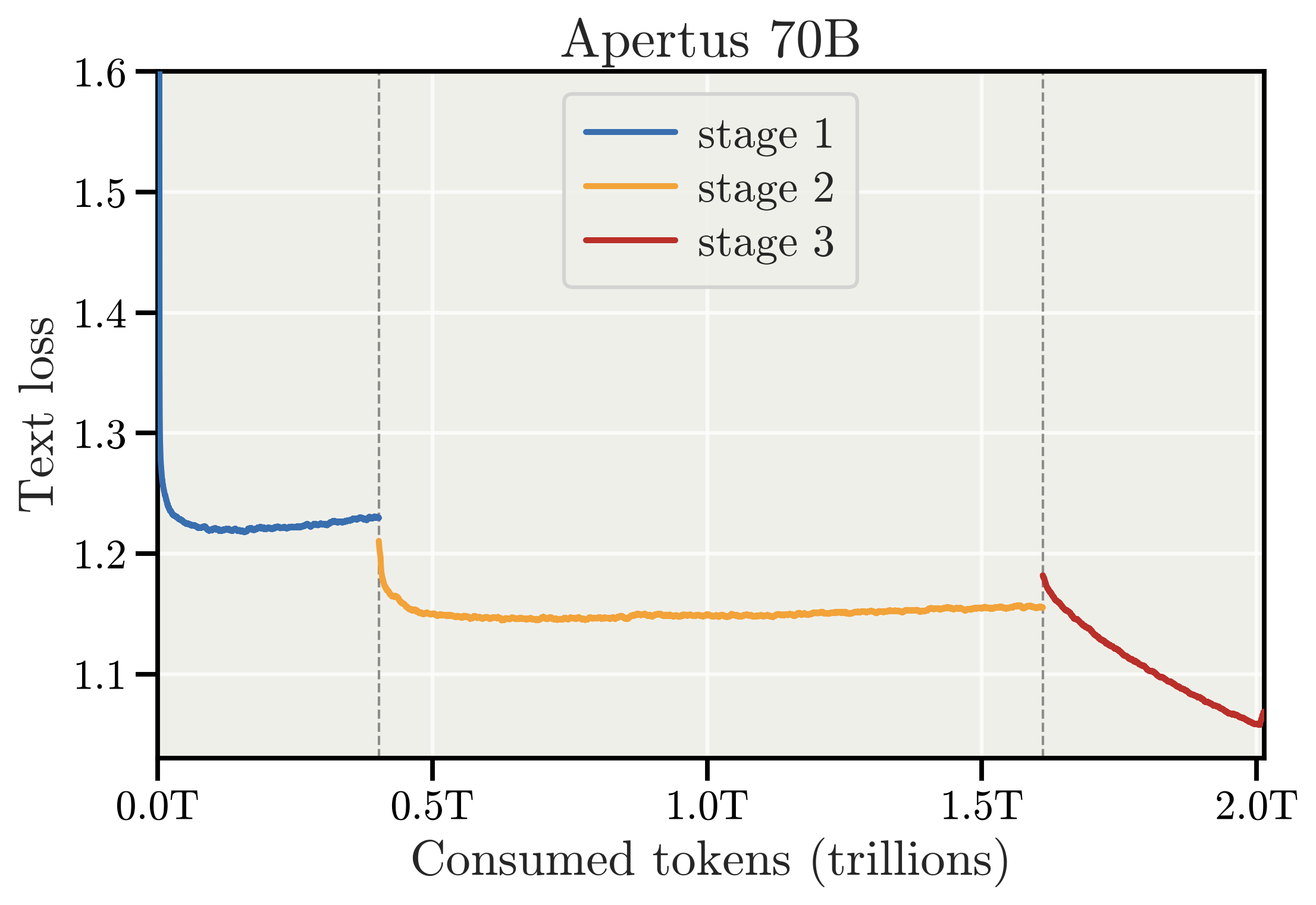}
    \caption{\apertusonefivelarge{}, text loss}
    \label{fig:apertus-70b-textloss}
  \end{subfigure}

  \medskip

  \begin{subfigure}{0.49\textwidth}
    \centering
    \includegraphics[width=\textwidth]{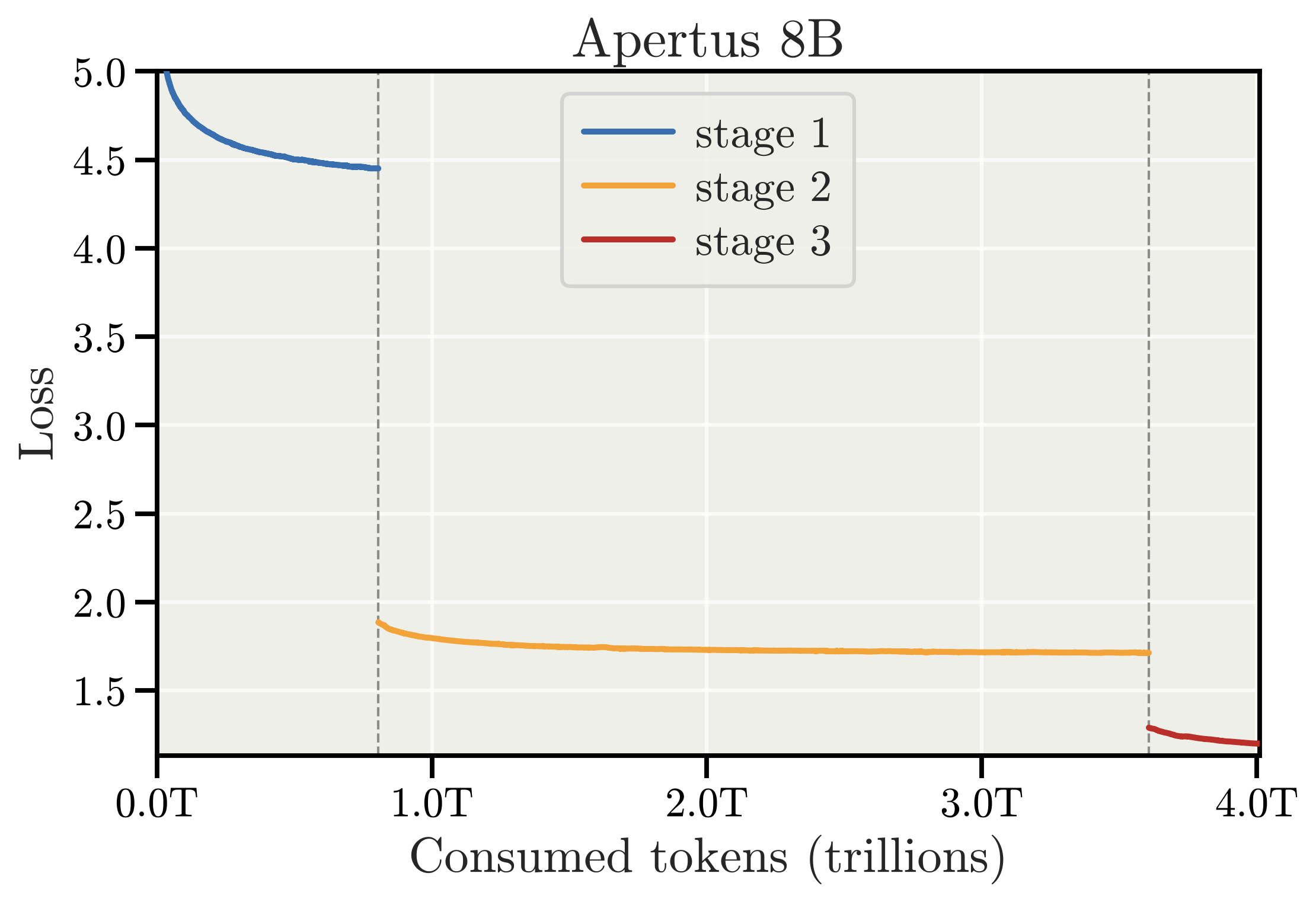}
    \caption{\apertusonefivesmall{}, combined loss}
    \label{fig:apertus-8b-loss}
  \end{subfigure}
  \hfill
  \begin{subfigure}{0.49\textwidth}
    \centering
    \includegraphics[width=\textwidth]{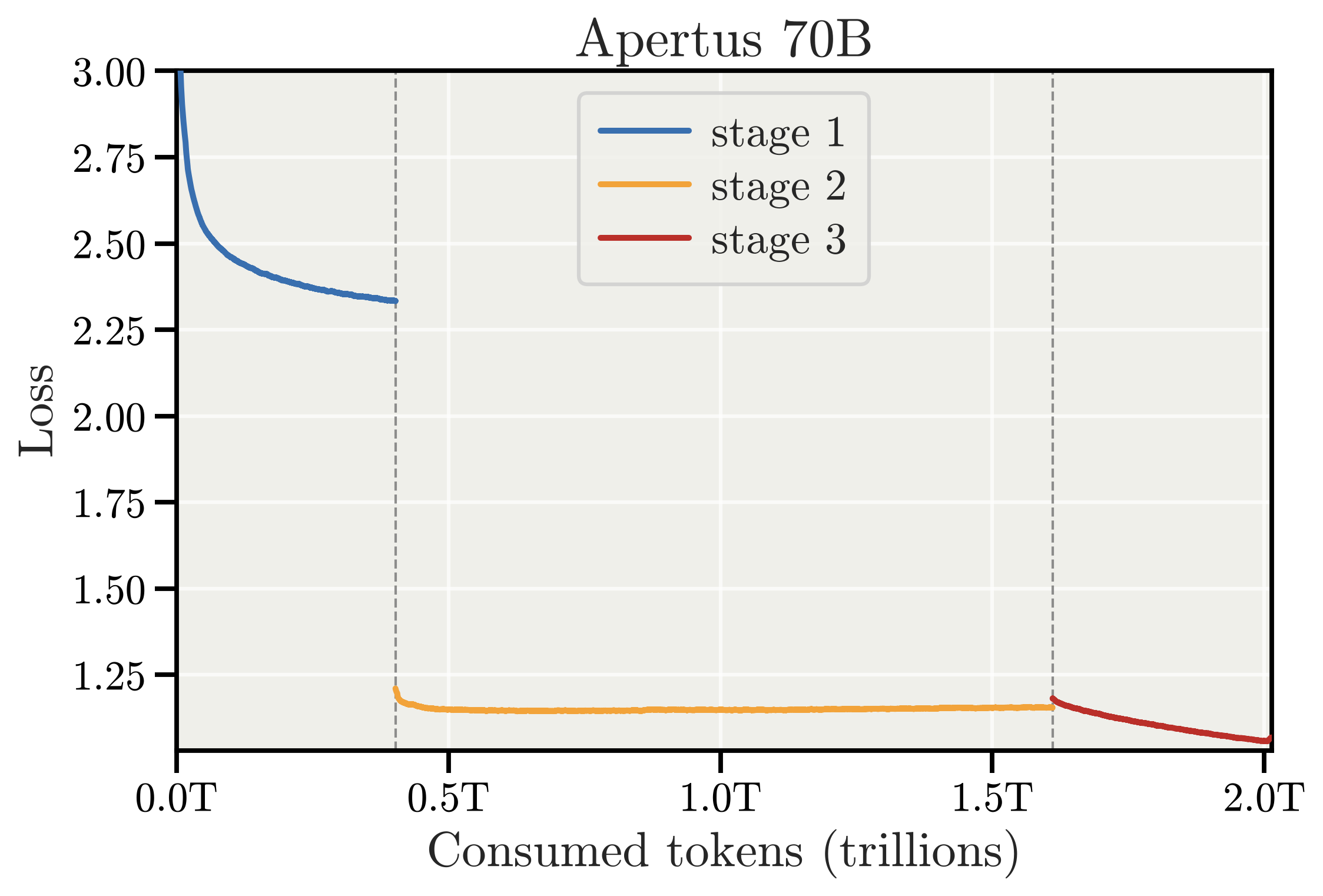}
    \caption{\apertusonefivelarge{}, combined loss}
    \label{fig:apertus-70b-loss}
  \end{subfigure}
  \caption{Per-stage loss against consumed tokens. \textbf{Top} (\subref{fig:apertus-8b-textloss},~\subref{fig:apertus-70b-textloss}): \emph{text loss}, computed over text tokens only in every stage and therefore comparable across stage boundaries. The text loss creeps upward through Stage~1 and then drops sharply over the Stage~3 cooldown. \textbf{Bottom} (\subref{fig:apertus-8b-loss},~\subref{fig:apertus-70b-loss}): \emph{combined loss} over all of each stage's prediction targets (\autoref{tab:cpt_stages}): image, audio, and text tokens in Stage~1, text only from Stage~2 onward. In Stage~1 the combined loss stays stable, so the text-loss creep visible in the top row is masked when text and multimodal targets are averaged together; the drop at the Stage~1$\to$2 boundary reflects the change of targets, not a change in quality. Averages are computed within each stage.}
  \label{fig:apertus-loss}
\end{figure}

We discard the first Stage~1 run for the 8B and restart it. That run had two problems. First, its peak learning rate was set to twice the \apertus{} 1.0 value, a choice that small-scale ablations had favored but that did not transfer to this setting. Second, its text mixture silently omitted the English quality-web, math, and code subsets: the corpus was spread across personal scratch directories with no single manifest, so the launch scripts only saw part of it. As a result, text capability degraded; the text loss eventually recovered, but the evaluations did not. The restart uses the original peak learning rate and the corrected mixture. We keep the text-to-multimodal ratio and accept a drop of about two points on general text evaluations as the cost of adding the new modalities; text evaluations recover by Stage~2. The 70B, by contrast, has enough capacity to absorb the new modalities with visibly less text-capability cost than the 8B (the Stage~1 creep is far milder in \autoref{fig:apertus-70b-textloss} than in \autoref{fig:apertus-8b-textloss}), and we choose its peak learning rate by continuing a pre-cooldown \apertus{} 1.0 checkpoint with preserved optimizer states on unchanged data, letting the loss settle, and comparing peak learning rates from there. This gives a reliable signal within tens of billions of tokens, and it selects twice the original peak. These two observations suggest that the text-capability cost of adding modalities is capacity-dependent.

Ablations narrow down what drives the Stage~1 text-loss creep. Returning to the original peak learning rate reduces the slow upward creep of the text loss during Stage~1 (top row, \autoref{fig:apertus-loss}) but does not remove it; we rule out weight-decay settings as the cause; freezing all non-embedding parameters at the start of training only delays the creep; and the creep appears at mixed text-to-multimodal ratios but not near either extreme, consistent with gradient interference between modalities~\citep{hao2025unixmitigatingmodalityconflict}. The main driver, however, is the schedule itself: \apertus{} 1.0 ends cooled down, and warming the learning rate back to its peak costs text performance until the next cooldown recovers it; both the Stage~1 rise and its reversal over the Stage~3 cooldown are visible in the top row of \autoref{fig:apertus-loss}. The same text-loss increase upon introducing multimodal data is reported for Kimi~K2.5, which for this reason keeps its learning-rate schedule unchanged~\citep{kimiteam2026kimik25}.

\autoref{fig:apertus-loss} shows the loss of both runs across the three stages: the top row on text tokens only, comparable across stages and exposing the Stage~1 creep and Stage~3 recovery discussed above, and the bottom row over each stage's own prediction targets.

The 70B gains little over the 8B on multimodal tasks despite being nine times larger (\autoref{sec:image_eval} and \autoref{sec:audio_eval}). We see two causes. First, the 70B sees less multimodal data: its run stops before consuming the full multimodal mixture, because its nodes are reallocated to the next \apertus{} pretraining campaign. Since multimodal ability improves throughout the mixture, this early stop costs the 70B much of the gain it would otherwise have made. Second, the base learning rate of the AdEMAMix optimizer used to pretrain \apertus{} 70B was likely too low; continued pretraining uses a learning rate proportional to the original, so a low base rate leaves little room for the model to adapt to a new modality.

We also ablated adding a small share of instruction-formatted data during the cooldown. This improved math, code, and instruction-following benchmarks. Replaying the instruction-formatted data five-fold further improved benchmarks but regressed safety behavior due to overfitting on the small refusal subsets, so the small share performed better. Both preference alignment and reinforcement learning improved long-context evaluations even though neither stage trains on long documents, suggesting that gains on short contexts carry over to long ones.

We also encountered various infrastructure and tooling incidents. During the 70B long-context extension, virtual pipelining halves throughput, so we disable it for that stage. Scaling the run to more nodes requires manually resharding optimizer states, as the memory-efficient resharding path fails at this size; we preserve the states rather than reinitializing them mid-run, since reinitializing would break the resumed optimization trajectory. Near the end of the 70B cooldown, some datasets are exhausted and the loss creeps slightly as the effective mixture shifts: because we blend proportionally by size, exhausting smaller datasets moves the mixture away from its target composition. The conversion used for evaluations also silently drops the position-embedding scaling configuration, so converted checkpoints underperform the checkpoints they come from, and part of the apparent regression is an artifact of conversion rather than the model.

Data tooling caused incidents as well. Because the corpus was assembled by a distributed effort across the team, some datasets were not stored on persistent storage and fell under automatic deletion policies on the Alps cluster; parts of the tokenized corpus were deleted mid-project. The data loss was recoverable, because every dataset could be regenerated from its preparation script. A bug in the decontamination script removed an entire dataset from the mixture over a single flagged sample; we caught it before training. Checkpoint inspection also exposes upstream data artifacts that aggregate validation checks miss: some captions in the upstream BLIP3 Grounding 50M dataset retain fragments of the recaptioning prompt and follow a single repeated response template, and the pretrained model occasionally reproduces those instructions or applies the template to unrelated images. This motivates direct inspection of generated samples and explicit prompt-fragment checks in addition to schema and aggregate quality checks.

\clearpage
\section{Post-Training}\label{sec:post_training}

Post-training proceeds in five stages (\autoref{fig:pipeline_overview}): supervised finetuning (SFT), long-context SFT, reinforcement learning (RL), preference alignment with DPO, and charter alignment with SDPO. Every stage trains on conversations in the same format, which we therefore describe first.

\subsection{Chat Template}\label{sec:chat_template}

The conversation format of \apertus{} 1.0 carries over: special tokens delimit system, developer, user, and assistant messages, and the developer message declares the tools available to the model~\citep{apertus2025apertusdemocratizingopencompliant}. Here we describe what \apertus{} 1.5 adds.

Two conventions are deliberate. The template renders an explicit empty system block when a conversation has no system message, so no user-controlled content can occupy the system slot. And the training conversations carry no knowledge-cutoff date, which keeps downstream finetunes free of a stale date. The released inference template differs in one respect: when a conversation has no system message, it inserts a short default system prompt that names the model and its modalities (\autoref{app:system_prompt}). Unlike the \apertus{} 1.0 default, this prompt contains neither a knowledge cutoff nor the current date.

Images and audio enter user messages as the placeholder tokens \texttt{<|image|>} and \texttt{<|audio|>}, which at tokenization time expand into the structure markers and content codes of \autoref{sec:image_tokenization} and \autoref{sec:audio_tokenization}. The literals \texttt{<image>} and \texttt{<audio>} are aliases: the tokenizer maps them to the same identifiers, so existing datasets that use these literals tokenize correctly without conversion.

Tool outputs are enclosed by the tokens \texttt{<|tool\_output\_start|>} and \texttt{<|tool\_output\_end|>}, new in this release, which repurpose two reserved identifiers of the \apertus{} 1.0 vocabulary. During training the delimiters identify tool outputs so that they are excluded from the loss: the model conditions on them without learning to generate them. Explicit tokens are necessary because the raw bracket characters that previously enclosed tool results cannot be masked reliably at the token level; brackets merge into hundreds of vocabulary tokens. The assistant emits tool calls as a JSON array between \texttt{<|tools\_prefix|>} and \texttt{<|tools\_suffix|>}: each call is an object that maps the tool name to its arguments, and parallel calls are elements of the same array. The outputs of parallel calls return between the tool-output delimiters, separated by commas. The tool declaration in the developer message is unchanged from \apertus{} 1.0: available tools render as TypeScript-style type declarations, and a conversation without tools declares them as disabled.

In thinking mode the model reasons between \texttt{<|inner\_prefix|>} and \texttt{<|inner\_suffix|>} before its visible answer. Thinking is off by default and enabled per request through the chat template. The two delimiters are deliberately not marked as special tokens, so they survive detokenization and downstream code can parse the reasoning span out of the decoded text. The literals \texttt{<think>} and \texttt{</think>} alias to the same identifiers. The two surface forms repeatedly confuse downstream tooling during development, with training data, reinforcement-learning parsers, and serving stacks latching onto different renderings; the token identifiers, not the rendered strings, are canonical. The template itself supports tool calls inside the thinking span; that the launch serving integrations expose tool calling and thinking only separately is a serving limitation, not a format constraint. \autoref{fig:chat_template_example} shows a rendered conversation that exercises the full format: the default system prompt, the developer message with the deliberation flag and a tool declaration, and an assistant turn that thinks, calls the tool, and answers after the tool output returns.

\begin{figure}[ht]
\begin{prompt}[title=Rendered conversation with thinking enabled and one tool declared,unbreakable]
\begin{promptverbatim}
<|system_start|>You are Apertus 1.5 Omni, a multimodal assistant
developed by the Swiss AI Initiative. Extended from Apertus 1 via
continued pretraining, you understand images and audio and respond
in text.<|system_end|><|developer_start|>Deliberation: enabled
Tool Capabilities:
// Return the current weather for a city.
type get_weather = (_: {
// City name
city: string
}) => any;<|developer_end|><|user_start|>What is the weather in
Bern?<|user_end|><|assistant_start|><|inner_prefix|>The user asks for the
current weather in Bern, so I call the weather
tool.<|tools_prefix|>[{"get_weather": {"city": "Bern"}}]<|tools_suffix|>
<|tool_output_start|>18 degrees Celsius, partly
cloudy<|tool_output_end|><|inner_suffix|>It is currently 18 degrees
Celsius and partly cloudy in Bern.<|assistant_end|>
\end{promptverbatim}
\end{prompt}
\caption{A conversation rendered by the inference chat template, with thinking enabled and one tool declared: the model reasons and calls the tool inside the thinking span, and answers after the tool output returns. The released models do not yet produce this combination (\autoref{sec:limitations}). Line breaks are added for display.}
\label{fig:chat_template_example}
\end{figure}

\subsection{Supervised Finetuning}\label{sec:sft}

We use SFT to develop assistant behavior over text, image, and speech inputs, with text output, using the shared conversation format of \autoref{sec:chat_template}. The mixture covers instruction following, multilingual interaction, reasoning, code, tool use, safety, and model identity. We train the models to provide reasoning traces when thinking is enabled and direct answers when it is disabled.

To curate the text mixture, we train an auxiliary nonthinking checkpoint of \apertusonefivesmall{} from its long-context pretrained checkpoint (\autoref{sec:long_context}), using text, image, and audio supervision with reasoning traces removed. We use this checkpoint for capability filtering: we evaluate selected verifiable text prompts and determine which receive verified reasoning traces.

For production SFT, we initialize \apertusonefivesmall{} and \apertusonefivelarge{} from their respective long-context pretrained checkpoints and use the text mixture curated with the auxiliary \apertusonefivesmall{} checkpoint. Only the curated data carries over from the auxiliary run; neither production model inherits its weights. After production SFT, we continue training both models with dedicated long-context SFT to address degraded long-context retrieval (\autoref{sec:long_context_sft}).

Both production SFT runs use a constant learning rate after 800~warmup iterations and a sequence length of 16,384~tokens: \apertusonefivesmall{} trains at $6 \times 10^{-5}$ with a global batch size of 256 for roughly 159B~tokens, and \apertusonefivelarge{} at $1 \times 10^{-5}$ with a global batch size of 1,024 for roughly 155B~tokens.\footnote{Launch scripts for the production and long-context SFT runs are available at \ghrepo{swiss-ai/pretrain-code}.} The auxiliary capability-filtering run instead decays a peak learning rate of $8 \times 10^{-5}$ linearly, at a global batch size of 1,024.

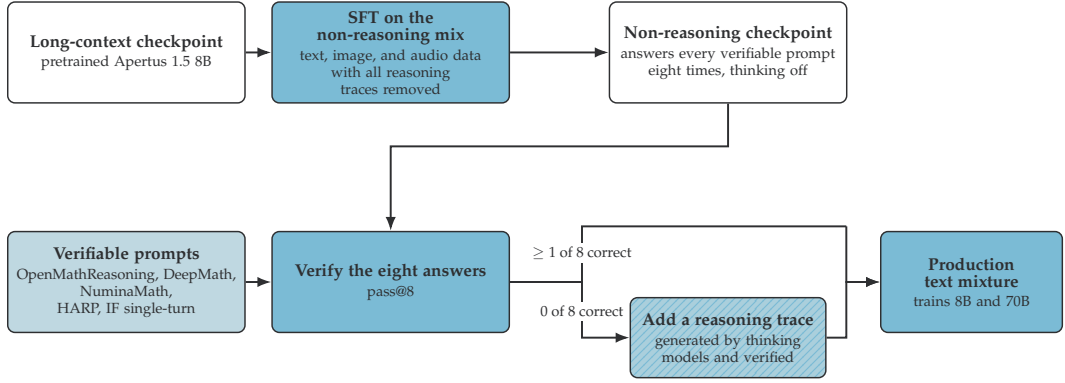
\begin{figure}[t]
\centering
\resizebox{\textwidth}{!}{%
\begin{tikzpicture}[x=1cm, y=1cm,
  every node/.style={text=apertusAnthracite},
  box/.style={draw=apertusAnthracite, line width=0.6pt, rounded corners=3pt,
              align=center, inner sep=3pt, font=\tiny,
              execute at begin node={\hyphenpenalty=10000\relax},
              minimum height=1.45cm, text width=3.2cm},
  stage/.style={box, fill=apertusSkyBlue},
  chip/.style={box, fill=apertusWhite},
  soft/.style={box, fill=apertusSkyBluePale},
  trace/.style={box, minimum height=1.15cm, text width=2.6cm,
                preaction={fill=apertusSkyBlueLight},
                pattern=north east lines, pattern color=apertusSkyBlue},
  flow/.style={-{Latex[length=2mm]}, draw=apertusAnthracite, line width=0.8pt},
  stream/.style={draw=apertusAnthracite, line width=0.8pt},
  edgelabel/.style={font=\tiny, fill=apertusWhite, inner sep=1.5pt},
]

\node[chip]  (lc)  at (1.95,5.5)  {{\scriptsize\bfseries Long-context checkpoint}\\[1pt] pretrained \apertusonefivesmall};
\node[stage] (sft) at (5.75,5.5)  {{\scriptsize\bfseries SFT on the\\ non-reasoning mix}\\[1pt] text, image, and audio data\\ with all reasoning traces removed};
\node[chip]  (nr)  at (10.6,5.5)  {{\scriptsize\bfseries Non-reasoning checkpoint}\\[1pt] answers every verifiable prompt\\ eight times, thinking off};
\draw[flow] (lc) -- (sft);
\draw[flow] (sft) -- (nr);

\node[soft]  (pool) at (1.95,2.2) {{\scriptsize\bfseries Verifiable prompts}\\[1pt] OpenMathReasoning, DeepMath,\\ NuminaMath, HARP, IF single-turn};
\node[stage] (ver)  at (5.75,2.2) {{\scriptsize\bfseries Verify the eight answers}\\[1pt] pass@8};
\node[trace] (gen)  at (10.6,1.4) {{\scriptsize\bfseries Add a reasoning trace}\\[1pt] generated by thinking\\ models and verified};
\node[stage, text width=2.4cm] (mix) at (14.1,2.2) {{\scriptsize\bfseries Production text mixture}\\[1pt] trains 8B and 70B};
\draw[flow] (pool) -- (ver);
\draw[flow] (nr.south) -- (10.6,4.15) -- (5.75,4.15) -- (ver.north);

\coordinate (split) at (8.5,2.2);
\coordinate (join)  at (12.3,2.2);
\draw[stream] (ver.east) -- (split) -- (8.5,3.0) -- (12.3,3.0);
\draw[flow]   (ver.east) -- (split) -- (8.5,1.4) -- (gen.west);
\node[edgelabel] at (8.5,2.62) {$\geq 1$ of $8$ correct};
\node[edgelabel] at (8.5,1.78) {$0$ of $8$ correct};
\draw[flow] (12.3,3.0) -- (join) -- (mix.west);
\draw[flow] (gen.east) -- (12.3,1.4) -- (join) -- (mix.west);
\end{tikzpicture}}
\caption{Capability filtering. We train an auxiliary nonthinking \apertusonefivesmall{} checkpoint to evaluate verifiable text prompts. Prompts solved at least once in eight attempts enter the production text mixture without reasoning traces; those that fail all eight receive verified traces. Both production models use the resulting mixture.}
\label{fig:capability_filtering}
\end{figure}

\paragraph{Capability Filtering.}\label{sec:sft_capability_filtering}
As outlined in the introduction to \autoref{sec:sft}, we train an auxiliary nonthinking \apertusonefivesmall{} checkpoint from its long-context pretrained checkpoint (\autoref{sec:long_context}), using text, image, and audio supervision with reasoning traces removed. We use this checkpoint to sample eight nonthinking answers for each selected verifiable prompt and check the final answers, extracted with the verifiable answer tool, against the reference. Prompts solved at least once (pass@8) remain in the mixture without a reasoning trace; prompts that fail all eight attempts receive a verified reasoning trace. \autoref{fig:capability_filtering} illustrates this procedure. Our ablations motivate this selective use of reasoning traces: with thinking on, trace supervision adds seven to 18~points on mathematics but costs 12 to 14 on instruction following, mostly because about a quarter of traces degenerate into repetition (\autoref{sec:sft_ablations}).

We apply capability filtering to the single-turn subset of our instruction-following (IF) mixture, OpenMathReasoning, DeepMath-103K, NuminaMath, and HARP. The share of prompts that remain without reasoning traces varies substantially by source: 99\% for instruction following and roughly 10--40\% for mathematics, with 11.9\% for the 400k-prompt OpenMathReasoning subset, 38.3\% for DeepMath-103K, 22.7\% for NuminaMath, and 15.1\% for HARP. We ablate $k = 1, \dots, 8$; for $k$ below eight, the filtering rate on the mathematics datasets is negligible, so we select pass@8 for the final run.

\paragraph{Data.} The text component of the final production SFT mixture contains 3,707,851~samples (\autoref{tab:sft_text_mixture}). Relative to \apertus{} 1.0, it shifts weight toward instruction following, mathematics, and code, adds newer multilingual chat data, carries refusal-style safety data whose responses we regenerate and verify as described below, and carries roughly 250,000~samples that reference the \apertus{} identity, mostly through their system prompts, though no dedicated identity question-answer pairs. The reasoning traces come partly from openly released datasets whose traces are distilled from open reasoning models, and partly from traces we generate ourselves on Alps. For mathematics and instruction following, we generate verified reasoning traces with thinking enabled, using multiple models: Kimi-K2.6, Qwen3.6-27B, Gemma-4-31B, and MiniMax-M2.7. We deduplicate text prompts, and for the capability-filtering phase we filter out prompts that could lead to verifiable answers, such as math prompts, using an LLM-as-a-judge (Kimi-K2.6). The text mixture remains thin on long multi-turn conversations, with only roughly 30,000~samples of three or more turns.

The training mixture for the auxiliary checkpoint that we subsequently use for capability filtering of text data includes an image pool of 20~datasets with roughly 58B~tokens and an audio pool of six datasets with roughly 1.1B~tokens (\autoref{tab:sft_image_pool}, \autoref{tab:audio_data_sft}). For this checkpoint's multimodal supervision, we follow a zero-reasoning policy: we regenerate samples whose sources contain reasoning traces with Qwen3.6-27B and thinking disabled. The final production SFT mixture uses the expanded image pool in \autoref{tab:vision_data_sft}, including reasoning-bearing additions described in \autoref{sec:sft_image}, and the same prepared audio pool (\autoref{tab:audio_data_sft}).

\begin{table}[t]
\setlength{\tabcolsep}{4pt}
\caption{Text component of the final production SFT mixture by source and category.
All datasets are decontaminated against evaluation benchmarks. Numbers indicate
example count after filtering. Data Ratio is each category's share of
the text SFT pool by example count.}
\label{tab:sft_text_mixture}
\centering
\small
\begin{tabular}{@{}llrr@{}}
\toprule
\textbf{Category} & \textbf{Dataset Source} & \textbf{Samples} & \textbf{Data Ratio} \\
\midrule
\multirow{6}{*}{Foundation}
 & OLMo2 OASST1 (Tulu-3)~\citep{kopf2023OpenAssistant} & 7,028 & \multirow{6}{*}{6.14\%} \\
 & Llama-Nemotron Chat~\citep{bercovich2025llamanemotron} & 20,585 & \\
 & Llama-Nemotron Safety~\citep{bercovich2025llamanemotron} & 13,765 & \\
 & Apertus FAQ~\citep{apertus2025apertusdemocratizingopencompliant} & 1,580 & \\
 & IF-augmented (single- \& multi-turn) & 184,580 & \\
 & Subtotal & 227,538 & \\
\addlinespace
\multirow{9}{*}{Math \& Reasoning}
 & OpenMathReasoning~\citep{moshkov2025aimo2} & 666,344 & \multirow{9}{*}{25.64\%} \\
 & DeepMath-103K~\citep{he2025deepmath} & 101,950 & \\
 & NuminaMath 1.5~\citep{numina_math_datasets} & 83,818 & \\
 & SciRIFF~\citep{wadden2025sciriff} & 66,322 & \\
 & Table-GPT~\citep{li2024tablegpt} & 13,222 & \\
 & FeTaQA~\citep{nan2022fetaqa} & 10,330 & \\
 & RiddleSense~\citep{lin2021riddlesense} & 4,531 & \\
 & HARP~\citep{yue2024harp} & 4,128 & \\
 & Subtotal & 950,645 & \\
\addlinespace
\multirow{8}{*}{Code \& Functions}
 & OpenCodeReasoning v2~\citep{ahmad2025opencodereasoning} & 512,428 & \multirow{8}{*}{19.79\%} \\
 & Glaive Function Calling v2~\citep{glaive2023functioncalling} & 52,727 & \\
 & Jupyter Agent~\citep{jupyteragent2025dataset} & 29,529 & \\
 & GSM8K (TIR)~\citep{cobbe2021gsm8k} & 7,473 & \\
 & Toucan 1.5M & 117,586 & \\
 & EnvScaler SFT Trajectories & 9,022 & \\
 & OpenSeeker v1 & 4,949 & \\
 & Subtotal & 733,714 & \\
\addlinespace
\multirow{4}{*}{Multilingual}
 & SmolTalk2 Multilingual (8 lang)~\citep{bakouch2025smollm3} & 653,834 & \multirow{4}{*}{25.83\%} \\
 & EuroBlocks Multilingual (2512)~\citep{ramos2026eurollm22b} & 286,381 & \\
 & EuroBlocks Aya (Synthetic 1124)~\citep{martins2025eurollm9b} & 17,467 & \\
 & Subtotal & 957,682 & \\
\addlinespace
\multirow{7}{*}{Regional}
 & WikiQA~\citep{yang2015wikiqa} & 482,799 & \multirow{7}{*}{15.35\%} \\
 & Romansh Varieties~\citep{apertus2025apertusdemocratizingopencompliant} & 46,078 & \\
 & German$\rightarrow$Romansh MT & 30,000 & \\
 & African Languages~\citep{apertus2025apertusdemocratizingopencompliant} & 7,339 & \\
 & Swiss German Dialects~\citep{apertus2025apertusdemocratizingopencompliant} & 2,674 & \\
 & Swiss Charter Q\&A~\citep{apertus2025apertusdemocratizingopencompliant} & 88 & \\
 & Subtotal & 568,978 & \\
\addlinespace
\multirow{4}{*}{Domain-Specific}
 & MedMCQA~\citep{pal2022medmcqa} & 156,355 & \multirow{4}{*}{7.26\%} \\
 & MIRIAD~\citep{zheng2025miriad} & 99,556 & \\
 & Medical-o1~\citep{chen2024huatuogpto1} & 13,383 & \\
 & Subtotal & 269,294 & \\
\addlinespace
Total & & 3,707,851 & 100.00\% \\
\bottomrule
\end{tabular}%
\end{table}

\paragraph{Tool-Use Data.} We select tool-use datasets by whether their trajectories are execution-verified, rather than filtered only by an LLM judge, and drop candidates that do not meet this bar. The mixture combines a curated subset of Toucan~\citep{xu2025toucan} (119,287~samples), EnvScaler~\citep{song2026envscaler} (9,022), OpenSeeker~\citep{du2026openseeker} (4,949), an execution-verified subset of the Jupyter Agent dataset~\citep{jupyteragent2025dataset} (roughly 30,000 sandbox-verified notebooks), Glaive function calling~\citep{glaive2023functioncalling} (roughly 52,000), and GSM8K~\citep{cobbe2021gsm8k} trajectories with calculator tool use (roughly 7,400). We validate and clean the tool-use datasets to ensure correct conversation formatting, declared tools, valid call arguments, and matching nonempty tool outputs. Because a model that sees tools only in tool-calling samples never learns to ignore them, we additionally inject tool declarations into 131,557~conversations that do not call any tool, teaching the model to answer directly even when tools are available. A cleanup pass strips 950,544 calculator-annotation blocks that appear in samples without a corresponding tool declaration.

\paragraph{Multilingual and Safety Data.} We use updated versions of EuroBlocks-SFT-2512~\citep{ramos2026eurollm22b} and Llama-Nemotron Safety~\citep{bercovich2025llamanemotron} to improve our safety and multilingual capabilities. For Llama-Nemotron Safety, we observe over-refusals of benign prompts and, on some occasions, harmful responses to dangerous prompts; therefore, we regenerate responses with GPT-OSS-120B-Safeguard and verify their validity with a Qwen3-235B judge to improve assistant behavior and mitigate over-refusal. In the EuroBlocks-SFT-2512 dataset, we observe that some of the assistant responses contain traces of the original EuroLLM identity training with the model name present in the provided outputs. Since we cannot remove something like that verbatim, we use a pool of several models (GPT-OSS-120B, MiniMax-M2, Qwen3.5-397B) randomly sampled to regenerate responses with the \apertus{} identity provided in the system prompt. Of the 1M original samples, we use only 286k non-English conversations to diversify our multilingual training mix.

\paragraph{Decontamination.} We decontaminate the SFT mixture against a list of 595 evaluation benchmarks with n-gram matching, and manually verify the flagged samples of every affected dataset rather than dropping hits blindly, since thresholds tuned for English text over-trigger on multimodal boilerplate. The true positives concentrate in multilingual datasets that contain translated benchmark items: in PangeaInstruct, roughly 71\% of the 9,114 flagged samples are genuine benchmark translations, estimated from a manual inspection of 100 random hits. For EuroBlocks 2512, almost all of the removed samples are unrelated non-contaminated multilingual samples. For our instruction-following mix, the majority of ``flagged'' samples contain an instruction format similar to some from the IFEval benchmark, which does not necessarily mean direct contamination. We still decide to keep the original similarity threshold of 0.5 (as used in \apertus{} 1.0) to ensure we catch contaminated samples at the cost of false positives, which is still not critical. We remove contaminated samples by identifier. For image data, text-based matching does not transfer at all: templated prompts dominate, so nearly every n-gram hit is a false positive, and matching must happen at the image level instead. We therefore match the perceptual hashes of the training images (\autoref{sec:multimodal_data_processing}) against roughly 950,000 benchmark images; this catches near-duplicates of natural images but produces false positives in low-entropy domains such as satellite and medical imagery, which require manual thresholding. We provide full details with counts of dropped samples for every dataset from the text SFT mixture in \autoref{app:sft_text_decontamination} (\autoref{tab:sft_text_decontamination}).

\subsubsection{Image}\label{sec:sft_image}

The image pool used to train the auxiliary checkpoint for capability filtering of text data spans 20~datasets with 17,534,672~sequences and roughly 58.1B~tokens; \autoref{tab:sft_image_pool} lists them. We assemble the pool under the zero-reasoning policy of that checkpoint's training: we exclude or strip sources whose assistant turns carry reasoning markup, and for template-heavy sources we regenerate the assistant answers with Qwen3.6-27B and thinking disabled. The permissively licensed subset of LLaVA-OneVision-1.5~\citep{llava-onevision-1-5} and SenseNova-SI-8M~\citep{sensenova-si} together hold roughly 80\% of the tokens in \autoref{tab:sft_image_pool}; the remainder covers document, chart, and table understanding; captioning and open question answering; GUI pointing; and domain imagery from remote sensing, pathology, and memes. The production phase additionally trains on reasoning-bearing image data: Innovator-VL~\citep{wen2026innovatorvl} ships with the boundaries of its thinking spans stripped by its creators, so we regenerate the reasoning traces by reformatting the samples with Qwen3.6-27B and keep only outputs whose answers match the original with a ROUGE-L score above 0.9; in the pool of \autoref{tab:vision_data_sft}, roughly 2.7 million of the dataset's 24.8 million samples carry restored reasoning. We remove structured answer tags from source datasets, so that all supervision matches the \apertus{} chat format; the processing of the image SFT datasets otherwise follows the multimodal data pipeline of \autoref{sec:multimodal_data}. The share of text data packed into image-SFT batches matters little in ablations, with benchmark movements small and split between 2.5\% and 15\%. For three FineVision subsets we regenerate the assistant turns with Qwen3.6-27B: for LNQA we keep the questions and replace the template-like single-fact answers with image-grounded answers of two to four sentences, and for emo-visual-data and Memotion 2 we replace the original Chinese descriptions and sentiment labels with two-turn English explanations of each meme, dropping samples that the generator flags as unsafe. In the \apertusonefivesmall{} production run, text makes up 13.6\% of the sequences and 7.8\% of the tokens. Besides Innovator-VL, the production image pool carries thinking spans in Nemotron-Image-Training-v3 (22\% of sequences), SenseNova-SI-8M (5\%), and RSRCC (all sequences).

The production candidate pool contains 57,130,672~sequences and 157.8B~input tokens. After we create the training and development splits and complete final packing, the pool of the \apertusonefivesmall{} run contains 43,915,058~sequences and 143.4B~input tokens, and \autoref{tab:vision_data_sft} gives its per-dataset composition.

\begin{table}[!htbp]
\caption{Image pool used to train the auxiliary checkpoint for capability filtering of text data: 20~datasets, 17.5M~sequences, 58.1B~tokens. The production SFT pool appears in \autoref{tab:vision_data_sft}.}
\label{tab:sft_image_pool}
\centering
\small
\setlength{\tabcolsep}{4pt}
\begin{tabularx}{\textwidth}{@{}Yrrl@{}}
\toprule
\textbf{Dataset} & \textbf{Sequences} & \textbf{Tokens (M)} & \textbf{Processing} \\
\midrule
\texttt{llava\_ov\_permissive}\newline\citepsmall{llava-onevision-1-5} & 10,019,044 & 24,859.2 & Permissive subset \\
\texttt{sensenova\_si\_8m}\newline\citepsmall{sensenova-si} & 5,205,020 & 21,859.4 & Reasoning stripped \\
\texttt{google\_maptrace}\newline\citepsmall{panagopoulou2025maptrace} & 1,055,208 & 7,585.8 & As released \\
\texttt{lnqa\_recap}\newline\citepsmall{pont-tuset2020localized,wiedmann2025finevision} & 302,777 & 1,034.5 & Answers regenerated \\
\texttt{pixmo\_cap\_qa}\newline\citepsmall{molmo-pixmo} & 163,359 & 768.3 & As released \\
\texttt{CoSyn\_400k}\newline\citepsmall{cosyn-400k} & 87,057 & 359.1 & As released \\
\texttt{bigearthnet}\newline\citepsmall{herzog2026bigearthnettxt} & 229,114 & 281.9 & As released \\
\texttt{google\_rsrcc}\newline\citepsmall{kazoom2026rsrcc} & 103,867 & 233.4 & Reasoning stripped \\
\texttt{pixmo\_ask\_model\_anything}\newline\citepsmall{molmo-pixmo} & 58,830 & 203.4 & As released \\
\texttt{mmevol\_judged}\newline\citepsmall{mmevol} & 90,216 & 197.7 & Judge filtered \\
\texttt{tcm\_shizhen\_vision}\newline\citepsmall{chen2025shizhengpt} & 63,806 & 185.6 & As released \\
\texttt{molmo2\_syn\_multiimage}\newline\citepsmall{molmo2} & 30,286 & 160.3 & As released \\
\texttt{molmopoint\_guisyn}\newline\citepsmall{molmopoint} & 20,939 & 97.5 & As released \\
\texttt{vdr\_cooking\_recipes}\newline\citepsmall{racineai2025vdrCookingRecipes} & 11,649 & 80.0 & As released \\
\texttt{chinesememe\_recap\_en}\newline\citepsmall{llmredteam2024emovisualdata,wiedmann2025finevision} & 42,678 & 73.2 & Recaptioned \\
\texttt{molmo2\_multiimage\_qa}\newline\citepsmall{molmo2} & 13,141 & 63.4 & As released \\
\texttt{path\_vqa}\newline\citepsmall{path-vqa} & 19,654 & 32.3 & As released \\
\texttt{multihiertt}\newline\citepsmall{multihiertt} & 5,358 & 22.9 & As released \\
\texttt{memotion\_recap\_en}\newline\citepsmall{ramamoorthy2022memotion2,wiedmann2025finevision} & 5,857 & 10.0 & Recaptioned \\
\texttt{spatialsense\_gold}\newline\citepsmall{spatialsense} & 6,812 & 6.4 & Answers regenerated \\
\addlinespace
Total & 17,534,672 & 58,114.2 & \\
\bottomrule
\end{tabularx}
\end{table}

\begin{figure}[t]
    \centering
    \includegraphics[width=\linewidth]{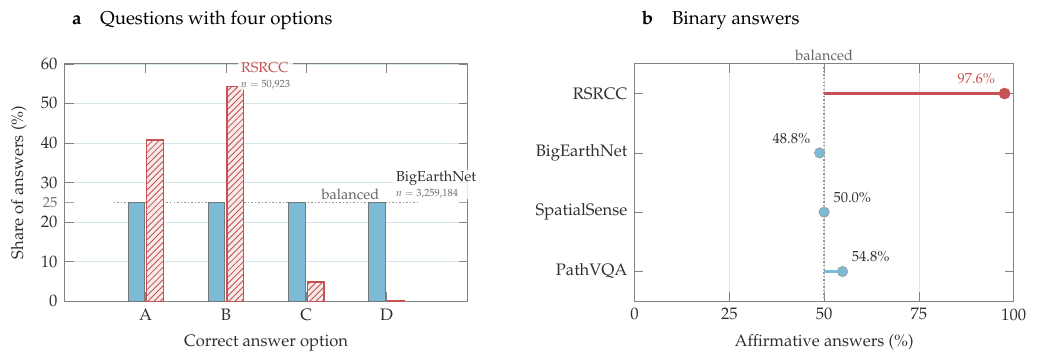}
    \caption{Representative categorical answer distributions in image SFT sources with deterministic extraction. \textbf{Left:} RSRCC strongly favors the first two answer options among questions with four options, whereas BigEarthNet is nearly uniform. \textbf{Right:} RSRCC also strongly favors affirmative binary answers, unlike BigEarthNet, SpatialSense, and PathVQA. These source distributions can create answer priors but do not by themselves establish a causal effect on evaluation accuracy.}
    \label{fig:sft_categorical_balance}
\end{figure}

\paragraph{Categorical Answer Balance.} We inspect categorical answers in image SFT sources whose schemas permit deterministic extraction. Their distributions vary substantially by source. In RSRCC, 97.6\% of 52,944 binary answers are ``yes.'' Among 50,923~questions with four options, 40.8\% use answer A, 54.3\% use B, 4.8\% use C, and 0.1\% use D. By contrast, the correct answers in BigEarthNet are almost uniformly distributed across the four options, and its binary answers are close to balanced; SpatialSense is exactly balanced between ``yes'' and ``no'' (\autoref{fig:sft_categorical_balance}). The RSRCC distribution can induce answer priors and can inflate evaluation scores on data drawn from the same source.

\subsubsection{Audio}\label{sec:sft_speech}

The audio pool used to train the auxiliary checkpoint for capability filtering of text data is speech only, with no music or environmental-sound supervision: six datasets with 1,116,511~sequences and roughly 1.1B~tokens, all speech in and text out. We synthesize and filter these datasets as described in \autoref{sec:audio_training_data}. The production SFT runs draw from the same six-dataset pool listed in \autoref{tab:audio_data_sft}. These counts describe prepared data available to the runs, rather than sampling weights, repetition factors, or tokens actually consumed. The later long-context SFT stage draws from the extended pool in \autoref{tab:audio_data_lc_sft}; \autoref{tab:long_context_sft_mix} gives its per-model training counts.

\subsubsection{Variety-Aware Translation Case Study}\label{sec:variety_aware_translation}

Generating text in the correct variety of a multi-variety language can be a challenge for language models, especially when some varieties are underrepresented in the training data. \citet{apertus2025apertusdemocratizingopencompliant} studied this problem for Romansh, a Swiss national language with six distinct varieties. While \apertus{} 1.0 outperformed comparable models at translation from German into Romansh, it still struggled to adhere to the distinct varieties of the language. We perform a case study to investigate how a small amount of variety-aware post-training data affects this behavior.

We follow a back-translation approach~\citep{sennrich-etal-2016-improving}. We translate a sample of Romansh text into German with machine translation, and during supervised finetuning we train \apertus{} to translate the German text back into the specific Romansh variety. The monolingual Romansh text comes from the \hfdata{swiss-ai/apertus-pretrain-romansh} collection released with \apertus{} 1.0~\citep{apertus2025apertusdemocratizingopencompliant}. We identify the variety of each text with the language identification system of \citet{model-etal-2026-robust}, and translate into German with the machine translation system of \citet{vamvas-et-al-2026-translation}. We include 5,000~samples per variety in the finetuning mixture, and release the SFT dataset at \hfdata{swiss-ai/apertus-pretrain-romansh-backtranslated-sft}.

\autoref{tab:translation_autoeval_chrf} reports the effect on translation from German into the six Romansh varieties, and the human evaluation in \autoref{fig:translation_humeval} additionally covers German into Sursilvan.

\subsection{Long-Context Supervised Finetuning}\label{sec:long_context_sft}

\begin{table}[t]
\caption{Long-context SFT data mixture. The mix each model's run trains on, with tokens and samples per source; the runs draw subsets of the pools, so the totals stay below the pool sizes. Length range is the observed range from minimum to maximum sequence length within the class. The short-context text and audio rows are length-filtered slices of the production pools of \autoref{sec:sft}; all three vision rows (short- and long-context) are length buckets of the same pool in \autoref{tab:vision_data_lc_sft}. Those tables give the per-dataset composition.}
\label{tab:long_context_sft_mix}
\centering
\small
\setlength{\tabcolsep}{4pt}
\begin{tabular}{@{}lrrrr@{}}
\toprule
& \multicolumn{2}{c}{\textbf{8B}} & \multicolumn{2}{c@{}}{\textbf{70B}} \\
\cmidrule(lr){2-3} \cmidrule(l){4-5}
\textbf{Dataset} & \textbf{Tokens} & \textbf{Samples} & \textbf{Tokens} & \textbf{Samples} \\
\midrule
\multicolumn{5}{@{}l}{\textit{Text, 26--108K}} \\
SFT mixture subset (\autoref{tab:sft_text_mixture}) & 1{,}870{,}000{,}000 & 1{,}119{,}142 & 2{,}410{,}000{,}000 & 1{,}438{,}897 \\
\addlinespace
\multicolumn{5}{@{}l}{\textit{Tool (short-context), 119--16.4K}} \\
Toucan~\citep{xu2025toucan} & 33{,}700{,}000 & 15{,}162 & 43{,}400{,}000 & 19{,}494 \\
EnvScaler~\citep{song2026envscaler} & 9{,}300{,}000 & 763 & 11{,}900{,}000 & 981 \\
OpenSeeker~\citep{du2026openseeker} & 946{,}200 & 88 & 1{,}200{,}000 & 113 \\
\addlinespace
\multicolumn{5}{@{}l}{\textit{Vision (short-context), 108--16.4K}} \\
Long-context vision pool (\autoref{tab:vision_data_lc_sft}) & 21{,}550{,}000{,}000 & 7{,}415{,}947 & 27{,}710{,}000{,}000 & 9{,}534{,}789 \\
\addlinespace
\multicolumn{5}{@{}l}{\textit{Audio (short-context), 105--8.19K}} \\
\makecell[l]{Production speech pool (\autoref{tab:audio_data_sft})\\and AudioMCQ GeminiCoT} & 725{,}600{,}000 & 730{,}285 & 933{,}000{,}000 & 938{,}938 \\
\addlinespace
\multicolumn{5}{@{}l}{\textit{Text (long-context), 64.7K--262K}} \\
CWE~\citep{olmo3techreport} & 184{,}200{,}000 & 1{,}493 & 236{,}800{,}000 & 1{,}920 \\
DocOrder & 92{,}900{,}000 & 791 & 119{,}400{,}000 & 1{,}016 \\
ArtificialNeedles~\citep{artificial-needles} & 66{,}300{,}000 & 395 & 85{,}300{,}000 & 508 \\
\addlinespace
\multicolumn{5}{@{}l}{\textit{Tool (long-context), 131K--260K}} \\
OpenSeeker~\citep{du2026openseeker} & 19{,}300{,}000 & 122 & 24{,}800{,}000 & 157 \\
\addlinespace
\multicolumn{5}{@{}l}{\textit{Vision (long-context, lower), 16.4K--131K}} \\
\href{https://huggingface.co/datasets/sensenova/SenseNova-SI-8M}{SenseNova-SI-8M} & 1{,}940{,}000{,}000 & 24{,}234 & 2{,}500{,}000{,}000 & 31{,}158 \\
\href{https://huggingface.co/datasets/nvidia/Nemotron-Image-Training-v3}{Nemotron Image Training v3} & 1{,}080{,}000{,}000 & 15{,}246 & 1{,}390{,}000{,}000 & 19{,}602 \\
\href{https://huggingface.co/datasets/allenai/Molmo2-SynMultiImageQA}{Molmo2 SynMultiImageQA} & 147{,}300{,}000 & 3{,}363 & 189{,}400{,}000 & 4{,}324 \\
\href{https://huggingface.co/datasets/InnovatorLab/Innovator-VL-Instruct-46M}{Innovator-VL Instruct} & 44{,}500{,}000 & 1{,}508 & 57{,}200{,}000 & 1{,}940 \\
\href{https://huggingface.co/datasets/google/MapTrace}{MapTrace} & 6{,}000{,}000 & 327 & 7{,}700{,}000 & 420 \\
\href{https://huggingface.co/datasets/mvp-lab/LLaVA-OneVision-1.5-Instruct-Data}{LLaVA-OneVision 1.5} & 3{,}900{,}000 & 194 & 5{,}000{,}000 & 250 \\
\href{https://huggingface.co/datasets/allenai/Molmo2-MultiImageQA}{Molmo2 MultiImageQA} & 1{,}000{,}000 & 51 & 1{,}300{,}000 & 65 \\
\href{https://huggingface.co/datasets/allenai/MolmoPoint-GUISyn}{MolmoPoint GUISyn} & 33{,}900 & 2 & 43{,}600 & 2 \\
\addlinespace
\multicolumn{5}{@{}l}{\textit{Vision (long-context, upper), 131K--262K}} \\
\href{https://huggingface.co/datasets/nvidia/Nemotron-Image-Training-v3}{Nemotron Image Training v3} & 2{,}560{,}000{,}000 & 14{,}146 & 3{,}300{,}000{,}000 & 18{,}188 \\
\href{https://huggingface.co/datasets/sensenova/SenseNova-SI-8M}{SenseNova-SI-8M} & 1{,}300{,}000{,}000 & 8{,}813 & 1{,}670{,}000{,}000 & 11{,}330 \\
\href{https://huggingface.co/datasets/allenai/Molmo2-SynMultiImageQA}{Molmo2 SynMultiImageQA} & 18{,}800{,}000 & 133 & 24{,}200{,}000 & 172 \\
\addlinespace
\multicolumn{5}{@{}l}{\textit{Audio (long-context), 8.15K--52.7K}} \\
marco\_longspeech\_sft~\citep{yang2026longspeech} & 3{,}560{,}000{,}000 & 132{,}759 & 4{,}570{,}000{,}000 & 170{,}690 \\
teleantifraud\_matching\_sft~\citep{ma2025TeleAntiFraud} & 5{,}800{,}000 & 594 & 7{,}500{,}000 & 764 \\
vocalized\_sft (\autoref{tab:audio_data_sft}) & 1{,}100{,}000 & 110 & 1{,}400{,}000 & 141 \\
\addlinespace
Total & 35{,}230{,}000{,}000 & 9{,}485{,}668 & 45{,}300{,}000{,}000 & 12{,}195{,}859 \\
\bottomrule
\end{tabular}
\end{table}

Preserving long-context retrieval through post-training is a known challenge: short-context SFT systematically biases attention toward local patterns, degrading the long-range retrieval established during context extension~\citep{zhou2026attentionamnesiahybridllms}. After performing SFT at 16,384~tokens we confirm this behavior: the RULER~\citep{hsieh2024rulerwhatsrealcontext} score at 262,144~tokens drops from 54.8 to 8.3 in \apertusonefivesmall{}, while short-context scores remain unaffected. We address this with a dedicated long-context SFT stage at the full sequence length of 262,144~tokens, following the staged recipe of Qwen~2.5~\citep{Yang2024Qwen25TR} and Nemotron~3 Nano Omni~\citep{nvidia2026nemotron3nanoomni}. Each model keeps its production SFT learning rate, held constant: $6 \times 10^{-5}$ for \apertusonefivesmall{} and $1 \times 10^{-5}$ for \apertusonefivelarge{}, after 195 and 100~warmup steps and at global batch sizes of 32 and 64~sequences, respectively. Once this stage finishes, we evaluate the resulting checkpoints across context lengths and select the best one to carry forward into reinforcement learning.

\paragraph{Data.} The mixture blends 30.3\% long-context multimodal data, 63.2\% short-context multimodal data, 5.4\% short-context text data, and 1.0\% long-context text data, weighted by each source's token count. The multimodal share is vision and audio, each split into short- and long-context length buckets. The text share replays a subset of the production SFT mixture (\autoref{tab:sft_text_mixture}) at short context and reuses the synthetic verifiable tasks of \autoref{sec:long_context} at long context, and tool-use data enters at both lengths. \autoref{tab:long_context_sft_mix} gives the per-source counts and length ranges. The short-context audio subset draws from the ordinary SFT speech pool (\autoref{tab:audio_data_sft}) and AudioMCQ GeminiCoT, while the long-context audio rows include Marco Longspeech and longer TeleAntiFraud and Vocalized SFT samples. The broader available pool appears in \autoref{tab:audio_data_lc_sft}; its inventory does not imply that every listed source enters each run.

\subsection{Improving Instruction Following}

Instruction following is one of the target skills we aim to improve for \apertus{} 1.5. IF scores improve over the course of post-training, as seen in \autoref{tab:lineage_dev} and \autoref{tab:if_dpo_ablation}.

\paragraph{Verifiable IF.} We curate the verifiable IF SFT data through different synthetic data pipelines; the data consists of the following subsets:
\begin{itemize}
    \item \textit{Single-turn verified}. We sample the prompts from the capability-filtering SFT prompt pool and the Dolci Instruct SFT Pool (used for the OLMo 3 SFT run~\citep{olmo3techreport}), and concatenate them with constraints from IFTrain~\citep{pyatkin2026generalizing} and IFEval~\citep{zhou2023instruction}. We either prepend or append the constraints to the instruction. We sample the completions from different open-weight models, such as Qwen3-235B, Kimi-K2.5, MiniMax-M2.7, Qwen3.5-397B, and GPT-OSS-120B-Safeguard. We then verify each prompt-completion pair twice: once for whether it passes all the verifiers of the constraints in the prompt, and then with an LLM-as-a-judge for whether the prompt and the constraints contradict each other and whether the answer is coherent.
    \item \textit{Multi-turn verified}. The multi-turn subset consists of multiple turns, where the first user turn is an instruction specifying the constraint and saying that this constraint should be followed throughout the whole conversation. The following user turns are prompts, sampled from the capability-filtering SFT prompt pool and the Dolci Instruct SFT Pool, with no constraints. We verify the completions similarly, by checking whether they fulfill the initial constraint and whether they are coherent.
\end{itemize}

\paragraph{Improving Non-verifiable Instruction Following.} In real-life user-AI conversations, instructions can be presented in a form which is hard to verify, e.g., ``\textit{When talking about table tennis, focus more on specific forehand serve technique of this player...}'', which can result in numerous correct responses, but the verification method would be based on an LLM-as-a-judge rather than a classic IFEval-like hard-coded constraint check. Moreover, conversations in real life are long and multi-turn with instructions and preferences shifting mid-dialogue. To target this area more, we construct a dataset with a multi-agent pipeline in which three roles, such as a planner, a responder, and an evaluator, collaborate to generate and validate each dialogue. The planner, given a seed context, a target challenge category, and a one-shot example from MultiChallenge~\citep{deshpande-etal-2025-multichallenge}, produces a JSON plan specifying the challenge type, an $N$-turn user question sequence, a hidden trap design, and concrete detection criteria. A trap is a hidden test planted in the conversation (e.g., an early instruction or detail the user later implicitly relies on) that the assistant fails (``triggers'') if it forgets or mishandles it, and the dataset keeps only dialogues in which the trap is set up but not triggered. The responder simulates the assistant under test by following the planned questions, and the evaluator inspects the full dialogue to verify that the trap is present and whether it is triggered, and to score quality on a 10-point scale across safety, accuracy, naturalness, and authenticity. We use seed contexts from Nemotron-Instruction-Following-Chat-v1~\citep{nvidia2025nvidianemotron3efficient}, MiniMax-M2.5 as a generator model (planning and responding), and Qwen3-235B as a judge (evaluator) to mitigate model-specific bias.

For each seed we generate four plans (one per challenge category) and three trajectories per plan, then retain only dialogues in which the trap is correctly introduced but not triggered by the assistant, selecting the top-1 scored trajectory per seed (top-2 for eight-turn chats) to balance quality and diversity. The generated 30,000~dialogues span one to eight conversation turns, with an average of three turns. The dataset shows improved performance on tasks involving more vague constraints compared to standard IF benchmarks, while its multi-turn nature assists with improving on Multi-IF.

\paragraph{Reinforcement Learning.} We sample the IF RL prompts from the capability-filtering SFT prompt pool and concatenate them with constraints from IFTrain~\citep{pyatkin2026generalizing} and IFEval~\citep{zhou2023instruction}. We either prepend or append the constraints to the instruction.

\paragraph{Preference Alignment.} Instruction following also profits from the preference stage of \autoref{sec:preference_alignment}, although no constraint checker is involved: prompts with formatting and content constraints are part of the Dolci-Instruct-DPO prompt set, and the judge scores each rollout on how well it serves the request. To isolate this contribution, we apply online DPO directly to the SFT checkpoint of \apertusonefivesmall{}, skipping reinforcement learning, with the recipe of \autoref{sec:preference_alignment} except for eight rollouts per prompt instead of 12 (\autoref{tab:if_dpo_ablation}). Judge-based preference optimization alone matches the verifier-driven reinforcement-learning stage on IFEval and comes within two points of it on Multi-IF, even though instruction following is a verifiable domain. The gain is specific to on-policy training: offline DPO from the same checkpoint leaves IFEval unchanged and adds roughly three points on Multi-IF. The two stages are complementary rather than redundant, as online DPO on top of the RL checkpoint adds a further 3.8~points on IFEval and 1.5 on Multi-IF. At 70B, the preference stage keeps IFEval and Multi-IF within one point of the reinforcement-learning checkpoint while restoring response length from 253 to 495~tokens and raising AlpacaEval from 25.9 to 59.4, so the added verbosity does not come at the expense of constraint adherence.

\begin{table}[!htbp]

\caption{Instruction following of \apertusonefivesmall{} under different post-training paths from the same SFT checkpoint, thinking disabled. The SFT column is the shared starting point, the next two apply a single alignment stage to it, and the last two are the checkpoints of the release lineage (\autoref{tab:lineage_dev}). Best value per row in bold.}
\label{tab:if_dpo_ablation}
\centering
\small
\begin{tabular}{@{}lrrrrr@{}}
\toprule
\textbf{Benchmark} & \textbf{SFT} & \textbf{+Offline DPO} & \textbf{+Online DPO} & \textbf{+RL} & \textbf{+RL +Online DPO} \\
\midrule
IFEval   & 80.2 & 80.0 & 86.1 & 85.8 & \textbf{89.6} \\
Multi-IF & 74.0 & 77.2 & 81.4 & 83.0 & \textbf{84.5} \\
\bottomrule
\end{tabular}
\end{table}

\subsection{Reinforcement Learning}\label{sec:reasoning_rl}

The reinforcement learning stage optimizes verifiable rewards with a fully asynchronous trainer on our fork of verl~\citep{sheng2025hybridflow}; \autoref{sec:engineering_challenges} describes the infrastructure and \autoref{app:training_configurations} the algorithmic configuration. Rewards come from checking or executing the model's output rather than from judging it: mathematical tasks check against ground truth using Math-Verify~\citep{Kydlicek_Math-Verify_Math_Verification}, instruction-following prompts run formal constraint checkers~\citep{pyatkin2026generalizing}, code runs against test cases in the sandbox, with a binary reward for a perfect pass rate, and other tasks use exact match against gold answers. The model emits final answers through a dedicated answer tool, which keeps the tool-calling format of \autoref{sec:sft} in distribution during the stage. We evaluate a held-out slice of standard benchmarks continuously during training.

Two production configurations train in parallel, one with thinking disabled throughout and one on a mixture of prompts with thinking enabled and disabled; the mixed configuration performs better and feeds the released models. Always-on thinking is not among them: it produces overlong reasoning traces that slow training, and the model rarely calls tools when thinking is enabled, which starves tool-use training of signal. We exclude tool-use prompts themselves from the production run after the hosted tool servers become a throughput bottleneck, a compromise of this release: the tool-use behavior of the released models comes from SFT rather than RL. The run uses no length penalty, to avoid advantage sign flips.

\begin{figure}[t]
    \centering
    \includegraphics[width=\linewidth]{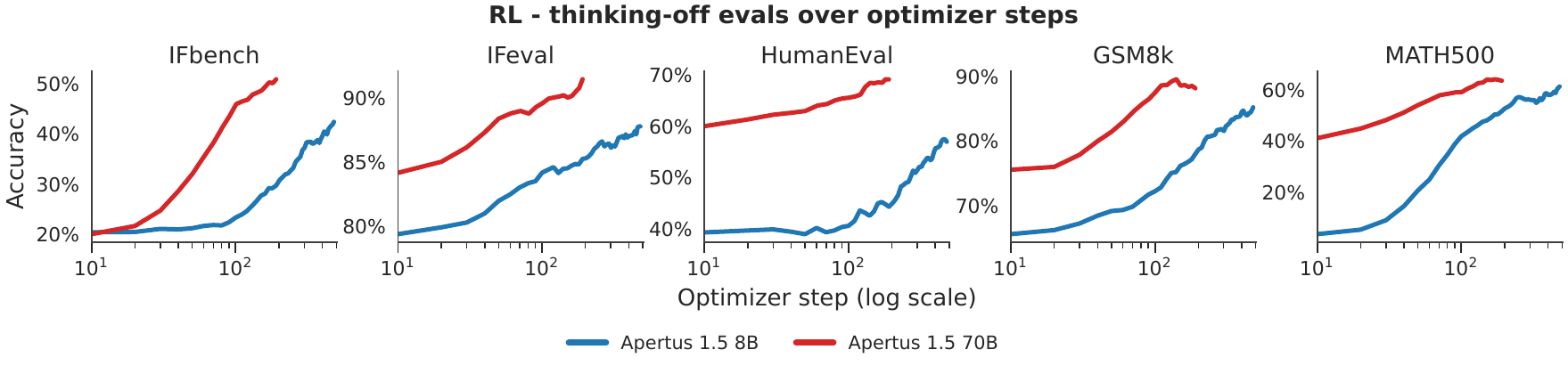}
    \caption{Validation accuracies (\emph{mean@4}) during the reinforcement learning run. Optimizer steps are shown on a log scale. Validation is performed with thinking mode disabled and 4,096~maximum output tokens, and relies on the verifiable answer for math domains. For faster train-time evaluation, benchmarks are randomly subsampled to 100~prompts each, so values may not exactly reflect the complete downstream evaluation.}
    \label{fig:rl-val-acc}
\end{figure}

RL starts from the selected long-context SFT checkpoints, after 4,200~long-context SFT steps for the 8B model and 2,700 for the 70B model. The 8B RL run trains for 480~optimizer steps, corresponding to roughly one million trajectories, and the 70B run for 180~steps. \autoref{fig:rl-val-acc} reports a few evaluation scores during the RL runs. We collect values with thinking mode disabled, 4,096~maximum output tokens, four attempts per prompt, and the dedicated answer tool where it applies. Especially at the 8B scale, evaluation scores do not show a plateau, hinting that longer training could have yielded further improvement.

\begin{table}[t]
\caption{Prompt composition of the \apertus{} 1.5 reinforcement-learning mixture. Counts refer to samples used in the production mixture.}
\label{tab:rl_data_mixture}
\centering
\small
\begin{tabular}{@{}llr@{}}
\toprule
\textbf{Domain} & \textbf{Dataset} & \textbf{Samples} \\
\midrule
Mathematics & Big-Math-RL-Verified & 50,000 \\
\ & DeepMath-103K & 50,000 \\
Instruction following & In-house single-turn & 60,000 \\
\ & In-house hard & 20,000 \\
Code & TACO & 13,000 \\
\ & CodeContests & 13,000 \\
\ & Codeforces & 6,700 \\
\ & APPS & 4,500 \\
Multilingual reasoning & In-house translated r-gym & 50,000 \\
Vision & In-house BlindTasks-RL & 50,000 \\
Tabular & TableGPT & 27,000 \\
Other reasoning & RiddleSense & 3,500 \\
Legal MCQ & LEXam & 1,600 \\
\midrule
\multicolumn{2}{r}{Total} & 349,300 \\
\bottomrule
\end{tabular}
\end{table}

\paragraph{Data.} The prompt pool spans mathematics with verified answers (Big-Math-RL-Verified~\citep{bigmath2025}, DeepMath-103K~\citep{he2025deepmath}), instruction following with checkable constraints from in-house prompt sets and from IFTrain~\citep{pyatkin2026generalizing}, competitive programming (TACO~\citep{taco2023}, APPS~\citep{hendrycks2021apps}, CodeContests~\citep{li2022codecontests}, Codeforces~\citep{openr1codeforces}), procedurally generated reasoning tasks translated into 12~languages (Reasoning Gym~\citep{reasoninggym2025}, translation of traces using the framework introduced by~\citet{yukhymenko-etal-2026-recovered}), riddles~\citep{lin2021riddlesense}, tabular data manipulation tasks~\citep{li2024tablegpt}, legal examination multiple-choice questions (LEXam~\citep{lexam2025}), and vision tasks, with thinking controlled per prompt. Because a small subset of LEXam is part of the RL training mixture, we do not consider it as a held-out evaluation benchmark. \autoref{tab:rl_data_mixture} reports the final data mixture distribution, and we release its configuration with our verl fork.
\subsection{Alignment}\label{sec:preference_alignment}

\looseness=-1
We deploy a two-stage approach to align the models with human preferences. First, we use Direct Preference Optimization (DPO)~\citep{Rafailov2023dpo} to align models with standard qualitative aspects such as truthfulness, helpfulness, honesty, and instruction following. Second, we use Self-Distillation Policy Optimization (SDPO)~\citep{hubotter2026reinforcement, buening2026aligninglanguagemodelsuser} to align the models with the \apertus{} charter~\citep{apertus2025apertusdemocratizingopencompliant}.

\subsubsection{Direct Preference Optimization}

In the first step, we apply DPO to align the models with standard qualitative aspects such as truthfulness, helpfulness, honesty, and instruction following. In particular, we leverage both the standard algorithm and its online variant~\citep{guo2024oaif}. While the former expects a static dataset as input, which we describe below, the latter generates completions for any new batch of prompts from the current version of the policy, calls a judge to select the qualitatively good and bad responses, and updates the weights accordingly. In what follows, we refer to the original DPO algorithm proposed by \citet{Rafailov2023dpo} as offline DPO to distinguish the two approaches. Our training recipe differs for \apertusonefivesmall{} and \apertusonefivelarge{}. We align \apertusonefivesmall{} with online DPO alone. However, we observe that this approach is not immediately applicable to \apertusonefivelarge{}, as it leads to qualitatively short and brittle responses. Therefore, we include standard DPO training on a static dataset before training with online DPO, which manages to improve the model's capability across the board. \autoref{app:dpo_ablations} in \autoref{app:alignment_ablations} details the hyperparameter sweeps and ablations driving these decisions.

\paragraph{Preference Data.} Both online and offline DPO draw from the same prompt set, a filtered and decontaminated subset of Dolci-Instruct-DPO~\citep{olmo3techreport}, whose sources span reasoning, code, instruction following, multilingual chat, mathematics, and safety. We keep the prompts and discard the original responses and preference labels. Decontamination against our evaluation suite and a 3,600-token length filter leave 245,898~preference pairs. For the offline data we generate responses with a pool of 24~models spanning 0.5B to 675B parameters, so that the pool spans a wide range of response quality. We then score these prompt-completion pairs with Qwen3-235B-A22B-Instruct-2507~\citep{Yang2025Qwen3TR}, using the rubric-based prompts of ActiveUltraFeedback~\citep{melikidze2026activeultrafeedback}. We elicit scores on a one-to-five scale for truthfulness, helpfulness, honesty, and instruction following, taking the expected value under the judge's distribution over the five rating tokens. We take the final score as the average over the four aspects. We build each preference pair by MaxMin selection, taking the highest-scoring response as chosen and the lowest-scoring as rejected. \citet{melikidze2026activeultrafeedback} compare this rule against a variety of response-pairing strategies and find MaxMin the strongest of them, which is consistent with the Delta Learning Hypothesis, under which the gap between the two sides of a pair, rather than the absolute quality of either side, drives the gain~\citep{geng2025deltalearning}.
Neither stage trains on multimodal or long-context data, and both run with thinking disabled. \autoref{app:preference_data} gives the rubric prompts and reports the per-source composition of the prompt set, the model pool, and the score distribution of the resulting pairs.

\paragraph{Offline DPO.}
We begin preference alignment with offline DPO~\citep{Rafailov2023dpo} for \apertusonefivelarge{} starting from its reinforcement-learning checkpoint (\autoref{sec:reasoning_rl}) and using the dataset described above. Our implementation uses a length-normalized version~\citep{meng2024simpo, lambert2025tulu3pushingfrontiers} with KL regularization coefficient $\beta = 25$, learning rate of $1 \times 10^{-6}$, linear learning rate scheduler with a warmup ratio of 0.1, the AdamW optimizer~\citep{Loshchilov2019DecoupledWD} ($\beta_1 = 0.9$, $\beta_2 = 0.999$, $\epsilon = 10^{-8}$) with no weight decay, global batch size of 128, for one epoch at a maximum sequence length of 4,096~tokens.

\paragraph{Online DPO.} We apply online DPO to both \apertusonefivesmall{} and \apertusonefivelarge{}, starting from the reinforcement-learning (\autoref{sec:reasoning_rl}) and the final offline DPO checkpoints, respectively. This training uses the prompts from the Dolci-Instruct-DPO dataset while discarding the offline responses. For each batch of prompts we sample rollouts from the current policy with SGLang~\citep{zheng2024sglang} at temperature 1.0 and unrestricted top-$p$ and top-$k$, score them, and form one MaxMin pair per prompt. We then use this batch of prompt-completion pairs to update the weights of the models with the DPO loss. We sample 12~rollouts per prompt for \apertusonefivesmall{} and eight for \apertusonefivelarge{}, where the smaller count is a concession to the cost of generation at that scale. To speed up training and reduce the scoring latency, we deploy Qwen3.6-27B~\citep{qwen3.6-27b} as a judge and use only the helpfulness aspect instead of the average of four. \autoref{app:judge-ablations} ablates both departures: the four aspect scores are highly correlated, and Qwen3.6-27B matches or exceeds the offline stage's much larger judge on external preference benchmarks despite the size gap.

For both model sizes, online DPO uses KL penalty coefficient $\beta = 0.1$, a linear schedule with a warmup ratio of 0.1, the AdamW optimizer~\citep{Loshchilov2019DecoupledWD} ($\beta_1 = 0.9$, $\beta_2 = 0.999$, $\epsilon = 10^{-8}$) with a weight decay of 0.01, and prompt and response lengths of 2,048~tokens each. The prompt cap removes a further 3,210~prompts, so an online epoch covers 242,688 of them. We train for one epoch on our verl fork with FSDP~\citep{zhao2023pytorchfsdp} and SGLang rollouts. \apertusonefivesmall{} trains at a learning rate of $5 \times 10^{-6}$ with 256~prompts per batch for 948~steps on 24~nodes, 16 for training and eight for judge annotation. \apertusonefivelarge{} trains at a learning rate of $2.5 \times 10^{-6}$ with 512~prompts per batch for 474~steps on 144~nodes, 128 for training and 16 for annotation. Both runs complete in under 12~hours. \autoref{tab:dpo_config} lists the full configurations.

\paragraph{Results.} Preference alignment raises the overall average of our development benchmark suite from 49.3\% to 54.8\% for \apertusonefivesmall{} and from 55.3\% to 61.2\% for \apertusonefivelarge{}. The gains are broad. Knowledge rises by 13.0 and 2.3~points at the two scales, mathematics by 4.1 and 4.8, reasoning by 2.6 and 2.7, and safety by 3.9 and 5.4, while cultural knowledge rises by 8.7~points for \apertusonefivelarge{}. Chat improves by 8.7 and 25.9~points, tool use by 13.1 and 12.5, and charter alignment by 22.5 and 21.3. Degenerate outputs become rarer at both scales, falling from 1.7\% to 0.6\% and from 10.6\% to 3.3\%. The gains carry to thinking mode although no stage trains with thinking enabled: the thinking-mode average rises by 4.2~points for \apertusonefivesmall{} and 6.6 for \apertusonefivelarge{}. \autoref{tab:lineage_dev}, \autoref{tab:lineage_heldout}, \autoref{tab:lineage_think}, and \autoref{app:lineage_full} provide detailed evaluation results, discussed in \autoref{sec:post_training_lineage}.

\subsubsection{Charter Alignment with SDPO}\label{sec:sdpo}

The final post-training stage aligns the model with the \apertus{} charter~\citep{apertus2025apertusdemocratizingopencompliant} through text-only self-distillation policy optimization (SDPO)~\citep{hubotter2026reinforcement, buening2026aligninglanguagemodelsuser}.
During training, the student samples a response to a prompt. A judge model receives that response together with the \apertus{} charter and writes textual feedback on how the response can be improved to follow the charter more closely. We form the teacher signal by calculating the log-probabilities of the initial student response with the textual feedback added to the original user prompt. Conditioning on the feedback in context shifts these log-probabilities toward a more charter-aligned version of the same response. We then distill the teacher log-probabilities back into the student, which sees the original prompt alone, with a top-$k$ reverse KL loss. The stage is deliberately short and lightweight, so that the capabilities built by the preceding stages carry over intact.

The two releases differ at this stage: \apertusonefivesmall{} includes SDPO, whereas \apertusonefivelarge{} ships the preference-aligned DPO checkpoint of \autoref{sec:preference_alignment}. In ablations, initially SDPO regressed the capabilities of the larger model, most visibly on math capabilities, and we dropped the stage from the release for that reason. Investigating after the release, we traced the regression not to SDPO itself but to an inconsistent tokenizer configuration between the DPO and the SDPO runs. Training is stable once the configuration is fixed, though the gains remain modest next to those observed at 8B scale, increasing charter alignment by 2.4\% while most other capabilities are unchanged, the exceptions being a 1.1\% gain on coding and a drop of 1.2\% on chat capabilities.

\paragraph{Data.} We train on value-laden safety prompts, where the charter shapes the response most. The prompts come from the \hfdata{gretelai/gretel-safety-alignment-en-v1} dataset, of which we keep the prompts alone and discard the pre-generated responses. Decontamination against our evaluation suite leaves 8,354~prompts. We compare this against Dolci-Instruct-DPO~\citep{olmo3techreport}, RealToxicityPrompts~\citep{RealToxicityPrompts2020}, and PRISM~\citep{kirk2024prism}, and find that the Gretel safety prompts align the model best. Growing the dataset beyond this size does not have a measurable advantage, so we train on it alone to keep the stage cheap.

\paragraph{Feedback.} The judge is Qwen3.6-27B~\citep{qwen3.6-27b}, chosen over self-reflection, in which the student itself, the DPO checkpoint, critiques its own responses. Both variants train comparably and self-reflection aligns better with the charter, reaching a win rate of 73.2\% in comparison to 67.4\%, but it costs roughly 3\% on math, so we adopt the Qwen judge. We further tune the feedback and teacher templates by hand over successive runs, guided by benchmark performance and by the rollouts of the preceding run. \autoref{app:training_configurations} gives both prompt templates. The remaining hyperparameters matter little at this scale, since the small learning rate and the short schedule limit how far the model can change.

\paragraph{Training.} We train in verl with vLLM~\citep{kwon2023vllm} rollouts and thinking disabled throughout. Starting from the DPO checkpoint, we run 117~steps at a batch size of 64, sampling 16~rollouts per prompt, at a learning rate of $1 \times 10^{-7}$ with 10~warmup steps, and distill the top-64 tokens plus one tail bucket that absorbs the remainder of the distribution. The run takes two hours on 32~nodes of four GH200 GPUs each.

\subsection{SFT Ablations}\label{sec:sft_ablations}

Multimodal SFT corpora usually run to tens of millions of conversations, an order of magnitude above our text mixture of 3.7M, and most of them carry reasoning traces~\citep{wen2026innovatorvl, sensenova-si, llava-onevision, nvidia2026nemotron3nanoomni}. Introducing them changes two things at once: text becomes a minority of the mixture, and a reasoning trace, a synthetic derivation written by a larger model that precedes the visible answer and that our chat template places in the thinking span (\autoref{sec:chat_template}), enters about a third of the training conversations.

\begin{samepage}
This motivates four questions:
\begin{enumerate}
    \renewcommand{\labelenumi}{Q\arabic{enumi}.}
    \item How does adding multimodal SFT data change the text capabilities relative to text-only SFT?
    \item What does supervising reasoning traces add, on text and on multimodal tasks?
    \item Do the gains from trace supervision persist when reasoning is disabled at inference time, or do they require the model to reason?
    \item How do off-policy reasoning traces from stronger models during SFT shape what subsequent RL can achieve?
\end{enumerate}
\end{samepage}

\paragraph{Q1--Q3.}
We build the SFT ablations from the cleaned text mixture and a selected subset of the multimodal SFT pool, after decontamination against the evaluation benchmarks, exact deduplication, and answer-format normalization. We construct paired versions of each eligible example with and without reasoning traces, referred to as the CoT and No-CoT arms. Within each pair, the prompt, any multimodal inputs, and the visible answer are identical; only the reasoning trace differs. Traces are retained from the source data where available and generated with Qwen3.8-27B for selected vision examples; a trace is kept only when its conclusion agrees with the visible answer.

\autoref{tab:sft_ablation_setup} summarizes the four training configurations. Text No-CoT versus MM No-CoT isolates the effect of adding multimodal data, while MM No-CoT versus MM CoT isolates reasoning supervision. Comparing Text CoT with MM CoT further measures the contribution of multimodal reasoning traces beyond text reasoning alone. All configurations initialize from the final \apertusonefivesmall{} long-context pretraining checkpoint (\autoref{sec:long_context}) with a fresh optimizer state and are trained for two epochs. We use a sequence length of 8,192 with best-fit packing, a global batch size of 1,024~sequences (8.4M~tokens per update), and a constant learning rate of $8\times10^{-5}$ after 28~warmup updates. The multimodal configurations are trained with three seeds and the text-only configurations with one. We evaluate every configuration with thinking forced off or on, banning the opening token or prefilling it, since after SFT enabling thinking alone does not reliably make the model reason.

\begin{table}[!htbp]
\caption{Configuration of the SFT ablation. Conversations are reported per epoch and tokens over the full two-epoch run. CoT and No-CoT arms use the same conversations and differ only in whether reasoning traces are retained.}
\label{tab:sft_ablation_setup}
\centering
\small
\begin{tabular}{@{}lllrrr@{}}
\toprule
\textbf{Arm} & \textbf{Training Data} & \textbf{Traces} &
\textbf{Conversations} & \textbf{Tokens} & \textbf{Updates} \\
\midrule
Text No-CoT & Text & No & 3.48M & 4.4B & 535 \\
Text CoT    & Text & Yes & 3.48M & 10.2B & 1,211 \\
MM No-CoT   & Text + multimodal & No & 7.40M & 19.6B & 2,352 \\
MM CoT      & Text + multimodal & Yes & 7.40M & 27.5B & 3,290 \\
\bottomrule
\end{tabular}
\end{table}

\begin{figure}[!htbp]
\centering
\begingroup
\sffamily
\sansmath
\pgfplotsset{
  sft axis/.style={
    scale only axis,
    width=3.23cm,
    height=3.55cm,
    ymin=-4.15,ymax=5.9,
    ytick={-4,-2,0,2,4},
    yticklabels={$-4$,$-2$,$0$,$+2$,$+4$},
    axis x line*=bottom,
    axis y line*=left,
    axis line style={draw=apertusAnthracite,line width=0.55pt},
    tick style={draw=apertusAnthracite,line width=0.55pt},
    tick align=outside,
    major tick length=2.5pt,
    tick label style={font=\footnotesize,text=apertusAnthracite},
    xticklabel style={align=center},
    label style={font=\small,text=apertusAnthracite},
    title style={at={(0.5,1.055)},anchor=south,align=center,font=\small,
      text=apertusAnthracite},
    clip=false,
  },
  sft mean/.style={
    ybar,bar width=15pt,bar shift=0pt,
    fill=apertusSkyBlue,draw=apertusAnthracite,line width=0.45pt,
    forget plot,
  },
  sft seed/.style={
    only marks,mark=*,mark size=1.7pt,
    mark options={fill=apertusAnthracite,draw=white,line width=0.35pt},
    forget plot,
  },
  sft zero/.style={draw=apertusAnthracite,opacity=0.5,
    line width=0.65pt,forget plot},
}
\tikzset{sft value/.style={font=\small,text=apertusAnthracite,inner sep=0pt}}
\input{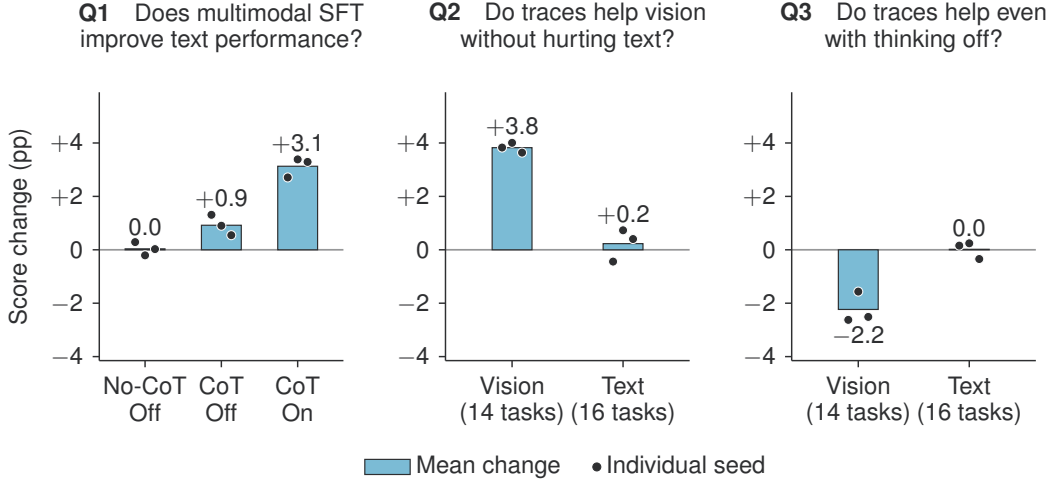}

\begin{tikzpicture}
  \begin{groupplot}[
    sft axis,
    group style={group size=3 by 1,horizontal sep=1.35cm},
  ]
    \nextgroupplot[
      title={\textbf{Q1}\quad Does multimodal SFT\\improve text performance?},
      xmin=-0.6,xmax=2.6,
      xtick={0,1,2},
      xticklabels={{No-CoT\\Off},{CoT\\Off},{CoT\\On}},
      ylabel={Score change (pp)},
      legend to name=sftablationlegend,
      legend columns=2,
      legend cell align=left,
      legend style={draw=none,fill=none,font=\footnotesize,
        text=apertusAnthracite,/tikz/every even column/.append style={column sep=10pt}},
    ]
      \addplot[sft zero] coordinates {(-0.6,0) (2.6,0)};
      \SFTPanelA
      \addlegendimage{area legend,fill=apertusSkyBlue,
        draw=apertusAnthracite,line width=0.45pt}
      \addlegendentry{Mean change}
      \addlegendimage{only marks,mark=*,mark size=1.5pt,
        mark options={fill=apertusAnthracite,draw=apertusAnthracite}}
      \addlegendentry{Individual seed}

    \nextgroupplot[
      title={\textbf{Q2}\quad Do traces help vision\\without hurting text?},
      xmin=-0.6,xmax=1.6,
      xtick={0,1},
      xticklabels={{Vision\\(14 tasks)},{Text\\(16 tasks)}},
    ]
      \addplot[sft zero] coordinates {(-0.6,0) (1.6,0)};
      \SFTPanelB

    \nextgroupplot[
      title={\textbf{Q3}\quad Do traces help even\\with thinking off?},
      xmin=-0.6,xmax=1.6,
      xtick={0,1},
      xticklabels={{Vision\\(14 tasks)},{Text\\(16 tasks)}},
    ]
      \addplot[sft zero] coordinates {(-0.6,0) (1.6,0)};
      \SFTPanelC
  \end{groupplot}
  \node[anchor=north,inner sep=0pt]
    at ($(group c2r1.south)+(0,-1.12cm)$)
    {\pgfplotslegendfromname{sftablationlegend}};
\end{tikzpicture}
\endgroup
\caption{SFT ablations of multimodal supervision and reasoning, one panel per question. Q1 compares text benchmark scores for MM No-CoT versus Text No-CoT and MM CoT versus Text CoT, evaluated under matched response modes. Q2 compares MM CoT with thinking on against MM No-CoT with thinking off. Q3 compares MM CoT against MM No-CoT, both with thinking off. Bars report mean score differences in percentage points; dots denote the three multimodal seeds. Comparisons use a single text-only reference seed in Q1 and paired seeds in Q2 and Q3. Scores are unweighted means over 16~text benchmarks, excluding MGSM (\autoref{tab:sft_ablation_details} gives the reason), and 14~vision benchmarks. Without thinking, decoding is greedy with a 16,384-token budget; thinking is sampled at temperature 0.6 and top-p 0.95 with a 32,768-token budget.}
\label{fig:sft_ablation_questions}
\end{figure}

\autoref{fig:sft_ablation_questions} answers Q1 to Q3 with one panel each. For Q1, adding multimodal SFT data preserves text performance; when multimodal reasoning traces are retained, text performance even improves on average over the model trained only on text reasoning. For Q2, reasoning-trace supervision provides a clear benefit on multimodal tasks when reasoning is enabled at inference time, while its effect on aggregate text performance remains negligible. However, for Q3, the multimodal benefit turns into a small loss when reasoning is disabled: trace supervision lowers vision scores by 2.2~points on average, consistently across all three seeds, while text is unaffected. The averages hide large per-benchmark effects (\autoref{app:sft_ablation_details} in \autoref{app:sft_ablations}): with thinking on, trace supervision adds seven to 18~points on mathematics but costs 12 to 14 on IFEval and Multi-IF, mostly because about a quarter of traces degenerate into repetition and never finish.

\paragraph{Takeaway.} Multimodal supervision can be integrated at scale during SFT without sacrificing text capabilities, and reasoning-trace supervision pays off primarily when the model is allowed to reason at inference time.

\paragraph{Q4.}
Whether reasoning supervision during SFT provides a better starting point for subsequent RL remains unsettled. In text reasoning, prior work finds both that SFT can constrain later exploration~\citep{kang2026quagmires} and that stronger reasoning SFT can improve the initialization for RL~\citep{liu2026acereasonnemotron}; off-policy reasoning traces can also be beneficial when combined with on-policy exploration~\citep{yan2025learning}. Evidence from multimodal reasoning is similarly mixed~\citep{chen2025sft,wei2025advancingmultimodalreasoningreinforcement}, but comes largely from encoder-based VLMs.

We test the same question in our unified discrete-token model. Here, the shift from SFT to RL introduces two distinct mismatches. First, the SFT reasoning traces are generated off-policy by stronger teacher models, whereas RL uses rollouts from the student itself. Second, much of the SFT reasoning supervision is conditioned on multimodal inputs, whereas the RL of this ablation is entirely text-only (\autoref{tab:rl_ablation_setup}). Thus, the multimodal reasoning trajectories learned during SFT are never directly revisited during RL.

\begin{figure}[!t]
\centering
\input{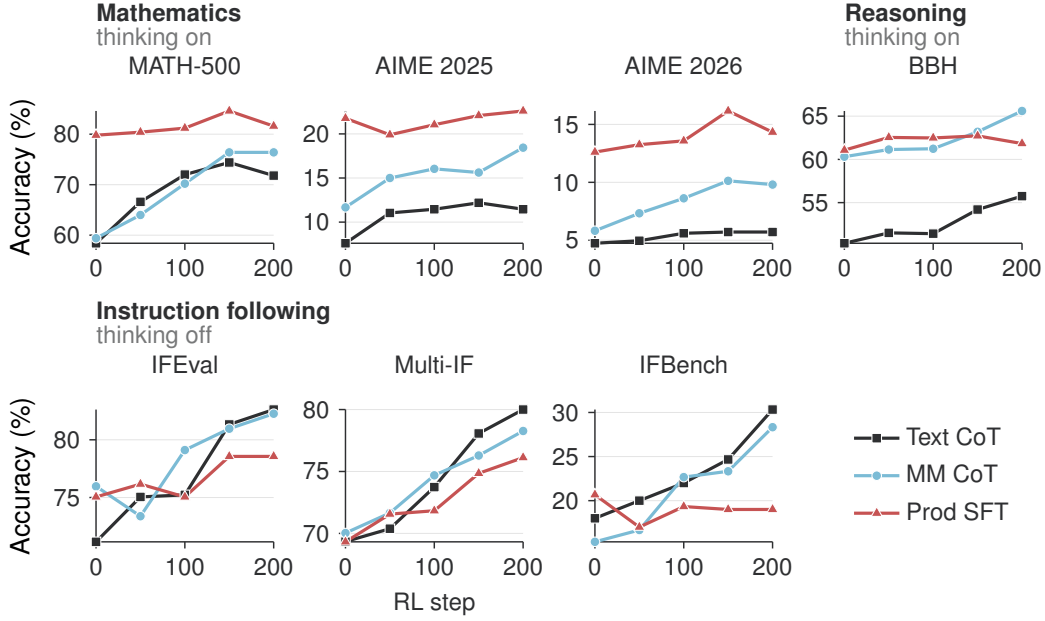}
\caption{Held-out benchmarks during RL from three SFT checkpoints. Each group of panels comes from the runs on one prompt set, evaluated in the thinking mode used for training (\autoref{tab:rl_ablation_setup}). Each curve is a single RL run; step~0 is the SFT checkpoint, and we evaluate every 50~steps. We report accuracy on MATH-500, \emph{mean@32} on AIME, zero-shot exact match on BBH, and prompt-level strict accuracy on IFEval, Multi-IF, and IFBench. A response without a final answer counts as wrong. AIME~2026 excludes problem~1, which appears in DAPO-Math-17k.}
\label{fig:rl_ablation_benchmarks}
\end{figure}

\begin{figure}[!b]
\centering
\input{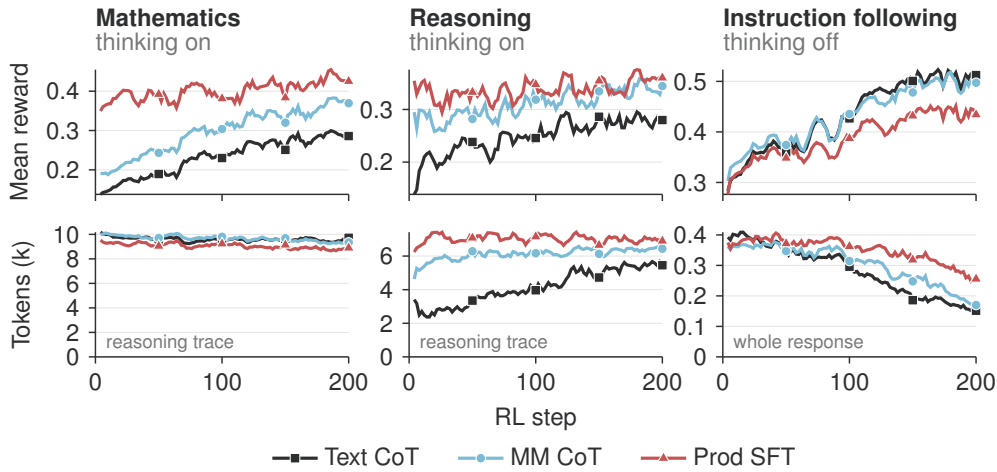}
\caption{Training dynamics of the runs in \autoref{fig:rl_ablation_benchmarks}. The top row shows the mean reward of each step, and the bottom row the tokens per rollout, counting the reasoning trace when thinking is on and the whole response when it is off. Curves are 10-step trailing means from step~3, because the first two steps of asynchronous rollout collect the fastest and then the slowest rollouts, producing unrepresentative reward and length statistics.}
\label{fig:rl_ablation_training}
\end{figure}

We make the controlled comparison between Text CoT and the seed-42 MM CoT checkpoint from the preceding SFT ablation. We additionally include the production SFT checkpoint of \autoref{sec:sft} as a retrospective reference. Unlike the two ablation checkpoints, it is trained on a broader, more heterogeneous mixture without the same pairing and normalization controls, and is therefore not a controlled arm. From each checkpoint, we run RL using our fork of NeMo RL~\citep{nvidia2025nemorl}\footnote{Available at \ghrepo{swiss-ai/Nemo-RL}.} separately on mathematics, reasoning, and instruction-following prompts (\autoref{tab:rl_ablation_setup}). Instruction following is trained with thinking off, since our SFT mixture supervises it almost entirely without reasoning traces and explicit reasoning can hurt instruction following~\citep{li2025thinking,fu2026scaling}. \autoref{fig:rl_ablation_benchmarks} tracks held-out performance, while \autoref{fig:rl_ablation_training} shows the corresponding reward and response-length dynamics.

\begin{table}[!htbp]
\caption{Training configuration of the RL ablation, with one row per prompt set. The sets are drawn from DAPO-Math-17k~\citep{yu2025dapo}, Reasoning Gym~\citep{reasoninggym2025}, and \hfdata{nvidia/Nemotron-RL-instruction\_following}. Each run lasts 200~steps of GRPO~\citep{shao2024deepseekmath}, with 64~prompts and eight rollouts per step. We use binary rewards, a leave-one-out baseline, DAPO asymmetric clipping, no KL penalty, and a constant learning rate of $2\times10^{-6}$. Rollouts are asynchronous, at most two policy versions stale, and capped at 14,336~new tokens with thinking on and 3,840 with thinking off. Mathematics is scored with Math-Verify~\citep{Kydlicek_Math-Verify_Math_Verification}; for reasoning tasks, Qwen3.6-35B-A3B is used as a fallback judge only when the task verifier rejects an answer.}
\label{tab:rl_ablation_setup}
\centering
\small
\begin{tabular}{@{}llrll@{}}
\toprule
\textbf{Prompt Set} & \textbf{Source} & \textbf{Prompts} & \textbf{Reward} & \textbf{Thinking} \\
\midrule
Mathematics & DAPO-Math-17k & 15,984 & Math-Verify & On \\
Reasoning & Reasoning Gym & 14,912 & Verifier, judge fallback & On \\
Instruction following & Nemotron-RL IF & 46,390 & All constraints met & Off \\
\bottomrule
\end{tabular}
\end{table}

Across mathematics and reasoning, MM CoT retains or increases its advantage over Text CoT throughout RL, despite the latter substantially increasing its reasoning length; on instruction following, where SFT contains little trace supervision, the two converge. These results suggest that the gains introduced by multimodal reasoning supervision are not washed out by subsequent text-only RL, but can persist through on-policy optimization. \autoref{fig:rl_ablation_benchmarks} also shows the production SFT checkpoint as a reference.

\paragraph{Takeaway.} Multimodal reasoning traces do more than raise the starting point for text-only RL: RL improves MM CoT more than Text CoT on mathematics, and on reasoning, where both improve equally, MM CoT keeps its lead from SFT.

\clearpage
\section{Evaluations}\label{sec:evaluation}

This chapter evaluates the \apertus{} 1.5 models. \autoref{sec:evaluation_setup} fixes the shared methodology: the harnesses, the benchmark suites, the judge models, and the pitfalls we control for. We then follow the models through the pipeline: the base models after continued pretraining, the post-trained models on general text capabilities including thinking mode, and their multilingual and translation behavior. The new modalities follow, image and audio understanding, and then the remaining capability areas: long context, tool use, and security and safety. Qualitative spot-testing closes the chapter with the behavior that benchmarks miss.

\subsection{Evaluation Setup}\label{sec:evaluation_setup}

We evaluate text capabilities with our fork of the LM Evaluation Harness~\citep{eval-harness}, driven by a launcher layer that fixes task lists, prompt formats, and the serving configuration, so that runs are comparable across checkpoints. Visual and audio benchmarks run on forks of LMMs-Eval~\citep{zhang2025lmmseval} and VLMEvalKit~\citep{duan2024vlmevalkit} behind a common dispatcher. The forks and the task configurations we report are public. The text suite starts from the 27~benchmarks reported for \apertus{} 1.0 and grows to a release suite of roughly 50~tasks spanning general knowledge, instruction following and chat quality, mathematics, coding, reasoning, multilingual and cultural knowledge, bias, toxicity and safety, system-prompt guardrails, function calling, and alignment with the \apertus{} charter. Additions for this release include machine translation, hallucination, over-refusal, function calling, multi-turn instruction following, and the charter-alignment benchmark. Many knowledge and mathematics benchmarks use chain-of-thought variants scored on extracted answers, matching contemporary open-model practice. Before a benchmark enters the suite, we validate the implementation by reproducing published scores of openly released models; we exclude from the reported set tasks whose published scores we cannot reproduce, or whose answer extraction proves unreliable.

Benchmarks that require a judge use openly released models served on Alps: a Llama-3.3-70B-Instruct judge (e.g., for \emph{AlpacaEval}), Llama-2-13B-cls for \emph{HarmBench}~\citep{mazeika2024harmbench}, and a Llama-Guard-4 safety classifier for \emph{RealToxicityPrompts}~\citep{RealToxicityPrompts2020}. Where the original benchmark prescribes a different judge, absolute scores are not directly comparable with the original leaderboard, and the affected sections say so.

We evaluate post-trained models through the chat template. Thinking mode is deactivated by default, and we report it separately. During development we additionally monitor a held-out set of benchmarks outside the training distribution (\autoref{tab:lineage_heldout}) to detect overfitting of the post-training recipe to the development suite.

\paragraph{Known Pitfalls.} Several classes of error distort benchmark numbers, and we control for them explicitly. Harness defaults are tuned for base models: few-shot prompting, greedy decoding, and tight output-token budgets under-report post-trained models, and especially thinking models, whose answers are truncated or missed by answer extraction. Small benchmarks need repeated sampled runs before differences are meaningful. Tool-use scores depend on the parser matching the call format the model emits. Serving-side effects, such as prompt-formatting differences and checkpoint-conversion errors, shift scores silently; the two serving backends are not numerically identical, so no reported series mixes them.

The full list of benchmarks, including relevant settings for replication, is given in~\autoref{app:evaluation_details}.

\subsection{Base Models}

\autoref{tab:pretrain-general-small} to \autoref{tab:pretrain-safety-large} compare the \apertus{} 1.5 base checkpoints against their \apertus{} 1.0 predecessors on the identical suite, with open base models as context: the difference quantifies what continued pretraining adds in text capability while introducing the new modalities. \autoref{sec:pretraining_retrospective} discusses how these capabilities evolve over the course of training.

The evaluations show that the final \apertus{} 1.5 base model improves on a wide range of tasks compared to \apertus{} 1.0. \apertusonefivesmall{} specifically compares well with fully open peer models. In categories such as general language understanding and factual knowledge, \apertus{} base models have mostly closed the gap to other open-source models. Previously a weak point, performance for reasoning, math, and coding has strongly improved compared to the 1.0 model releases. \apertus{} models remain strong in multilingual tasks, despite small regressions.

\begin{table}[!htbp]
\caption{\textbf{Pretraining Evaluation (8B-class models).} Performance (\%) on \textit{general language understanding} tasks, higher is better. \textbf{Bold} = best overall; \textit{italic} = best among fully open models.}
\label{tab:pretrain-general-small}
\centering
\small
\setlength{\tabcolsep}{4pt}
\begin{tabular}{@{}lrrrrrr@{}}
\toprule
 & \multicolumn{4}{c}{\textbf{Fully open}} & \multicolumn{2}{c}{\textbf{Open-weight}} \\
\cmidrule(lr){2-5}\cmidrule(lr){6-7}
\textbf{Benchmark} & \textbf{\apertus{} 1.0} & \textbf{\apertus{} 1.5} & \textbf{OLMo 3} & \textbf{Marin} & \textbf{Qwen3.5} & \textbf{Llama 3.1} \\
Parameters & 8B & 8B & 7B & 8B & 9B & 8B \\
\midrule
HellaSwag     & 78.8          & 78.1          & 74.1 & \textbf{82.4} & 79.3          & 79.3 \\
PIQA          & 80.2          & 80.3          & 78.7 & \textbf{81.9} & 80.5          & 81.2 \\
CommonsenseQA & 62.3          & 62.3          & 58.7 & \textit{70.3} & \textbf{83.6} & 70.4 \\
LAMBADA       & \textit{70.1} & 68.8          & 65.1 & 68.8          & 70.5          & \textbf{72.3} \\
XNLI          & 44.1          & \textit{45.3} & 39.6 & 42.4          & 39.8          & \textbf{45.6} \\
XCOPA         & \textbf{65.6} & 64.9          & 55.1 & 57.0          & 64.1          & 62.5 \\
\midrule
\textbf{Avg}  & 66.9          & 66.6          & 61.9 & \textit{67.1} & \textbf{69.6} & 68.6 \\
\bottomrule
\end{tabular}
\end{table}

\begin{table}[!htbp]
\caption{\textbf{Pretraining Evaluation (70B-class models).} Performance (\%) on \textit{general language understanding} tasks, higher is better. \textbf{Bold} = best overall; \textit{italic} = best among fully open models.}
\label{tab:pretrain-general-large}
\centering
\small
\setlength{\tabcolsep}{3pt}
\begin{tabular}{@{}lrrrrrrrr@{}}
\toprule
 & \multicolumn{6}{c}{\textbf{Fully open}} & \multicolumn{2}{c}{\textbf{Open-weight}} \\
\cmidrule(lr){2-7}\cmidrule(lr){8-9}
\textbf{Benchmark} & \makecell[rb]{\textbf{\apertus{}}\\\textbf{1.0}} & \makecell[rb]{\textbf{\apertus{}}\\\textbf{1.5}} & \textbf{OLMo 3} & \textbf{EuroLLM} & \textbf{Marin} & \textbf{K2-V2} & \textbf{Qwen3.5} & \makecell[rb]{\textbf{Nemotron}\\\textbf{3.5}} \\
Parameters & 70B & 70B & 32B & 22B & 32B & 70B & 27B & 30B-A3B \\
\midrule
HellaSwag     & 83.0          & 82.2          & 81.5 & 79.0 & \textbf{85.0} & 83.7          & 83.4          & 63.7 \\
PIQA          & 83.0          & 82.3          & 81.3 & 81.4 & 83.3          & \textbf{83.9} & 82.3          & 83.1 \\
CommonsenseQA & 53.8          & 66.8          & 71.1 & 60.5 & \textit{75.5} & 73.3          & \textbf{87.4} & 77.7 \\
LAMBADA       & 72.9          & 72.9          & 70.5 & 70.9 & 68.7          & \textbf{76.1} & 74.2          & 72.8 \\
XNLI          & 45.9          & \textbf{46.3} & 42.4 & 44.1 & 42.3          & 44.1          & 45.3          & 43.4 \\
XCOPA         & \textbf{69.9} & 68.0          & 59.5 & 61.3 & 58.6          & 63.8          & 67.3          & 62.1 \\
\midrule
\textbf{Avg}  & 68.1          & 69.8          & 67.7 & 66.2 & 68.9          & \textit{70.8} & \textbf{73.3} & 67.1 \\
\bottomrule
\end{tabular}
\end{table}

\begin{table}[!htbp]
\caption{\textbf{Pretraining Evaluation (8B-class models).} Performance (\%) on \textit{factual knowledge} tasks, higher is better. \textbf{Bold} = best overall; \textit{italic} = best among fully open models.}
\label{tab:pretrain-knowledge-small}
\centering
\small
\setlength{\tabcolsep}{4pt}
\begin{tabular}{@{}lrrrrrr@{}}
\toprule
 & \multicolumn{4}{c}{\textbf{Fully open}} & \multicolumn{2}{c}{\textbf{Open-weight}} \\
\cmidrule(lr){2-5}\cmidrule(lr){6-7}
\textbf{Benchmark} & \textbf{\apertussmall{}} & \textbf{\apertusonefivesmall{}} & \textbf{OLMo 3} & \textbf{Marin} & \textbf{Qwen3.5} & \textbf{Llama 3.1} \\
Parameters & 8B & 8B & 7B & 8B & 9B & 8B \\
\midrule
MMLU CoT    & 59.3          & \textit{62.9} & 59.4          & 62.6          & \textbf{76.7} & 60.7 \\
Global-MMLU & 54.4          & \textit{56.0} & 44.3          & 48.5          & \textbf{77.1} & 52.7 \\
MMLU-Pro    & 29.6          & \textit{38.5} & 33.0          & 30.9          & \textbf{58.4} & 36.1 \\
AGIEval     & 36.1          & 37.2          & 35.0          & \textit{37.3} & \textbf{58.0} & 35.5 \\
SQuADv2     & 32.1          & 17.6          & \textit{36.9} & 24.5          & \textbf{38.7} & 23.6 \\
TruthfulQA  & 51.3          & 49.0          & 46.9          & \textbf{53.7} & 51.8          & 44.2 \\
\midrule
\textbf{Avg} & \textit{43.8} & 43.5          & 42.6          & 42.9          & \textbf{60.1} & 42.1 \\
\bottomrule
\end{tabular}
\end{table}

\begin{table}[!htbp]
\caption{\textbf{Pretraining Evaluation (70B-class models).} Performance (\%) on \textit{factual knowledge} tasks, higher is better. \textbf{Bold} = best overall; \textit{italic} = best among fully open models.}
\label{tab:pretrain-knowledge-large}
\centering
\small
\setlength{\tabcolsep}{3pt}
\begin{tabular}{@{}lrrrrrrrr@{}}
\toprule
 & \multicolumn{6}{c}{\textbf{Fully open}} & \multicolumn{2}{c}{\textbf{Open-weight}} \\
\cmidrule(lr){2-7}\cmidrule(lr){8-9}
\textbf{Benchmark} & \makecell[rb]{\textbf{\apertus{}}\\\textbf{1.0}} & \makecell[rb]{\textbf{\apertus{}}\\\textbf{1.5}} & \textbf{OLMo 3} & \textbf{EuroLLM} & \textbf{Marin} & \textbf{K2-V2} & \textbf{Qwen3.5} & \makecell[rb]{\textbf{Nemotron}\\\textbf{3.5}} \\
Parameters & 70B & 70B & 32B & 22B & 32B & 70B & 27B & 30B-A3B \\
\midrule
MMLU CoT    & 61.8 & 67.0          & 71.4          & 62.9 & \textit{71.6} & 70.5          & \textbf{84.9} & 75.4 \\
Global-MMLU & 56.3 & \textit{59.3} & 49.1          & 54.9 & 54.2          & 56.1          & \textbf{78.6} & 69.4 \\
MMLU-Pro    & 36.0 & 43.8          & \textit{48.0} & 36.8 & 45.3          & 45.4          & \textbf{78.0} & 67.2 \\
AGIEval     & 41.2 & 43.9          & \textit{46.6} & 40.1 & 43.9          & 40.4          & \textbf{62.7} & 50.5 \\
SQuADv2     & 18.8 & 19.1          & \textit{39.1} & 35.6 & 30.1          & 22.8          & \textbf{43.5} & 30.1 \\
TruthfulQA  & 52.5 & 50.8          & \textit{54.9} & 49.3 & 53.3          & \textit{54.9} & \textbf{58.3} & 55.5 \\
\midrule
\textbf{Avg} & 44.4 & 47.3          & \textit{51.5} & 46.6 & 49.7          & 48.4          & \textbf{67.7} & 58.0 \\
\bottomrule
\end{tabular}
\end{table}

\begin{table}[!htbp]
\caption{\textbf{Pretraining Evaluation (8B-class models).} Performance (\%) on \textit{reasoning, mathematics, and code} tasks, higher is better. GSM8K-P = GSM8K Platinum; Minerva = Minerva MATH; MGSM-En = MGSM English. \textbf{Bold} = best overall; \textit{italic} = best among fully open models.}
\label{tab:pretrain-math-code-small}
\centering
\small
\setlength{\tabcolsep}{4pt}
\begin{tabular}{@{}lrrrrrr@{}}
\toprule
 & \multicolumn{4}{c}{\textbf{Fully open}} & \multicolumn{2}{c}{\textbf{Open-weight}} \\
\cmidrule(lr){2-5}\cmidrule(lr){6-7}
\textbf{Benchmark} & \textbf{\apertus{} 1.0} & \textbf{\apertus{} 1.5} & \textbf{OLMo 3} & \textbf{Marin} & \textbf{Qwen3.5} & \textbf{Llama 3.1} \\
Parameters & 8B & 8B & 7B & 8B & 9B & 8B \\
\midrule
\multicolumn{7}{@{}l}{\textit{Reasoning}} \\
BBH       & 48.8 & 60.2          & \textit{63.9} & 55.8          & \textbf{84.4} & 65.2 \\
ACP       & 22.1 & 28.1          & 31.8          & \textit{32.0} & \textbf{52.7} & 35.1 \\
\addlinespace
\multicolumn{7}{@{}l}{\textit{Mathematics}} \\
GSM8K     & 47.2 & 66.2          & \textit{67.6} & 60.4          & \textbf{87.0} & 50.5 \\
GSM8K-P   & 48.2 & 67.8          & \textit{71.1} & 63.6          & \textbf{89.9} & 52.4 \\
Minerva   & 14.0 & 29.6          & \textit{40.4} & 24.6          & \textbf{50.4} & 19.7 \\
MGSM-En   & 5.2  & \textit{40.1} & 31.5          & 12.9          & \textbf{55.3} & 16.5 \\
\addlinespace
\multicolumn{7}{@{}l}{\textit{Code}} \\
HumanEval & 48.6 & \textit{81.8} & 81.4          & 58.5          & \textbf{93.5} & 75.7 \\
MBPP      & 37.6 & \textit{50.6} & \textit{50.6} & 39.2          & \textbf{62.2} & 48.6 \\
\midrule
\textbf{Avg} & 34.0 & 53.1        & \textit{54.8} & 43.4          & \textbf{71.9} & 45.5 \\
\bottomrule
\end{tabular}
\end{table}

\begin{table}[!htbp]
\caption{\textbf{Pretraining Evaluation (70B-class models).} Performance (\%) on \textit{reasoning, mathematics, and code} tasks, higher is better. GSM8K-P = GSM8K Platinum; Minerva = Minerva MATH; MGSM-En = MGSM English. \textbf{Bold} = best overall; \textit{italic} = best among fully open models.}
\label{tab:pretrain-math-code-large}
\centering
\small
\setlength{\tabcolsep}{3pt}
\begin{tabular}{@{}lrrrrrrrr@{}}
\toprule
 & \multicolumn{6}{c}{\textbf{Fully open}} & \multicolumn{2}{c}{\textbf{Open-weight}} \\
\cmidrule(lr){2-7}\cmidrule(lr){8-9}
\textbf{Benchmark} & \makecell[rb]{\textbf{\apertus{}}\\\textbf{1.0}} & \makecell[rb]{\textbf{\apertus{}}\\\textbf{1.5}} & \textbf{OLMo 3} & \textbf{EuroLLM} & \textbf{Marin} & \textbf{K2-V2} & \textbf{Qwen3.5} & \makecell[rb]{\textbf{Nemotron}\\\textbf{3.5}} \\
Parameters & 70B & 70B & 32B & 22B & 32B & 70B & 27B & 30B-A3B \\
\midrule
\multicolumn{9}{@{}l}{\textit{Reasoning}} \\
BBH       & 58.5 & 68.2 & \textit{78.8} & 59.4 & 71.4 & 69.3 & \textbf{91.7} & 77.9 \\
ACP       & 32.8 & 32.0 & \textit{41.7} & 34.4 & 40.6 & 41.6 & \textbf{77.0} & 47.1 \\
\addlinespace
\multicolumn{9}{@{}l}{\textit{Mathematics}} \\
GSM8K     & 54.0 & 70.7 & \textit{80.9} & 55.8 & 69.8 & 68.6 & \textbf{87.3} & 85.4 \\
GSM8K-P   & 58.0 & 72.9 & \textit{82.7} & 59.4 & 70.6 & 70.1 & \textbf{91.2} & 88.2 \\
Minerva   & 17.6 & 32.7 & \textit{46.1} & 24.8 & 38.1 & 28.9 & \textbf{72.0} & -- \\
MGSM-En   & 21.6 & 49.2 & \textit{51.2} & 46.4 & 37.8 & 36.7 & 69.0          & \textbf{77.4} \\
\addlinespace
\multicolumn{9}{@{}l}{\textit{Code}} \\
HumanEval & 55.9 & 84.6 & \textit{92.3} & 66.8 & 78.4 & 79.6 & \textbf{98.7} & 81.6 \\
MBPP      & 42.4 & 56.4 & \textit{66.6} & 43.8 & 55.8 & 59.0 & \textbf{75.8} & 74.4 \\
\midrule
\textbf{Avg} & 42.6 & 58.3 & \textit{67.5} & 48.9 & 57.8 & 56.7 & \textbf{82.8} & 76.0 \\
\bottomrule
\end{tabular}
\end{table}

\begin{table}[!htbp]
\caption{\textbf{Pretraining Evaluation (8B-class models).} Performance (\%) on \textit{multilingual} tasks (``-ml'' = multilingual variant), higher is better. \textbf{Bold} = best overall; \textit{italic} = best among fully open models.}
\label{tab:pretrain-multilingual-small}
\centering
\small
\setlength{\tabcolsep}{4pt}
\begin{tabular}{@{}lrrrrrr@{}}
\toprule
 & \multicolumn{4}{c}{\textbf{Fully open}} & \multicolumn{2}{c}{\textbf{Open-weight}} \\
\cmidrule(lr){2-5}\cmidrule(lr){6-7}
\textbf{Benchmark} & \textbf{\apertus{} 1.0} & \textbf{\apertus{} 1.5} & \textbf{OLMo 3} & \textbf{Marin} & \textbf{Qwen3.5} & \textbf{Llama 3.1} \\
Parameters & 8B & 8B & 7B & 8B & 9B & 8B \\
\midrule
XNLI          & 44.1          & \textit{45.3} & 39.6 & 42.4          & 39.8 & \textbf{45.6} \\
XCOPA         & \textbf{65.6} & 64.9          & 55.1 & 57.0          & 64.1 & 62.5 \\
ARC-ml        & \textbf{38.4} & 37.5          & 25.5 & 26.4          & 37.9 & 33.1 \\
HellaSwag-ml  & \textbf{55.0} & 53.6          & 37.4 & 42.1          & 51.9 & 49.1 \\
LAMBADA-ml    & \textbf{52.2} & 48.9          & 40.2 & 44.7          & 50.3 & 49.6 \\
TruthfulQA-ml & 46.3          & 46.7          & 45.4 & \textbf{47.2} & 46.0 & 45.0 \\
\midrule
\textbf{Avg}  & \textbf{50.3} & 49.5          & 40.5 & 43.3          & 48.3 & 47.5 \\
\bottomrule
\end{tabular}
\end{table}

\begin{table}[!htbp]
\caption{\textbf{Pretraining Evaluation (70B-class models).} Performance (\%) on \textit{multilingual} tasks (``-ml'' = multilingual variant), higher is better. \textbf{Bold} = best overall; \textit{italic} = best among fully open models.}
\label{tab:pretrain-multilingual-large}
\centering
\small
\setlength{\tabcolsep}{3pt}
\begin{tabular}{@{}lrrrrrrrr@{}}
\toprule
 & \multicolumn{6}{c}{\textbf{Fully open}} & \multicolumn{2}{c}{\textbf{Open-weight}} \\
\cmidrule(lr){2-7}\cmidrule(lr){8-9}
\textbf{Benchmark} & \makecell[rb]{\textbf{\apertus{}}\\\textbf{1.0}} & \makecell[rb]{\textbf{\apertus{}}\\\textbf{1.5}} & \textbf{OLMo 3} & \textbf{EuroLLM} & \textbf{Marin} & \textbf{K2-V2} & \textbf{Qwen3.5} & \makecell[rb]{\textbf{Nemotron}\\\textbf{3.5}} \\
Parameters & 70B & 70B & 32B & 22B & 32B & 70B & 27B & 30B-A3B \\
\midrule
XNLI          & 45.9          & \textbf{46.3} & 42.4 & 44.1          & 42.3 & 44.1 & 45.3          & 43.4 \\
XCOPA         & \textbf{69.9} & 68.0          & 59.5 & 61.3          & 58.6 & 63.8 & 67.3          & 62.1 \\
ARC-ml        & 39.6          & \textit{40.2} & 34.0 & 35.9          & 29.1 & 34.7 & \textbf{43.6} & 39.4 \\
HellaSwag-ml  & \textbf{57.6} & 57.1          & 47.2 & 52.9          & 45.2 & 52.4 & 57.3          & 41.8 \\
LAMBADA-ml    & 54.9          & 53.6          & 48.4 & \textit{55.7} & 48.1 & 50.7 & \textbf{56.7} & 52.7 \\
TruthfulQA-ml & 48.4          & \textit{48.8} & 46.7 & 46.1          & 47.2 & 48.3 & \textbf{51.1} & 48.8 \\
\midrule
\textbf{Avg}  & \textit{52.7} & 52.3          & 46.4 & 49.3          & 45.1 & 49.0 & \textbf{53.6} & 48.0 \\
\bottomrule
\end{tabular}
\end{table}

\begin{table}[!htbp]
\caption{\textbf{Pretraining Evaluation (8B-class models).} Performance (\%) on \textit{bias, safety, and cultural knowledge} tasks, higher is better (higher ToxiGen/BBQ = less toxic/biased). \textbf{Bold} = best overall; \textit{italic} = best among fully open models.}
\label{tab:pretrain-safety-small}
\centering
\small
\setlength{\tabcolsep}{4pt}
\begin{tabular}{@{}lrrrrrr@{}}
\toprule
 & \multicolumn{4}{c}{\textbf{Fully open}} & \multicolumn{2}{c}{\textbf{Open-weight}} \\
\cmidrule(lr){2-5}\cmidrule(lr){6-7}
\textbf{Benchmark} & \textbf{\apertus{} 1.0} & \textbf{\apertus{} 1.5} & \textbf{OLMo 3} & \textbf{Marin} & \textbf{Qwen3.5} & \textbf{Llama 3.1} \\
Parameters & 8B & 8B & 7B & 8B & 9B & 8B \\
\midrule
BBQ           & 56.7 & 57.0 & \textit{59.4} & 58.4          & \textbf{63.0} & 58.6 \\
ToxiGen       & 52.2 & 42.7 & \textit{58.8} & 48.3          & \textbf{59.8} & 43.0 \\
WMDP          & 50.2 & 53.2 & 54.5          & \textit{54.9} & \textbf{63.7} & 54.3 \\
CulturalBench & 69.7 & 69.9 & 70.6          & \textit{72.8} & \textbf{78.1} & 67.2 \\
\midrule
\textbf{Avg}  & 57.2 & 55.7 & \textit{60.8} & 58.6          & \textbf{66.2} & 55.8 \\
\bottomrule
\end{tabular}
\end{table}

\begin{table}[!htbp]
\caption{\textbf{Pretraining Evaluation (70B-class models).} Performance (\%) on \textit{bias, safety, and cultural knowledge} tasks, higher is better (higher ToxiGen/BBQ = less toxic/biased). \textbf{Bold} = best overall; \textit{italic} = best among fully open models.}
\label{tab:pretrain-safety-large}
\centering
\small
\setlength{\tabcolsep}{3pt}
\begin{tabular}{@{}lrrrrrrrr@{}}
\toprule
 & \multicolumn{6}{c}{\textbf{Fully open}} & \multicolumn{2}{c}{\textbf{Open-weight}} \\
\cmidrule(lr){2-7}\cmidrule(lr){8-9}
\textbf{Benchmark} & \makecell[rb]{\textbf{\apertus{}}\\\textbf{1.0}} & \makecell[rb]{\textbf{\apertus{}}\\\textbf{1.5}} & \textbf{OLMo 3} & \textbf{EuroLLM} & \textbf{Marin} & \textbf{K2-V2} & \textbf{Qwen3.5} & \makecell[rb]{\textbf{Nemotron}\\\textbf{3.5}} \\
Parameters & 70B & 70B & 32B & 22B & 32B & 70B & 27B & 30B-A3B \\
\midrule
BBQ           & 55.1 & 58.6 & 61.8          & 57.4 & 61.2          & \textit{62.7} & \textbf{63.0} & 59.5 \\
ToxiGen       & 44.1 & 47.2 & \textbf{61.9} & 39.7 & 50.3          & 43.2          & 58.6          & 45.5 \\
WMDP          & 53.3 & 55.9 & \textit{60.5} & 52.5 & 59.5          & 59.2          & \textbf{70.0} & 63.6 \\
CulturalBench & 69.9 & 72.6 & 74.7          & 59.5 & \textit{77.2} & 75.5          & \textbf{83.7} & 64.3 \\
\midrule
\textbf{Avg}  & 55.6 & 58.6 & \textit{64.7} & 52.3 & 62.1          & 60.2          & \textbf{68.8} & 58.2 \\
\bottomrule
\end{tabular}
\end{table}
\subsection{Post-Trained Models}

\autoref{tab:posttrain_8b} and \autoref{tab:posttrain_70b} report the release evaluation runs of the final checkpoints against \apertus{} 1.0 and contemporary peers, all with thinking disabled; the thinking-mode comparison follows in \autoref{tab:posttrain_thinking}. Each table splits the peers into fully open models, which release weights, data, and training recipes, and open-weight models. We evaluate our 8B model against other fully open models: \emph{OLMo 3 7B Instruct}~\citep{olmo3techreport}, \emph{Marin 8B Instruct}~\citep{marin2025marin8binstruct}, and \emph{EuroLLM 9B Instruct-2512}~\citep{ramos2026eurollm22b}. We also report scores for the open-weight models \emph{Qwen3.5-9B}~\citep{qwen3_5} and \emph{Llama 3.1 8B Instruct}~\citep{grattafiori2024llama3herdmodels}. 

We report results for \apertusonefivelarge{} along with other open models above 22B parameters in size. We compare to fully open models \emph{OLMo 3.1 32B Instruct}~\citep{olmo3techreport} by AI2, \emph{K2-V2 Instruct}~\citep{k2team2025k2v2360openreasoningenhancedllm}, and \emph{EuroLLM}~\citep{ramos2026eurollm22b}. Additionally, we report results for the open-weight models \emph{Llama 3.3 70B Instruct}~\citep{grattafiori2024llama3herdmodels}, \emph{Gemma 4 31B}~\citep{gemmateam2026gemma4technicalreport}, \emph{Qwen 3.5 27B}~\citep{qwen3_5}, and \emph{Nemotron 3.5 Lightning 30B-A3B}~\citep{nvidia2025nvidianemotron3efficient}. 

The generational step is clear: \apertusonefivesmall{} improves over \apertus{} 1.0 8B Instruct on every reported benchmark except TruthfulQA and GPQA (\autoref{tab:posttrain_8b}). Among the fully open models of its class, the 8B leads on knowledge, instruction following, and cultural benchmarks, while OLMo~3 leads on mathematics and code; across all compared models of the class, including open-weight ones, it is the strongest on instruction following (IFEval, Multi-IF, and IFBench) and AGIEval. We report separate scores for the subsequent steps of post-training in the following \autoref{sec:post_training_lineage}.

\apertusonefivelarge{} improves over its predecessor on every benchmark and outperforms EuroLLM-22B on nearly all benchmarks, but trails the strongest open-weight contemporaries on most axes (\autoref{tab:posttrain_70b}).

We omit the judged Arena-Hard scores from this snapshot, as the judge model is unavailable in these runs; we report AIME with the thinking comparison in \autoref{tab:posttrain_thinking}.

\begin{table}[t]
\caption{Release evaluation of the 8B-class post-trained models, thinking disabled for all models including thinking-capable peers. All values in percent, higher is better, best per group in bold. Instruction following is scored by prompt-level strict accuracy and code by pass@1.}
\label{tab:posttrain_8b}
\centering
\small
\setlength{\tabcolsep}{4pt}
\begin{tabular}{@{}lrrrrrrr@{}}
\toprule
 & \multicolumn{5}{c}{\textbf{Fully open}} & \multicolumn{2}{c}{\textbf{Open-weight}} \\
\cmidrule(lr){2-6}\cmidrule(lr){7-8}
\textbf{Benchmark} & \textbf{\apertus{} 1.0} & \textbf{\apertus{} 1.5} & \textbf{OLMo 3} & \textbf{Marin} & \textbf{EuroLLM} & \textbf{Qwen3.5} & \textbf{Llama 3.1} \\
Parameters & 8B & 8B & 7B & 8B & 9B & 9B & 8B \\
\midrule
\multicolumn{8}{@{}l}{\textit{Knowledge}} \\
MMLU & 60.6 & \textbf{70.5} & 67.1 & 63.5 & 62.0 & \textbf{83.7} & 72.4 \\
MMLU-Pro & 35.8 & 48.8 & \textbf{52.7} & 41.0 & 38.5 & \textbf{76.0} & 48.4 \\
AGIEval & 38.5 & \textbf{45.4} & 37.1 & 36.4 & 41.7 & \textbf{39.0} & 38.0 \\
ARC-Challenge & 70.2 & \textbf{85.0} & 75.3 & 68.5 & 73.7 & \textbf{88.6} & 81.8 \\
TruthfulQA & 56.6 & 56.2 & \textbf{57.5} & 55.3 & 47.4 & \textbf{59.1} & 55.0 \\
\addlinespace
\multicolumn{8}{@{}l}{\textit{Instruction Following and Chat}} \\
IFEval & 68.4 & \textbf{87.1} & 82.8 & 65.2 & 66.0 & \textbf{84.3} & 75.0 \\
Multi-IF & 66.8 & \textbf{83.7} & 70.5 & 59.6 & 56.4 & \textbf{80.0} & 66.5 \\
IFBench & 19.7 & \textbf{35.3} & 28.3 & 17.7 & 14.7 & \textbf{34.3} & 25.7 \\
AlpacaEval (LC) & 12.5 & 47.2 & 57.3 & 25.2 & \textbf{64.9} & \textbf{71.8} & 36.9 \\
\addlinespace
\multicolumn{8}{@{}l}{\textit{Mathematics}} \\
GSM8K & 62.7 & 78.5 & \textbf{83.5} & 81.2 & 62.7 & \textbf{94.8} & 84.0 \\
MATH & 19.0 & 41.2 & \textbf{59.4} & 31.5 & 31.1 & \textbf{71.4} & 36.6 \\
MATH-500 & 19.6 & 51.2 & \textbf{77.2} & 38.0 & 35.4 & \textbf{87.2} & 47.6 \\
\addlinespace
\multicolumn{8}{@{}l}{\textit{Code}} \\
HumanEval & 41.4 & 68.8 & \textbf{81.6} & 51.2 & 38.8 & \textbf{88.3} & 65.1 \\
MBPP & 40.2 & 51.8 & \textbf{57.8} & 41.2 & 41.2 & \textbf{70.2} & 60.2 \\
\addlinespace
\multicolumn{8}{@{}l}{\textit{Reasoning and Reading}} \\
BBH & 55.7 & \textbf{68.5} & 58.9 & 60.3 & 54.5 & \textbf{84.1} & 72.0 \\
GPQA & 27.9 & 27.2 & \textbf{31.5} & 25.9 & 28.3 & \textbf{57.6} & 26.6 \\
DROP & 44.7 & 54.2 & 55.5 & \textbf{60.1} & 45.8 & \textbf{67.9} & 60.4 \\
\addlinespace
\multicolumn{8}{@{}l}{\textit{Multilingual and Cultural}} \\
Global-MMLU & 55.5 & \textbf{59.4} & 46.4 & 48.7 & 55.0 & \textbf{74.6} & 56.9 \\
SwitzerlandQA & 63.1 & \textbf{64.6} & 53.4 & 53.7 & 64.1 & \textbf{67.9} & 60.2 \\
\bottomrule
\end{tabular}
\end{table}

\begin{table}[!htbp]
\caption{Release evaluation of the 70B-class post-trained models, thinking disabled for all models including thinking-capable peers. All values in percent, higher is better, best per group in bold. Instruction following is scored by prompt-level strict accuracy and code by pass@1.}
\label{tab:posttrain_70b}
\centering
\small
\setlength{\tabcolsep}{2pt}
\begin{tabular}{@{}lrrrrrrrrr@{}}
\toprule
 & \multicolumn{5}{c}{\textbf{Fully open}} & \multicolumn{4}{c}{\textbf{Open-weight}} \\
\cmidrule(lr){2-6}\cmidrule(lr){7-10}
\textbf{Benchmark} & \makecell[rb]{\textbf{\apertus{}}\\\textbf{1.0}} & \makecell[rb]{\textbf{\apertus{}}\\\textbf{1.5}} & \makecell[rb]{\textbf{OLMo}\\\textbf{3.1}} & \textbf{K2-V2} & \textbf{EuroLLM} & \makecell[rb]{\textbf{Llama}\\\textbf{3.3}} & \makecell[rb]{\textbf{Gemma}\\\textbf{4}} & \textbf{Qwen3.5} & \makecell[rb]{\textbf{Nemotron}\\\textbf{3.5}} \\
Parameters & 70B & 70B & 32B & 70B & 22B & 70B & 31B & 27B & 30B-A3B \\
\midrule
\multicolumn{10}{@{}l}{\textit{Knowledge}} \\
MMLU & 70.1 & 76.5 & 80.1 & \textbf{88.1} & 69.0 & 87.6 & \textbf{90.5} & 89.5 & 85.0 \\
MMLU-Pro & 44.7 & 56.4 & 66.0 & \textbf{73.3} & 49.1 & 71.1 & \textbf{84.8} & 81.9 & 56.4 \\
AGIEval & 40.4 & \textbf{50.6} & \textbf{50.6} & 35.8 & 46.0 & \textbf{54.2} & 50.1 & 38.0 & 45.9 \\
ARC-Challenge & 79.7 & 87.8 & \textbf{89.0} & 54.4 & 80.0 & 93.5 & \textbf{96.8} & 93.3 & 91.7 \\
TruthfulQA & 61.9 & 67.0 & \textbf{68.8} & 57.9 & 56.3 & \textbf{66.2} & 64.8 & 64.1 & 58.3 \\
\addlinespace
\multicolumn{10}{@{}l}{\textit{Instruction Following and Chat}} \\
IFEval & 65.1 & \textbf{87.8} & 86.7 & 31.1 & 71.2 & 90.2 & \textbf{90.9} & 88.0 & 75.6 \\
Multi-IF & 65.6 & \textbf{83.7} & 81.9 & 29.2 & 67.8 & 87.8 & \textbf{89.7} & 86.6 & 70.3 \\
IFBench & 20.7 & 25.7 & \textbf{38.0} & 25.7 & 19.0 & 41.3 & \textbf{50.3} & 45.3 & 31.0 \\
\makecell[lt]{AlpacaEval\\(LC)} & 26.0 & 67.2 & 57.6 & 34.4 & \textbf{70.7} & 49.8 & 70.1 & \textbf{78.2} & 62.6 \\
\addlinespace
\multicolumn{10}{@{}l}{\textit{Mathematics}} \\
GSM8K & 75.1 & 85.0 & \textbf{92.9} & 89.0 & 71.8 & 94.9 & \textbf{96.3} & \textbf{96.3} & 89.9 \\
MATH & 27.8 & 52.9 & 65.3 & \textbf{72.4} & 38.0 & 59.9 & \textbf{78.7} & 74.9 & 51.2 \\
MATH-500 & 34.4 & 64.0 & 79.6 & \textbf{85.8} & 43.2 & 75.0 & \textbf{96.4} & 89.6 & 50.4 \\
\addlinespace
\multicolumn{10}{@{}l}{\textit{Code}} \\
HumanEval & 46.6 & 70.3 & \textbf{86.9} & 74.5 & 51.6 & 85.0 & 92.0 & \textbf{96.2} & 55.5 \\
MBPP & 44.8 & 61.4 & \textbf{68.8} & 64.0 & 47.4 & 75.4 & 69.6 & \textbf{80.0} & 74.6 \\
\addlinespace
\multicolumn{10}{@{}l}{\textit{Reasoning and Reading}} \\
BBH & 64.1 & 66.0 & 63.2 & \textbf{85.2} & 60.0 & 87.5 & 92.0 & \textbf{93.3} & 83.3 \\
GPQA & 28.1 & 32.1 & 35.3 & \textbf{54.2} & 32.1 & 58.5 & \textbf{72.1} & 61.8 & 45.1 \\
DROP & 44.3 & 60.7 & \textbf{71.2} & 56.4 & 52.1 & 70.5 & 72.6 & \textbf{82.4} & 70.4 \\
\addlinespace
\multicolumn{10}{@{}l}{\textit{Multilingual and Cultural}} \\
Global-MMLU & 62.9 & 66.5 & 65.0 & \textbf{77.7} & 58.2 & 77.7 & \textbf{84.7} & 83.0 & 69.6 \\
SwitzerlandQA & 67.3 & 69.8 & 61.4 & \textbf{71.2} & 66.3 & 74.2 & \textbf{77.8} & 74.3 & 68.0 \\
\bottomrule
\end{tabular}
\end{table}

\subsubsection{Reasoning and Thinking Mode}

Thinking is a request-time flag of the chat template rather than a separate model, and we evaluate every checkpoint with thinking on and off, at a 32,768-token thinking budget. For small benchmarks with a high run-to-run variance, like AIME, we report \emph{mean@k} metrics across multiple seeds (e.g., with $k=8$ for AIME). \autoref{tab:posttrain_thinking} compares the \apertus{} models in both modes with thinking-native peers in their thinking mode. We compare to \emph{OLMo 3 7B Think}~\citep{olmo3techreport}, \emph{OLMo 3.1 32B Think}~\citep{olmo3techreport}, and \emph{K2 Think V2}~\citep{k2think2026k2think0126}. The improvement concentrates where reasoning is long: on the 8B, MATH-500 improves from 51.2 to 71.4, MATH from 41.2 to 57.3, and IFBench from 35.3 to 44.3, the strongest IFBench score among the compared 8B-class models; multimodal reasoning gains double digits, roughly 10~points on MMMU. Knowledge benchmarks move less, and the 70B even cedes GPQA under thinking.
For AIME 24/25, while still behind other peer models, \apertus{} has improved dramatically since version 1.0, when the models were unable to solve a substantial number of tasks. Thinking further pushes \apertusonefivesmall{} from 7.5 to 22.9 on AIME 24, and from 8.3 to 23.8 on AIME 25.

\begin{table}[!htbp]
\caption{Thinking-mode comparison on reasoning-heavy benchmarks: the \apertus{} 1.5 models with thinking off and on, and thinking-native peers in their thinking mode. All values in percent, best per row in bold. We report \emph{mean@8} for AIME, and \emph{mean@4} for other math and coding benchmarks.}
\label{tab:posttrain_thinking}
\centering
\small
\begin{tabular}{@{}lrrrrrrr@{}}
\toprule
 & \multicolumn{2}{c}{\makecell{\textbf{\apertus{} 1.5}\\\textbf{8B}}} & \multicolumn{2}{c}{\makecell{\textbf{\apertus{} 1.5}\\\textbf{70B}}} & \multicolumn{3}{c}{\textbf{Thinking-native peers}} \\
\cmidrule(lr){2-3}\cmidrule(lr){4-5}\cmidrule(lr){6-8}
\textbf{Benchmark} & \textbf{Off} & \textbf{On} & \textbf{Off} & \textbf{On} & \textbf{OLMo 3 7B} & \textbf{OLMo 3.1 32B} & \textbf{K2 Think V2 70B} \\
\midrule
\multicolumn{8}{@{}l}{\textit{Knowledge and Reasoning}} \\
MMLU & 70.5 & 74.5 & 76.5 & 77.1 & 74.0 & 82.7 & \textbf{88.0} \\
GPQA & 27.2 & 32.1 & 32.1 & 30.1 & 27.2 & 39.7 & \textbf{59.6} \\
\addlinespace
\multicolumn{8}{@{}l}{\textit{Instruction Following}} \\
IFBench & 35.3 & \textbf{44.3} & 25.7 & 32.7 & 20.0 & 23.0 & 36.7 \\
\addlinespace
\multicolumn{8}{@{}l}{\textit{Mathematics}} \\
GSM8K & 78.5 & 85.7 & 85.0 & 86.4 & 91.9 & \textbf{96.7} & 89.8 \\
MATH & 41.2 & 57.3 & 52.9 & 59.0 & 74.0 & 74.9 & \textbf{75.2} \\
MATH-500 & 51.2 & 71.4 & 64.0 & 73.6 & 86.2 & 88.2 & \textbf{90.2} \\
AIME 24 & 7.5 & 22.9 & 15.0 & 27.9 & 74.2 & 80.0 & \textbf{84.6} \\
AIME 25 & 8.3 & 23.8 & 7.0 & 21.7 & 69.1 & \textbf{78.8} & 74.6 \\
\bottomrule
\end{tabular}
\end{table}


\subsubsection{Post-Training Lineage}\label{sec:post_training_lineage}

Our post-training pipeline runs in four stages: supervised finetuning, including its long-context stage (\autoref{sec:sft} and \autoref{sec:long_context_sft}), reinforcement learning (\autoref{sec:reasoning_rl}), preference alignment with DPO (\autoref{sec:preference_alignment}), and charter alignment with SDPO (\autoref{sec:sdpo}). \autoref{tab:lineage_dev} follows both models through that sequence, each column adding one stage to the checkpoint of the column before it, so the difference between adjacent columns isolates what the intervening stage contributes. We split the suite in two: the development benchmarks, which guide our own iteration, and the held-out benchmarks of \autoref{tab:lineage_heldout}, which we never consult when selecting checkpoints and which therefore measure how far the gains generalize. \autoref{tab:lineage_think} repeats the exercise with thinking enabled, and \autoref{app:lineage_full} gives all 37~development benchmarks. In these tables, including \autoref{tab:lineage_full}, GPT-4.1 (\texttt{gpt-4.1-2025-04-14}) judges AlpacaEval and Arena-Hard v0.1, and o3-mini (\texttt{o3-mini-2025-01-31}) judges Arena-Hard v2.

Every stage adds to the overall average, which rises from 47.3\% to 55.0\% for \apertusonefivesmall{} and from 52.2\% to 61.2\% for \apertusonefivelarge{}, and the held-out averages move with them, from 48.2\% to 51.7\% and from 53.2\% to 56.4\%. The stages divide the labor. Reinforcement learning does most of the work on reasoning and instruction following, lifting ACPBench by 5.1 and 11.4~points at the two scales and IFEval by 5.6 and 7.3. Preference alignment then takes the conversational and agentic axes: AlpacaEval gains 12.1~points at 8B and 33.5 at 70B, Arena-Hard v0.1 gains 13.9 and 38.8, and tool use gains 13.1 and 12.5. It also carries most of the safety movement, with BBQ rising 12.4~points at 70B and ToxiGen 8.6 at 8B. SDPO is deliberately narrow, moving charter alignment by 9.9~points and AlpacaEval by 10.0, and leaving the remaining benchmarks within a few points.

Two effects run against the trend. Reinforcement learning shortens responses at 70B scale, from 367 to 253~tokens on average over the three chat benchmarks, and offline DPO raises them to 495, the behavior that motivates the offline stage of \autoref{sec:preference_alignment}. The alignment stages also make both models markedly more cautious: refusals on benign prompts that resemble harmful ones climb sharply through the alignment stages (\autoref{sec:overrefusal_hallucination}).

\begin{table}[!htbp]
\caption{Post-training lineage of the \apertus{} 1.5 models on the development benchmark suite, thinking disabled. Each column adds one stage to the one before it. A representative subset of each skill area is shown; \autoref{tab:lineage_full} gives all 37~benchmarks, and the overall average is taken over all of them. Best value per row in bold, taken separately within each model.}
\label{tab:lineage_dev}
\centering
\small
\begin{tabular}{@{}lrrrrrrr@{}}
\toprule
& \multicolumn{4}{c}{\textbf{\apertusonefivesmall{}}} & \multicolumn{3}{c}{\textbf{\apertusonefivelarge{}}} \\
\cmidrule(lr){2-5}\cmidrule(lr){6-8}
\textbf{Benchmark} & \textbf{SFT} & \textbf{+RL} & \textbf{+DPO} & \textbf{+SDPO} & \textbf{SFT} & \textbf{+RL} & \textbf{+DPO} \\
\midrule
\multicolumn{8}{@{}l}{\textit{Knowledge and Reasoning}} \\
MMLU & 65.8 & 66.6 & \textbf{70.7} & 70.6 & 75.1 & 72.0 & \textbf{77.2} \\
MMLU-Pro & 40.1 & 43.0 & \textbf{49.4} & 49.2 & 46.3 & 44.4 & \textbf{56.9} \\
BBH & 67.9 & 69.1 & \textbf{71.1} & 68.2 & 72.4 & \textbf{75.2} & 66.7 \\
ACPBench & 26.1 & 31.2 & \textbf{37.9} & 37.1 & 38.3 & 49.7 & \textbf{63.6} \\
DROP & 53.2 & \textbf{56.0} & 55.2 & 54.1 & 57.7 & 57.8 & \textbf{60.6} \\
\addlinespace
\multicolumn{8}{@{}l}{\textit{Instruction Following and Chat}} \\
IFEval & 80.2 & 85.8 & 89.6 & \textbf{90.0} & 83.5 & \textbf{90.8} & 89.8 \\
Multi-IF & 74.0 & 83.0 & 84.5 & \textbf{85.0} & 78.0 & \textbf{86.3} & 85.7 \\
AlpacaEval    & 20.0 & 20.6 & 32.7 & \textbf{42.7} & 22.6 & 25.9 & \textbf{59.4} \\
\addlinespace
\multicolumn{8}{@{}l}{\textit{Math}} \\
GSM8K & 77.0 & 79.5 & \textbf{80.7} & 78.6 & 78.6 & 84.7 & \textbf{84.8} \\
MATH & 35.2 & 39.7 & \textbf{44.6} & 41.4 & 38.0 & 45.8 & \textbf{52.8} \\
MATH-500 & 43.2 & 47.0 & \textbf{54.2} & 50.2 & 44.2 & 56.2 & \textbf{63.2} \\
\addlinespace
\multicolumn{8}{@{}l}{\textit{Coding}} \\
HumanEval & 85.3 & \textbf{86.0} & 83.2 & 85.5 & 89.6 & \textbf{91.8} & 88.6 \\
MBPP & 48.6 & 51.6 & 51.2 & \textbf{52.6} & 54.2 & 56.4 & \textbf{62.2} \\
\addlinespace
\multicolumn{8}{@{}l}{\textit{Multilingual and Cultural}} \\
Global-MMLU & 55.8 & 55.4 & 58.9 & \textbf{59.3} & 64.0 & 61.7 & \textbf{66.6} \\
MGSM & 53.3 & 57.5 & \textbf{66.7} & 66.2 & 59.8 & \textbf{70.3} & 67.7 \\
INCLUDE-44 & \textbf{56.9} & 56.7 & 56.6 & \textbf{56.9} & 60.4 & 59.9 & \textbf{61.6} \\
SwitzerlandQA & \textbf{65.7} & 64.4 & 64.6 & 64.6 & 68.5 & 67.6 & \textbf{69.8} \\
\addlinespace
\multicolumn{8}{@{}l}{\textit{Safety and Alignment}} \\
BBQ & 62.9 & 67.4 & \textbf{67.8} & 67.6 & 65.7 & 68.1 & \textbf{80.5} \\
ToxiGen & 53.8 & 73.3 & \textbf{81.9} & 81.7 & 77.7 & 83.8 & \textbf{86.8} \\
WMDP & 51.7 & 50.7 & \textbf{53.4} & \textbf{53.4} & 54.6 & 56.1 & \textbf{56.9} \\
Charter Alignment & 33.4 & 34.3 & 56.8 & \textbf{66.7} & 41.7 & 50.9 & \textbf{72.2} \\
\addlinespace
\multicolumn{8}{@{}l}{\textit{Tool Use}} \\
BFCL v3 & 32.2 & 39.8 & \textbf{52.9} & 48.2 & 43.3 & 43.5 & \textbf{56.0} \\
\midrule
\textbf{Overall Average} & 47.3 & 49.3 & 54.8 & \textbf{55.0} & 52.2 & 55.3 & \textbf{61.2} \\
\bottomrule
\end{tabular}
\end{table}

\begin{table}[!htbp]
\caption{Held-out benchmarks, excluded from development, for both models, thinking disabled. All six held-out benchmarks are shown. Best value per row in bold, taken separately within each model.}
\label{tab:lineage_heldout}
\centering
\small
\begin{tabular}{@{}lrrrrrrr@{}}
\toprule
& \multicolumn{4}{c}{\textbf{\apertusonefivesmall{}}} & \multicolumn{3}{c}{\textbf{\apertusonefivelarge{}}} \\
\cmidrule(lr){2-5}\cmidrule(lr){6-8}
\textbf{Benchmark} & \textbf{SFT} & \textbf{+RL} & \textbf{+DPO} & \textbf{+SDPO} & \textbf{SFT} & \textbf{+RL} & \textbf{+DPO} \\
\midrule
\multicolumn{8}{@{}l}{\textit{Knowledge and Reasoning}} \\
AGIEval & 33.1 & 34.9 & 44.9 & \textbf{45.5} & 38.1 & 49.1 & \textbf{50.5} \\
ARC-Challenge & 78.4 & 84.0 & \textbf{84.9} & 84.8 & 85.1 & 86.9 & \textbf{87.3} \\
GPQA & 29.5 & 29.7 & \textbf{31.2} & 26.8 & \textbf{32.8} & 32.1 & 31.2 \\
\addlinespace
\multicolumn{8}{@{}l}{\textit{Math}} \\
GSM8K-Platinum & 76.5 & 78.0 & \textbf{79.6} & 78.2 & 84.6 & \textbf{87.1} & 83.3 \\
\addlinespace
\multicolumn{8}{@{}l}{\textit{Multilingual}} \\
mLogiQA & 35.4 & \textbf{37.5} & 35.2 & 35.1 & 41.5 & 39.0 & \textbf{44.5} \\
Mult. ARC & 36.5 & 37.1 & \textbf{39.8} & 39.6 & 37.1 & 40.2 & \textbf{41.8} \\
\midrule
\textbf{Overall Average} & 48.2 & 50.2 & \textbf{52.6} & 51.7 & 53.2 & 55.7 & \textbf{56.4} \\
\bottomrule
\end{tabular}
\end{table}

\begin{table}[t]
\caption{Post-training lineage with thinking enabled, on all 19~benchmarks evaluated in thinking mode. Benchmarks marked $^\dagger$ are held out from development. The two averages cover the development and held-out benchmarks respectively, and neither is comparable with \autoref{tab:lineage_dev}, which spans a wider suite. Best value per row in bold, taken separately within each model.}
\label{tab:lineage_think}
\centering
\small
\begin{tabular}{@{}lrrrrrrr@{}}
\toprule
& \multicolumn{4}{c}{\textbf{\apertusonefivesmall{}}} & \multicolumn{3}{c}{\textbf{\apertusonefivelarge{}}} \\
\cmidrule(lr){2-5}\cmidrule(lr){6-8}
\textbf{Benchmark} & \textbf{SFT} & \textbf{+RL} & \textbf{+DPO} & \textbf{+SDPO} & \textbf{SFT} & \textbf{+RL} & \textbf{+DPO} \\
\midrule
\multicolumn{8}{@{}l}{\textit{Knowledge and Reasoning}} \\
MMLU & 69.9 & 70.7 & \textbf{74.2} & 73.2 & 73.4 & 76.6 & \textbf{78.6} \\
MMLU-Pro & 46.5 & 48.4 & 50.5 & \textbf{50.8} & 51.5 & 46.7 & \textbf{57.7} \\
AGIEval$^\dagger$ & 33.2 & 35.4 & 42.9 & \textbf{44.4} & 34.0 & \textbf{48.7} & 47.0 \\
BBH & 67.9 & 68.1 & 70.4 & \textbf{71.0} & 72.0 & \textbf{75.4} & 65.1 \\
ACPBench & 40.1 & 33.8 & 48.6 & \textbf{49.4} & 37.8 & 50.2 & \textbf{70.6} \\
DROP & 53.3 & 55.4 & \textbf{55.5} & 55.4 & 53.4 & 54.6 & \textbf{61.1} \\
GPQA$^\dagger$ & 29.9 & \textbf{33.0} & 31.7 & 32.1 & 29.7 & 32.8 & \textbf{37.3} \\
\addlinespace
\multicolumn{8}{@{}l}{\textit{Math}} \\
GSM8K & 79.8 & 86.4 & 84.7 & \textbf{86.7} & 84.8 & \textbf{88.2} & 85.4 \\
MATH & 51.6 & 55.8 & 59.3 & \textbf{60.3} & 56.6 & 55.2 & \textbf{59.7} \\
Minerva MATH & 60.9 & 66.2 & 73.0 & \textbf{74.3} & 67.0 & 65.9 & \textbf{74.0} \\
MATH-500 & 61.2 & 66.2 & \textbf{70.4} & 69.6 & 66.6 & 66.4 & \textbf{74.0} \\
GSM8K-Platinum$^\dagger$ & 84.3 & 90.2 & \textbf{91.1} & 89.2 & 82.5 & \textbf{90.8} & 88.5 \\
MathQA & 45.8 & 46.0 & 46.3 & \textbf{46.4} & 48.1 & 47.9 & \textbf{48.4} \\
AIME 24 & 3.3 & 0.0 & 16.7 & \textbf{23.3} & 3.3 & 20.0 & \textbf{26.7} \\
AIME 25 & 16.7 & 10.0 & \textbf{23.3} & 20.0 & 10.0 & 13.3 & \textbf{30.0} \\
\addlinespace
\multicolumn{8}{@{}l}{\textit{Coding}} \\
HumanEval & \textbf{51.4} & 38.0 & 32.6 & 26.2 & \textbf{65.0} & 37.5 & 59.7 \\
MBPP & 49.6 & 51.4 & 51.6 & \textbf{52.0} & 55.6 & 58.2 & \textbf{62.2} \\
\addlinespace
\multicolumn{8}{@{}l}{\textit{Multilingual}} \\
MGSM & 52.3 & 65.1 & \textbf{68.6} & 68.0 & 60.0 & 64.7 & \textbf{66.0} \\
mLogiQA$^\dagger$ & \textbf{40.0} & 32.8 & 32.1 & 31.5 & 42.2 & 33.4 & \textbf{44.4} \\
\midrule
\textbf{Average (development)} & 50.0 & 50.8 & 55.0 & \textbf{55.1} & 53.7 & 54.7 & \textbf{61.3} \\
\textbf{Average (held-out)$^\dagger$} & 46.8 & 47.9 & \textbf{49.5} & 49.3 & 47.1 & 51.4 & \textbf{54.3} \\
\bottomrule
\end{tabular}
\end{table}

\subsection{Multilingual and Cultural Evaluation}\label{sec:multilingual_eval}

Multilinguality is where \apertus{} stands strongest relative to its peers. The 13-point gap to Qwen3.5-9B on English knowledge benchmarks (\autoref{tab:posttrain_8b}) shrinks to two or three points on the multilingual and cultural slice (\autoref{tab:multilingual_8b}); \apertusonefivesmall{} is the strongest of all compared models on multilingual ARC and Multi-IF and among the strongest on multilingual TruthfulQA, and its lead over the fully open group is wide. Thinking transfers across languages, lifting Global-MMLU by eight points. The clearest weakness is XNLI, which regresses against \apertus{} 1.0 at both scales; \apertusonefivelarge{} improves on its predecessor everywhere else (\autoref{tab:multilingual_70b}) but trails the strongest open-weight contemporaries.

\begin{table}[!htbp]
\caption{Multilingual and cultural evaluation of the 8B-class post-trained models, thinking disabled. All values in percent, higher is better, best per group in bold. Global-MMLU, Multi-IF, and SwitzerlandQA repeat from \autoref{tab:posttrain_8b} for completeness. Multiple-choice benchmarks report accuracy, length-normalized for multilingual ARC.}
\label{tab:multilingual_8b}
\centering
\small
\setlength{\tabcolsep}{4pt}
\renewcommand{\arraystretch}{0.9}
\begin{tabular}{@{}lrrrrrrr@{}}
\toprule
 & \multicolumn{5}{c}{\textbf{Fully open}} & \multicolumn{2}{c}{\textbf{Open-weight}} \\
\cmidrule(lr){2-6}\cmidrule(lr){7-8}
\textbf{Benchmark} & \textbf{\apertus{} 1.0} & \textbf{\apertus{} 1.5} & \textbf{OLMo 3} & \textbf{Marin} & \textbf{EuroLLM} & \textbf{Qwen3.5} & \textbf{Llama 3.1} \\
Parameters & 8B & 8B & 7B & 8B & 9B & 9B & 8B \\
\midrule
\multicolumn{8}{@{}l}{\textit{Multilingual Knowledge}} \\
Global-MMLU & 55.5 & \textbf{59.4} & 46.4 & 48.7 & 55.0 & \textbf{74.6} & 56.9 \\
INCLUDE-44 & 54.1 & \textbf{56.9} & 38.4 & 38.6 & 51.5 & \textbf{70.9} & 53.2 \\
INCLUDE-45 & 38.9 & \textbf{39.5} & 32.1 & 34.6 & 34.4 & \textbf{44.5} & 33.8 \\
Mult.\ TruthfulQA & \textbf{52.1} & 51.7 & 50.6 & 47.6 & 47.2 & \textbf{52.0} & 50.8 \\
Mult.\ ARC & 39.3 & \textbf{42.2} & 29.8 & 29.5 & 36.5 & \textbf{39.2} & 35.6 \\
\addlinespace
\multicolumn{8}{@{}l}{\textit{Multilingual Reasoning and Instruction Following}} \\
MGSM & 47.0 & \textbf{66.2} & 48.5 & 42.9 & 38.8 & \textbf{67.8} & \textbf{67.8} \\
mLogiQA & 28.9 & 35.2 & 36.8 & 32.6 & \textbf{37.0} & \textbf{62.3} & 41.2 \\
XNLI & \textbf{47.0} & 43.8 & 39.3 & 40.4 & 42.4 & \textbf{43.9} & 43.4 \\
Multi-IF & 66.8 & \textbf{83.7} & 70.5 & 59.6 & 56.4 & \textbf{80.0} & 66.5 \\
\addlinespace
\multicolumn{8}{@{}l}{\textit{Cultural Knowledge}} \\
BLEnD & 62.9 & \textbf{67.3} & 59.6 & 62.0 & 66.5 & \textbf{69.5} & 67.3 \\
CulturalBench & 50.9 & 60.5 & 62.1 & \textbf{69.4} & 46.7 & \textbf{69.1} & 68.2 \\
SwitzerlandQA & 63.1 & \textbf{64.6} & 53.4 & 53.7 & 64.1 & \textbf{67.9} & 60.2 \\
\bottomrule
\end{tabular}
\end{table}

\begin{table}[!htbp]
\caption{Multilingual and cultural evaluation of the 70B-class post-trained models, thinking disabled. All values in percent, higher is better, best per group in bold. Multiple-choice benchmarks report accuracy, length-normalized for multilingual ARC.}
\label{tab:multilingual_70b}
\centering
\small
\setlength{\tabcolsep}{2pt}
\renewcommand{\arraystretch}{0.9}
\begin{tabular}{@{}lrrr@{\hspace{7.5pt}}rrrrrr@{}}
\toprule
 & \multicolumn{5}{c}{\textbf{Fully open}} & \multicolumn{4}{c}{\textbf{Open-weight}} \\
\cmidrule(lr){2-6}\cmidrule(lr){7-10}
\textbf{Benchmark} & \makecell[rb]{\textbf{\apertus{}}\\\textbf{1.0}} & \makecell[rb]{\textbf{\apertus{}}\\\textbf{1.5}} & \makecell[rb]{\textbf{OLMo}\\\textbf{3.1}} & \textbf{K2} & \textbf{EuroLLM} & \makecell[rb]{\textbf{Llama}\\\textbf{3.3}} & \makecell[rb]{\textbf{Gemma}\\\textbf{4}} & \textbf{Qwen3.5} & \textbf{Nemotron} \\
Parameters & 70B & 70B & 32B & 70B & 22B & 70B & 31B & 27B & 30B-A3B \\
\midrule
\multicolumn{10}{@{}l}{\textit{Multilingual Knowledge}} \\
Global-MMLU & 62.9 & 66.5 & 65.0 & \textbf{77.7} & 58.2 & 77.7 & \textbf{84.7} & 83.0 & 69.6 \\
INCLUDE-44 & 58.3 & 61.6 & 51.6 & \textbf{66.6} & 55.3 & 71.7 & \textbf{78.9} & 77.1 & 56.7 \\
INCLUDE-45 & 40.0 & 41.8 & 37.9 & \textbf{45.3} & 37.8 & 45.9 & \textbf{54.6} & 50.8 & 39.7 \\
Mult.\ TruthfulQA & 54.2 & \textbf{56.9} & 53.8 & 49.5 & 48.9 & 55.2 & \textbf{56.3} & 55.5 & 52.5 \\
Mult.\ ARC & 39.5 & \textbf{43.5} & 38.4 & 32.2 & 37.8 & \textbf{46.0} & 39.5 & 42.1 & 39.7 \\
\addlinespace
\multicolumn{10}{@{}l}{\textit{Multilingual Reasoning and Instruction Following}} \\
MGSM & 62.2 & \textbf{67.7} & 67.3 & 61.9 & 50.0 & \textbf{85.5} & 73.7 & 73.9 & 74.4 \\
mLogiQA & 40.2 & 44.1 & \textbf{50.5} & 46.8 & 40.4 & 57.5 & \textbf{74.1} & 68.5 & 49.9 \\
XNLI & \textbf{47.1} & 42.3 & 40.9 & 42.9 & 43.3 & 42.2 & 40.5 & \textbf{44.4} & 43.6 \\
Multi-IF & 65.6 & \textbf{83.7} & 81.9 & 29.2 & 67.8 & 87.8 & \textbf{89.7} & 86.6 & 70.3 \\
\addlinespace
\multicolumn{10}{@{}l}{\textit{Cultural Knowledge}} \\
BLEnD & 65.3 & \textbf{69.8} & 66.0 & 22.8 & 68.0 & 75.1 & \textbf{80.1} & 75.1 & 71.1 \\
CulturalBench & 60.1 & 68.9 & \textbf{70.3} & 34.4 & 58.4 & \textbf{82.6} & 69.7 & 77.1 & 44.1 \\
SwitzerlandQA & 67.3 & 69.8 & 61.4 & \textbf{71.2} & 66.3 & 74.2 & \textbf{77.8} & 74.3 & 68.0 \\
\bottomrule
\end{tabular}
\end{table}

\begin{table}[!htbp]
\caption{Automatic evaluation of the SwissGov translation tasks with COMETKiwi22.}
\label{tab:translation_autoeval_swissgov}
\centering
\includegraphics[width=0.8\linewidth]{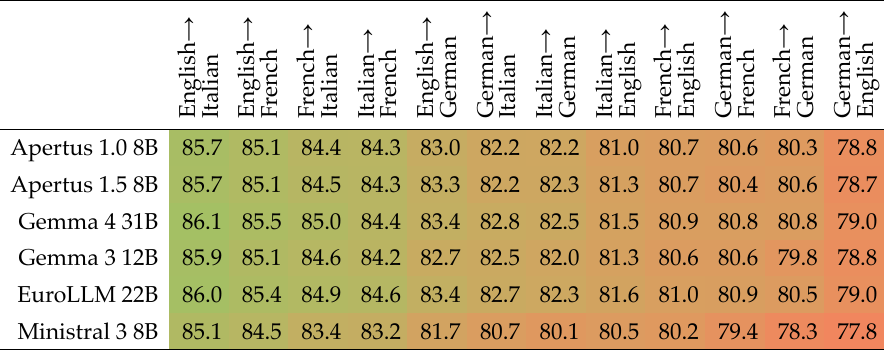}
\end{table}

\subsubsection{Machine Translation}

Being able to translate well is a major sign of a model's multilingual capability, and machine translation remains a challenging NLP task~\citep{zouhar2026translationbenchmark}. To this end, we evaluate several comparable models across many languages on WMT24++~\citep{kocmi-etal-2024-findings,deutsch-wmt24pp} and its Romansh extension~\citep{vamvas2025expandingwmt24benchmark}, WMT25~\citep{kocmi-etal-2025-findings}, and SwissGov~\citep{wastl-etal-2026-swissgov}. This evaluation pays particular attention to, but is not limited to, German, Swiss German, French, Italian, and Romansh.

\begin{table}[!htbp]
\caption{Automatic evaluation on the news domain of the WMT24, WMT24++, and WMT25 translation tasks with the ChrF metric.}
\label{tab:translation_autoeval_chrf}
\centering
\includegraphics[width=\linewidth]{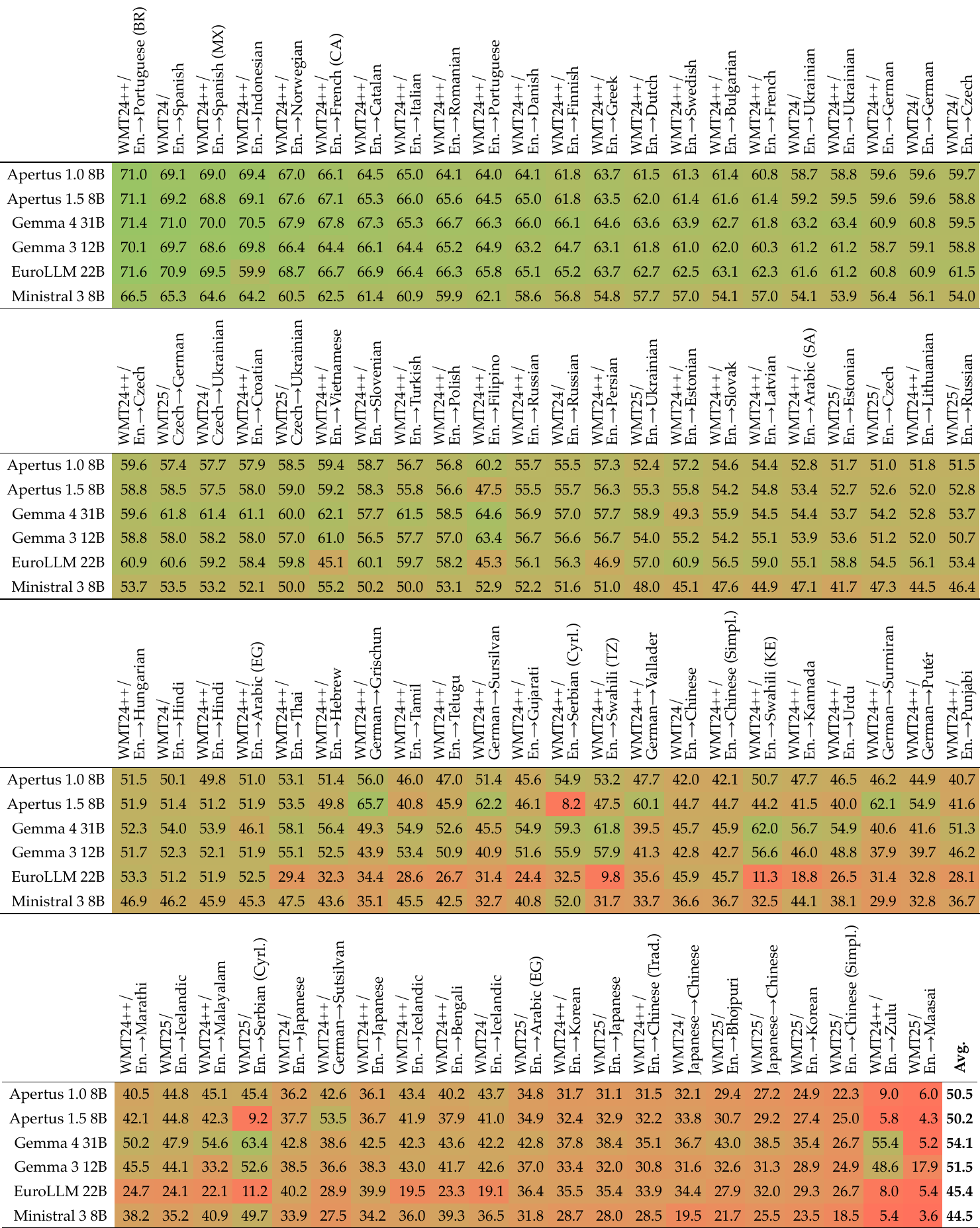}
\end{table}

\begin{table}[htbp]
    \centering
    \caption{Human evaluation results for the translation tasks on a 0\% to 100\% cESA scale.}
    \includegraphics[width=0.75\linewidth]{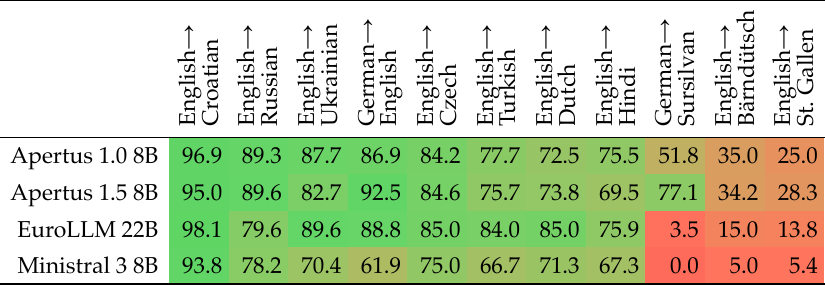}
    \label{fig:translation_humeval}
\end{table}

\paragraph{Automatic Evaluation.} The results of automatic evaluation in \autoref{tab:translation_autoeval_chrf} and \autoref{tab:translation_autoeval_swissgov}, based on the ChrF~\citep{popovic-2015-chrf} and COMETKiwi22~\citep{rei-etal-2022-cometkiwi} metrics respectively, show that the average performance of \apertus{} models correlates across languages with other models, indicating that they do not tend to have an inherent preference for a particular language. An exception is translation from German into Romansh varieties, where the targeted post-training~(\autoref{sec:variety_aware_translation}) enables \apertusonefivesmall{} to surpass comparable models by a large margin.

\paragraph{Human Evaluation.} Automatic evaluation can sometimes be misaligned with human perception of quality~\citep{lavie-etal-2025-findings,kocmi-etal-2025-findings}. To this end, we also run human evaluation of translation quality, where bilingual annotators (recruited from authors and their professional network) are blindly presented with multiple translations and rate them from 0\% to 100\%. We carry out the annotations using the cESA annotation protocol~\citep{zouhar2026contrastiveesahumanevaluation, zouhar2026pearmut} with four model outputs shown next to each other: \apertus{} 1.0 8B, \apertusonefivesmall{}, EuroLLM 22B, and Ministral 3 8B. In \autoref{fig:translation_humeval} the human evaluation confirms that \apertusonefivesmall{} avoids critical failure on Switzerland-specific language pairs. The data collected by human annotators are available publicly.\footnote{\ghrepo{swiss-ai/apertus-translate}}

\paragraph{Automatic Evaluation with SwissGov.} SwissGov is a multiparallel document-level dataset of Swiss government texts in three official national languages and English, providing a domain-specific evaluation that approximates a realistic use case for \apertus{}. For language pairs also covered by WMT, evaluation with SwissGov yields the same relative ranking, suggesting that the observed performance differences generalize beyond the news domain. 

\paragraph{Last Translation Benchmark.} We also evaluate Apertus models on the hard translation examples~\citep{zouhar2026translationbenchmark} over more than 100~languages and 3000+~texts.
Apertus v1.5 8B shows promising performance, achieving 1.32\% on LTBv1-eval (911~samples), matching the GPT-OSS-20B score and outperforming models like Claude Haiku 4.5 or Google Translate. In comparison, the Apertus v1 8B model (Apertus-8B-Instruct-2509) and the 70B model (Apertus-70B-Instruct-2509) achieve 0.77\% and 1.21\%, respectively, while the v1.5 70B model shows a worse result with 0.66\%. We believe that our bigger model is better adapted to English-centric, high-performance tasks, while the 8B model is a fitting candidate for translation. Moreover, the results give a strong signal toward more focus on multilingual and multicultural post-training, which we plan to target in future versions with greater effort.

\subsection{Image Understanding}\label{sec:image_eval}

The image suite spans around 70~unique benchmarks, from general visual question answering and OCR to spatial reasoning, remote sensing, and medical imagery; \autoref{tab:image_eval} reports a representative subset for the released checkpoints against openly released vision-language models, evaluated with our forks of LMMs-Eval~\citep{zhang2025lmmseval} and VLMEvalKit~\citep{duan2024vlmevalkit}. Decoding is greedy with a 16,384-token budget; thinking samples at temperature 0.6 and top-p 0.95 with a 32,768-token budget. Spatial-intelligence benchmarks follow the EASI protocol~\citep{easi2025}. Within the fully open group, \apertusonefivesmall{} leads on document and chart understanding (ChartQA, DocVQA) and general VQA (GQA), while Molmo2-8B leads general perception. The strongest open-weight models, Qwen3-VL and Qwen2.5-VL, stay ahead overall, most visibly on OCR-heavy tasks. Thinking lifts reasoning-heavy benchmarks by roughly 12~points (MMMU 39.8 to 52.2) but costs some extraction-style tasks about as much (ChartQA 78.7 to 66.0). \apertusonefivelarge{} adds little over the 8B on most image benchmarks.

\begin{table}[p]
\caption{Image understanding of the released checkpoints against open vision-language models, on the 33~image benchmarks of the released model card plus MM-Safety. 8Bth is the 8B model with thinking enabled. All values in percent, higher is better, best per group in bold. Dashes mark cells with no valid run. \textsuperscript{*}For MM-Safety we report the safety rate instead of the attack rate. \textsuperscript{\dag}EuroVLM answers AI2D in prose rather than the requested option letter in 74.6\% of responses, whereas every other model complies in essentially all of them; the harness credits only bare-letter replies, so this cell largely measures instruction following. A prose-tolerant extractor scores the same generations at 64.3.}
\label{tab:image_eval}
\centering
\small
\setlength{\tabcolsep}{2pt}
\setlength{\defaultaddspace}{0.3em}
\begin{tabular}{@{}l@{\hspace{4pt}}r@{\hspace{7pt}}r@{\hspace{7pt}}r@{\hspace{4pt}}r@{\hspace{3pt}}r@{\hspace{3pt}}r@{\hspace{3pt}}r@{\hspace{3.5pt}}r@{\hspace{3.5pt}}r@{\hspace{3pt}}r@{}}
\toprule
 & \multicolumn{5}{c}{\textbf{Fully open}} & \multicolumn{5}{c}{\textbf{Open-weight}} \\
\cmidrule(lr){2-6}\cmidrule(lr){7-11}
 & \multicolumn{3}{c}{\textbf{Apertus 1.5}} & \textbf{Molmo2} & \textbf{EuroVLM} & \makecell[rb]{\textbf{Qwen3-}\\\textbf{VL}} & \makecell[rb]{\textbf{Qwen2.5-}\\\textbf{VL}} & \makecell[rb]{\textbf{Gemma}\\\textbf{4}} & \textbf{Pixtral} & \textbf{Emu3} \\
\cmidrule(lr){2-4}
\textbf{Benchmark} & 8B & 8Bth & 70B & 8B & 9B & 8B & 7B & 12B & 12B & 8B \\
\midrule
\multicolumn{11}{@{}l}{\textit{General VQA and Perception}} \\
GQA & \textbf{59.2} & 52.8 & 58.4 & 56.6 & 47.9 & \textbf{61.6} & 60.9 & 38.0 & 58.8 & 60.3 \\
MMStar & 45.2 & 63.1 & 47.0 & \textbf{64.5} & 45.5 & 63.9 & 62.4 & \textbf{67.9} & 55.8 & 44.4 \\
RealWorldQA & 60.0 & 63.5 & 64.3 & \textbf{71.1} & 49.9 & 68.8 & \textbf{68.9} & 64.7 & 63.4 & 54.0 \\
VQAv2 & 76.9 & 71.5 & 75.3 & \textbf{77.7} & 71.7 & 81.8 & \textbf{83.0} & 56.5 & 78.8 & 73.7 \\
V*Bench & 63.4 & 68.6 & 60.2 & \textbf{72.8} & 56.5 & \textbf{84.8} & 80.1 & 51.3 & 52.9 & 52.4 \\
CountBench & 84.9 & 86.0 & \textbf{90.8} & 88.4 & 88.6 & \textbf{91.5} & 85.7 & \textbf{91.5} & 3.7 & 65.4 \\
\addlinespace
\multicolumn{11}{@{}l}{\textit{Documents and OCR}} \\
ChartQA & \textbf{78.7} & 66.0 & 72.8 & 73.1 & 71.2 & 85.2 & \textbf{86.0} & 51.0 & 83.8 & 69.1 \\
DocVQA & 84.8 & \textbf{86.2} & 82.6 & 69.1 & 71.3 & \textbf{95.7} & 95.1 & 80.3 & 88.9 & 72.8 \\
InfoVQA & 53.0 & \textbf{56.4} & 55.5 & 53.6 & 43.4 & \textbf{83.6} & 82.3 & 52.0 & 52.1 & 36.6 \\
\makecell[lt]{SEED-Bench-\\2-Plus} & 55.4 & 60.4 & 56.6 & \textbf{70.1} & 50.9 & \textbf{71.2} & 71.1 & 68.6 & 68.3 & 44.8 \\
OmniDocBench & 33.5 & 35.9 & \textbf{36.4} & 11.6 & 10.7 & \textbf{54.7} & 37.2 & 22.8 & 26.8 & 30.7 \\
\addlinespace
\multicolumn{11}{@{}l}{\textit{Visual Math and Logic}} \\
MathVista & 61.0 & \textbf{66.7} & 61.4 & 63.9 & 46.5 & \textbf{79.6} & 67.4 & 75.6 & 58.0 & 45.6 \\
MathVision & 25.0 & \textbf{27.4} & 23.9 & 26.4 & 14.6 & 41.6 & 25.6 & \textbf{57.1} & 17.5 & 12.8 \\
BabyVision & 10.6 & 5.2 & \textbf{11.6} & 10.6 & 9.5 & 6.4 & 12.1 & \textbf{18.6} & 13.1 & 8.0 \\
\addlinespace
\multicolumn{11}{@{}l}{\textit{STEM and Knowledge}} \\
AI2D & 74.3 & 76.8 & 74.1 & \textbf{84.5} & 15.8\textsuperscript{\dag} & \textbf{83.7} & 82.6 & 75.1 & 78.0 & 70.1 \\
MMMU & 39.8 & 52.2 & 44.2 & \textbf{52.6} & 36.8 & 53.9 & 51.9 & \textbf{63.7} & 50.0 & 29.8 \\
MMMU-Pro & 23.5 & \textbf{39.0} & 26.9 & 37.5 & 22.9 & 39.9 & 37.5 & \textbf{46.8} & 33.5 & 15.3 \\
ScienceQA & 84.7 & 90.6 & 85.1 & \textbf{94.2} & 52.5 & \textbf{92.9} & 88.8 & 86.5 & 88.8 & 72.9 \\
\addlinespace
\multicolumn{11}{@{}l}{\textit{Spatial Intelligence}} \\
EmbSpatial & 66.9 & \textbf{75.2} & 65.4 & 68.8 & 20.7 & \textbf{78.0} & 71.5 & 70.8 & 28.4 & 53.5 \\
MMSI & \textbf{32.5} & 26.4 & 28.9 & 29.3 & 25.2 & \textbf{30.5} & 27.3 & 29.4 & 24.2 & 21.8 \\
ViewSpatial & \textbf{43.6} & 43.2 & 41.2 & 41.5 & 15.4 & \textbf{41.9} & 37.3 & 38.3 & 16.1 & 21.5 \\
MindCube & \textbf{43.9} & 36.1 & 42.5 & 41.1 & 37.9 & 30.0 & 33.6 & \textbf{48.6} & 29.9 & 32.1 \\
\addlinespace
\multicolumn{11}{@{}l}{\textit{Remote Sensing}} \\
FRIEDA & \textbf{16.1} & 15.1 & 15.7 & 12.0 & 15.1 & \textbf{19.6} & 15.3 & 5.4 & 10.4 & 8.3 \\
GeoBench & 28.3 & 32.2 & 31.1 & \textbf{46.1} & 30.8 & \textbf{42.1} & 41.7 & 41.3 & 34.2 & 28.6 \\
VRSBench-VQA & 51.0 & 47.9 & \textbf{53.1} & 45.1 & 50.3 & \textbf{51.9} & 50.2 & 37.4 & 47.5 & 43.7 \\
\addlinespace
\multicolumn{11}{@{}l}{\textit{Medical Imaging}} \\
PathVQA & 68.8 & \textbf{70.7} & 62.4 & 63.4 & 61.6 & \textbf{71.0} & 65.8 & 69.5 & 64.8 & 55.0 \\
PMC-VQA & 52.7 & 51.3 & \textbf{55.3} & 48.5 & 45.0 & 54.8 & 51.3 & \textbf{58.1} & 49.4 & 45.5 \\
SLAKE & 77.4 & \textbf{79.6} & 70.9 & 63.7 & 71.4 & 77.9 & 71.4 & \textbf{80.0} & 69.7 & 52.2 \\
VQA-RAD & \textbf{68.1} & 66.5 & \textbf{68.1} & 65.3 & 67.3 & 75.7 & 74.5 & 72.5 & 71.3 & \textbf{78.5} \\
\addlinespace
\multicolumn{11}{@{}l}{\textit{Robustness and Bias}} \\
MMVP & 69.0 & 76.0 & 70.3 & \textbf{86.3} & 73.7 & 79.7 & 77.0 & \textbf{83.0} & 68.0 & 70.3 \\
VLMs-Are-Biased & 21.5 & 20.9 & 18.8 & \textbf{25.2} & 18.9 & 13.9 & \textbf{17.1} & 3.9 & 12.1 & 10.5 \\
VLMsAreBlind & \textbf{68.9} & 54.5 & 56.8 & 58.9 & 47.1 & 59.8 & 52.0 & \textbf{66.0} & 47.2 & 15.5 \\
\addlinespace
\multicolumn{11}{@{}l}{\textit{Hallucination and Safety}} \\
POPE & 86.1 & 86.0 & 86.8 & \textbf{88.7} & \textbf{88.7} & \textbf{88.8} & 87.8 & 81.8 & 85.5 & 81.9 \\
MM-Safety\textsuperscript{*} & 62.0 & 65.8 & 71.5 & \textbf{74.9} & 47.9 & 78.0 & 57.6 & \textbf{82.0} & 42.1 & 33.9 \\
\bottomrule
\end{tabular}
\end{table}

\subsection{Audio Understanding}\label{sec:audio_eval}

The audio suite centers on multilingual speech recognition, the primary target of the audio training, complemented by speech translation and by audio understanding beyond transcription, from audio and music question answering to vocal-sound classification. \autoref{tab:audio_eval} reports a representative subset against openly released audio models, namely Qwen2.5-Omni~\citep{xu2025qwen25omni}, Qwen2-Audio~\citep{chu2024qwen2audio}, and Kimi-Audio~\citep{kimiteam2025kimiaudio}, all evaluated with our fork of LMMs-Eval~\citep{zhang2025lmmseval}. It covers speech recognition on LibriSpeech~\citep{panayotov2015librispeech}, VoxPopuli~\citep{wang2021voxpopuli}, SPGISpeech~\citep{oneill2021spgispeech}, TED-LIUM~\citep{hernandez2018tedlium3}, and FLEURS~\citep{conneau2022fleurs}; speech translation on CoVoST2~\citep{wang2021covost2}; and audio understanding on MMAU~\citep{sakshi2025mmau}, MuChoMusic~\citep{weck2024muchomusic}, Clotho-AQA~\citep{lipping2022clothoaqa}, and VocalSound~\citep{gong2022vocalsound}.

We evaluate every model ourselves, on the same tasks with the same scoring and each task's instruction as the prompt, so that the values are comparable. Qwen2-Audio keeps the default of its chat template, and Kimi-Audio is prompted as in its authors' own evaluation (\ghrepo{MoonshotAI/Kimi-Audio-Evalkit}), with the instruction in the user turn and no system prompt, which its inference code (\ghrepo{MoonshotAI/Kimi-Audio}) does not accept. Qwen2.5-Omni receives the task-category prompts of its authors' examples (\ghrepo{QwenLM/Qwen2.5-Omni}) for speech recognition, speech translation, and vocal-sound classification; for audio question answering, for which its authors publish no evaluation prompt, it receives a short audio-understanding prompt, because with its default prompt it appends conversation to one-word answers. We report MMAU on its test-mini split, the split with public answers~\citep{sakshi2025mmau}. One CoVoST2 clip cannot be decoded in the pipeline of the three peers; we score their empty answers as wrong, so that every model is scored on the same samples.

On English speech recognition of short utterances, the dedicated audio models lead, by wide margins on LibriSpeech and SPGISpeech, while \apertusonefivelarge{} comes within 2.1~points of the best VoxPopuli error rate and ahead of Qwen2.5-Omni and Qwen2-Audio. On FLEURS, \apertusonefivelarge{} holds the best word error rates on Spanish and Ukrainian, and \apertusonefivesmall{} the best on Polish, where Qwen2.5-Omni reaches 87.1\%, Qwen2-Audio 98.6\%, and Kimi-Audio 65.3\%; the peers lead English, German, French, and Italian. On TED-LIUM long form, whose eight talks run 3 to 22~minutes, \apertusonefivesmall{} reaches 36.0\% and \apertusonefivelarge{} 47.3\%, against 61.8\% for Kimi-Audio, 67.7\% for Qwen2.5-Omni, and 99.6\% for Qwen2-Audio, which processes audio in 30-second windows and returns at most 26~words per talk. On CoVoST2 English-Chinese, the Qwen models lead with 44.3~BLEU for Qwen2.5-Omni and 38.1 for Qwen2-Audio, against 15.3 for \apertusonefivesmall{}, 13.1 for \apertusonefivelarge{}, and 0.3 for Kimi-Audio. On audio understanding beyond speech, \apertus{} trails the peers throughout, and its vocal-sound classification is at or below chance (16.0\% and 5.0\% against 16.7\% for six classes).

\begin{table}[t]
\caption{Audio evaluation of the released \apertus{} 8B and 70B checkpoints against open audio models. Word error rates (WER) are in percent, lower is better; accuracies are in percent and BLEU on a 0 to 100 scale, both higher is better. Best per row in bold. All models decode greedily, Qwen2-Audio with the repetition penalty~\citep{keskar2019ctrl} of 1.1 from its generation configuration, which its authors' evaluation code (\ghrepo{QwenLM/Qwen2-Audio}) also applies, and use each task's output cap; where a task sets none, all models share a cap of 4,096~tokens, which TED-LIUM long form also receives because its own cap of 256~tokens would cut off transcripts of up to 4,164~words; on that task, Qwen2.5-Omni uses PyTorch SDPA attention, whose fused kernels do not store the full attention matrix~\citep{dao2022flashattention,rabe2021selfattention}, because with the default eager attention its longest talks exceed GPU memory.}
\label{tab:audio_eval}
\centering
\small
\begin{tabular}{@{}lrrrrr@{}}
\toprule
\textbf{Benchmark} & \makecell[rb]{\textbf{\apertus{} 1.5}\\\textbf{8B}} & \makecell[rb]{\textbf{\apertus{} 1.5}\\\textbf{70B}} & \makecell[rb]{\textbf{Qwen2.5-}\\\textbf{Omni}} & \makecell[rb]{\textbf{Qwen2-}\\\textbf{Audio}} & \makecell[rb]{\textbf{Kimi-}\\\textbf{Audio}} \\
Parameters & 8B & 70B & 7B & 7B & 7B \\
\midrule
\multicolumn{6}{@{}l}{\textit{English Speech Recognition (WER, lower is better)}} \\
LibriSpeech test-clean & 6.6 & 6.2 & 2.3 & 4.2 & \textbf{1.3} \\
LibriSpeech test-other & 19.9 & 19.3 & 4.4 & 8.2 & \textbf{2.5} \\
VoxPopuli & 11.2 & 10.3 & 11.9 & 14.0 & \textbf{8.2} \\
SPGISpeech & 18.4 & 12.6 & 5.5 & 6.0 & \textbf{3.5} \\
TED-LIUM long form & \textbf{36.0} & 47.3 & 67.7 & 99.6 & 61.8 \\
\addlinespace
\multicolumn{6}{@{}l}{\textit{Multilingual Speech Recognition (FLEURS, WER, lower is better)}} \\
English & 15.2 & 14.1 & 7.4 & 8.4 & \textbf{4.4} \\
German & 16.2 & 17.8 & \textbf{8.3} & 16.8 & 23.7 \\
French & 22.2 & 21.0 & \textbf{10.6} & 13.8 & 17.7 \\
Italian & 7.6 & 7.3 & \textbf{6.8} & 12.1 & 13.1 \\
Spanish & 8.7 & \textbf{7.9} & 8.0 & 11.5 & 14.9 \\
Polish & \textbf{43.9} & 44.7 & 87.1 & 98.6 & 65.3 \\
Ukrainian & 34.9 & \textbf{31.4} & 57.6 & 99.3 & 108.2 \\
\addlinespace
\multicolumn{6}{@{}l}{\textit{Speech Translation (BLEU, higher is better)}} \\
CoVoST2 English-Chinese & 15.3 & 13.1 & \textbf{44.3} & 38.1 & 0.3 \\
\addlinespace
\multicolumn{6}{@{}l}{\textit{Audio Understanding and Classification (accuracy, higher is better)}} \\
MMAU & 55.1 & 55.9 & \textbf{67.8} & 57.7 & 66.4 \\
MuChoMusic & 51.2 & 57.3 & 62.5 & 67.6 & \textbf{68.7} \\
Clotho-AQA & 53.0 & 57.2 & \textbf{87.0} & 75.7 & 83.9 \\
VocalSound & 16.0 & 5.0 & 92.2 & \textbf{93.7} & 65.0 \\
\bottomrule
\end{tabular}
\end{table}

\subsection{Long Context}\label{sec:long_context_eval}

We evaluate long-context ability with two benchmarks: RULER~\citep{hsieh2024rulerwhatsrealcontext} measures synthetic retrieval and aggregation, while HELMET~\citep{yen2025helmetevaluatelongcontextlanguage} spans seven categories of downstream tasks.\footnote{RULER evaluation is run with \ghrepoicon{swiss-ai/evals-post-train}; HELMET with \ghrepoicon{swiss-ai/HELMET}.} \autoref{tab:long_context_eval} reports both at every stage of the \apertus{} 1.5 8B and 70B pipelines. The context extension is the decisive step: on RULER it puts \apertus{} 1.5 ahead of \apertus{} 1.0 at every length and at both sizes, and it carries the model out to 256k~tokens. Short-context SFT then undoes much of that, collapsing retrieval on long inputs, and the dedicated long-context SFT stage described in \autoref{sec:long_context_sft} restores it. RL and preference optimization add further gains on RULER even though neither stage trains on long documents, a point we return to in \autoref{sec:pretraining_retrospective}. HELMET, in the right-hand columns, shows the same collapse and the same recovery, with the preference-optimization stages strongest at every length. Through 128k the 70B pipeline stays ahead of the 8B on both benchmarks.

\begin{table*}[!t]
\caption{Long-context evaluation on RULER and HELMET, in percent, higher is better. RULER cells average 13~subtasks; HELMET cells average seven task-group scores, HELMET's own aggregation. Three HELMET settings use ROUGE in place of the reference LLM judge. Base and post-trained models are shown separately, fully open models above open-weight ones; bold marks the best value per column within a panel. ``--'' marks a length beyond the model's position ceiling.}
\label{tab:long_context_eval}
\centering
\footnotesize
\setlength{\tabcolsep}{3.5pt}
\begin{tabular}{@{}lrrrrrr@{\hspace{10pt}}rrrrr@{}}
\toprule
& \multicolumn{6}{c}{\textbf{RULER}} & \multicolumn{5}{c}{\textbf{HELMET}} \\
\cmidrule(lr){2-7} \cmidrule(l){8-12}
\textbf{Model} & \textbf{8k} & \textbf{16k} & \textbf{32k} & \textbf{64k} & \textbf{128k} & \textbf{256k} & \textbf{8k} & \textbf{16k} & \textbf{32k} & \textbf{64k} & \textbf{128k} \\
\midrule
\multicolumn{12}{@{}l}{\textbf{Base models}} \\
\multicolumn{12}{@{}l}{\quad\textit{Fully open}} \\
\apertus{} 1.0 8B & 82.0 & 75.7 & 71.2 & 55.5 & -- & -- & 49.2 & 47.2 & 44.5 & 37.3 & -- \\
\apertus{} 1.0 70B & 79.8 & 78.1 & 71.6 & 57.3 & -- & -- & 48.5 & 47.2 & 44.4 & 37.7 & -- \\
\apertus{} 1.5 8B & 85.9 & 81.3 & 76.1 & 67.5 & 61.5 & \textbf{54.8} & 45.1 & 42.1 & 40.7 & 37.6 & 36.5 \\
\apertus{} 1.5 70B & 86.8 & 85.5 & 78.8 & 71.9 & 61.8 & 51.0 & 49.6 & 49.0 & 44.2 & 41.1 & 37.8 \\
OLMo~3 7B & 85.5 & 77.8 & 70.9 & 62.4 & -- & -- & 48.7 & 47.0 & 44.6 & 39.7 & -- \\
OLMo~3 32B & \textbf{92.7} & 89.4 & 83.9 & 76.3 & -- & -- & 52.7 & 50.8 & \textbf{48.8} & 46.2 & -- \\
\addlinespace[2pt]
\multicolumn{12}{@{}l}{\quad\textit{Open-weight}} \\
Gemma-3-12B & 91.3 & 90.0 & \textbf{89.4} & \textbf{85.1} & \textbf{78.8} & -- & 50.0 & 49.3 & 47.5 & \textbf{46.6} & \textbf{44.1} \\
Qwen3-8B & 92.5 & \textbf{91.1} & 86.6 & -- & -- & -- & \textbf{52.8} & \textbf{50.9} & 48.3 & -- & -- \\
\midrule
\multicolumn{12}{@{}l}{\textbf{Post-trained models}} \\
\multicolumn{12}{@{}l}{\quad\textit{Fully open}} \\
\apertus{} 1.0 8B Instruct & 88.3 & 80.1 & 65.3 & 59.6 & -- & -- & 44.7 & 34.3 & 27.0 & 23.8 & -- \\
\apertus{} 1.0 70B Instruct & 88.5 & 85.6 & 80.6 & 69.4 & -- & -- & 51.4 & 46.4 & 39.4 & 30.9 & -- \\
\apertus{} 1.5 8B after SFT & 83.1 & 77.6 & 56.9 & 40.5 & 18.9 & 8.3 & 41.0 & 33.9 & 23.4 & 15.8 & 8.7 \\
\quad after long-context SFT & 86.8 & 82.6 & 76.3 & 65.9 & 59.5 & 44.9 & 42.9 & 36.7 & 29.2 & 23.8 & 21.1 \\
\quad after RL & 87.1 & 82.8 & 78.6 & 67.1 & 60.1 & 48.8 & 41.1 & 37.1 & 29.1 & 24.1 & 19.7 \\
\quad after DPO & 88.2 & 84.0 & 80.0 & 69.2 & 61.5 & 50.4 & 48.5 & 46.0 & 36.1 & 28.5 & 23.1 \\
\quad after SDPO & 87.4 & 85.0 & 80.7 & 68.3 & 61.1 & 50.5 & 48.3 & 45.9 & 36.3 & 28.7 & 23.1 \\
\apertus{} 1.5 70B after SFT & 90.9 & 88.5 & 79.3 & 70.5 & 54.1 & 12.0 & 46.6 & 42.8 & 38.6 & 34.4 & 28.7 \\
\quad after long-context SFT & 90.7 & 89.7 & 82.1 & 75.4 & 65.2 & 43.9 & 48.7 & 45.8 & 42.1 & 37.5 & 35.4 \\
\quad after RL & 90.5 & 89.1 & 82.2 & 75.6 & 66.0 & 45.0 & 51.7 & 47.8 & 42.5 & 36.5 & 34.7 \\
\quad after DPO & 91.7 & 90.8 & 85.3 & 77.0 & 65.9 & 45.6 & 56.1 & 54.7 & 50.0 & 45.3 & 42.1 \\
OLMo~3 7B Instruct & 58.3 & 45.9 & 36.3 & 26.1 & -- & -- & 39.6 & 35.3 & 32.1 & 26.8 & -- \\
OLMo~3.1 32B Instruct & 91.3 & 86.9 & 80.2 & 61.4 & -- & -- & 50.1 & 44.1 & 42.3 & 39.5 & -- \\
\addlinespace[2pt]
\multicolumn{12}{@{}l}{\quad\textit{Open-weight}} \\
Gemma-3-12B & 92.6 & 87.7 & 80.0 & 70.7 & 60.9 & -- & 56.2 & 51.8 & 44.4 & 38.2 & 33.0 \\
Gemma-3-27B & 94.2 & 93.3 & 91.4 & 83.7 & 65.5 & -- & 58.7 & 57.0 & 51.9 & 45.8 & 38.7 \\
Llama-3.1-8B & 93.8 & 93.4 & 87.3 & 84.6 & 78.8 & -- & 45.4 & 43.2 & 42.2 & 42.4 & 33.4 \\
Llama-3.3-70B & 95.4 & 95.2 & 94.1 & 84.7 & 56.6 & -- & 59.2 & 59.2 & 57.4 & 51.3 & 38.0 \\
Qwen3.5-9B & 96.1 & 95.9 & 96.0 & 95.8 & 94.1 & 93.6 & 59.7 & 59.9 & 59.2 & 56.9 & 55.4 \\
Qwen3.5-27B & \textbf{96.7} & \textbf{96.6} & \textbf{96.5} & \textbf{96.3} & \textbf{94.9} & \textbf{94.8} & \textbf{63.7} & \textbf{64.8} & \textbf{65.2} & \textbf{64.7} & \textbf{64.9} \\
\bottomrule
\end{tabular}
\end{table*}
\subsection{Tool Use}

We evaluate function calling on the single-turn categories of BFCL~v3, with the parser matched to the call format of the chat template (\autoref{tab:tool_use_eval}); the official leaderboard additionally includes multi-turn categories, so these numbers are not comparable to published entries. Both models improve over their predecessors, the 70B most visibly: its overall accuracy rises from 58.0 to 70.1, within 1.5~points of Llama-3.3-70B, and it repairs the parallel-call collapse of \apertus{} 1.0. The 8B remains weak on parallel calls, a known release limitation, but nearly doubles its predecessor's score on the irrelevance categories, where the declared tools do not fit the request and the model should answer in plain text instead of calling one. This is the intended effect of a supervised-finetuning design choice: tool declarations are also injected into conversations that call no tool (\autoref{sec:sft}). At launch, providers expose tool calling and thinking as mutually exclusive, a serving limitation discussed with the chat template.

\begin{table}[!htbp]
\caption{BFCL~v3 function calling by call type, over the single-turn categories alone. All values in percent, higher is better, best per group in bold. The irrelevance categories reward declining to call a tool when none applies.}
\label{tab:tool_use_eval}
\centering
\small
\setlength{\tabcolsep}{3.5pt}
\begin{tabular}{@{}lrrrrrrrrr@{}}
\toprule
 & \multicolumn{6}{c}{\textbf{Fully open}} & \multicolumn{3}{c}{\textbf{Open-weight}} \\
\cmidrule(lr){2-7}\cmidrule(lr){8-10}
 & \multicolumn{4}{c}{\textbf{\apertus{}}} & & & & & \\
\cmidrule(lr){2-5}
\textbf{Category} & \textbf{1.0} & \textbf{1.0} & \textbf{1.5} & \textbf{1.5} & \textbf{OLMo 3.1} & \textbf{EuroLLM} & \textbf{Qwen3.5} & \textbf{Llama 3.1} & \textbf{Llama 3.3} \\
Parameters & 8B & 70B & 8B & 70B & 32B & 9B & 9B & 8B & 70B \\
\midrule
Overall & 56.4 & 58.0 & 61.2 & 70.1 & \textbf{73.8} & 31.1 & 69.1 & 64.5 & \textbf{71.6} \\
\addlinespace
\multicolumn{10}{@{}l}{\textit{Single Calls}} \\
Simple & 90.3 & 91.8 & 78.8 & \textbf{93.3} & 85.3 & 79.8 & 87.0 & \textbf{94.3} & \textbf{94.3} \\
Multiple & 89.5 & 91.5 & 79.0 & \textbf{92.0} & 86.0 & 83.5 & 93.0 & 93.5 & \textbf{94.5} \\
Live simple & 61.6 & 72.9 & 60.9 & \textbf{79.8} & 62.8 & 52.7 & 69.0 & 75.2 & \textbf{82.2} \\
Live multiple & 56.4 & 64.0 & 47.3 & \textbf{72.1} & 57.7 & 48.8 & 68.8 & 70.1 & \textbf{78.6} \\
\addlinespace
\multicolumn{10}{@{}l}{\textit{Parallel Calls}} \\
Parallel & 80.0 & 24.0 & 64.0 & \textbf{87.5} & 81.5 & 0.5 & 87.0 & \textbf{89.0} & 88.5 \\
Parallel multiple & 71.0 & 9.5 & 55.5 & \textbf{82.0} & 74.0 & 0.0 & 88.0 & 88.5 & \textbf{89.5} \\
Live parallel & 56.3 & 12.5 & 31.3 & \textbf{81.3} & 50.0 & 0.0 & \textbf{81.3} & 43.8 & 68.8 \\
\addlinespace
\multicolumn{10}{@{}l}{\textit{Tool Rejection}} \\
Irrelevance & 36.3 & 50.0 & 69.6 & 67.9 & \textbf{94.2} & 41.3 & 60.4 & 44.2 & \textbf{65.4} \\
Live irrelevance & 29.0 & 46.7 & 66.4 & 45.7 & \textbf{85.3} & 32.8 & \textbf{51.9} & 30.5 & 39.5 \\
\bottomrule
\end{tabular}
\end{table}

\subsection{Security, Safety, and Interpretability}\label{sec:safety}

As \apertus{} is a fully open and massively multilingual foundation model, the safety evaluations of the \apertus{} family present a number of unique challenges, initially outlined for the \apertus{} 1.0 release~\citep{apertus2025apertusdemocratizingopencompliant}. While most current LLM literature, standards, and regulations focus on \textit{LLM deployments}, \apertus{} is an open foundation LLM, for which release risks have to be evaluated accordingly. First, any security mechanisms relying on alignment, such as refusal, can be removed by post-release training, so the intrinsic model usefulness for dangerous capabilities must be evaluated independently of alignment~\citep{DeepIgnorance2026}. Second, pre-release evaluations of models' safety are inherently distinct from performance evaluation. While benchmarks can serve as good general proxies of performance, security evaluations seek to evaluate the severity and likelihood of potential harms arising from the model release, with a single high-likelihood, high-impact risk being able to block the model release.

To ensure the released model's safety in light of those considerations, we approach model safety in three stages.

First, we design the post-training to ensure safety in non-adversarial usage scenarios. The SFT mixture carries only about 16,000~refusal-style safety samples out of 3.7 million (\autoref{sec:sft}), so safety behavior is concentrated in the alignment stages, most directly in the charter alignment of SDPO. Over-refusal is tracked as the counterweight throughout development. Of the safety suite, only BBQ, ToxiGen, and WMDP inform alignment iteration; every other safety benchmark is held out as an out-of-distribution check.

Second, we evaluate release-candidate models against standardized safety and security benchmarks. Notably, we use HarmBench Direct Requests and Human Jailbreaks (referred to as HarmBench**) as a proxy for the safety of the release candidates in non-adversarial use, thanks to good coverage of general harm categories~\citep{mazeika2024harmbench}. We use WMDP as a proxy for intrinsically dangerous knowledge, but without seeking to optimize it, due to overlap with general model capabilities~\citep{Li2024WMDP}. Finally, we use BBQ, ToxiGen, and RealToxicityPrompts*\footnote{For scalability, we randomly subsample RealToxicityPrompts and use the LLaMA-Guard3 judge model, as reported in the \apertus{} 1.0 technical report.} to measure the potential for unprompted and unexpected toxicity in English, and PolyGloToxicityPrompts to validate it in other languages~\citep{parrish-etal-2022-bbq, hartvigsen2022toxigen, RealToxicityPrompts2020, PolyglotToxicityPrompts2024}.

Finally, we perform a blocking manual evaluation of every release-candidate model and provide early access to independent third parties. As part of the manual evaluation, we notably evaluate CBRNE, dual-use, and child sexual abuse usefulness, ensuring a high coverage of risk criteria outlined in NIST-RMF GenAI 600.1~\citep{Roberts2024ArtificialIR} and ML Commons Risk 1.0~\citep{MLCommons2025AILuminate} (cf.\ \autoref{tab:framework-comparison} in \autoref{app:sec_other_frameworks} for detailed mapping). Consistent with the concern outlined above for open-weight model release, to ensure that we evaluate the intrinsic capabilities of the models rather than alignment, we create and evaluate abliterated versions of each release-candidate model~\citep{abliteration2024} and provide access to them to external evaluators. Due to the nature of evaluation of usefulness for inherently malicious use, the manual evaluation criteria per se can be useful to malicious parties and hence are not publicly released. However, more detailed information about the manual evaluation and third parties is available in \autoref{app:sec-manual-evals}, and the pre-release manual review report is available upon request.

\subsubsection{Safety Benchmark Performance}

We report safety benchmark performance in \autoref{tab:safety_eval} for the released checkpoints, their predecessors, and open-weight anchors, using the served classifier and judge models of \autoref{sec:evaluation_setup}. The alignment stages move the metrics in the intended direction for \apertus{} 1.5 vs 1.0: ToxiGen rises by 11.4\% for the 8B and 24.3\% for the 70B, BBQ improves at both sizes with the 70B reaching the best score in the comparison, and the judged toxicity of generated continuations is low throughout. WMDP rises slightly alongside general capability (2.1\% and 1.8\%  for 8B and 70B respectively).
On HarmBench, the released 8B resists direct harmful requests (1.3\% attack success) but remains vulnerable to human-written jailbreaks (31.5\%), for an average request success rate of 26.5\%. This is consistent with the main defense against malicious misuse of the \apertus{} family being capability limitations, rather than alignment, and is still an improvement of 8.7\% and 2.8\% for the 8B and 70B respectively. Finally, we evaluate \apertus{} 1.5 release candidates for multimodal jailbreak robustness, leveraging MM-SafetyBench~\citep{Liu2024MMSafetyBenchAB}, obtaining scores of 38.0\% and 28.5\% for the 8B and 70B respectively (lower is better). Comparatively, Qwen3-VL-8B, Molmo2-8B, Gemma4-12B, and EuroVLM-9B achieve 22.0\%, 25.1\%, 18.0\%, and 52.1\%, respectively, positioning the released models within the normal distribution of similarly sized open-weight models (full data in \autoref{tab:image_eval}; note that the values are inverted in the table to facilitate comparison).

\begin{table}[!htbp]
\caption{Safety evaluation of the released checkpoints against their predecessors and open-weight anchors. All values in percent; WMDP reports accuracy on hazardous-knowledge questions without a preferred direction, and is excluded from bolding.}
\label{tab:safety_eval}
\centering
\small
\setlength{\tabcolsep}{3.5pt}
\begin{tabular}{@{}lrrrrrrr@{}}
\toprule
 & \multicolumn{4}{c}{\textbf{\apertus{}}} & & & \\
\cmidrule(lr){2-5}
\textbf{Benchmark} & \textbf{1.0 8B} & \textbf{1.5 8B} & \textbf{1.0 70B} & \textbf{1.5 70B} & \textbf{OLMo 3} & \textbf{Qwen3.5} & \textbf{Llama 3.3} \\
Parameters & 8B & 8B & 70B & 70B & 7B & 9B & 70B \\
\midrule
\multicolumn{8}{@{}l}{\textit{Higher Is Better}} \\
BBQ & 65.2 & 67.6 & 68.6 & \textbf{80.5} & 73.2 & 75.3 & 72.0 \\
ToxiGen & 70.4 & 81.8 & 62.5 & 86.8 & 83.6 & 74.8 & \textbf{87.3} \\
\addlinespace
\multicolumn{8}{@{}l}{\textit{Lower Is Better}} \\
PolyGloToxicityPrompts (toxicity) & 4.7 & 7.9 & 3.7 & 7.0 & 7.0 & \textbf{3.9} & 6.3 \\
RealToxicityPrompts* (toxicity) &  0.2 & 0.5 & 0.2 & 0.5 & \textbf{0.3} & 0.5 & 1.1 \\
HarmBench** (attack success) & 35.2 & 26.5 & 31.9 & 29.1 & 21.0 & \textbf{1.6} & 39.2 \\
\addlinespace
\multicolumn{8}{@{}l}{\textit{No Preferred Direction}} \\
WMDP (accuracy) & 51.4 & 53.5 & 54.9 & 56.7 & 49.8 & 45.8 & 57.1 \\
\bottomrule
\end{tabular}
\end{table}

\subsubsection{Over-Refusal and Hallucination}\label{sec:overrefusal_hallucination}

Two additions to the evaluation suite measure failure modes adjacent to safety. We measure over-refusal, the refusal of benign requests, on the hard subset of OR-Bench~\citep{cui2025orbench} as a counterweight to the refusal-based safety benchmarks. Refusal rates rise from 3.6\% after SFT to 81.5\% after alignment at 8B and from 2.9\% to 67.6\% at 70B (\autoref{tab:lineage_full}). Most of the rise comes at the DPO stage; at 70B, RL already raises the rate to 24.7\%. A share of the benchmark's prompts is legitimately refusable, so we report this trend rather than an absolute over-refusal rate (\autoref{sec:limitations}). We measure hallucination with HalluLens~\citep{bang2025hallulens}. Our judge is a locally served Qwen3.5-27B, whereas the original benchmark uses a Llama-3.1-405B, so absolute scores are not comparable with the official leaderboard; a cross-check with a Llama-3.3-70B judge moves the rates by at most two points. \autoref{tab:hallulens_eval} shows a split result. Where the failure is answering about things that do not exist, \apertus{} 1.5 improves sharply: abstention on fabricated entities more than triples at 8B, and long-form generation becomes both more precise and more willing to abstain, gaining six points of F1 at 8B and two at 70B. Short-form hallucination on questions about real entities does not improve. Both releases attempt an answer on more than eight of 10~questions and hallucinate on roughly six to seven of them, slightly more than \apertus{} 1.0 at 70B scale. Reducing this without inflating over-refusal remains open for a future release.

\begin{table}[!htbp]
\caption{HalluLens hallucination suite of the released checkpoints. All values in percent. The PreciseWiki rates partition all questions, so hallucination, correct, and refusal sum to roughly 100; the abstention rows reward declining to answer questions about nonexistent entities. The best value per row and model size is in bold on rows with a direction. The judge is a served Qwen3.5-27B, whereas the original benchmark uses a Llama-3.1-405B judge, so scores are not comparable with the official leaderboard.}
\label{tab:hallulens_eval}
\centering
\small
\setlength{\tabcolsep}{5pt}
\begin{tabular}{@{}lrrrr@{}}
\toprule
 & \multicolumn{2}{c}{\textbf{8B}} & \multicolumn{2}{c}{\textbf{70B}} \\
\cmidrule(lr){2-3}\cmidrule(lr){4-5}
\textbf{Metric} & \textbf{\apertus{} 1.0} & \textbf{\apertus{} 1.5} & \textbf{\apertus{} 1.0} & \textbf{\apertus{} 1.5} \\
\midrule
\multicolumn{5}{@{}l}{\textit{PreciseWiki (short-form factoid questions)}} \\
Hallucination $\downarrow$ & 71.9 & \textbf{71.5} & \textbf{54.0} & 59.6 \\
Correct $\uparrow$ & \textbf{18.8} & 15.2 & \textbf{27.4} & 23.3 \\
Refusal & 9.4 & 13.3 & 18.7 & 17.1 \\
\addlinespace
\multicolumn{5}{@{}l}{\textit{Nonexistent-Entity Abstention}} \\
Fabricated entities $\uparrow$ & 27.0 & \textbf{87.1} & 61.5 & \textbf{87.9} \\
Mixed entities $\uparrow$ & 18.5 & \textbf{38.2} & 15.7 & \textbf{29.0} \\
\addlinespace
\multicolumn{5}{@{}l}{\textit{LongWiki (long-form generation)}} \\
Precision $\uparrow$ & 25.5 & \textbf{33.2} & 31.1 & \textbf{35.9} \\
Recall $\uparrow$ & 17.7 & \textbf{23.0} & 27.1 & \textbf{28.9} \\
F1 $\uparrow$ & 19.5 & \textbf{25.6} & 27.6 & \textbf{29.8} \\
Abstained & 1.9 & 27.1 & 5.2 & 10.6 \\
\bottomrule
\end{tabular}
\end{table}

\subsubsection{Interpretability}
\label{sec:interpretability}

We do not claim to understand the internal behavior of the released models, and the results above make the gap explicit: the preceding benchmarks tell us about the frequency of failures, not \textit{why} they arise. To be open about potential risks, we treat interpretability as one safeguard in a ``defense-in-depth'' strategy~\citep{ISRSAA2026}, alongside evaluation, alignment, and manual safety review. 

With this release, we launch \apertus{} Claritas\footnote{\url{https://apertus-claritas.org}}, a public platform that collects interpretability studies on the \apertus{} models, including exploratory and negative results. Importantly, open weights and intermediate checkpoints let us study \textit{when} an unsafe behavior forms~\citep{biderman2026position}, rather than only after training. The studies to date use the \apertus{} 1.0 and 1.5 checkpoints; three of their findings feed into this release.

\paragraph{Exploding Activations.} Hallucination probes trained on the activations of the 8B initially fail where the same probes succeed on Llama-3.1-8B~\citep{obeso2025realtimedetectionhallucinatedentities}. The cause is activation norms that grow roughly a hundredfold toward the deeper layers. With layer normalization and precision fixes, the probe reaches an AUC of 0.90 (\ghrepo{swiss-ai/feature-probes}). The diagnosis motivates revisiting how the activation-function parameters are regularized in continued pretraining (\autoref{sec:optimizer_recipe}).

\paragraph{Degeneration Probes.} Linear probes on the hidden states of the 8B detect repetitive collapse before it surfaces in the generated text, most reliably from layer 15 onward (\ghrepo{swiss-ai/degeneration-probe}). This points toward intervening during generation rather than diagnosing afterward, and complements the degeneration rates tracked across the post-training stages (\autoref{sec:post_training_lineage}).

\paragraph{Multilingual Geometry.} Ablations on the translation directions of the 8B show that low-resource languages lean on the representations of related high-resource languages: for example, suppressing the high-resource direction degrades Galician and Afrikaans translation, while the reverse intervention has almost no effect (\ghrepo{swiss-ai/multilingual-geometry}). This gives a mechanism behind the coverage picture of \autoref{sec:multilingual_eval} and suggests that long-tail quality needs more than additional data.

Further platform studies cover error-predicting probes with calibrated inference-time steering (\ghrepo{swiss-ai/apertus-probes})~\citep{mera2025to}, emotion persona representations~\citep{vanderben2026modelshappinessemotionvectors}, and bias auditing of latent generation paths~\citep{pelossi2026exposingunsaidvisualizinghidden}.

\subsection{Qualitative Spot-Testing}

Hand testing during development repeatedly surfaces behavior that the benchmark suite misses, including degeneration on out-of-distribution prompts and weak multi-turn instruction following; this section documents that practice and its findings. User reports on \apertus{} 1.0 concentrated on a failure mode the benchmarks had not measured: degeneration in long conversations, particularly under language switching. Hand testing of \apertus{} 1.5 adds identity confusion, where the model misattributes itself to another provider in some languages (\autoref{sec:limitations}). The SFT mixture references the \apertus{} identity (\autoref{sec:sft}) but contains no dedicated identity data, and occasional misattribution appears, plausibly from model-generated content in the web-scale pretraining data. The degeneration reports feed back into the evaluation suite as an automated degeneration metric, an n-gram repetition ratio computed on the generations of a small set of benchmarks of varying difficulty. Two further practices support it: every release candidate passes an automated compatibility suite of roughly 65~checks before distribution, covering prompt and begin-of-sequence invariants, degeneration probes, multimodal inputs, reasoning-channel separation, and the tool-calling and streaming paths; and we bisect behavioral regressions by serving every post-training stage side by side and comparing them on the same prompts.

\clearpage
\section{Infrastructure, Scaling, and Efficiency}\label{sec:infrastructure}

Whereas the \apertus{} 1.0 report documents the Alps Research Infrastructure in greater detail, this chapter focuses on what the 1.5 campaign adds. The introduction of new modalities and an expanded vocabulary brings additional infrastructure costs, while asynchronous reinforcement learning introduces a new class of workload. These runs share a production system whose scheduling environment, operating system, and storage infrastructure continue to evolve over the course of the project. We describe the platform and its operation (\autoref{sec:infra_platform}), the throughput and interruption records of the training runs (\autoref{sec:training_run_performance}), their energy footprint (\autoref{sec:energy_carbon}), and the engineering approaches used to address the challenges encountered (\autoref{sec:engineering_challenges}).

\subsection{Infrastructure}\label{sec:infra_platform}

Both models are trained on the Alps supercomputer at CSCS using a flexible reservation of GH200 nodes that grows from approximately 150 to 600~nodes as production ramps up. This allocation is shared across the \apertus{} 1.5 training campaign and related project workloads, including preparatory work for subsequent post-training stages. The production runs use 256~nodes for the 8B model and 512~nodes for the 70B model.

Alps, being a shared production system serving multiple users and projects, must also accommodate unrelated workloads alongside the \apertus{} campaign. For approximately two weeks in the middle of the project, one such full-scale workload requires the entire machine. During this period, training continues in the available night and weekend reservation windows, with schedules adjusted at short notice to match system availability.

All training jobs run in containers based on NGC images and integrated with the Alps software and networking environment through site-provided mechanisms. Library versions are pinned where required, and project-specific patches are applied on top of the validated software stack. During the project, the cluster operating system and GPU driver are also upgraded through staged rollouts: changes are first deployed to a small test partition and subsequently rolled out cluster-wide.

The cluster has undergone no major changes since \apertus{} 1.0 and is documented in detail there~\citep{apertus2025apertusdemocratizingopencompliant}.
\subsection{Full Training Run Performance}\label{sec:training_run_performance}

CSCS engineering teams contribute specialized expertise in analyzing and optimizing the performance of the Apertus workloads at scale. During the pretraining preparatory phase, they profile the workloads across the system and apply targeted tuning to its network, storage, memory, and compute components; this work continues as the runs scale up and new bottlenecks emerge. In parallel, CSCS and the Apertus core team jointly tune the 70B pretraining configuration, including sequence parallelism, virtual pipelining, and larger gradient-communication buckets. Together with node-level memory and clock tuning, these optimizations achieve roughly 750~tokens per second per GPU at 256~nodes, on par with the \apertus{} 1.0 70B run despite the doubled sequence length and enlarged vocabulary. Scaling from 256 to 512~nodes retains roughly 90\% parallel efficiency.

\autoref{tab:run_performance} reports the measured performance of the production phases, aggregated from the logged run segments; each SLURM job opens a new segment, so the segment count records how often scheduling windows and failures partition a phase. The 8B sustains roughly 4,900 to 5,500~tokens per second per GPU (270 to 300~TFLOPs per GPU, a model FLOPs utilization (MFU) of 0.27 to 0.30 against the nominal 989~TFLOPs bf16 peak of the GH200, which understates efficiency on Alps, where each GH200 module runs under a 660~W power cap (\autoref{sec:energy_carbon}) and cannot reach that peak), and the 70B 430 to 710 (0.20 to 0.33). The long-context extension pays the expected attention cost: throughput decreases with each doubling of the context window, although cross-document attention masking partially alleviates this degradation. Sustained rates overstate delivered progress: dividing consumed tokens by segment wall time gives an effective rate roughly 30\% lower for the 70B, the gap consisting of job startup, checkpoint saving and loading, and stalls before failures.

\begin{table}[!htbp]
\caption{Measured performance of the production training phases, aggregated from the logged wandb run segments. Tokens are per phase, from the differences of the cumulative counter; throughput and TFLOPs per GPU are means over the longest segment of each phase; wall time sums the process time of all segments of the phase.}
\label{tab:run_performance}
\centering
\small
\setlength{\tabcolsep}{5pt}
\begin{tabular}{@{}llrrrrr@{}}
\toprule
\textbf{Phase} & \textbf{Nodes} & \textbf{Tokens} & \textbf{Tokens/s/GPU} & \textbf{TFLOPs/GPU} & \textbf{Segments} & \textbf{Wall (h)} \\
\midrule
8B stage 1 & 256 & 0.81T & 5,474 & 300 & 11 & 44 \\
8B stage 2 & 256 & 2.80T & 4,899 & 269 & 19 & 155 \\
8B stage 3 & 256 & 0.40T & 5,438 & 298 & 5 & 21 \\
\addlinespace
8B extension 32k & 256 & 0.06T & 3,510 & 260 & 1 & 6 \\
8B extension 65k & 256 & 0.06T & 2,671 & 267 & 1 & 7 \\
8B extension 131k & 256 & 0.06T & 1,449 & 220 & 2 & 12 \\
8B extension 262k & 256 & 0.06T & 712 & 181 & 2 & 20 \\
\addlinespace
70B stage 1 & 512 & 0.40T & 713 & 325 & 21 & 158 \\
70B stage 2 & 512 & 1.21T & 562 & 257 & 65 & 463 \\
70B stage 3 & 512 & 0.40T & 428 & 195 & 15 & 113 \\
\addlinespace
70B extension 32k & 512 & 0.06T & 422 & 233 & 3 & 18 \\
70B extension 65k & 512 & 0.06T & 365 & 249 & 2 & 24 \\
70B extension 131k & 512 & 0.05T & 219 & 206 & 4 & 35 \\
70B extension 262k & 512 & 0.05T & 119 & 166 & 20 & 56 \\
\bottomrule
\end{tabular}
\end{table}

\subsection{Energy Consumption}\label{sec:energy_carbon}

We estimate the energy of the training runs with the same method as the Apertus 1.0 report: measured GPU-hours times an assumed draw of 560~W per GH200 module, below the configured 660~W power cap~\citep{apertus2025apertusdemocratizingopencompliant}. The logged segments of \autoref{tab:run_performance} total roughly 225,000~GPU-hours for the 8B continued-pretraining stages, 1.50~million for the 70B continued-pretraining stages, and the remainder for the long-context extensions of both models, together roughly 2~million GPU-hours or 1.1~GWh, against roughly 5~GWh for the from-scratch Apertus~1.0 pretraining. The estimate covers the logged production segments only; it excludes the phases missing from \autoref{tab:run_performance}, post-training, evaluation, ablations, and idle reservation time. During the whole project duration, 9~million~GPU-hours have been allocated, which translates to an estimate of 5~GWh, excluding the idle times and the resources consumed by the other projects within the Swiss AI initiative that have synergies with the Apertus project but were not part of the training campaign. Alps draws on carbon-free hydropower, and CSCS cools the infrastructure with lake water in a closed cycle.

\subsection{Engineering Challenges and Solutions}\label{sec:engineering_challenges}

\paragraph{Software Environment and Node Validation.}

The first weeks were characterized by further work on tuning the software environment for production at scale. During this phase, multi-node jobs exposed two interacting issues: a system-level network library affecting collective operations during checkpointing, and rendezvous failures in the campaign-specific container environment following an update to the system software stack. To reduce this coupling, we evolved the container environment to incorporate the required network components directly where appropriate. This deliberately trades some of the low-friction integration provided by the CSCS software environment for greater control over the campaign-specific stack, making long-running production workloads less sensitive to system upgrades.

We also incorporated into every job preamble a node-vetting tool developed by CSCS and made available to all users. Before training starts, it validates the execution environment, GPU health, and intra- and inter-node communication, allowing unsuitable nodes or configurations to be detected before they affect a distributed run. Finally, we hardened just-in-time (JIT) compilation for large-scale execution by placing compilation caches per process on node-local storage. This avoids cross-process and cross-node contention and ensures that generated kernels remain consistent with the software environment in which they are compiled.

\paragraph{Communication at Scale.}

Two performance issues encountered during the \apertus{} 1.5 campaign have close analogues in the \apertus{} 1.0 experience, allowing previous diagnostic work to accelerate their resolution. In the first case, throughput dropped by roughly an order of magnitude at 256~nodes, but only after restoring from a checkpoint. CSCS engineers traced this distinct failure mode to a memory-registration cache limit in the network stack; the corresponding mitigation was subsequently incorporated into the shared software environment and documented publicly. At scales beyond 256~nodes, we additionally enabled an alternative rendezvous protocol whose implementation had been corrected by the interconnect vendor (HPE) during the \apertus{} 1.0 campaign, avoiding a separate collapse in collective communication performance at large scale.

\paragraph{Storage Capacity and Data Placement.}

Storage capacity and data placement are among the first operational constraints of the runs. Preparing the multimodal corpus required substantially more space than initially available within the project allocation, leading to a coordinated cleanup of existing data and an increase in storage quotas before preprocessing could proceed at full scale. As production matured, data was placed according to its I/O characteristics: checkpoints were written to storage suited to large sequential transfers, while tokenized training datasets resided on an IOPS-optimized tier better matched to the fine-grained access patterns of training. Checkpoint size is dominated by optimizer state, which accounts for roughly 90\% of the footprint: for the 70B model, removing optimizer state reduces a checkpoint from approximately 1.2~TB to 132~GB. We therefore retain full optimizer state only at phase boundaries and prune it from intermediate checkpoints.

The campaign also reinforces the importance of matching storage tiers and scratch policies to access patterns and retention needs. Tokenized datasets were initially moved to backed-up disk storage to avoid scratch-retention limits, but the additional metadata activity introduced by the backup process became visible as training throughput variability at scale. Frequently accessed training data was therefore placed on SSD-backed scratch storage using the recommended striping policy, which distributes file blocks across multiple storage targets and avoids concentrating large-file traffic on a single server. The same consideration applies to all large objects accessed concurrently by distributed jobs: at 512-node startup, insufficiently striped container images and checkpoints can concentrate I/O on individual storage servers, a pattern clearly visible in the storage-system telemetry. During the campaign, CSCS also introduced a project-scoped storage tier combining scratch-like performance with longer-lived data retention, providing a better fit for such workloads.

\paragraph{Network Safeguards for Data Ingestion.}

Collecting web-scale multimodal data from the Internet on a large supercomputing infrastructure introduces network and security considerations that differ substantially from those of conventional training workloads. Many permissively licensed image datasets provide URL manifests rather than the image objects themselves. Corpus construction therefore requires dereferencing large numbers of URLs across a very broad set of Internet domains, producing highly distributed outbound HTTP(S) traffic with characteristics similar to large-scale web crawling.

A small fraction of these destinations are present on security blacklists. During early ingestion runs, requests to such endpoints repeatedly triggered CSCS security monitoring, while retries against denied destinations eventually caused a shared outbound IP address to be blacklisted by Internet Service Providers. Because Internet egress is shared, such an event can affect not only the \apertus{} campaign but also the other unrelated scientific workloads running concurrently on Alps. Filtering against a single widely used hosts list proved insufficient, and ingestion was paused while we developed a more robust filtering pipeline together with the CSCS network and security teams.

For each ingestion run, the pipeline retrieves fresh copies of multiple public threat-intelligence feeds, including FireHOL IP sets, URLhaus, Spamhaus, and additional feeds used by the CSCS monitoring infrastructure. These are compiled into IP-network, domain, and URL access-control lists consumed by a local web proxy.\footnote{Available at \ghrepo{swiss-ai/multimodal-data} under \texttt{download/blacklist}.} All downloader traffic passes through the proxy, including requests generated by HTTP redirects, and destination hostnames are resolved before the resulting addresses are checked against the IP-based blacklists. Because HTTPS path information is not visible to the proxy without TLS interception, URL-level HTTPS indicators are conservatively promoted to host-level blocks. In the first production run, the resulting policy rejected fewer than 0.01\% of candidate downloads, indicating that a very small fraction of destinations accounts for the relevant security risk.

\paragraph{Reinforcement Learning Infrastructure.}

Reinforcement learning introduces a distinct infrastructure workload. This stage runs on our fork of \texttt{verl} with a fully asynchronous trainer. Enabling the rollout stack on the ARM-based Grace Hopper superchip (GH200) platform and the Alps interconnect requires upstream contributions across the inference engines, training framework, and attention kernels. At scale, two performance bottlenecks become dominant. The first is the distribution of updated model weights from the trainer to the inference workers, which we addressed by porting a high-performance transfer engine to the interconnect's \texttt{libfabric} interface in close collaboration with CSCS engineering teams. The second is the loading of sharded checkpoints from the parallel filesystem, whose aggregate I/O becomes increasingly significant as the number of workers grows.

Reward environments are decoupled from the GPU compute nodes and run as CPU-only services on an on-premises Kubernetes cluster built on commodity hardware. The code sandbox executes its test harnesses within the Kubernetes pods, keeping this workload outside the GPU allocation and isolating it from the training infrastructure. End-to-end rollout throughput comes within approximately 5\% of comparable open models executed on the same software stack. The remaining difference is attributed primarily to the implementation of the xIELU activation function.

\clearpage
\section{Responsible Release and Governance}\label{sec:governance}

This chapter documents the release: the license and the gated distribution (\autoref{sec:license}), the known limitations of the models (\autoref{sec:limitations}), what the openness of weights, data, recipes, and values enables and what it risks (\autoref{sec:broader_impact}), and the external evaluation and community work around the release (\autoref{sec:community}). We describe practices and published documents and make no claim about their legal sufficiency.

\subsection{License and Acceptable Use}\label{sec:license}

We release model weights and training and evaluation code under the Apache~2.0 license. All training data is fully available: every dataset is public, and \autoref{sec:pretraining_data} and the legal spreadsheet identify each one in the version we use. Third-party data can be obtained from its original providers; we also publish the SFT mixture as a single dataset, with each component retaining its source license. Datasets we construct ourselves, such as the instruction-following prompt sets and preference pairs in \autoref{sec:post_training}, are published on Hugging Face under open licenses specified in the corresponding dataset cards. We maintain the license, acceptable-use policy, and privacy policy in \ghrepo{swiss-ai/apertus-legal}. The models are distributed on Hugging Face behind a gated download that records assent to the acceptable-use policy.

\textbf{Transparency Documentation.} We publish two documents prepared for the EU framework for general-purpose AI models: the Public Summary of Training Content, following the European Commission's template, and a Code of Practice document covering transparency and copyright policy. We publish the Code of Practice voluntarily, and neither document asserts that any obligation is satisfied. Both are versioned in \ghrepo{swiss-ai/apertus-legal}; the legal spreadsheet of \autoref{sec:data_compliance} is versioned in \ghrepo{swiss-ai/apertus-data-transparency}.

\subsection{Limitations}\label{sec:limitations}

\textbf{Factual Reliability.} The models hallucinate on short-form factual questions at the rate of their predecessors: on PreciseWiki they attempt an answer on more than eight of 10~questions and hallucinate on roughly six to seven of them, slightly more than \apertus{} 1.0 at 70B scale (\autoref{tab:hallulens_eval}). The gains of this release lie in recognizing questions about things that do not exist and in long-form precision, not in factoid recall.

\textbf{Over-Refusal.} The alignment stages make the models cautious, and prompts that merely resemble harmful ones are often declined; the effect is visible across the post-training lineage (\autoref{sec:post_training_lineage}) and in manual testing. On the hard subset of OR-Bench, refusal rates rise sharply through the alignment stages (\autoref{sec:overrefusal_hallucination}); because a share of its prompts is legitimately refusable, this shows the trend but not an absolute over-refusal rate. Reducing the caution without weakening the safety behavior remains open.

\textbf{Language Coverage.} Multilinguality is the models' strongest relative position, but not uniformly so. XNLI regresses against \apertus{} 1.0 at both scales (\autoref{tab:multilingual_8b}), and the instruction-following and reasoning data concentrate on a few dozen languages, so quality in the long tail of the pretraining languages is not evaluated.

\textbf{Multimodal Scope.} The models understand images and audio but do not generate them. The 70B trails the 8B on several multimodal benchmarks because its run stops before consuming the full multimodal mixture and its inherited learning rate leaves it less room to adapt (\autoref{sec:pretraining_retrospective}). The audio training data skews toward parliamentary and political speech (\autoref{sec:pretraining_data}), and for image and audio data we rely on the curation and filtering of the dataset creators (\autoref{sec:data_compliance}).

\textbf{Grounding.} Preference alignment improves most multimodal capability groups, among them spatial reasoning, medical and document understanding, and visual math, but systematically degrades referring-expression grounding, by about 17~points across the RefCOCO family. The diagnosis is precise: the output format remains valid and the model still locates the object, but the predicted boxes drift systematically looser, a regression in extent calibration. We accept this trade-off for the release; we note grounding-aware rewards in alignment for future iterations.

\textbf{Identity.} The post-trained models sometimes misstate their identity, claiming to be created by other providers, particularly in lower-resource languages; \apertus{} 1.0 did not show this behavior. We check the post-training data carefully: it conveys no other provider's identity. This points at subtler inheritance: a model can adopt a teacher's identity from distillation data with all identity markers removed~\citep{zhong2026identitytransfer}, and behavioral traits transmit through hidden signals in data more generally~\citep{cloud2026subliminal}, so model-generated content in the web corpus and the openly released reasoning traces in our mixture remain plausible carriers. Instilling a robust self-identity despite such data is an open problem.

\textbf{Tool Calling and Thinking.} Tool calling and thinking do not combine in the released models. The chat template supports tool calls inside the reasoning span (\autoref{fig:chat_template_example}), but the supervised mixture contains no conversation that both reasons and calls a tool, and we exclude tool-use prompts from the reinforcement-learning run (\autoref{sec:reasoning_rl}). The models behave accordingly: with thinking enabled and the reasoning span forced, they produce no tool calls; with thinking enabled but not forced, they tend to skip the reasoning span and answer as in non-thinking mode. At launch the serving stacks additionally mis-parse tool calls emitted under thinking, dropping the call terminator. Inference providers therefore expose the two features as mutually exclusive. More generally, the post-training of this release is not designed for agentic use: tool-use behavior comes from supervised finetuning alone, we exclude tool-use prompts from reinforcement learning, and we do not evaluate multi-turn tool use (\autoref{tab:tool_use_eval} covers the single-turn categories of BFCL). The 8B further remains weak on parallel calls, often producing a single call where several are requested.

\subsection{Broader Impact}\label{sec:broader_impact}

\apertus{} 1.5 releases the weights, the data recipe, the training and evaluation code, and the charter its alignment follows, and trains only on publicly available data (\autoref{sec:pretraining_data}). We enumerate every dataset, its license, and its filtering, and we publish the data ablations, so the training data and its processing can be inspected in full. The recipes of \autoref{sec:pretraining_recipe} and \autoref{sec:post_training} are the ones we ran, so institutions that cannot train such models from scratch can continue, finetune, or distill them on their own infrastructure. The models can be deployed without dependence on a vendor, which matters for public administrations, regulated sectors, and research that needs a model whose behavior is fixed and documented.

Openness extends to the values to which models are aligned. For most released models, the alignment data and objectives are undisclosed, so users observe a behavior but not the intent behind it. The \apertus{} charter, introduced with \apertus{} 1.0~\citep{apertus2025apertusdemocratizingopencompliant}, states the principles the models are aligned to; the charter-alignment stage of \autoref{sec:sdpo} trains the released 8B against it with a published recipe and judge prompts (\autoref{app:training_configurations}); and the charter-alignment score of \autoref{sec:post_training_lineage} measures how far each released checkpoint follows it. A deployer who disagrees with a principle can see what to change, and retrain.

Roughly 40\% of the pretraining data of \apertus{} 1.0 is non-English, the post-training data of this release adds multilingual instruction data, and the evaluations of \autoref{sec:multilingual_eval} place the models at their strongest relative position on multilingual and cultural benchmarks, including the national languages of Switzerland and Romansh (\autoref{sec:variety_aware_translation}). For communities whose languages are underrepresented in commercial models, an open model that can be adapted further is a direct benefit, within the coverage limits of \autoref{sec:limitations}.

We keep the data measures of \apertus{} 1.0: we respect opt-outs retroactively, use no paywalled or private sources, remove personally identifiable information and toxic text, train with the Goldfish objective to limit verbatim memorization, and publish the provenance of every dataset (\autoref{sec:data_compliance}, \autoref{sec:optimizer_recipe}).

Open weights can be finetuned to remove the safety behavior of \autoref{sec:safety}, and released models can be misused at scale. We address this with alignment trained into the released checkpoints, the acceptable-use policy (\autoref{sec:license}), the data-protection request channel, and pre-release evaluation by external organizations (\autoref{sec:community}); \autoref{sec:limitations} reports the measured safety limitations.
\subsection{Community and External Evaluation}\label{sec:community}

Beyond releasing weights, data, and recipes (\autoref{sec:broader_impact}), we open the development process to outside contributors: an outreach and community function connects external contributors and deployment engineers with the development process and collects their findings to improve model compatibility. \citet{smith2026reimagining} argue that inclusive design and capacity building are conditions for responsible openness. The public website\footnote{\url{https://apertus-ai.org}} includes the charter, usage guides, benchmark resources, model documentation, and links to downstream projects; newsletters, social media, and multilingual media monitoring complement it. Direct involvement of the engineering team has included presentations to science and industry, round-tables with companies, support of hackathons, and hands-on workshops.

One month before the Apertus 1.5 release, in addition to the third-party safety evaluation of \autoref{sec:safety}, 15~organizations, represented by roughly 50~engineers and experts, evaluated the 8B and 70B weights for deployment readiness: quantization, alignment, multimodal input, tokenizer distribution, and general compatibility. Their feedback led to quantizations and community builds for several inference frameworks.

The pre-release surfaced problems that internal testing had missed, and we thank Infomaniak and OnPrem.ai in particular. Infomaniak contributed a vLLM integration that runs the image and audio tokenizers on the GPU. Our review of it found that encoding several media items as one zero-padded batch changes their codes, because neither tokenizer satisfies $\operatorname{crop}(T(\operatorname{pad}(x))) = T(x)$, so the integration now encodes each item separately. OnPrem.ai traced a tool-calling failure in the 70B pre-release checkpoint, where a faulty generation configuration let the model continue past a complete tool call and fabricate the tool's result, and confirmed the corrected configuration in the release candidate. Their quantization studies showed that 8-bit (FP8) weights and activations are near-lossless. 4-bit NVFP4 in the attention layers of the 8B model produces NaN logits, which they attribute to the quadratic term of xIELU amplifying quantization noise. Keeping attention at FP8 and quantizing only the MLP to NVFP4 fits the 70B model into 49~GiB, so it runs on a single 96~GB GPU with more than 200k~tokens of context.

We also measured the serving performance of the 70B model at five inference providers and on our own deployment with a single run of the open-source tool \href{https://vllm-project.github.io/guidellm/main/}{GuideLLM} from CSCS on 16 September 2026. \autoref{tab:apertus-performance} lists latency, throughput, and time to first token, with providers anonymized; requests originate at CSCS, which gives our own deployment a network advantage.

\begin{table}[t]
\caption{Serving performance of \apertusonefivelarge{} on a simulated workload, by anonymized inference provider. Latency in seconds and TTFT (time to first token) in milliseconds, lower is better; throughput in tokens per second, higher is better.}
\label{tab:apertus-performance}
\centering
\small
\begin{tabular}{@{}lrrrr@{}}
\toprule
 & & & \multicolumn{2}{c}{\textbf{Throughput (tokens/s)}} \\
\cmidrule(lr){4-5}
\textbf{Provider} & \textbf{Latency (s)} & \textbf{TTFT (ms)} & \textbf{Input} & \textbf{Output} \\
\midrule
P0 (ours) & 1.90 & 103.6 & 179.7 & 67.9 \\
P3 & 2.09 & 87.9 & 157.6 & 61.5 \\
P5 & 2.18 & 171.2 & 150.9 & 58.9 \\
P1 & 2.31 & 469.8 & 143.6 & 56.4 \\
P4 & 3.63 & 81.6 & 92.8 & 35.3 \\
P2 & 4.30 & 223.6 & 79.5 & 29.9 \\
\bottomrule
\end{tabular}
\end{table}

For replies of the same length, latency varies by a factor of 2.3 between deployments (1.9 to 4.3~s) and time to first token by a factor of almost six (82 to 470~ms). Comparing answer quality across providers proved harder than measuring speed: most of our benchmarks score answers by log-likelihood, which requires a completions endpoint that several providers and our own gateway do not expose, and rate limits slow the remaining generation-based tests. Therefore, a quality comparison between providers remains open.

\clearpage
\section{Conclusion}\label{sec:conclusion}

\apertus{} 1.5 extends the \apertus{} 1.0 models from text to image and audio understanding through continued pretraining on 4T and 2T additional tokens, extends the context window fourfold to 262,144~tokens, and introduces the second generation of our post-training pipeline, which adds a thinking mode, more reliable instruction following, and more effective tool calling. The released models improve over their predecessors on nearly every benchmark; among fully open models of its class, the 8B leads on knowledge, instruction following, and cultural benchmarks, and both models stand strongest relative to their peers on multilingual evaluations. Thinking lifts the 8B from 51.2 to 71.4 on MATH-500; its instruction following reaches 87.1 on IFEval and 83.7 on Multi-IF, above every model we compare against; single-turn function calling on the 70B rises from 58.0 to 70.1, within 1.5~points of Llama-3.3-70B; and among fully open multimodal models the 8B leads document and chart understanding and stays ahead on long-form and lower-resource speech recognition, where dedicated audio models degrade.

The second generation of the post-training pipeline also produced most of the lessons of this release. Adding modalities through the tokenizer, with the language-model head restricted to text, upgraded the models without an architectural change. A dedicated long-context stage after supervised finetuning preserved the context extension. Filtering reasoning data by what the model already solves, and at 70B scale preceding online preference optimization with an offline stage, were decisive for the post-trained models. The 70B run, stopped before it consumed its multimodal mixture, showed that multimodal capability follows exposure.

The next generation of \apertus{} starts from these recipes, with the remaining gaps of \autoref{sec:limitations}, among them short-form hallucination and the combination of tool calling with thinking, as its first targets. As with the first release, the weights, intermediate checkpoints, data recipes, training and evaluation code, and the charter that guides the alignment are public.

\phantomsection
\section*{Acknowledgments}
\addcontentsline{toc}{section}{Acknowledgments}

This work was supported as part of the Swiss AI Initiative by a grant from the Swiss National Supercomputing Centre (CSCS). JV acknowledges support from Lia Rumantscha, RTR, and Fundaziun Patrimoni Cultural RTR. JV and MW acknowledge funding by the Swiss National Science Foundation (project InvestigaDiff; no.~10000503). We thank Ahmetcan Yavuz, Not Battesta Soliva, Manuel Tuor, Oleg Lavrovsky, Arundhati Balasubramaniam, and Marius Huber for participating in the human evaluation study for machine translation.

\newpage
\phantomsection
\addcontentsline{toc}{section}{References}
\begin{small}
\bibliography{refs}
\bibliographystyle{iclr2025_conference}
\end{small}

\clearpage
\appendix

\setcounter{table}{0}
\setcounter{figure}{0}

\renewcommand\thefigure{\thesection.\arabic{figure}}
\renewcommand\thetable{\thesection.\arabic{table}}
\counterwithin*{figure}{section}
\counterwithin*{table}{section}

\section*{Appendix}
\raggedbottom
\section{Contributions}\label{app:contributions}

Contributors by effort area, ordered by contribution within each area; bold marks core team members.

\noindent\begingroup\raggedright%
\renewcommand{\au}[2]{#1}\renewcommand{\cau}[2]{\textbf{#1}}%
\effortrow{Multimodality}{\cau{Yixuan~Xu}{ethz}, \cau{Raphael~Kreft}{ethz}, \cau{Tommy~Chu}{epfl}, \cau{Anunay~Yadav}{epfl}, \cau{Nicola~Irmiger}{ethz}, \au{Yuhei~Fukuhara}{epfl}}

\subeffortrow{Geospatial Data}{\au{Li~Mi}{ethz}, \au{Haoliang~Shang}{ethz}}
\subeffortrow{Swiss German Audio Data}{\au{Jérome~Landolt}{FHNW}, \au{Fiona~Panicali}{FHNW}, \au{Bryan~Schmid}{FHNW}, \au{Flynn~Guerra}{FHNW}, \au{Daniel~Perruchoud}{FHNW}, \au{Michael~Graber}{FHNW}}

\effortrow{Data and Compliance}{\cau{Sven~Najem\mbox{-}Meyer}{epfl}, \cau{Bettina~Messmer}{epfl}, \cau{Vinko~Sabolčec}{epfl}, \cau{Yassine~Turki}{epfl}, 
\au{Matin~Ansaripour}{epfl}, \au{Roman~Machacek}{bern}, \au{Konstantinos~Pitas}{ethz}, \au{David~Brüggemann}{sdsc}, \au{Qianqian~Zou}{sdsc}
}
\effortrow{Pretraining and Long Context}{\cau{Allen~Huang}{epfl}, \cau{Dani~Tamayo}{epfl}, \au{Lorenzo~Noci}{ethz}, \au{Alexander~Hägele}{epfl}, \au{Mikolaj~Boronski}{epfl}, \au{Faruk~Zahiragić}{epfl}, \au{Yusif~Askari}{epfl}}
\effortrow{Supervised Finetuning}{\cau{Hanna~Yukhymenko}{ethz}, \cau{Valentina~Pyatkin}{ethz}, \au{Max~Neuwinger}{ethz}, \cau{Matteo~Santelmo}{epfl}, \cau{Juan~García~Giraldo}{epfl}, \cau{Dani~Tamayo}{epfl}, \au{Zixuan~Chen}{ethz}, \au{Haofeng~Yang}{ethz}, \au{Jingchun~Xiao}{ethz}, \au{Nathan~Ranchin}{ethz,epfl}, \au{Bich~Ngoc~Doan}{epfl}, \au{Rada~Kamysheva}{ethz}, \au{Luca~Baumann}{ethz}, \au{Anna~Kosovskaia}{ethz}, \cau{Eduard~Ďurech}{ethz}}
\effortrow{Reinforcement Learning}{
    \cau{Eduard~Ďurech}{ethz},
    \cau{Valentina~Pyatkin}{ethz},
    \cau{Matteo~Santelmo}{epfl},
    \cau{Juan~García~Giraldo}{epfl},
    \cau{Oscar~Alessandro~Tazza}{ethz}
}
\effortrow{RL Environments}{\cau{Eduard~Ďurech}{ethz}}
\subeffortrow{Code Gym}{\au{Roman~Machacek}{bern}, \au{Martin~Majsec}{epfl}, \au{Alex~Padula}{ethz}, \au{Khanh~Nguyen}{epfl}, \au{Bartosz~Szostakiewicz}{ethz}, \cau{Robert~Smith}{ethz}}
\subeffortrow{Reasoning Gym}{\au{Mauro~Pellonara}{epfl}, \au{Camille~Challier}{epfl}, \au{Clément~Charmillot}{epfl}, \cau{Hanna~Yukhymenko}{ethz}}
\subeffortrow{Vision RL}{\au{Mukhammadali~Sayfiddinov}{ethz}, \au{Badr~Al~Mahouri}{epfl}, \au{Joshua~Oyebanji}{epfl}, \au{Alessandro~Bombarda}{epfl}}
\subeffortrow{QA Gym}{\au{Oscar~Wohlfahrt}{epfl}, \au{Florian~Kolly}{epfl}, \au{Nicolas~Cuveillier}{epfl}, \au{Marko~Stojanovic}{epfl}, \cau{Valentina~Pyatkin}{ethz}}
\subeffortrow{Tool Gym}{\au{Loïc~Deslarzes}{epfl}, \au{Michal~Tešnar}{ethz}, \au{Francesco~Monti}{epfl}, \au{Morgane~Magnin}{epfl}, \au{Dana~Shayakhmetova}{epfl}, \au{Suei\mbox{-}Wen~Chen}{ethz}}

\subeffortrow{IF Gym}{\au{Emma~Anastassova}{epfl}, \au{Mattea~Busato}{epfl}, \au{Taliesin~Perez}{epfl}, \au{Simon~Pierre}{epfl}, \cau{Valentina~Pyatkin}{ethz}}
\subeffortend
\effortrow{Alignment}{\cau{Davit~Melikidze}{ethz}, \cau{Marian~Schneider}{ethz}, \au{Simon~Matrenok}{epfl}, \au{Thomas~Kleine~Buening}{ethz}, \au{Barna~Pasztor}{ethz}, \au{Vinzenz~Thoma}{ethz}, \au{Jessica~Lam}{ethz}}

\effortrow{Applications and Inference}{\cau{Robert~Smith}{ethz}, \cau{Aryan~Ahadinia}{epfl}, \cau{Oleg~Lavrovsky}{epfl}, \cau{Davit~Melikidze}{ethz}, \au{Ahmad~Fraij}{ethz}, \au{Helen~El~Khoury}{epfl}, \au{Szabina~Horváth\mbox{-}Mikulas}{epfl}, \au{Burak~Sezen}{ethz}, \au{Xiaozhe~Yao}{ethz}}

\effortrow{Framework Integrations}{\cau{Raphael~Kreft}{ethz}, \cau{Anunay~Yadav}{epfl}, \cau{Aryan~Ahadinia}{epfl}, \cau{Robert~Smith}{ethz}}

\effortrow{Evaluations and Safety}{\au{Yannick~Metz}{ethz}, \au{Frederike~Lübeck}{ethz}, \cau{Davit~Melikidze}{ethz}, \au{Alexander~Sternfeld}{hevs}, \au{Vinzenz~Thoma}{ethz}, \au{Vilém~Zouhar}{ethz}, \cau{Hanna~Yukhymenko}{ethz}, \au{Jannis~Vamvas}{uzh}, \au{Andrei~Kucharavy}{hevs}, \au{Michelle~Wastl}{uzh}, \au{Arthur~Wuhrmann}{epfl}}
\effortrow{Interpretability}{\au{Anna~Hedström}{ethz}, \au{Rita~Sevastjanova}{ethz}, \au{Arundhati~Balasubramaniam}{ethz}, \au{Tymoteusz~Kwiecinski}{ethz}, \au{Lorenzo~Baggi}{ethz}, \au{Marco~De~Negri}{ethz}, \au{Luca~Sartori}{ethz}, \au{Moritz~Reihs}{ethz}, \au{Lasse~B.~Strand}{ethz}, \au{Thilo~Spinner}{ethz}}
\effortrow{Infrastructure}{\au{Fabian~Bösch}{cscs}, \au{Jonathan~Coles}{cscs}, \au{Lukas~Drescher}{cscs}, \au{David~Isztl}{cscs}, \au{Miguel~Gila}{cscs}, \au{Sebastian~Keller}{cscs}, \au{Mark~Klein}{cscs}, \au{Henrique~Mendonça}{cscs}, \au{Fawzi~Roberto~Mohamed}{cscs}, \au{Gennaro~Oliva}{cscs}, \au{Elia~Palme}{cscs}, \au{Marco~Passerini}{cscs}, \au{Tim~Robinson}{cscs}, \au{Stefano~Schuppli}{cscs}, \au{Joost~VandeVondele}{cscs}}
\effortrow{Management and Advisors}{\cau{Sean~F.~Carroll}{epfl}, \au{Joost~VandeVondele}{cscs}, \au{Thomas~Schulthess}{cscs}, \au{Elliott~Ash}{ethz}, \au{Andreas~Krause}{ethz}, \au{Ana~Klimovic}{ethz}, \au{Martin~Rajman}{epfl}, \au{Torsten~Hoefler}{ethz}, \au{Alexander~Ilic}{ethz}, \au{Mennatallah~El\mbox{-}Assady}{ethz}, \au{Timo~Kehrer}{bern}}
\effortrow{Project Leadership}{\cau{Imanol~Schlag}{ethz}, \cau{Martin~Jaggi}{epfl}, \cau{Antoine~Bosselut}{epfl}}
\par\endgroup
\clearpage\flushbottom

\clearpage
\section{Vision Datasets by Training Stage}\label{sec:vision_dataset_tables}

The stagewise tables give the vision pool of each stage: for every dataset, the prepared samples and input tokens available to the run. The mix a run trains on, that is, the tokens trained on including repetition, appears per modality in \autoref{tab:token-budget} for the pretraining stages, per dataset in \autoref{app:long_context_mix} for the long-context extension, and per source in \autoref{tab:long_context_sft_mix} for long-context SFT. Stages~1 and~2 appear in \autoref{tab:vision_data_consolidated} and \autoref{tab:vision_data_stage_two}, the cooldown and long-context pretraining pools in \autoref{tab:vision_data_cooldown} and \autoref{tab:vision_data_lc}, and the SFT and long-context SFT pools in \autoref{tab:vision_data_sft} and \autoref{tab:vision_data_lc_sft}. \autoref{tab:vision_data_stage_two} and \autoref{tab:vision_data_cooldown} report the \apertusonefivesmall{} pools. The \apertusonefivelarge{} cooldown additionally reuses datasets already listed for Stage~2 in \autoref{tab:vision_data_stage_two}, which we do not repeat. Vision reinforcement learning adds 50,000 in-house BlindTasks-RL prompts, which \autoref{tab:rl_data_mixture} reports separately. The preference-optimization stages use no multimodal inputs. The SFT tables here report the production pools; the smaller image pool of the capability-filtering SFT phase appears in \autoref{tab:sft_image_pool}, whereas the audio pool is the same in both phases.


\newcommand{\declarevisiondataset}[2]{\expandafter\newcommand\csname visiondataset@#1\endcsname{#2}}
\newcommand{\visiondataset}[1]{\csname visiondataset@#1\endcsname}
\declarevisiondataset{blip3o-long-caption}{\href{https://huggingface.co/datasets/BLIP3o/BLIP3o-Pretrain-Long-Caption/tree/e4d07091a466d1a1e35a9b0c61caddc78d14a059}{BLIP3o Long Caption}}
\declarevisiondataset{commoncatalog-cc-by}{\href{https://huggingface.co/datasets/common-canvas/commoncatalog-cc-by/tree/80f50fe4a1ca937f37a11be3f8eee5199d776ff3}{CommonCatalog CC-BY}}
\declarevisiondataset{docci}{\href{https://huggingface.co/datasets/google/docci/tree/a0a43eaf34676ffd008fb6565dd8c2ba00d09100}{DOCCI}}
\declarevisiondataset{facecaption-15m}{\href{https://huggingface.co/datasets/OpenFace-CQUPT/FaceCaption-15M/tree/3ed92d90f7fc7199b47c4da17c6863b1a175f380}{FaceCaption-15M}}
\declarevisiondataset{fine-t2i}{\href{https://huggingface.co/datasets/ma-xu/fine-t2i/tree/28fdd5663ee202b5cafc01d6ed08a03f14957854}{Fine-T2I}}
\declarevisiondataset{hq-50k}{\href{https://huggingface.co/datasets/YangQiee/HQ-50K/tree/1c501522f77594726b00b5943f1804c600e0230d}{HQ-50K}}
\declarevisiondataset{innovator-vl-instruct}{\href{https://huggingface.co/datasets/InnovatorLab/Innovator-VL-Instruct-46M/tree/1e7bac4db314fcc99e43601632c6ad7de71b3a6c}{Innovator-VL Instruct}}
\declarevisiondataset{laion-aesthetics-12m}{\href{https://huggingface.co/datasets/dclure/laion-aesthetics-12m-umap/tree/06928317703bcfa6099c7fc0f13e11bb295e7769}{LAION Aesthetics 12M}}
\declarevisiondataset{llava-onevision-1-5-mid-training}{\href{https://huggingface.co/datasets/mvp-lab/LLaVA-OneVision-1.5-Mid-Training-85M/tree/c5218cad785eba7d218137e8ce4997bda568a050}{LLaVA-OV 1.5 Mid-Training}}
\declarevisiondataset{lnqa-recap}{\href{https://huggingface.co/datasets/HuggingFaceM4/FineVision/tree/3c380a731a3429c1d04693d6ec16d7e683def84c}{LNQA (recap)}}
\declarevisiondataset{megalith-10m}{\href{https://huggingface.co/datasets/madebyollin/megalith-10m/tree/1e65a79953396f6d05f60eba4cc564541ad4be8c}{Megalith-10M}}
\declarevisiondataset{mint-1t-html}{\href{https://huggingface.co/datasets/mlfoundations/MINT-1T-HTML/tree/906a8b85cea61198ff7339c4dd711ad0b5361847}{MINT-1T HTML}}
\declarevisiondataset{mmammoth}{\href{https://huggingface.co/datasets/MAmmoTH-VL/MAmmoTH-VL-Instruct-12M/tree/bac8f77cb8a8f9c4d0de407c6e3a589bd722562a}{MAmmoTH-VL}}
\declarevisiondataset{nemotron-image-training-v3}{\href{https://huggingface.co/datasets/nvidia/Nemotron-Image-Training-v3/tree/7656391d4d4cb11ec3722b34f10d499435de0460}{Nemotron-Image-Training-3}}
\declarevisiondataset{open-images-v7}{\href{https://huggingface.co/datasets/bitmind/open-images-v7/tree/4518ecd40f8f9ef66ee4356be438f840c714e95a}{Open Images v7}}
\declarevisiondataset{pd12m}{\href{https://huggingface.co/datasets/Spawning/pd12m-full/tree/7bda6bfe13c6a39fd0a286adced5df1228041720}{PD12M}}
\declarevisiondataset{pin-200m}{\href{https://huggingface.co/datasets/m-a-p/PIN-200M/tree/0518ee384f3afbbc594f68c428667b86847327e8}{PIN-200M}}
\declarevisiondataset{pixmo-ask-model-anything}{\href{https://huggingface.co/datasets/allenai/pixmo-ask-model-anything/tree/739ffd8b417737da31e5069b9c8ff6a72a7d58b6}{PixMo Ask Model Anything}}
\declarevisiondataset{pixmo-cap}{\href{https://huggingface.co/datasets/allenai/pixmo-cap/tree/edce6390d9d5be6c8db0d863fbe62718c88988a4}{PixMo Cap}}
\declarevisiondataset{pixmo-cap-qa}{\href{https://huggingface.co/datasets/allenai/pixmo-cap/tree/edce6390d9d5be6c8db0d863fbe62718c88988a4}{PixMo Cap QA}}
\declarevisiondataset{product-catalogue}{\href{https://huggingface.co/datasets/Shopify/product-catalogue/tree/002ca15536bb015ebc47b99cc2d716d7040bb09e}{Product Catalogue}}
\declarevisiondataset{recap-datacomp-1b}{\href{https://huggingface.co/datasets/UCSC-VLAA/Recap-DataComp-1B/tree/457c44d98651bcfdfb3cc8695f5e60a0d2705e78}{Recap-DataComp-1B}}
\declarevisiondataset{uno-1m}{\href{https://huggingface.co/datasets/bytedance-research/UNO-1M/tree/f25bb61db6d6d66d82f41d1e613c0e04ba342e84}{UNO-1M}}
\declarevisiondataset{visual-genome}{\href{https://huggingface.co/datasets/ranjaykrishna/visual\_genome/tree/65bc9e7e7353fff750326c9523e384701934e530}{Visual Genome}}
\declarevisiondataset{argimi-ardian-finance-10-k}{\href{https://huggingface.co/datasets/artefactory/Argimi-Ardian-Finance-10k-text-image/tree/d019db455ff58bc14cae72422c4fc3ef0c301ea7}{Argimi-Ardian Finance 10-K}}
\declarevisiondataset{bigdocs-7-5m}{\href{https://huggingface.co/datasets/ServiceNow/BigDocs-7.5M/tree/dae4403c28307bd5328920740e81ce5232819e74}{BigDocs-7.5M}}
\declarevisiondataset{cosyn-400k}{\href{https://huggingface.co/datasets/HuggingFaceM4/FineVision/tree/3c380a731a3429c1d04693d6ec16d7e683def84c}{CoSyn-400K}}
\declarevisiondataset{handwriting-data}{\href{https://huggingface.co/datasets/swiss-ai/handwriting\_data/tree/850e7569decaec09b1b1a8c959c9560c9059dc58}{Handwriting Data}}
\declarevisiondataset{handwriting-documents}{\href{https://huggingface.co/datasets/swiss-ai/handwriting\_data/tree/850e7569decaec09b1b1a8c959c9560c9059dc58}{Handwriting documents}}
\declarevisiondataset{handwriting-slides}{\href{https://huggingface.co/datasets/swiss-ai/handwriting\_data/tree/850e7569decaec09b1b1a8c959c9560c9059dc58}{Handwriting slides}}
\declarevisiondataset{latex-formulas-80m}{\href{https://huggingface.co/datasets/OleehyO/latex-formulas-80M/tree/5cd783320b0092caa85720a85d86595f42df043b}{LaTeX Formulas 80M}}
\declarevisiondataset{mint-1t-arxiv}{\href{https://huggingface.co/datasets/mlfoundations/MINT-1T-ArXiv/tree/7c5b00ffd5b563071010c3bf2082b4a8f836eb72}{MINT-1T ArXiv}}
\declarevisiondataset{mint-1t-pdf}{\href{https://huggingface.co/collections/mlfoundations/mint-1t}{MINT-1T PDF}}
\declarevisiondataset{mit-10m}{\href{https://huggingface.co/datasets/liboaccn/MIT-10M/tree/bcba6b2651771c69f93e000486c2baa0896d32c3}{MIT-10M}}
\declarevisiondataset{multihiertt}{\href{https://huggingface.co/datasets/HuggingFaceM4/FineVision/tree/3c380a731a3429c1d04693d6ec16d7e683def84c}{MultiHiertt}}
\declarevisiondataset{textatlas5m}{\href{https://huggingface.co/datasets/CSU-JPG/TextAtlas5M/tree/f9f2a0f5000fbb078f718197acb45cfb9ceed551}{TextAtlas5M}}
\declarevisiondataset{vdr-cooking-recipes}{\href{https://huggingface.co/datasets/racineai/VDR\_Cooking\_Recipes/tree/0be4bc6a1d7f4483fc74d2dfd5c9dc1c2f55eb17}{VDR Cooking Recipes}}
\declarevisiondataset{websight-v0-2}{\href{https://huggingface.co/datasets/HuggingFaceM4/WebSight/tree/b11f8172f89c992b56ac702319e02c428cca4a4e}{WebSight v0.2}}
\declarevisiondataset{chartnet}{\href{https://huggingface.co/datasets/ibm-granite/ChartNet/tree/52832dd476c9a7a7b01c20245e952958cef1e2b2}{ChartNet}}
\declarevisiondataset{data-insights}{\href{https://ourworldindata.org/}{Data Insights}}
\declarevisiondataset{datikz-v4}{\href{https://huggingface.co/datasets/nllg/DaTikZ-V4/tree/33734c83608211682be11001a1618856fc1979dd}{DaTikZ-V4}}
\declarevisiondataset{llava-onevision-1-5-instruct}{\href{https://huggingface.co/datasets/mvp-lab/LLaVA-OneVision-1.5-Instruct-Data/tree/0efb7ad47f8ba36e1d262545cadec4cd8960dc5f}{LLaVA-OV 1.5 Instruct}}
\declarevisiondataset{molmo2-multiimageqa}{\href{https://huggingface.co/datasets/allenai/Molmo2-MultiImageQA/tree/f47ca3644d394b548be07a68d5a6fc0275924f08}{Molmo2-MultiImageQA}}
\declarevisiondataset{molmo2-synmultiimageqa}{\href{https://huggingface.co/datasets/allenai/Molmo2-SynMultiImageQA/tree/5a00b7a26f0ad416ae7dd030094270e0a3adfb0f}{Molmo2-SynMultiImageQA}}
\declarevisiondataset{owid-articles}{\href{https://ourworldindata.org/}{OWID Articles}}
\declarevisiondataset{owid-grapher}{\href{https://ourworldindata.org/}{OWID Grapher}}
\declarevisiondataset{waffle}{\href{https://tau-vailab.github.io/WAFFLE/}{WAFFLE}}
\declarevisiondataset{annotated-ultrasound-liver-images}{\href{https://www.kaggle.com/datasets/orvile/annotated-ultrasound-liver-images-dataset}{Ultrasound Liver Images}}
\declarevisiondataset{brain-tumor-mri}{\href{https://www.kaggle.com/datasets/masoudnickparvar/brain-tumor-mri-dataset}{Brain Tumor MRI}}
\declarevisiondataset{breast-ultrasound-busi}{\href{https://www.kaggle.com/datasets/sabahesaraki/breast-ultrasound-images-dataset}{Breast Ultrasound (BUSI)}}
\declarevisiondataset{busi}{\href{https://www.kaggle.com/datasets/sabahesaraki/breast-ultrasound-images-dataset}{BUSI}}
\declarevisiondataset{covid-us}{\href{https://github.com/nrc-cnrc/COVID-US}{COVID-US}}
\declarevisiondataset{dailymed-spl}{\href{https://dailymed.nlm.nih.gov/dailymed/spl-resources-all-drug-labels.cfm}{DailyMed SPL}}
\declarevisiondataset{ddti}{\href{https://www.kaggle.com/datasets/dasmehdixtr/ddti-thyroid-ultrasound-images}{DDTI}}
\declarevisiondataset{ddti-thyroid-ultrasound}{\href{https://www.kaggle.com/datasets/dasmehdixtr/ddti-thyroid-ultrasound-images}{DDTI Thyroid Ultrasound}}
\declarevisiondataset{diabetic-retinopathy-2015}{\href{https://www.kaggle.com/datasets/sovitrath/diabetic-retinopathy-2015-data-colored-resized}{Diabetic Retinopathy 2015}}
\declarevisiondataset{ebhi-seg}{\href{https://github.com/dataset-ninja/ebhi-seg}{EBHI-SEG}}
\declarevisiondataset{gmai-vl-permissive-subset}{\href{https://huggingface.co/datasets/General-Medical-AI/GMAI-VL-5.5M/tree/d68bd3c1fb46d123b3ff3509301da9ffeb8ea07e}{GMAI-VL (permissive)}}
\declarevisiondataset{isic}{\href{https://www.kaggle.com/datasets/abhii1929/isic-skin-disease-image-dataset-4-classes}{ISIC}}
\declarevisiondataset{medmax}{\href{https://huggingface.co/datasets/mint-medmax/medmax\_data/tree/2a63fb4d5e57bbf30a130976825fadf95ea95ebb}{MedMax}}
\declarevisiondataset{medmnist}{\href{https://zenodo.org/records/10519652}{MedMNIST}}
\declarevisiondataset{medpix-2}{\href{https://zenodo.org/records/12624810}{MedPix-2}}
\declarevisiondataset{medtrinity-25m}{\href{https://huggingface.co/datasets/UCSC-VLAA/MedTrinity-25M/tree/89e5c684794e5c4cc1af9e8f1a7798af7c937dbf}{MedTrinity-25M}}
\declarevisiondataset{multicare-case-images}{\href{https://huggingface.co/datasets/openmed-community/multicare-case-images/tree/c8517124928d2fe3651ee6cb6c560fce66e02344}{MultiCare Case Images}}
\declarevisiondataset{multicare-images}{\href{https://huggingface.co/datasets/openmed-community/multicare-images/tree/5c954c4fbf9abdcb55053488dab6c1ef142796b5}{MultiCare Images}}
\declarevisiondataset{nct-crc-he-100k}{\href{https://zenodo.org/records/1214456}{NCT-CRC-HE-100K}}
\declarevisiondataset{nih-chest-x-ray}{\href{https://huggingface.co/datasets/alkzar90/NIH-Chest-X-ray-dataset/tree/36778e3b0e4f4b4fad31d1728d6190f3eda5b543}{NIH Chest X-Ray}}
\declarevisiondataset{open-ufi}{\href{https://springernature.figshare.com/articles/dataset/Open\_UFI\_and\_clinical\_IQA/26936446}{Open UFI}}
\declarevisiondataset{openpmc-18m}{\href{https://huggingface.co/datasets/vector-institute/open-pmc-18m/tree/b5a67783ec3e1bf91809a5efc4b72fbedacacdf6}{OpenPMC-18M}}
\declarevisiondataset{pathvqa}{\href{https://huggingface.co/datasets/flaviagiammarino/path-vqa/tree/1685832883334b5bb5beaf4e4b333fdeecaa4ad9}{PathVQA}}
\declarevisiondataset{pmc-oa}{\href{https://huggingface.co/datasets/axiong/pmc\_oa/tree/1d2296e9c022a24e82a47e524d53f0915b98c926}{PMC OA}}
\declarevisiondataset{pubmedvision}{\href{https://huggingface.co/datasets/FreedomIntelligence/PubMedVision/tree/3c84e04b38bceb5341419b9a4f8ca37ba790cb84}{PubMedVision}}
\declarevisiondataset{rfmid-2}{\href{https://zenodo.org/records/7505822}{RFMiD 2}}
\declarevisiondataset{scin}{\href{https://huggingface.co/datasets/google/scin/tree/996257142f7517fb8991a28cfba46ec4e3f530a9}{SCIN}}
\declarevisiondataset{shizhengpt-instruction-tuning}{\href{https://huggingface.co/datasets/FreedomIntelligence/TCM-Instruction-Tuning-ShizhenGPT/tree/718596f04350cc6c4c7735f8d62899a5a1649cb8}{ShizhenGPT IT}}
\declarevisiondataset{shizhengpt-pretraining-book}{\href{https://huggingface.co/datasets/FreedomIntelligence/TCM-Pretrain-Data-ShizhenGPT/tree/db4874ce4e322f47432fe322c558e516c5aad71e}{ShizhenGPT pretrain (book)}}
\declarevisiondataset{shizhengpt-pretraining-web}{\href{https://huggingface.co/datasets/FreedomIntelligence/TCM-Pretrain-Data-ShizhenGPT/tree/db4874ce4e322f47432fe322c558e516c5aad71e}{ShizhenGPT pretrain (web)}}
\declarevisiondataset{slid-e}{\href{https://figshare.com/articles/dataset/SLID-E/26172919}{SLID-E}}
\declarevisiondataset{bigearthnet}{\href{https://huggingface.co/datasets/wangyi111/Copernicus-Bench/tree/a287ab1b414d2bff99557166988571c5885ed81a}{BigEarthNet}}
\declarevisiondataset{bigearthnet-v2}{\href{https://huggingface.co/datasets/BIFOLD-BigEarthNetv2-0/BigEarthNet.txt/tree/72d865f2146f0a85b720f7f3ca1cdbaeafc3d316}{BigEarthNet v2}}
\declarevisiondataset{dfc2020}{\href{https://huggingface.co/datasets/wangyi111/Copernicus-Bench/tree/a287ab1b414d2bff99557166988571c5885ed81a}{DFC2020}}
\declarevisiondataset{flair-hub}{\href{https://huggingface.co/datasets/IGNF/FLAIR-HUB/tree/8275163f72f0eed69050a703925791b7c3577f10}{FLAIR-HUB}}
\declarevisiondataset{geochat-instruct}{\href{https://huggingface.co/datasets/MBZUAI/GeoChat\_Instruct/tree/8eb13307eabc7fa9c1f8b0e61e372a327ccd68b1}{GeoChat Instruct}}
\declarevisiondataset{ign}{\href{https://geoservices.ign.fr/planign}{IGN}}
\declarevisiondataset{maptrace}{\href{https://huggingface.co/datasets/google/MapTrace/tree/8dd60adfde2f189768f27204c78ec44af07a67bf}{MapTrace}}
\declarevisiondataset{rsrcc}{\href{https://huggingface.co/datasets/google/RSRCC/tree/7898de7bfd08bc404d9a92e1caaa9dce91b0c3ea}{RSRCC}}
\declarevisiondataset{rsteller}{\href{https://huggingface.co/datasets/SlytherinGe/RSTeller/tree/a03b35f1bc9a3ac14ae93724d175c2611f1bba5b}{RSTeller}}
\declarevisiondataset{skyscript}{\href{https://github.com/wangzhecheng/SkyScript}{SkyScript}}
\declarevisiondataset{swissimage-10-cm}{\href{https://www.swisstopo.admin.ch/en/orthoimage-swissimage-10\#Download}{SWISSIMAGE 10 cm}}
\declarevisiondataset{swisstopo-map-sat}{\href{https://www.geo.admin.ch/en/wms-available-services-an-data}{Swisstopo (map-sat)}}
\declarevisiondataset{mint-1t-arxiv-2}{\href{https://huggingface.co/datasets/mlfoundations/MINT-1T-ArXiv/tree/7c5b00ffd5b563071010c3bf2082b4a8f836eb72}{MINT-1T ArXiv}}
\declarevisiondataset{nasa-images}{\href{https://www.nasa.gov/}{NASA Images}}
\declarevisiondataset{treeoflife-10m}{\href{https://huggingface.co/datasets/imageomics/TreeOfLife-10M/tree/91debffb7146c32c89d76feb1eb575b555e2ecc7}{TreeOfLife-10M}}
\declarevisiondataset{art-museums-pd-440k}{\href{https://huggingface.co/datasets/Mitsua/art-museums-pd-440k/tree/fba945da78b36262eb9272067197cc28d06cffbf}{Art Museums PD-440K}}
\declarevisiondataset{chinesememe-english-recap}{\href{https://huggingface.co/datasets/HuggingFaceM4/FineVision/tree/3c380a731a3429c1d04693d6ec16d7e683def84c}{ChineseMeme (English)}}
\declarevisiondataset{crello}{\href{https://huggingface.co/datasets/cyberagent/crello/tree/7997e2f434ee4aa73cf4cdf22c5954cb175872e1}{Crello}}
\declarevisiondataset{memotion-english-recap}{\href{https://huggingface.co/datasets/HuggingFaceM4/FineVision/tree/3c380a731a3429c1d04693d6ec16d7e683def84c}{Memotion (English recap)}}
\declarevisiondataset{smithsonian-open-access}{\href{https://www.si.edu/openaccess}{Smithsonian Open Access}}
\declarevisiondataset{egopat3dv2}{\href{https://huggingface.co/datasets/ai4ce/EgoPAT3Dv2/tree/9f20d0b0f6f48022bc2e10c46f219e3b89c44681}{EgoPAT3Dv2}}
\declarevisiondataset{holoassist}{\href{https://holoassist.github.io/}{HoloAssist}}
\declarevisiondataset{blip3-grounding-50m}{\href{https://huggingface.co/datasets/Salesforce/blip3-grounding-50m/tree/4d622c4f19b8a6b91ad914caf343306e363ff79b}{BLIP3 Grounding 50M}}
\declarevisiondataset{llava-onevision-2-osd}{\href{https://huggingface.co/datasets/mvp-lab/LLaVA-OneVision-2-Data/tree/e73747a5}{LLaVA-OV 2 (OSD)}}
\declarevisiondataset{llava-onevision-2-roborefer-sim}{\href{https://huggingface.co/datasets/mvp-lab/LLaVA-OneVision-2-Data/tree/e73747a5}{LLaVA-OV 2 (RoboRefer)}}
\declarevisiondataset{mmevol-judged}{\href{https://huggingface.co/datasets/HuggingFaceM4/FineVision/tree/3c380a731a3429c1d04693d6ec16d7e683def84c}{MMEvol (judged)}}
\declarevisiondataset{molmopoint-guisyn}{\href{https://huggingface.co/datasets/allenai/MolmoPoint-GUISyn/tree/24bb3e990bb1e48796d80ade9dbc858dda695e75}{MolmoPoint-GUISyn}}
\declarevisiondataset{pixmo-point-explanations}{\href{https://huggingface.co/datasets/allenai/pixmo-point-explanations/tree/08a566fa00747e4c1c7e8481c350763b469c209c}{PixMo Point Explanations}}
\declarevisiondataset{sensenova-si-8m}{\href{https://huggingface.co/datasets/sensenova/SenseNova-SI-8M/tree/cfc5b062b304e9ee57ad1e8323f3d83fe94d33d1}{SenseNova-SI-8M}}
\declarevisiondataset{sensenova-si-8m-cot}{\href{https://huggingface.co/datasets/sensenova/SenseNova-SI-8M/tree/cfc5b062b304e9ee57ad1e8323f3d83fe94d33d1}{SenseNova-SI-8M (CoT)}}
\declarevisiondataset{spatialsense-gold}{\href{https://huggingface.co/datasets/HuggingFaceM4/FineVision/tree/3c380a731a3429c1d04693d6ec16d7e683def84c}{SpatialSense (gold)}}

\newcommand{\visiontablesplitrule}{\cmidrule{1-3}\cmidrule{5-7}}
\newcommand{\visiontablehead}{%
  \textbf{Dataset} & \textbf{Samples} & \textbf{Tokens} &
  & \textbf{Dataset} & \textbf{Samples} & \textbf{Tokens} \\
  \visiontablesplitrule
}

\begin{table}[!htbp]
\renewcommand{\arraystretch}{0.95}
\setlength{\tabcolsep}{2pt}
\caption{Vision pool of Stage~1.}\label{tab:vision_data_consolidated}
\centering
\small
\begin{tabular*}{\textwidth}{@{\extracolsep{\fill}}>{\raggedright\arraybackslash}p{4.0cm}>{\raggedleft\arraybackslash}p{1.3cm}>{\raggedleft\arraybackslash}p{1.2cm}@{}p{0.35cm}@{}>{\raggedright\arraybackslash}p{4.0cm}>{\raggedleft\arraybackslash}p{1.3cm}>{\raggedleft\arraybackslash}p{1.2cm}@{}}
\toprule
\visiontablehead
\visiondataset{bigdocs-7-5m} & 2.2M & 6.1B & & \visiondataset{medtrinity-25m} & 4.4M & 2.0B \\
\visiondataset{bigearthnet} & 24.0K & 6.0M & & \visiondataset{mint-1t-arxiv-2} & 4.6M & 14.4B \\
\visiondataset{busi} & 732 & 934.7K & & \visiondataset{mint-1t-html} & 125.0M & 228.0B \\
\visiondataset{covid-us} & 9.7K & 11.1M & & \visiondataset{mint-1t-pdf} & 58.8M & 93.4B \\
\visiondataset{ddti} & 347 & 278.6K & & \visiondataset{mmammoth} & 1.5M & 3.1B \\
\visiondataset{dfc2020} & 5.1K & 1.5M & & \visiondataset{open-ufi} & 700 & 3.7M \\
\visiondataset{handwriting-documents} & 15.0K & 49.9M & & \visiondataset{openpmc-18m} & 17.3M & 4.8B \\
\visiondataset{handwriting-slides} & 2.5K & 4.3M & & \visiondataset{pmc-oa} & 1.3M & 364.9M \\
\visiondataset{holoassist} & 266.6K & 126.6M & & \visiondataset{rfmid-2} & 754 & 2.4M \\
\visiondataset{isic} & 216.3K & 18.2M & & \visiondataset{scin} & 10.4K & 31.6M \\
\visiondataset{laion-aesthetics-12m} & 4.4M & 8.7B & & \visiondataset{slid-e} & 3.0K & 5.3M \\
\visiondataset{medmax} & 446.0K & 537.8M & & \visiondataset{swissimage-10-cm} & 1.3M & 5.1B \\
\visiondataset{medmnist} & 236.6K & 52.3M & & \visiondataset{treeoflife-10m} & 765.5K & 4.4B \\
\bottomrule
\end{tabular*}
\end{table}
\begin{table}[!htbp]
\renewcommand{\arraystretch}{0.95}
\setlength{\tabcolsep}{2pt}
\caption{Vision pool of Stage~2.}\label{tab:vision_data_stage_two}
\centering
\small
\begin{tabular*}{\textwidth}{@{\extracolsep{\fill}}>{\raggedright\arraybackslash}p{4.0cm}>{\raggedleft\arraybackslash}p{1.3cm}>{\raggedleft\arraybackslash}p{1.2cm}@{}p{0.35cm}@{}>{\raggedright\arraybackslash}p{4.0cm}>{\raggedleft\arraybackslash}p{1.3cm}>{\raggedleft\arraybackslash}p{1.2cm}@{}}
\toprule
\visiontablehead
\visiondataset{annotated-ultrasound-liver-images} & 735 & 1.8M & & \visiondataset{medtrinity-25m} & 6.9M & 5.1B \\
\visiondataset{art-museums-pd-440k} & 10.6M & 26.7B & & \visiondataset{megalith-10m} & 8.9M & 26.4B \\
\visiondataset{blip3-grounding-50m} & 33.7M & 47.5B & & \visiondataset{mit-10m} & 11.8M & 30.5B \\
\visiondataset{blip3o-long-caption} & 29.3M & 36.9B & & \visiondataset{molmo2-multiimageqa} & 564.3K & 2.6B \\
\visiondataset{brain-tumor-mri} & 7.2K & 6.7M & & \visiondataset{multicare-case-images} & 96.7K & 106.8M \\
\visiondataset{breast-ultrasound-busi} & 773 & 1.3M & & \visiondataset{multicare-images} & 161.6K & 132.9M \\
\visiondataset{commoncatalog-cc-by} & 14.6M & 84.9B & & \visiondataset{nct-crc-he-100k} & 107.2K & 25.5M \\
\visiondataset{covid-us} & 30.1K & 44.9M & & \visiondataset{nih-chest-x-ray} & 112.1K & 470.4M \\
\visiondataset{datikz-v4} & 427.8K & 419.8M & & \visiondataset{open-images-v7} & 1.9M & 6.4B \\
\visiondataset{ddti-thyroid-ultrasound} & 347 & 429.1K & & \visiondataset{openpmc-18m} & 17.5M & 8.1B \\
\visiondataset{diabetic-retinopathy-2015} & 35.1K & 8.5M & & \visiondataset{pd12m} & 12.5M & 83.6B \\
\visiondataset{ebhi-seg} & 2.2K & 534.1K & & \visiondataset{pin-200m} & 16.6M & 38.0B \\
\visiondataset{egopat3dv2} & 6.8K & 29.2M & & \visiondataset{pixmo-cap} & 706.5K & 3.4B \\
\visiondataset{facecaption-15m} & 10.6M & 4.1B & & \visiondataset{pmc-oa} & 1.6M & 601.8M \\
\visiondataset{fine-t2i} & 168.4K & 1.2B & & \visiondataset{product-catalogue} & 36.7K & 137.2M \\
\visiondataset{flair-hub} & 241.1K & 299.6M & & \visiondataset{recap-datacomp-1b} & 153.3M & 243.4B \\
\visiondataset{geochat-instruct} & 302.1K & 540.9M & & \visiondataset{rsteller} & 1.3M & 1.2B \\
\visiondataset{handwriting-data} & 4.1K & 19.9M & & \visiondataset{shizhengpt-pretraining-book} & 11.6K & 51.9M \\
\visiondataset{ign} & 14.4K & 18.3M & & \visiondataset{shizhengpt-pretraining-web} & 384.1K & 1.5B \\
\visiondataset{latex-formulas-80m} & 45.6M & 14.6B & & \visiondataset{skyscript} & 5.1M & 1.8B \\
\visiondataset{llava-onevision-1-5-mid-training} & 90.0M & 153.1B & & \visiondataset{textatlas5m} & 4.9M & 6.8B \\
\visiondataset{maptrace} & 241.9K & 1.0B & & \visiondataset{uno-1m} & 1.2M & 4.8B \\
\visiondataset{medmax} & 859.7K & 1.2B & & \visiondataset{websight-v0-2} & 1.5M & 12.2B \\
\visiondataset{medpix-2} & 405 & 1.0M & & & &  \\
\bottomrule
\end{tabular*}
\end{table}

\begin{table}[!htbp]
\renewcommand{\arraystretch}{0.95}
\setlength{\tabcolsep}{2pt}
\caption{Vision pool of the cooldown.}\label{tab:vision_data_cooldown}
\centering
\small
\begin{tabular*}{\textwidth}{@{\extracolsep{\fill}}>{\raggedright\arraybackslash}p{4.0cm}>{\raggedleft\arraybackslash}p{1.3cm}>{\raggedleft\arraybackslash}p{1.2cm}@{}p{0.35cm}@{}>{\raggedright\arraybackslash}p{4.0cm}>{\raggedleft\arraybackslash}p{1.3cm}>{\raggedleft\arraybackslash}p{1.2cm}@{}}
\toprule
\visiontablehead
\visiondataset{art-museums-pd-440k} & 10.6M & 26.7B & & \visiondataset{mint-1t-arxiv} & 4.4M & 16.7B \\
\visiondataset{chartnet} & 4.3M & 20.1B & & \visiondataset{nasa-images} & 23.0K & 152.1M \\
\visiondataset{crello} & 23.1K & 61.0M & & \visiondataset{owid-articles} & 102 & 533.8K \\
\visiondataset{dailymed-spl} & 77.4K & 328.0M & & \visiondataset{owid-grapher} & 3.5K & 8.4M \\
\visiondataset{data-insights} & 398 & 3.1M & & \visiondataset{pixmo-cap} & 706.5K & 3.4B \\
\visiondataset{docci} & 9.6K & 72.3M & & \visiondataset{smithsonian-open-access} & 41.7K & 314.8M \\
\visiondataset{hq-50k} & 36.7K & 255.8M & & \visiondataset{swisstopo-map-sat} & 280.5K & 1.1B \\
\visiondataset{latex-formulas-80m} & 29.4M & 3.1B & & \visiondataset{visual-genome} & 102.0K & 299.9M \\
\visiondataset{medtrinity-25m} & 6.9M & 5.1B & & & & \\
\bottomrule
\end{tabular*}
\end{table}

\begin{table}[!htbp]
\renewcommand{\arraystretch}{0.95}
\setlength{\tabcolsep}{2pt}
\caption{Vision pool of the long-context extension; \autoref{app:long_context_mix} gives the mix each run trains on.}\label{tab:vision_data_lc}
\centering
\small
\begin{tabular*}{\textwidth}{@{\extracolsep{\fill}}>{\raggedright\arraybackslash}p{4.0cm}>{\raggedleft\arraybackslash}p{1.3cm}>{\raggedleft\arraybackslash}p{1.2cm}@{}p{0.35cm}@{}>{\raggedright\arraybackslash}p{4.0cm}>{\raggedleft\arraybackslash}p{1.3cm}>{\raggedleft\arraybackslash}p{1.2cm}@{}}
\toprule
\visiontablehead
\visiondataset{argimi-ardian-finance-10-k} & 138.7K & 37.4B & & \visiondataset{molmo2-multiimageqa} & 466.2K & 7.4B \\
\visiondataset{chartnet} & 764.6K & 6.6B & & \visiondataset{nasa-images} & 225 & 1.9M \\
\visiondataset{data-insights} & 18 & 148.4K & & \visiondataset{owid-articles} & 301 & 7.1M \\
\visiondataset{hq-50k} & 100 & 829.0K & & \visiondataset{pin-200m} & 938.9K & 11.3B \\
\visiondataset{medpix-2} & 41 & 507.2K & & \visiondataset{shizhengpt-pretraining-book} & 6.5K & 110.0M \\
\visiondataset{mint-1t-arxiv} & 23.4K & 193.4M & & \visiondataset{shizhengpt-pretraining-web} & 113.9K & 1.9B \\
\visiondataset{waffle} & 11.8K & 75.6M & & & & \\
\bottomrule
\end{tabular*}
\end{table}

\begin{table}[!htbp]
\renewcommand{\arraystretch}{0.95}
\setlength{\tabcolsep}{2pt}
\caption{Vision pool of SFT, after splitting and packing for the \apertusonefivesmall{} production run.}\label{tab:vision_data_sft}
\centering
\small
\begin{tabular*}{\textwidth}{@{\extracolsep{\fill}}>{\raggedright\arraybackslash}p{4.0cm}>{\raggedleft\arraybackslash}p{1.3cm}>{\raggedleft\arraybackslash}p{1.2cm}@{}p{0.35cm}@{}>{\raggedright\arraybackslash}p{4.0cm}>{\raggedleft\arraybackslash}p{1.3cm}>{\raggedleft\arraybackslash}p{1.2cm}@{}}
\toprule
\visiontablehead
\visiondataset{bigearthnet-v2} & 160.4K & 191.8M & & \visiondataset{molmopoint-guisyn} & 25.3K & 177.8M \\
\visiondataset{chinesememe-english-recap} & 29.9K & 51.1M & & \visiondataset{multihiertt} & 3.8K & 16.5M \\
\visiondataset{cosyn-400k} & 143.8K & 863.7M & & \visiondataset{nemotron-image-training-v3} & 4.1M & 14.7B \\
\visiondataset{innovator-vl-instruct} & 24.8M & 65.2B & & \visiondataset{pathvqa} & 13.8K & 22.2M \\
\visiondataset{llava-onevision-1-5-instruct} & 7.2M & 19.0B & & \visiondataset{pixmo-ask-model-anything} & 41.8K & 145.9M \\
\visiondataset{llava-onevision-2-osd} & 546.7K & 1.2B & & \visiondataset{pixmo-cap-qa} & 117.1K & 557.3M \\
\visiondataset{llava-onevision-2-roborefer-sim} & 172.1K & 471.9M & & \visiondataset{pixmo-point-explanations} & 11.6K & 48.0M \\
\visiondataset{lnqa-recap} & 211.9K & 724.0M & & \visiondataset{rsrcc} & 72.7K & 169.8M \\
\visiondataset{maptrace} & 1.3M & 11.4B & & \visiondataset{sensenova-si-8m} & 4.8M & 27.5B \\
\visiondataset{memotion-english-recap} & 4.1K & 7.1M & & \visiondataset{shizhengpt-instruction-tuning} & 46.8K & 149.2M \\
\visiondataset{mmevol-judged} & 64.7K & 151.1M & & \visiondataset{spatialsense-gold} & 4.8K & 4.5M \\
\visiondataset{molmo2-multiimageqa} & 17.1K & 132.3M & & \visiondataset{vdr-cooking-recipes} & 8.2K & 56.0M \\
\visiondataset{molmo2-synmultiimageqa} & 45.0K & 427.0M & & & & \\
\bottomrule
\end{tabular*}
\end{table}

\begin{table}[!htbp]
\renewcommand{\arraystretch}{0.95}
\setlength{\tabcolsep}{2pt}
\caption{Vision pool of long-context SFT; \autoref{tab:long_context_sft_mix} gives the mix each run trains on.}\label{tab:vision_data_lc_sft}
\centering
\small
\begin{tabular*}{\textwidth}{@{\extracolsep{\fill}}>{\raggedright\arraybackslash}p{4.0cm}>{\raggedleft\arraybackslash}p{1.3cm}>{\raggedleft\arraybackslash}p{1.2cm}@{}p{0.35cm}@{}>{\raggedright\arraybackslash}p{4.0cm}>{\raggedleft\arraybackslash}p{1.3cm}>{\raggedleft\arraybackslash}p{1.2cm}@{}}
\toprule
\visiontablehead
\visiondataset{bigearthnet-v2} & 34.7K & 41.5M & & \visiondataset{molmopoint-guisyn} & 5.5K & 38.7M \\
\visiondataset{chinesememe-english-recap} & 6.5K & 11.1M & & \visiondataset{multihiertt} & 825 & 3.6M \\
\visiondataset{cosyn-400k} & 31.1K & 186.5M & & \visiondataset{nemotron-image-training-v3} & 939.1K & 8.8B \\
\visiondataset{gmai-vl-permissive-subset} & 653.7K & 341.3M & & \visiondataset{pathvqa} & 3.0K & 4.8M \\
\visiondataset{innovator-vl-instruct} & 5.4M & 14.2B & & \visiondataset{pixmo-ask-model-anything} & 9.0K & 32.0M \\
\visiondataset{llava-onevision-1-5-instruct} & 1.6M & 4.1B & & \visiondataset{pixmo-cap-qa} & 25.3K & 120.3M \\
\visiondataset{llava-onevision-2-osd} & 118.3K & 259.3M & & \visiondataset{pubmedvision} & 1.3M & 1.9B \\
\visiondataset{llava-onevision-2-roborefer-sim} & 37.2K & 102.1M & & \visiondataset{rsrcc} & 15.7K & 36.7M \\
\visiondataset{lnqa-recap} & 45.9K & 156.7M & & \visiondataset{sensenova-si-8m} & 1.0M & 8.0B \\
\visiondataset{maptrace} & 292.0K & 2.5B & & \visiondataset{sensenova-si-8m-cot} & 37.4K & 3.0B \\
\visiondataset{memotion-english-recap} & 887 & 1.5M & & \visiondataset{shizhengpt-instruction-tuning} & 10.1K & 32.1M \\
\visiondataset{mmevol-judged} & 14.0K & 32.7M & & \visiondataset{spatialsense-gold} & 1.0K & 974.2K \\
\visiondataset{molmo2-multiimageqa} & 3.8K & 30.1M & & \visiondataset{vdr-cooking-recipes} & 1.8K & 12.1M \\
\visiondataset{molmo2-synmultiimageqa} & 15.1K & 349.1M & & & &  \\
\bottomrule
\end{tabular*}
\end{table}

\clearpage
\clearpage
\section{Audio Datasets by Training Stage}\label{sec:audio_dataset_tables}

Tables~\autorefnum{tab:audio_data_stage_one}, \autorefnum{tab:audio_data_stage_two_cooldown}, \autorefnum{tab:audio_data_lc}, \autorefnum{tab:audio_data_sft}, and \autorefnum{tab:audio_data_lc_sft} give the audio pool of each training stage. \autoref{app:long_context_mix} and \autoref{tab:long_context_sft_mix} give the corresponding per-run mixtures for the long-context extension and long-context SFT.


\newcommand{\declareaudiodataset}[2]{\expandafter\newcommand\csname audiodataset@#1\endcsname{#2}}
\newcommand{\audiodataset}[1]{\csname audiodataset@#1\endcsname}
\declareaudiodataset{aishell-1-3-4-blend}{\href{https://www.openslr.org/33/}{AISHELL-1/3/4}}
\declareaudiodataset{aozora-hurigana}{\href{https://github.com/ndl-lab/hurigana-speech-corpus-aozora}{Aozora Hurigana}}
\declareaudiodataset{audiomcq-strongac-geminicot}{\href{https://huggingface.co/datasets/Harland/AudioMCQ-StrongAC-GeminiCoT/tree/6d549d883b2b4a7e323e8786596277e876c01052}{AudioMCQ GeminiCoT}}
\declareaudiodataset{audioset-balanced-unbalanced}{\href{https://huggingface.co/datasets/agkphysics/AudioSet/tree/0c609e8302cf139307f639c57652032af0a88041}{AudioSet (bal./unbal.)}}
\declareaudiodataset{cc-podcasts}{\href{https://huggingface.co/datasets/shuyuej/CC-Podcasts/tree/044c5e8b545e306cec2d9bc6e4a3db5ded17625b}{CC-Podcasts}}
\declareaudiodataset{common-voice-24}{\href{https://commonvoice.mozilla.org/}{Common Voice 24}}
\declareaudiodataset{coral-v3}{\href{https://huggingface.co/datasets/CoRal-project/coral-v3/tree/01f7c93c21fc9dec87fe9f7149c79569cc433f08}{Coral V3}}
\declareaudiodataset{eurospeech}{\href{https://huggingface.co/datasets/disco-eth/EuroSpeech/tree/d244daad20b9189245f361c057df5b130d75c113}{EuroSpeech}}
\declareaudiodataset{f1-team-radio}{\href{https://huggingface.co/datasets/MikCil/f1-team-radio/tree/a0b99e1a325d92d63b574541a24902a660a352ea}{F1 Team Radio}}
\declareaudiodataset{farsi-asr}{\href{https://huggingface.co/datasets/farsi-asr/farsi-asr-dataset/tree/59f14ee2ffce2f9e454d80f2b6d29016dd513492}{Farsi ASR}}
\declareaudiodataset{gemeinderat-zurich}{\href{https://www.cs.technik.fhnw.ch/i4ds-datasets}{Gemeinderat Zurich}}
\declareaudiodataset{ghana-english-speech}{\href{https://huggingface.co/datasets/ghananlpcommunity/ghana-english-speech-600hrs/tree/9c0b9da2d02cfea61c082582f2c628e39aebeeca}{Ghana English Speech}}
\declareaudiodataset{gigaspeech}{\href{https://huggingface.co/datasets/speechcolab/gigaspeech/tree/63c0836b643dc6136a608de041e56b67c12649b3}{GigaSpeech}}
\declareaudiodataset{gigaspeech-2}{\href{https://huggingface.co/datasets/speechcolab/gigaspeech2/tree/aeedc8cde7fd37c4164ab1d1b5551dd12ee931fe}{GigaSpeech 2}}
\declareaudiodataset{granary-yodas-ytc}{\href{https://huggingface.co/datasets/nvidia/Granary/tree/ed9782b0a927000dfd4d3b3a7e3d3e08a83360d5}{Granary (YODAS/YTC)}}
\declareaudiodataset{heysquad-human}{\href{https://huggingface.co/datasets/yijingwu/HeySQuAD\_human/tree/2a9aff6032a76185b98eff703144a2a896313e49}{HeySQuAD human}}
\declareaudiodataset{heysquad-machine}{\href{https://huggingface.co/datasets/yijingwu/HeySQuAD\_machine/tree/129220a5a6d27eb2c70be4ecc2e4bbcad4cca436}{HeySQuAD machine}}
\declareaudiodataset{infore2-audiobooks}{\href{https://huggingface.co/datasets/doof-ferb/infore2\_audiobooks/tree/836a9a79e28c6e598a5c308de7f047613dcf9ec5}{Infore2 Audiobooks}}
\declareaudiodataset{kathbath}{\href{https://huggingface.co/datasets/ai4bharat/Kathbath/tree/5b9e92849222026d9141acba4e8434fe816396bf}{Kathbath}}
\declareaudiodataset{kazakh-speech}{\href{https://huggingface.co/datasets/Flamme-VRM/kazakh-speech-dataset/tree/343bb8647d6dadef5c7db0710fd0d34340cfc488}{Kazakh Speech}}
\declareaudiodataset{legco-speech}{\href{https://huggingface.co/datasets/laubonghaudoi/legco-speech/tree/d7b41c4be0bea9d58b9fc98e7ad3e14dce3771eb}{LegCo Speech}}
\declareaudiodataset{libriheavy}{\href{https://huggingface.co/datasets/mythicinfinity/libriheavy/tree/784858eb1793bfcf2fd859f7a7ce2dae67dfe5f7}{LibriHeavy}}
\declareaudiodataset{localized-narratives-audio}{\href{https://storage.googleapis.com/localized-narratives}{Localized Narratives Audio}}
\declareaudiodataset{marco-longspeech}{\href{https://huggingface.co/datasets/ATH-MaaS/Marco\_Longspeech/tree/bb50d8c11fe9aa9fc5e899a0d941e3b40acbc740}{Marco Longspeech}}
\declareaudiodataset{mediaspeech}{\href{https://www.openslr.org/108/}{MediaSpeech}}
\declareaudiodataset{mls}{\href{https://huggingface.co/datasets/facebook/multilingual\_librispeech/tree/2e83e61823b4c47dcbcb1980bb88601274127609}{MLS}}
\declareaudiodataset{mtg-jamendo}{\href{https://huggingface.co/datasets/rkstgr/mtg-jamendo/tree/8265518f6b5127d386a85ab5c380d867ff9ae70b}{MTG-Jamendo}}
\declareaudiodataset{multimed}{\href{https://huggingface.co/datasets/leduckhai/MultiMed/tree/459d0ab6db332904f9d7b76a8baabf3333958fa8}{MultiMed}}
\declareaudiodataset{omnilingual-asr}{\href{https://huggingface.co/datasets/facebook/omnilingual-asr-corpus/tree/8648ba8946377697b427ae952076e49fc0e5e44d}{OmniLingual ASR}}
\declareaudiodataset{parlament-parla}{\href{https://huggingface.co/datasets/projecte-aina/parlament\_parla\_v3/tree/5fabb9fb9cf76ecf8a6cf915dd41d276d48fd0ff}{ParlaMent Parla}}
\declareaudiodataset{parlaspeech-rs}{\href{https://huggingface.co/datasets/classla/ParlaSpeech-RS/tree/a806c1fb6e2bfb99c75f4a811b8efe537e70d631}{ParlaSpeech-RS}}
\declareaudiodataset{people-s-speech-supervised}{\href{https://huggingface.co/datasets/MLCommons/peoples\_speech/tree/f10597c5d3d3a63f8b6827701297c3afdf178272}{People's Speech (sup.)}}
\declareaudiodataset{russian-librispeech}{\href{https://www.openslr.org/96/}{Russian LibriSpeech}}
\declareaudiodataset{seamlessalign}{\href{https://huggingface.co/datasets/ai4bharat/SeamlessAlign/tree/8c812b0c895ffc3e1d8f63f3c0122d9cf332a286}{SeamlessAlign}}
\declareaudiodataset{slopalspeech}{\href{https://huggingface.co/datasets/NaiveNeuron/SloPalSpeech/tree/c8a0ee1733fc1f28dee7fb80fec24a2153280569}{SloPalSpeech}}
\declareaudiodataset{spc-r}{\href{https://huggingface.co/datasets/i4ds/spc\_r/tree/86a54db1b27c10154cc371576fcc97017f434f79}{SPC-R}}
\declareaudiodataset{suno}{\href{https://huggingface.co/datasets/nyuuzyou/suno/tree/dd95495c415eea043c250f12da595de2ad4cad7f}{Suno}}
\declareaudiodataset{tcm-shizhen-speech}{\href{https://huggingface.co/datasets/FreedomIntelligence/TCM-Instruction-Tuning-ShizhenGPT/tree/718596f04350cc6c4c7735f8d62899a5a1649cb8}{TCM Shizhen Speech}}
\declareaudiodataset{teleantifraud}{\href{https://huggingface.co/datasets/JimmyMa99/TeleAntiFraud/tree/0872e54b584b28d34e0911dffcf696f0b2e5e49a}{TeleAntiFraud}}
\declareaudiodataset{tonewebinars}{\href{https://huggingface.co/datasets/Vikhrmodels/ToneWebinars/tree/a9bda19fb4bf501a744dc0a9c1c7d239b4c43ea4}{ToneWebinars}}
\declareaudiodataset{tuda-de}{\href{https://huggingface.co/datasets/uhhlt/Tuda-De/tree/26ad28a9ac73b664c2f7c7c93969f9507ea6aadc}{Tuda-De}}
\declareaudiodataset{unsupervised-people-s-speech}{\href{https://huggingface.co/datasets/MLCommons/unsupervised\_peoples\_speech/tree/d917e17e86f6abc4fa4d83e958c8f4173f45f0e7}{People's Speech (unsup.)}}
\declareaudiodataset{vietspeech}{\href{https://huggingface.co/datasets/NhutP/VietSpeech/tree/0f18688afa9b64ea515d3fdaad850999b86a0f52}{VietSpeech}}
\declareaudiodataset{vimedcss}{\href{https://huggingface.co/datasets/tensorxt/ViMedCSS/tree/b6959a18a08739464733930a872e7125c03e6558}{ViMedCSS}}
\declareaudiodataset{vocalized-sft}{Vocalized SFT}
\declareaudiodataset{voiceassistant-400k}{\href{https://huggingface.co/datasets/gpt-omni/VoiceAssistant-400K/tree/65dab707fcdd3d43dcfb834398aa9fed4116be3a}{VoiceAssistant-400K}}
\declareaudiodataset{voxpopuli}{\href{https://github.com/facebookresearch/voxpopuli}{VoxPopuli}}
\declareaudiodataset{wenetspeech}{\href{https://huggingface.co/datasets/wenet-e2e/wenetspeech/tree/5fe6045424583e9480cc6ca2a09df6932e3d51fd}{WenetSpeech}}
\declareaudiodataset{zeroth-korean}{\href{https://huggingface.co/datasets/Bingsu/zeroth-korean/tree/bd173fe2c8ed0dccd47acb4eda77542593651622}{Zeroth Korean}}
\declareaudiodataset{zoengjyutgaai}{\href{https://huggingface.co/datasets/CanCLID/zoengjyutgaai/tree/829b30725811cabe9b8e2fd0513966470b304f12}{Zoengjyutgaai}}

\newcommand{\audiotablesplitrule}{\cmidrule{1-3}\cmidrule{5-7}}
\newcommand{\audiotablehead}{%
  \textbf{Dataset} & \textbf{Samples} & \textbf{Tokens} &
  & \textbf{Dataset} & \textbf{Samples} & \textbf{Tokens} \\
  \audiotablesplitrule
}

\begin{table}[!htbp]
\renewcommand{\arraystretch}{0.95}
\setlength{\tabcolsep}{2pt}
\caption{Audio pool of Stage~1.}
\label{tab:audio_data_stage_one}
\centering
\small
\begin{tabular*}{\textwidth}{@{\extracolsep{\fill}}>{\raggedright\arraybackslash}p{4.0cm}>{\raggedleft\arraybackslash}p{1.3cm}>{\raggedleft\arraybackslash}p{1.2cm}@{}p{0.35cm}@{}>{\raggedright\arraybackslash}p{4.0cm}>{\raggedleft\arraybackslash}p{1.3cm}>{\raggedleft\arraybackslash}p{1.2cm}@{}}
\toprule
\audiotablehead
\audiodataset{audioset-balanced-unbalanced} & 1.7M & 700.3M & & \audiodataset{suno} & 235.4K & 1.4B \\
\audiodataset{common-voice-24} & 2.0M & 427.0M & & \audiodataset{unsupervised-people-s-speech} & 15.5M & 51.2B \\
\audiodataset{gemeinderat-zurich} & 72.1K & 172.7M & & \audiodataset{voxpopuli} & 31.9M & 56.7B \\
\audiodataset{mtg-jamendo} & 19.5K & 107.6M & & & & \\
\bottomrule
\end{tabular*}
\end{table}

\begin{table}[!htbp]
\renewcommand{\arraystretch}{0.95}
\setlength{\tabcolsep}{2pt}
\caption{Audio pool of Stage~2 and the cooldown.}
\label{tab:audio_data_stage_two_cooldown}
\centering
\small
\begin{tabular*}{\textwidth}{@{\extracolsep{\fill}}>{\raggedright\arraybackslash}p{4.0cm}>{\raggedleft\arraybackslash}p{1.3cm}>{\raggedleft\arraybackslash}p{1.2cm}@{}p{0.35cm}@{}>{\raggedright\arraybackslash}p{4.0cm}>{\raggedleft\arraybackslash}p{1.3cm}>{\raggedleft\arraybackslash}p{1.2cm}@{}}
\toprule
\audiotablehead
\audiodataset{aishell-1-3-4-blend} & 226.7K & 48.8M & & \audiodataset{mediaspeech} & 10.0K & 6.3M \\
\audiodataset{aozora-hurigana} & 2.6M & 703.6M & & \audiodataset{multimed} & 34.1K & 15.2M \\
\audiodataset{cc-podcasts} & 238.5K & 437.4M & & \audiodataset{omnilingual-asr} & 183.9K & 373.9M \\
\audiodataset{common-voice-24} & 11.7M & 2.6B & & \audiodataset{parlament-parla} & 241.7K & 115.8M \\
\audiodataset{coral-v3} & 295.2K & 80.6M & & \audiodataset{parlaspeech-rs} & 159.7K & 117.7M \\
\audiodataset{eurospeech} & 2.1M & 2.1B & & \audiodataset{people-s-speech-supervised} & 2.1M & 1.4B \\
\audiodataset{f1-team-radio} & 14.3K & 6.8M & & \audiodataset{russian-librispeech} & 31.3K & 14.6M \\
\audiodataset{farsi-asr} & 329.7K & 78.7M & & \audiodataset{spc-r} & 66.1K & 59.3M \\
\audiodataset{ghana-english-speech} & 405.9K & 96.6M & & \audiodataset{seamlessalign} & 6.0M & 2.4B \\
\audiodataset{gigaspeech} & 7.8M & 1.6B & & \audiodataset{slopalspeech} & 243.3K & 450.1M \\
\audiodataset{gigaspeech-2} & 19.0M & 3.9B & & \audiodataset{tonewebinars} & 285.2K & 325.3M \\
\audiodataset{granary-yodas-ytc} & 74.1M & 41.9B & & \audiodataset{tuda-de} & 50.8K & 19.2M \\
\audiodataset{infore2-audiobooks} & 265.8K & 68.3M & & \audiodataset{vimedcss} & 7.3K & 3.8M \\
\audiodataset{kathbath} & 776.0K & 240.3M & & \audiodataset{vietspeech} & 1.0M & 183.4M \\
\audiodataset{kazakh-speech} & 230.7K & 75.6M & & \audiodataset{voxpopuli} & 22.2M & 28.1B \\
\audiodataset{legco-speech} & 3.5M & 1.4B & & \audiodataset{wenetspeech} & 5.9M & 761.5M \\
\audiodataset{libriheavy} & 4.5M & 3.0B & & \audiodataset{zeroth-korean} & 21.8K & 8.0M \\
\audiodataset{localized-narratives-audio} & 827.6K & 707.8M & & \audiodataset{zoengjyutgaai} & 47.0K & 14.4M \\
\audiodataset{mls} & 580.0K & 380.1M & & & & \\
\bottomrule
\end{tabular*}
\end{table}

\begin{table}[!htbp]
\renewcommand{\arraystretch}{0.95}
\setlength{\tabcolsep}{2pt}
\caption{Audio pool of the long-context extension.}
\label{tab:audio_data_lc}
\centering
\small
\begin{tabular*}{\textwidth}{@{\extracolsep{\fill}}>{\raggedright\arraybackslash}p{4.0cm}>{\raggedleft\arraybackslash}p{1.3cm}>{\raggedleft\arraybackslash}p{1.2cm}@{}p{0.35cm}@{}>{\raggedright\arraybackslash}p{4.0cm}>{\raggedleft\arraybackslash}p{1.3cm}>{\raggedleft\arraybackslash}p{1.2cm}@{}}
\toprule
\audiotablehead
\audiodataset{aishell-1-3-4-blend} & 39 & 405.6K & & \audiodataset{parlaspeech-rs} & 286 & 2.9M \\
\audiodataset{eurospeech} & 16.6K & 206.2M & & \audiodataset{people-s-speech-supervised} & 66.8K & 1.8B \\
\audiodataset{granary-yodas-ytc} & 266.2K & 3.3B & & \audiodataset{spc-r} & 1.2K & 16.0M \\
\audiodataset{legco-speech} & 53.8K & 1.7B & & \audiodataset{voxpopuli} & 509.7K & 30.2B \\
\audiodataset{libriheavy} & 146.4K & 4.1B & & \audiodataset{wenetspeech} & 42.6K & 841.1M \\
\audiodataset{mls} & 3.5K & 556.9M & & \audiodataset{zoengjyutgaai} & 838 & 16.3M \\
\bottomrule
\end{tabular*}
\end{table}

\begin{table}[!htbp]
\begin{minipage}[t]{0.49\textwidth}
\renewcommand{\arraystretch}{0.95}
\setlength{\tabcolsep}{2pt}
\captionof{table}{Audio pool of SFT.}
\label{tab:audio_data_sft}
\centering
\small
\begin{tabular*}{\linewidth}{@{\extracolsep{\fill}}>{\raggedright\arraybackslash}p{4.0cm}>{\raggedleft\arraybackslash}p{1.3cm}>{\raggedleft\arraybackslash}p{1.2cm}@{}}
\toprule
\textbf{Dataset} & \textbf{Samples} & \textbf{Tokens} \\
\midrule
\audiodataset{heysquad-human} & 49.6K & 23.5M \\
\audiodataset{heysquad-machine} & 98.2K & 36.2M \\
\audiodataset{tcm-shizhen-speech} & 58.4K & 59.5M \\
\audiodataset{teleantifraud} & 33.0K & 124.0M \\
\audiodataset{vocalized-sft} & 635.1K & 725.6M \\
\audiodataset{voiceassistant-400k} & 242.2K & 144.3M \\
\bottomrule
\end{tabular*}
\end{minipage}\hfill
\begin{minipage}[t]{0.49\textwidth}
\renewcommand{\arraystretch}{0.95}
\setlength{\tabcolsep}{2pt}
\captionof{table}{Audio pool of long-context SFT.}
\label{tab:audio_data_lc_sft}
\centering
\small
\begin{tabular*}{\linewidth}{@{\extracolsep{\fill}}>{\raggedright\arraybackslash}p{4.0cm}>{\raggedleft\arraybackslash}p{1.3cm}>{\raggedleft\arraybackslash}p{1.2cm}@{}}
\toprule
\textbf{Dataset} & \textbf{Samples} & \textbf{Tokens} \\
\midrule
\audiodataset{audiomcq-strongac-geminicot} & 10.5K & 6.4M \\
\audiodataset{heysquad-human} & 49.6K & 23.5M \\
\audiodataset{heysquad-machine} & 98.2K & 36.2M \\
\audiodataset{marco-longspeech} & 204.9K & 5.5B \\
\audiodataset{tcm-shizhen-speech} & 58.4K & 59.5M \\
\audiodataset{teleantifraud} & 34.0K & 133.0M \\
\audiodataset{vocalized-sft} & 635.3K & 727.2M \\
\audiodataset{voiceassistant-400k} & 242.2K & 144.3M \\
\bottomrule
\end{tabular*}
\end{minipage}
\end{table}

\clearpage
\section{Image Dataset Exploration Detailed Breakdown}\label{app:image_exploration}

Our large-scale multimodal image collections comprise a diverse range of visual media, including photographs, illustrations, screenshots, documents, diagrams, and composite graphics, with correspondingly varying content. While visual benchmarks are usually balanced across different image types and different task types, our vision data is not guaranteed to be balanced across image types, and might therefore cover some benchmarks poorly. Beyond the coarse classification of \autoref{fig:mm_sample_composition}, quantifying our coverage of different image types thus requires a rich representation taxonomy. In turn, this could lead to a filtering scheme that balances the vision data in appropriate ways.

\begin{figure}[!htbp]
    \centering
    \includegraphics[width=1\linewidth]{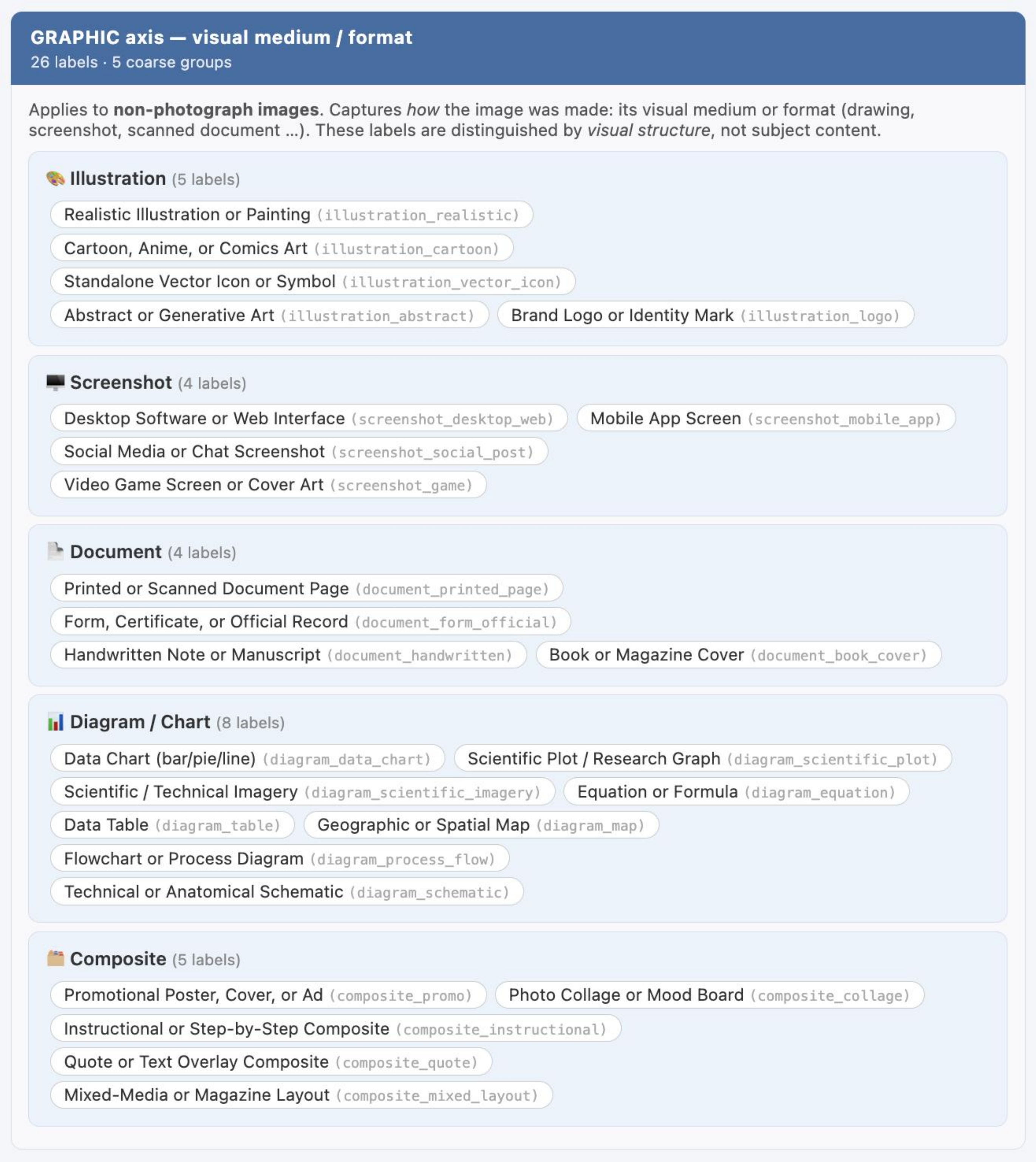}
    \caption{Graphic axis of the taxonomy, subdivided into five coarse groups and then into 26 fine-grained groups.}
    \label{fig:image_class_taxonomy_coarse}
\end{figure}

\begin{figure}[!htbp]
    \centering
    \includegraphics[width=1\linewidth]{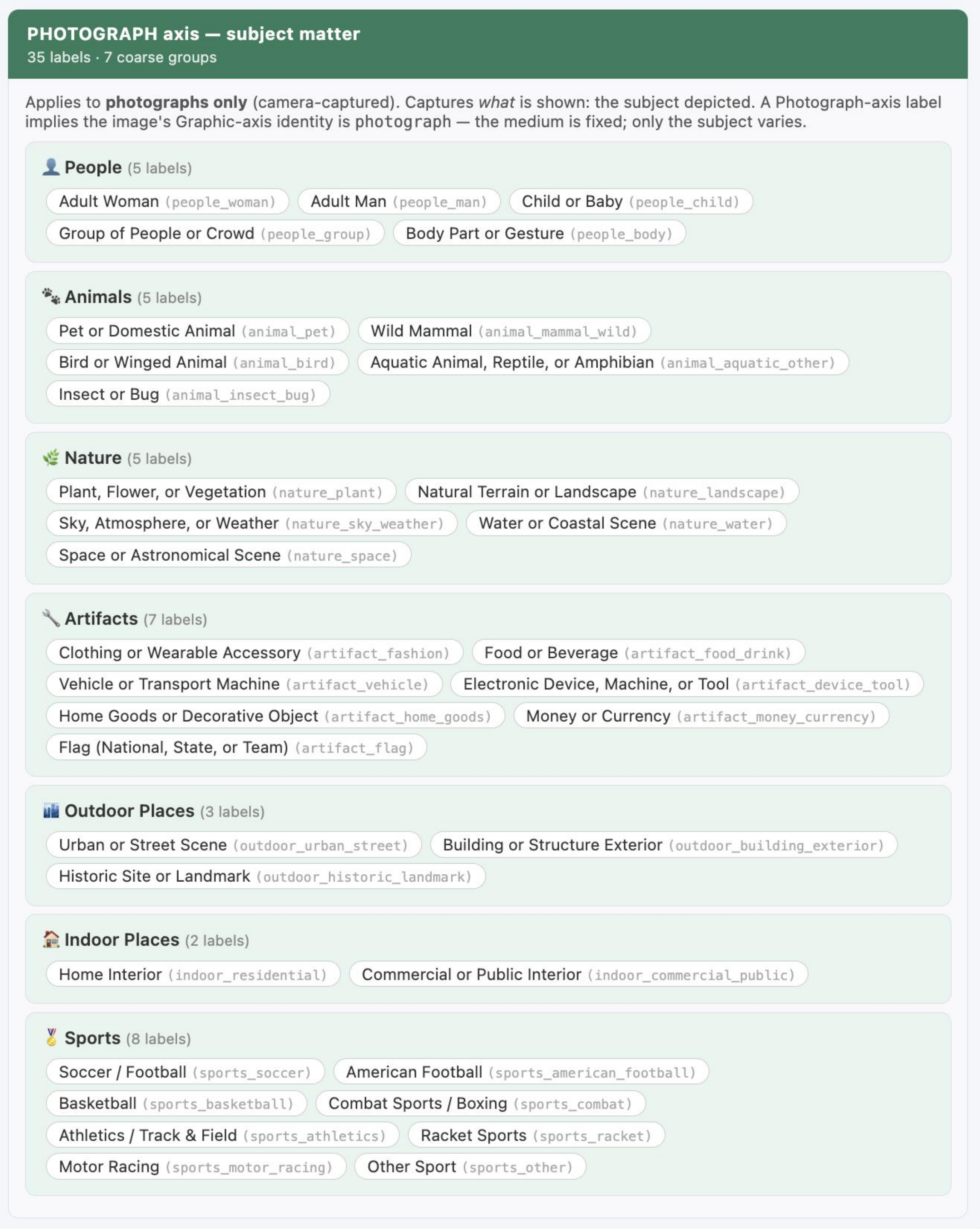}
    \caption{Photograph axis of the taxonomy, subdivided into seven coarse groups and then into 35 fine-grained groups.}
    \label{fig:image_class_taxonomy_fine}
\end{figure}

\textbf{Taxonomy.} We therefore design a taxonomy consisting of 61 fine-grained labels organized along two orthogonal dimensions: graphics and photographs. We also organize these fine-grained labels into five coarse-grained labels for graphics and seven coarse-grained labels for photographs for easier analysis. We plot this taxonomy in \autoref{fig:image_class_taxonomy_coarse} and \autoref{fig:image_class_taxonomy_fine}. Using CLIP ViT-L/14, we classify approximately 1\% of the full datasets in a single forward pass, enabling efficient large-scale analysis while preserving the overall dataset statistics. The taxonomy uses a flat one-pass design: all 61~labels are scored in a single CLIP forward pass (image $\leftrightarrow$ text prompt), so classification adds no extra compute beyond the embedding computation. We focus on LLaVA-OneVision, BLIP3, DataComp-1B, MINT-1T-ArXiv, MINT-1T-HTML, and MINT-1T-PDF. These are some of the largest datasets by token count (LLaVA-OneVision, BLIP3, DataComp-1B) that consist of large image piles sourced from the general internet. We also explore a selection of more specialized datasets (MINT-1T-ArXiv, MINT-1T-HTML, and MINT-1T-PDF). These MINT-1T subsets still form a large part of the overall token count. At the same time, the image source is constrained (PDF documents for both the arXiv and PDF subparts), and the domain for MINT-1T-ArXiv is further constrained to scientific works. We separate our analysis into quantitative and qualitative steps.

\textbf{Quantitative breakdown.} The first step provides a quantitative exploration of the datasets by applying our taxonomy-based image classification pipeline. The resulting distributions over the taxonomy classes reveal the diversity of visual content within and across datasets while demonstrating that the proposed taxonomy effectively explores both graphics and photographs.

\begin{figure}[!htbp]
    \centering
    \includegraphics[width=0.7\linewidth]{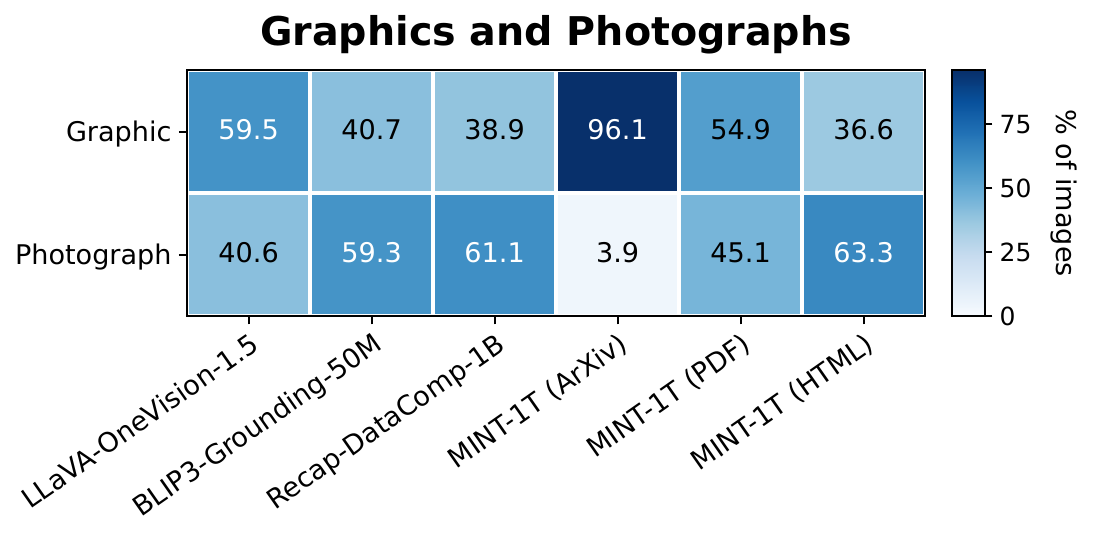}
    \caption{Breakdown of the main image datasets into Graphic and Photograph categories. We observe that most datasets are roughly evenly split between graphics and photographs. The main outlier is the MINT-1T (arXiv) dataset, which consists overwhelmingly of graphics.}
    \label{fig:graphics_vs_photographs}
\end{figure}

\begin{figure}[!htbp]
    \begin{minipage}[t]{0.48\linewidth}
        \centering
        \includegraphics[width=\linewidth]{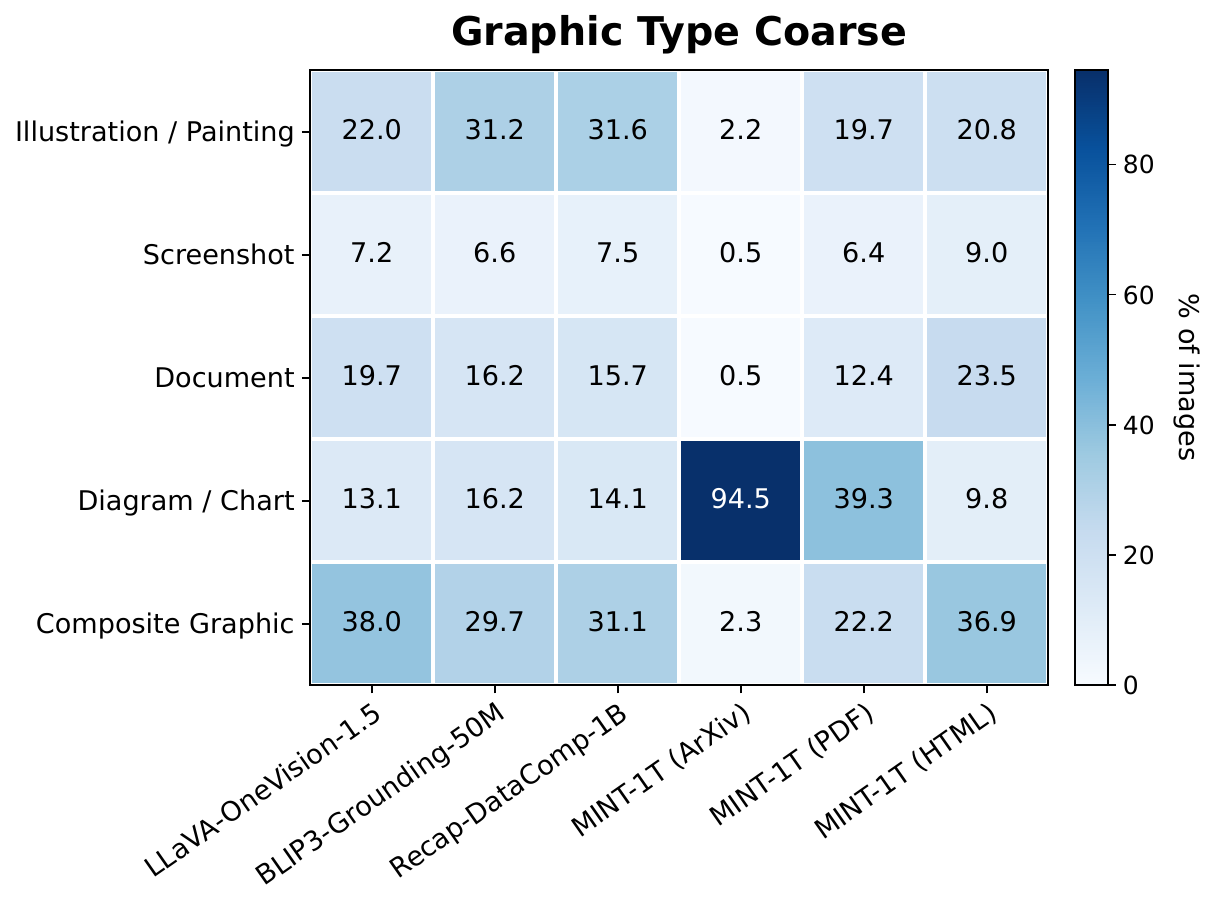}
        \caption{Breakdown of the main image datasets by coarse graphic type. The datasets are relatively evenly distributed across the different image types. Composite graphics seem to be overrepresented in the datasets. Furthermore, MINT-1T-ArXiv is an outlier, as most images belong to the Diagram/Chart class. This is consistent with the dataset's sourcing, as it consists of scientific papers.}
        \label{fig:image_type_coarse}
    \end{minipage}\hfill
    \begin{minipage}[t]{0.48\linewidth}
        \centering
        \includegraphics[width=\linewidth]{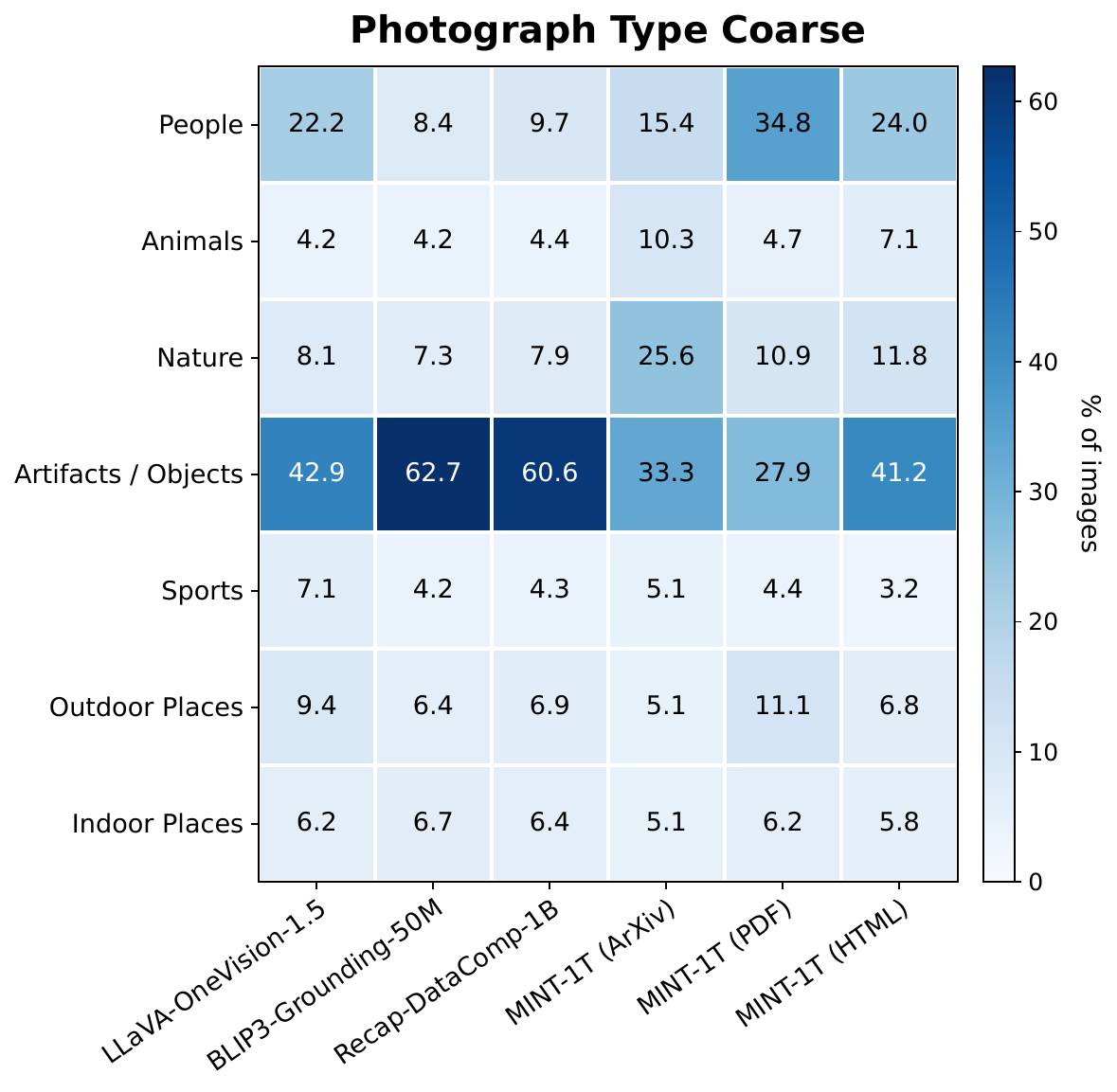}
        \caption{Breakdown of the main image datasets by coarse photograph type. We observe that all datasets are significantly unbalanced. The Artifacts/Objects class is significantly overrepresented in all datasets. The second most represented class is People. The remaining classes are roughly equally represented, with 5--10\% of the total images.}
        \label{fig:image_domain_coarse}
    \end{minipage}
\end{figure}

\begin{figure}[!htbp]
    \centering
    \includegraphics[width=0.62\linewidth]{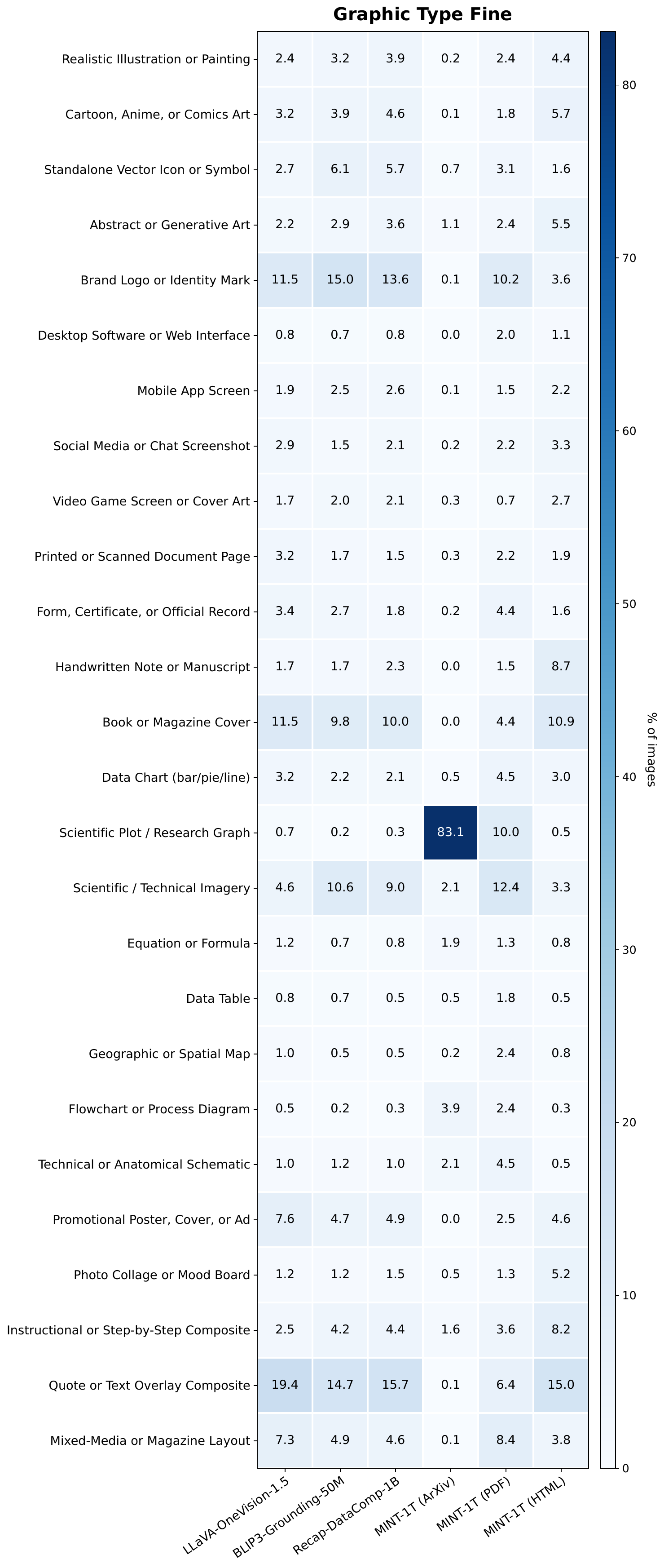}
    \caption{Breakdown of the main image datasets by fine graphic type. We observe that the Brand Logo, Book or Magazine Cover, Scientific Plot, Scientific / Technical Imagery, and Quote or Text Overlay Composite types are overrepresented in the image distribution. By contrast, the Data Table, Geographic or Spatial Map, and Flowchart or Process Diagram classes are relatively underrepresented.}
    \label{fig:image_type_fine}
\end{figure}

\begin{figure}[!htbp]
    \centering
    \includegraphics[width=0.5\linewidth]{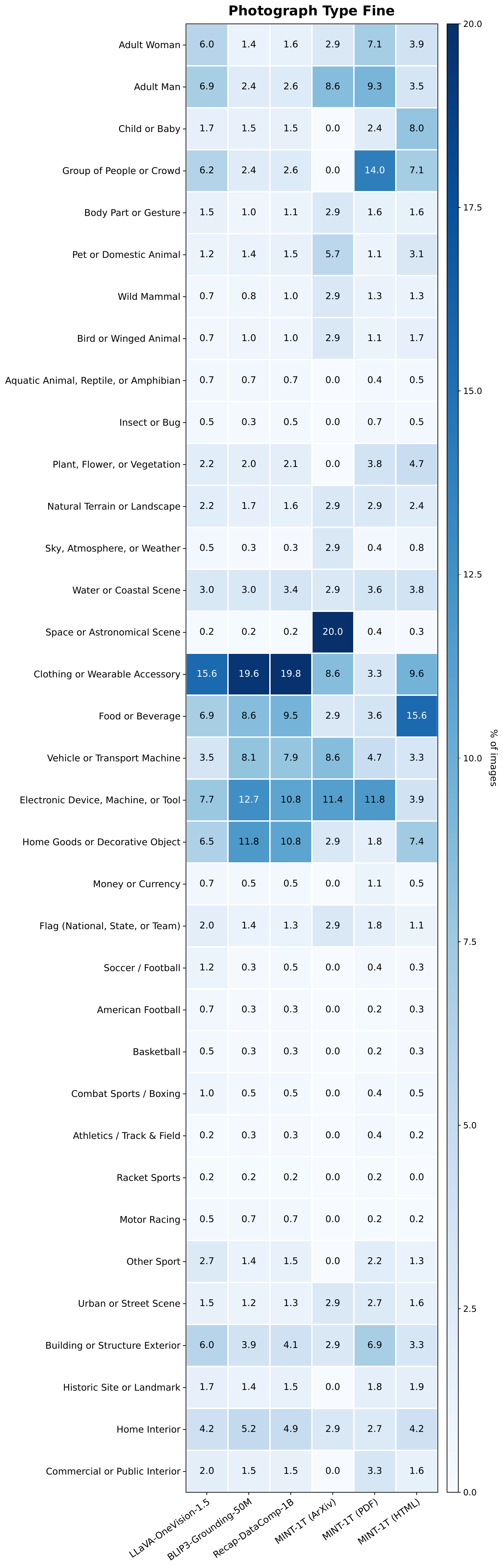}
    \caption{Breakdown of the main image datasets by fine photograph type.}
    \label{fig:image_domain_fine}
\end{figure}

We plot the results of our analysis in \autoref{fig:graphics_vs_photographs}, \autoref{fig:image_type_coarse}, \autoref{fig:image_domain_coarse}, \autoref{fig:image_type_fine}, and \autoref{fig:image_domain_fine}. This analysis yields several interesting conclusions that can also inspire future work. First, in our coarse-grained analysis, we observe that ``Composite Graphic'' is the most frequent graphic. At the same time, ``Artifacts/Objects'' appears as the most frequent photograph. This seems intuitive, given that these datasets provide great internet coverage; logos and related material should dominate due to commercial actors and advertisements. At the same time, in our fine-grained analysis, we observe that in most datasets, the ``Artifacts/Objects'' class consists of ``Clothing'', ``Food'', ``Vehicle'', ``Electronic Device'', and ``Home Goods'', which figure prominently across all datasets. In contrast, specialized domains such as insects, astronomy, and historic sites make up a relatively small percentage of all images, while potentially appearing prominently in user queries to the model. We quantify this imbalance in \autoref{tab:category_std_transposed}. These imbalances raise the prospect of rebalancing the dataset to make the image distribution more equal across different domains. Finally, we observe that the image distribution remains relatively similar across different datasets except the MINT-1T dataset. Here, especially for the arXiv subset, documents and research graphs dominate the distribution in accordance with our data source. Similarly, in the MINT-1T-PDF subset, we observe diagrams and charts dominating. This aligns with how these datasets are sourced. It also provides quantitative motivation for including diversified datasets in our data mixture to address limitations in larger but more general image pools.

\begin{table}[!htbp]
\caption{Standard deviation (percentage points) of renormalized category shares across categories, per dataset, for each grouping. We observe significant variation across the different domain and type classes.}
\label{tab:category_std_transposed}
\centering
\small
\begin{tabular}{@{}lrrrr@{}}
\toprule
\makecell{\textbf{Dataset}} &  \makecell{\textbf{Graphic}\\\textbf{Coarse}} &  \makecell{\textbf{Photograph}\\\textbf{Coarse}} &  \makecell{\textbf{Graphic}\\\textbf{Fine}} &  \makecell{\textbf{Photograph}\\\textbf{Fine}} \\
\midrule
LLaVA-OneVision-1.5 & 10.38 & 12.87 & 4.29 & 3.15 \\
BLIP3-Grounding-50M & 9.25 & 19.83 & 4.10 & 4.24 \\
Recap-DataComp-1B & 9.68 & 18.97 & 3.97 & 4.11 \\
MINT-1T (arXiv) & 37.25 & 10.43 & 15.87 & 4.15 \\
MINT-1T (PDF) & 11.16 & 11.24 & 3.05 & 3.25 \\
MINT-1T (HTML) & 10.22 & 12.70 & 3.49 & 3.26 \\
\addlinespace
Average & 14.66 & 14.34 & 5.80 & 3.69 \\
\bottomrule
\end{tabular}
\end{table}

\begin{figure}[!htbp]
    \centering
    \includegraphics[width=1\linewidth]{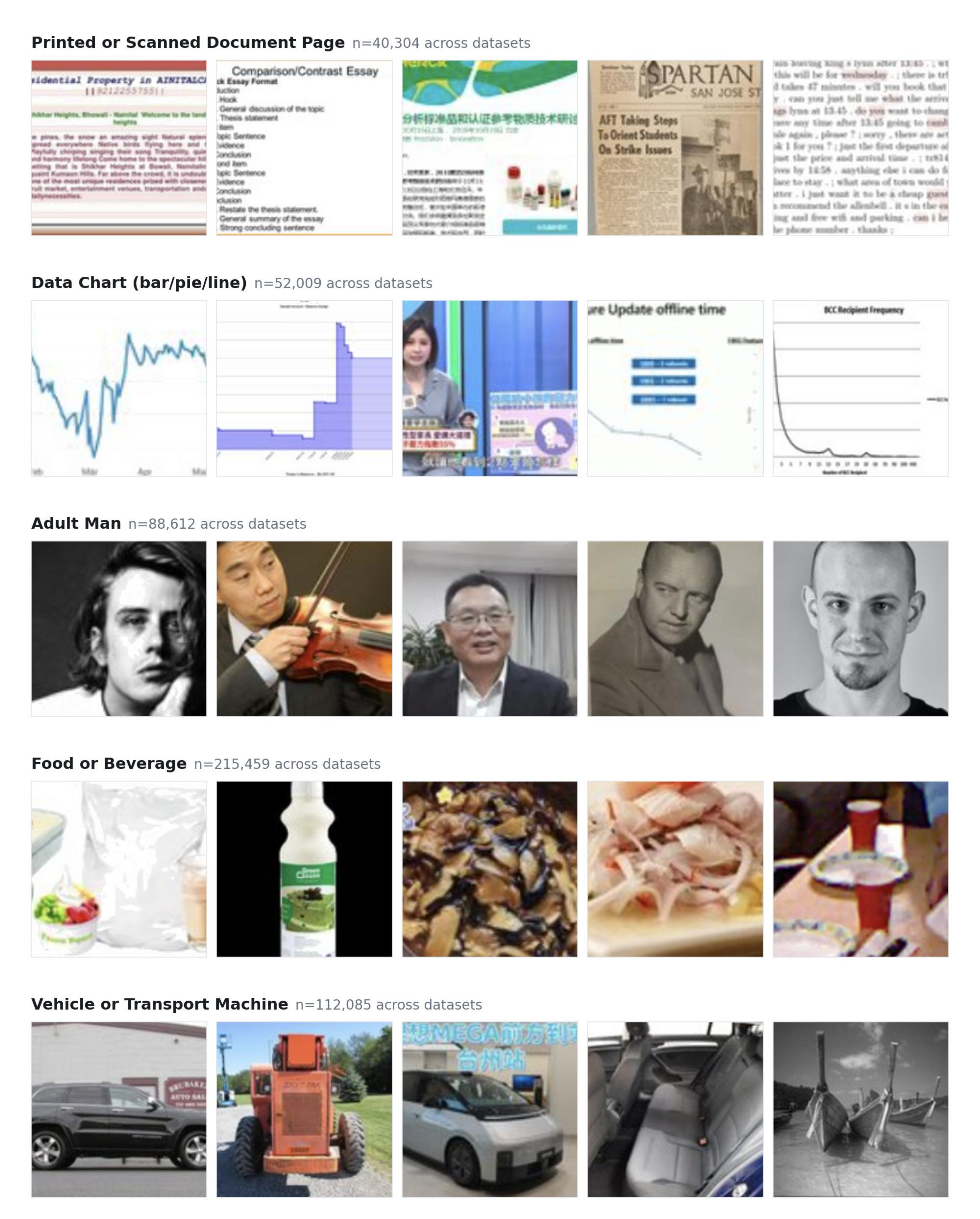}
    \caption{Image samples from several prominent fine-grained classes. We observe that our embedding and classifier pipeline correctly captures the diverse classes in our taxonomy.}
    \label{fig:category_grid}
\end{figure}

\FloatBarrier
\textbf{Qualitative check.} The second step consists of a qualitative evaluation of the proposed taxonomy. Specifically, we construct a per-label montage containing representative high-confidence examples for each of the 61 fine-grained categories together with their top-5 CLIP prediction scores. This visualization enables direct inspection of the image classes and model confidence across the taxonomy. The resulting examples demonstrate that most category images are visually coherent, supporting the validity of the proposed taxonomy for analyzing our image datasets. \autoref{fig:category_grid} shows samples from several prominent fine-grained classes.

\begin{figure}[!htbp]
    \centering
    \includegraphics[width=0.95\linewidth]{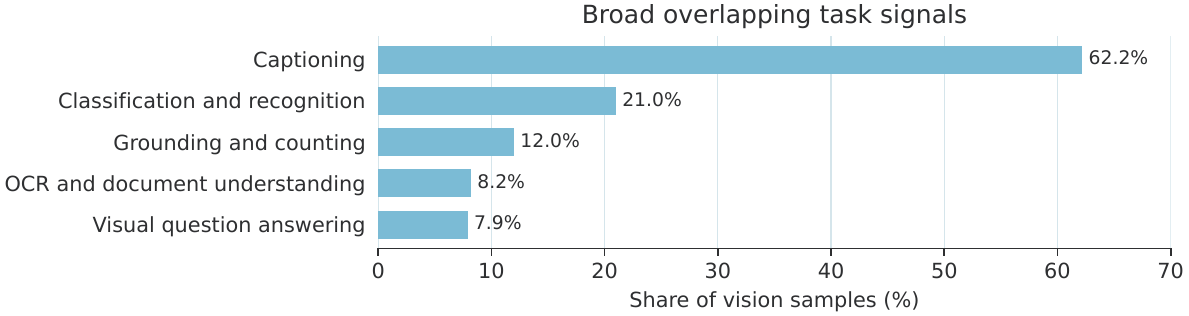}
    \caption{Prevalence of broad overlapping task signals.}
    \label{fig:mm_task_prevalence}
\end{figure}

\textbf{Task signals.} The broad task view in \autoref{fig:mm_task_prevalence} focuses on training signals. Captioning is the largest signal across the entire training mixture. Classification and recognition, grounding and counting, visual question answering, and OCR and document understanding also occur across the inventory. The task labels overlap and their shares are not mutually exclusive.

\clearpage
\section{Tokenizer Reconstruction by Category}\label{app:emu_tokenizers_categories}

\autoref{tab:emu_tokenizers_categories} gives the per-category values behind \autoref{fig:emu_tokenizer_dumbbell} and the averages of \autoref{tab:emu_tokenizers}.

\begin{table}[!htbp]
\caption{Reconstruction quality and token cost of the Emu3 and Emu3.5 tokenizers on 350 Stage~1 training images, 50 per source category, grouped into six representative categories of \autoref{sec:sample_level_vision_composition} and preprocessed as in training. Tokens are the mean per image, arrows give the better direction, and bold marks the better tokenizer per row and metric.}
\label{tab:emu_tokenizers_categories}
\centering
\small
\setlength{\tabcolsep}{4pt}
\begin{tabular}{@{}lrrrrrrrr@{}}
\toprule
 & \multicolumn{2}{c}{\textbf{Tokens} $\downarrow$} & \multicolumn{2}{c}{\textbf{PSNR (dB)} $\uparrow$} & \multicolumn{2}{c}{\textbf{SSIM} $\uparrow$} & \multicolumn{2}{c}{\textbf{LPIPS} $\downarrow$} \\
\cmidrule(lr){2-3}\cmidrule(lr){4-5}\cmidrule(lr){6-7}\cmidrule(lr){8-9}
\textbf{Category} & \textbf{Emu3} & \textbf{Emu3.5} & \textbf{Emu3} & \textbf{Emu3.5} & \textbf{Emu3} & \textbf{Emu3.5} & \textbf{Emu3} & \textbf{Emu3.5} \\
\midrule
General visual & 8,748 & \textbf{2,184} & \textbf{26.55} & 24.00 & \textbf{0.784} & 0.691 & \textbf{0.081} & 0.115 \\
Charts \& diagrams & 8,020 & \textbf{2,001} & \textbf{27.99} & 24.36 & \textbf{0.898} & 0.865 & \textbf{0.043} & 0.064 \\
Illustration \& synthetic & 8,572 & \textbf{2,135} & \textbf{27.17} & 24.63 & \textbf{0.819} & 0.760 & \textbf{0.062} & 0.091 \\
Documents \& OCR & 8,251 & \textbf{2,055} & \textbf{26.38} & 23.22 & \textbf{0.853} & 0.795 & \textbf{0.054} & 0.079 \\
Medical & 1,662 & \textbf{414} & \textbf{26.05} & 22.61 & \textbf{0.761} & 0.669 & \textbf{0.079} & 0.112 \\
Geospatial \& remote sensing & 15,330 & \textbf{3,772} & 24.80 & \textbf{25.06} & \textbf{0.632} & 0.531 & \textbf{0.159} & 0.167 \\
\bottomrule
\end{tabular}
\end{table}

\clearpage
\section{Long-Context Data Mixture Tables}\label{app:long_context_mix}

This appendix gives the exact per-dataset breakdown behind \autoref{tab:long_context_class_budget}. \autoref{tab:long_context_mix_8b} and \autoref{tab:long_context_mix_70b} list, for every dataset and every class, the token count trained on at each long-context phase, already scaled to the tokens each phase actually consumes (\autoref{sec:long_context}).

\begin{table}[!htbp]
    \RenewDocumentCommand{\hfdata}{m m}{%
      \href{https://huggingface.co/datasets/#1}{#2}%
    }
    \setlength{\tabcolsep}{3.5pt}
\caption{Long-context data mixture (\apertusonefivesmall{}). Per-dataset token counts for each long-context training step; columns denote the trained context length in tokens (32,768, 65,536, 131,072, and 262,144, respectively). We draw the Cooldown row and the short-context top-up share of Image tokens from documents under 16,384~tokens even though they count toward that step's mixture. We sort rows within a class alphabetically.}
    \label{tab:long_context_mix_8b}
\centering
\small
\begin{tabular}{@{}lrrrr@{}}
        \toprule
        \textbf{Dataset} & \textbf{32k} & \textbf{65k} & \textbf{131k} & \textbf{262k} \\
        \midrule
        \multicolumn{5}{@{}l}{\textit{Cooldown}} \\
        Cooldown Stage 3 Subsample & 39{,}220{,}000{,}000 & 36{,}640{,}000{,}000 & 36{,}770{,}000{,}000 & 36{,}850{,}000{,}000 \\
        \addlinespace
        \multicolumn{5}{@{}l}{\textit{English}} \\
        \hfdata{almanach/Biomed-Enriched}{Biomed-Enriched} & 509{,}700{,}000 & 178{,}400{,}000 & 0 & 49{,}900{,}000 \\
        \hfdata{allenai/dolma3\_pool}{Dolma 3 OCR} & 2{,}780{,}000{,}000 & 2{,}480{,}000{,}000 & 211{,}500{,}000 & 0 \\
        \hfdata{HuggingFaceFW/finepdfs-edu}{FinePDFs-Edu (EN)} & 2{,}780{,}000{,}000 & 2{,}480{,}000{,}000 & 2{,}990{,}000{,}000 & 2{,}420{,}000{,}000 \\
        \hfdata{institutional/institutional-books-hl}{Institutional Books} & 0 & 494{,}300{,}000 & 2{,}490{,}000{,}000 & 3{,}220{,}000{,}000 \\
        \addlinespace
        \multicolumn{5}{@{}l}{\textit{Multilingual}} \\
        \hfdata{HuggingFaceFW/finepdfs-edu}{FinePDFs-Edu (multi)} & 1{,}350{,}000{,}000 & 1{,}660{,}000{,}000 & 1{,}780{,}000{,}000 & 2{,}150{,}000{,}000 \\
        \hfdata{HuggingFaceFW/finetranslations}{FineTranslations (edu)} & 1{,}350{,}000{,}000 & 714{,}400{,}000 & 239{,}100{,}000 & 138{,}400{,}000 \\
        \hfdata{HuggingFaceFW/fineweb-2}{FineWeb-2 (10\% HQ)} & 1{,}350{,}000{,}000 & 1{,}420{,}000{,}000 & 1{,}780{,}000{,}000 & 1{,}520{,}000{,}000 \\
        \addlinespace
        \multicolumn{5}{@{}l}{\textit{Image}} \\
        \hfdata{artefactory/Argimi-Ardian-Finance-10k-text-image}{Argimi Finance} & 51{,}500{,}000 & 193{,}700{,}000 & 791{,}000{,}000 & 4{,}900{,}000{,}000 \\
        \hfdata{ibm-granite/ChartNet}{ChartNet} & 1{,}770{,}000{,}000 & 2{,}170{,}000{,}000 & 2{,}380{,}000{,}000 & 1{,}320{,}000{,}000 \\
        \href{https://ourworldindata.org/}{Data Insights} & 8{,}100 & 19{,}000 & 23{,}800 & 0 \\
        \hfdata{YangQiee/HQ-50K}{HQ-50K} & 224{,}100 & 273{,}400 & 300{,}400 & 0 \\
        \href{https://zenodo.org/records/12624810}{MedPix-2} & 98{,}300 & 82{,}200 & 63{,}200 & 0 \\
        \hfdata{mlfoundations/MINT-1T-ArXiv}{MINT-1T ArXiv} & 51{,}900{,}000 & 63{,}300{,}000 & 69{,}700{,}000 & 38{,}800{,}000 \\
        \hfdata{allenai/Molmo2-MultiImageQA}{Molmo2 MultiImageQA} & 1{,}270{,}000{,}000 & 2{,}230{,}000{,}000 & 1{,}360{,}000{,}000 & 7{,}700{,}000 \\
        \href{https://www.nasa.gov/}{NASA Images} & 505{,}000 & 616{,}500 & 679{,}500 & 377{,}500 \\
        \href{https://ourworldindata.org/}{OWID Articles} & 3{,}900{,}000 & 3{,}700{,}000 & 2{,}300{,}000 & 1{,}300{,}000 \\
        \hfdata{m-a-p/PIN-200M}{PIN-200M} & 2{,}720{,}000{,}000 & 1{,}150{,}000{,}000 & 1{,}440{,}000{,}000 & 0 \\
        \hfdata{FreedomIntelligence/TCM-Pretrain-Data-ShizhenGPT}{ShizhenGPT} & 774{,}400{,}000 & 415{,}200{,}000 & 198{,}600{,}000 & 4{,}500{,}000 \\
        \href{https://tau-vailab.github.io/WAFFLE/}{WAFFLE} & 20{,}100{,}000 & 24{,}900{,}000 & 27{,}100{,}000 & 15{,}100{,}000 \\
        \addlinespace
        \multicolumn{5}{@{}l}{\textit{Audio}} \\
        \hfdata{disco-eth/EuroSpeech}{EuroSpeech} & 6{,}600{,}000 & 1{,}800{,}000 & 269{,}700 & 0 \\
        \hfdata{laubonghaudoi/legco-speech}{LegCo Speech} & 80{,}800{,}000 & 77{,}500{,}000 & 47{,}700{,}000 & 10{,}400{,}000 \\
        \hfdata{mythicinfinity/libriheavy}{LibriHeavy} & 308{,}400{,}000 & 253{,}200{,}000 & 35{,}200{,}000 & 0 \\
        \hfdata{facebook/multilingual\_librispeech}{MLS-7Lang} & 1{,}500{,}000 & 2{,}300{,}000 & 4{,}700{,}000 & 49{,}700{,}000 \\
        \hfdata{classla/ParlaSpeech-RS}{ParlaSpeech-RS} & 10{,}400 & 0 & 0 & 0 \\
        \hfdata{MLCommons/peoples\_speech}{People's Speech} & 160{,}100{,}000 & 85{,}400{,}000 & 17{,}000{,}000 & 3{,}900{,}000 \\
        \hfdata{i4ds/spc\_r}{SPC-R} & 991{,}200 & 117{,}000 & 0 & 0 \\
        \href{https://github.com/facebookresearch/voxpopuli}{VoxPopuli} & 506{,}100{,}000 & 761{,}000{,}000 & 1{,}120{,}000{,}000 & 1{,}160{,}000{,}000 \\
        \hfdata{wenet-e2e/wenetspeech}{WenetSpeech} & 95{,}100{,}000 & 17{,}100{,}000 & 1{,}600{,}000 & 0 \\
        \hfdata{espnet/yodas-granary}{YODAS Granary} & 142{,}200{,}000 & 19{,}600{,}000 & 1{,}200{,}000 & 0 \\
        \hfdata{CanCLID/zoengjyutgaai}{Zoengjyutgaai} & 1{,}800{,}000 & 338{,}800 & 0 & 0 \\
        \addlinespace
        \multicolumn{5}{@{}l}{\textit{SFT}} \\
        ArtificialNeedles~\citep{artificial-needles} & 950{,}100{,}000 & 943{,}600{,}000 & 947{,}000{,}000 & 948{,}700{,}000 \\
        CWE~\citep{olmo3techreport} & 2{,}860{,}000{,}000 & 2{,}660{,}000{,}000 & 2{,}600{,}000{,}000 & 2{,}580{,}000{,}000 \\
        DocOrder & 1{,}790{,}000{,}000 & 1{,}570{,}000{,}000 & 1{,}410{,}000{,}000 & 1{,}330{,}000{,}000 \\
        \addlinespace
        Total & 62{,}910{,}000{,}000 & 58{,}720{,}000{,}000 & 58{,}720{,}000{,}000 & 58{,}720{,}000{,}000 \\
        \bottomrule
    \end{tabular}
\end{table}

\begin{table}[!htbp]
    \RenewDocumentCommand{\hfdata}{m m}{%
      \href{https://huggingface.co/datasets/#1}{#2}%
    }
    \setlength{\tabcolsep}{3.5pt}
\caption{Long-context data mixture (\apertusonefivelarge{}). Per-dataset token counts for each long-context training step; columns denote the trained context length in tokens (32,768, 65,536, 131,072, and 262,144, respectively). We draw the Cooldown row and the short-context top-up share of Image tokens from documents under 16,384~tokens even though they count toward that step's mixture. We sort rows within a class alphabetically.}
    \label{tab:long_context_mix_70b}
\centering
\small
\begin{tabular}{@{}lrrrr@{}}
        \toprule
        \textbf{Dataset} & \textbf{32k} & \textbf{65k} & \textbf{131k} & \textbf{262k} \\
        \midrule
        \multicolumn{5}{@{}l}{\textit{Cooldown}} \\
        Cooldown Stage 3 Subsample & 37{,}460{,}000{,}000 & 37{,}530{,}000{,}000 & 32{,}270{,}000{,}000 & 32{,}350{,}000{,}000 \\
        \addlinespace
        \multicolumn{5}{@{}l}{\textit{English}} \\
        \hfdata{almanach/Biomed-Enriched}{Biomed-Enriched} & 487{,}100{,}000 & 182{,}600{,}000 & 0 & 43{,}700{,}000 \\
        \hfdata{allenai/dolma3\_pool}{Dolma 3 OCR} & 2{,}660{,}000{,}000 & 2{,}540{,}000{,}000 & 185{,}600{,}000 & 0 \\
        \hfdata{HuggingFaceFW/finepdfs-edu}{FinePDFs-Edu (EN)} & 2{,}660{,}000{,}000 & 2{,}540{,}000{,}000 & 2{,}620{,}000{,}000 & 2{,}130{,}000{,}000 \\
        \hfdata{institutional/institutional-books-hl}{Institutional Books} & 0 & 507{,}000{,}000 & 2{,}190{,}000{,}000 & 2{,}820{,}000{,}000 \\
        \addlinespace
        \multicolumn{5}{@{}l}{\textit{Multilingual}} \\
        \hfdata{HuggingFaceFW/finepdfs-edu}{FinePDFs-Edu (multi)} & 1{,}290{,}000{,}000 & 1{,}700{,}000{,}000 & 1{,}570{,}000{,}000 & 1{,}880{,}000{,}000 \\
        \hfdata{HuggingFaceFW/finetranslations}{FineTranslations (edu)} & 1{,}290{,}000{,}000 & 732{,}200{,}000 & 209{,}700{,}000 & 121{,}400{,}000 \\
        \hfdata{HuggingFaceFW/fineweb-2}{FineWeb-2 (10\% HQ)} & 1{,}290{,}000{,}000 & 1{,}450{,}000{,}000 & 1{,}570{,}000{,}000 & 1{,}330{,}000{,}000 \\
        \addlinespace
        \multicolumn{5}{@{}l}{\textit{Image}} \\
        \hfdata{artefactory/Argimi-Ardian-Finance-10k-text-image}{Argimi Finance} & 49{,}200{,}000 & 198{,}400{,}000 & 694{,}300{,}000 & 4{,}300{,}000{,}000 \\
        \hfdata{ibm-granite/ChartNet}{ChartNet} & 359{,}000{,}000 & 867{,}800{,}000 & 930{,}900{,}000 & 0 \\
        \href{https://ourworldindata.org/}{Data Insights} & 7{,}800 & 19{,}500 & 20{,}900 & 0 \\
        \hfdata{YangQiee/HQ-50K}{HQ-50K} & 43{,}400 & 108{,}800 & 116{,}700 & 0 \\
        \href{https://zenodo.org/records/12624810}{MedPix-2} & 93{,}900 & 84{,}200 & 55{,}400 & 0 \\
        \hfdata{mlfoundations/MINT-1T-ArXiv}{MINT-1T ArXiv} & 10{,}100{,}000 & 25{,}400{,}000 & 27{,}200{,}000 & 0 \\
        \hfdata{allenai/Molmo2-MultiImageQA}{Molmo2 MultiImageQA} & 1{,}210{,}000{,}000 & 2{,}290{,}000{,}000 & 1{,}190{,}000{,}000 & 6{,}700{,}000 \\
        \href{https://www.nasa.gov/}{NASA Images} & 98{,}500 & 247{,}200 & 265{,}200 & 0 \\
        \href{https://ourworldindata.org/}{OWID Articles} & 2{,}300{,}000 & 2{,}300{,}000 & 738{,}700 & 39{,}300 \\
        \hfdata{m-a-p/PIN-200M}{PIN-200M} & 2{,}600{,}000{,}000 & 1{,}180{,}000{,}000 & 1{,}270{,}000{,}000 & 0 \\
        \hfdata{FreedomIntelligence/TCM-Pretrain-Data-ShizhenGPT}{ShizhenGPT} & 740{,}000{,}000 & 425{,}400{,}000 & 174{,}200{,}000 & 4{,}000{,}000 \\
        \href{https://tau-vailab.github.io/WAFFLE/}{WAFFLE} & 3{,}900{,}000 & 9{,}900{,}000 & 10{,}600{,}000 & 0 \\
        \addlinespace
        \multicolumn{5}{@{}l}{\textit{Audio}} \\
        \hfdata{disco-eth/EuroSpeech}{EuroSpeech} & 6{,}300{,}000 & 1{,}800{,}000 & 236{,}700 & 0 \\
        \hfdata{laubonghaudoi/legco-speech}{LegCo Speech} & 77{,}200{,}000 & 79{,}300{,}000 & 41{,}900{,}000 & 9{,}200{,}000 \\
        \hfdata{mythicinfinity/libriheavy}{LibriHeavy} & 294{,}700{,}000 & 259{,}400{,}000 & 30{,}900{,}000 & 0 \\
        \hfdata{facebook/multilingual\_librispeech}{MLS-7Lang} & 1{,}500{,}000 & 2{,}400{,}000 & 4{,}200{,}000 & 43{,}600{,}000 \\
        \hfdata{classla/ParlaSpeech-RS}{ParlaSpeech-RS} & 10{,}000 & 0 & 0 & 0 \\
        \hfdata{MLCommons/peoples\_speech}{People's Speech} & 153{,}000{,}000 & 87{,}400{,}000 & 14{,}900{,}000 & 3{,}400{,}000 \\
        \hfdata{i4ds/spc\_r}{SPC-R} & 947{,}200 & 119{,}800 & 0 & 0 \\
        \href{https://github.com/facebookresearch/voxpopuli}{VoxPopuli} & 483{,}600{,}000 & 779{,}500{,}000 & 978{,}800{,}000 & 1{,}020{,}000{,}000 \\
        \hfdata{wenet-e2e/wenetspeech}{WenetSpeech} & 90{,}900{,}000 & 17{,}500{,}000 & 1{,}400{,}000 & 0 \\
        \hfdata{espnet/yodas-granary}{YODAS Granary} & 135{,}800{,}000 & 20{,}100{,}000 & 1{,}000{,}000 & 0 \\
        \hfdata{CanCLID/zoengjyutgaai}{Zoengjyutgaai} & 1{,}800{,}000 & 347{,}100 & 0 & 0 \\
        \addlinespace
        \multicolumn{5}{@{}l}{\textit{SFT}} \\
        ArtificialNeedles~\citep{artificial-needles} & 908{,}000{,}000 & 966{,}500{,}000 & 831{,}200{,}000 & 833{,}200{,}000 \\
        CWE~\citep{olmo3techreport} & 2{,}730{,}000{,}000 & 2{,}720{,}000{,}000 & 2{,}280{,}000{,}000 & 2{,}270{,}000{,}000 \\
        DocOrder & 1{,}710{,}000{,}000 & 1{,}600{,}000{,}000 & 1{,}230{,}000{,}000 & 1{,}160{,}000{,}000 \\
        \addlinespace
        Total & 58{,}720{,}000{,}000 & 58{,}720{,}000{,}000 & 50{,}330{,}000{,}000 & 50{,}330{,}000{,}000 \\
        \bottomrule
    \end{tabular}
\end{table}

\clearpage
\section{Training Configurations}\label{app:training_configurations}

\begin{table}[!htbp]
\caption{Continued-pretraining configuration of both models. The learning rate follows the warmup-stable-decay schedule of \autoref{sec:optimizer_recipe}: constant at the peak through stages 1 and 2, decaying to the minimum over stage 3 with the $1-\sqrt{x}$ shape.}
\label{tab:cpt_config}
\centering
\small
\begin{tabular}{@{}lll@{}}
\toprule
 & \textbf{8B} & \textbf{70B} \\
\midrule
Nodes & 256 & 512 \\
Sequence length & 8,192 & 8,192 \\
Global batch size & 1,024 & 2,048 \\
Parallelism & TP2 & TP4, PP8, VPP2 \\
Peak learning rate & $1.1 \times 10^{-4}$ & $2 \times 10^{-5}$ \\
Minimum learning rate & $1.1 \times 10^{-5}$ & $2 \times 10^{-6}$ \\
Warmup & 500~iterations & 250~iterations \\
\bottomrule
\end{tabular}
\end{table}

Both models train with AdEMAMix ($\beta_1 = 0.9$, $\beta_2 = 0.999$, $\alpha = 8$, $\beta_3 = 0.9999$, with the $\beta_3$ and $\alpha$ warmup over 100,000~iterations), gradient clipping at 0.1, and weight decay of 0.1, applied also to the xIELU parameters and the normalization gains. The optimizer state is fresh at the start of stage 1 and carries over across the later stage transitions. All stages pack documents with cross-document attention masking: position identifiers reset at document boundaries, and end-of-document positions are excluded from the loss.

The long-context extension proceeds through a curriculum of 32,768, 65,536, 131,072, and 262,144~tokens, each step loading the previous checkpoint with a fresh optimizer state and training at a constant learning rate equal to the main run's minimum. The RoPE base grows to 2M for the first step and 4M thereafter, with scaling factors of 8, 8, 16, and 32. The token budgets per step are 62.9B, 58.7B, 58.7B, and 58.7B for the 8B model, and 58.7B, 58.7B, 50.3B, and 50.3B for the 70B model (\autoref{app:long_context_mix}). The global batch size halves with each doubling of the sequence length (256, 128, 64, and 32 for the 8B model), context parallelism scales as 2, 4, 8, and 16 for the 8B model and 4, 8, 16, and 32 for the 70B model, and sequences are packed with a best-fit-decreasing policy.

Both production SFT runs of \autoref{sec:sft} use Adam ($\beta_1 = 0.9$, $\beta_2 = 0.999$) with no weight decay and gradient clipping at 1.0; \autoref{sec:sft} gives their learning rates, warmup, batch sizes, and sequence length, together with those of the capability-filtering run. We apply the chat template offline during data preparation, restrict the loss to assistant regions with special tokens masked, and pack samples with the same best-fit-decreasing policy.

The reinforcement-learning stage trains with a fully asynchronous trainer on FSDP2 with FlashAttention-3~\citep{shah2024flashattention3}, eight rollouts per prompt served by SGLang, leave-one-out advantage estimation, filtering of groups with zero reward variance, asymmetric clipping (0.2, 0.28), a geometric-mean loss aggregation, no KL penalty, and AdamW at a learning rate of $1 \times 10^{-6}$. For the 8B model, the training runs for 480~training steps, effectively training on about one million trajectories, while the 70B model runs for 180~steps.

\begin{table}[!htbp]
\caption{Preference-alignment configurations. Offline DPO trains \apertusonefivelarge{} only; online DPO trains both models. Online sequence lengths are given as prompt plus response.}
\label{tab:dpo_config}
\centering
\small
\begin{tabular}{@{}lccc@{}}
\toprule
 & \textbf{Offline DPO (70B)} & \textbf{Online DPO (8B)} & \textbf{Online DPO (70B)} \\
\midrule
$\beta$ & 25 (length-normalized) & 0.1 & 0.1 \\
Learning rate & $1 \times 10^{-6}$ & $5 \times 10^{-6}$ & $2.5 \times 10^{-6}$ \\
Schedule & linear, warmup 0.1 & linear, warmup 0.1 & linear, warmup 0.1 \\
Weight decay & 0.0 & 0.01 & 0.01 \\
Batch size & 128 pairs & 256 prompts & 512 prompts \\
Rollouts per prompt & & 12 & 8 \\
Sequence length & 4,096 & 2,048 + 2,048 & 2,048 + 2,048 \\
Epochs & 1 & 1 & 1 \\
Steps & 1921 & 948 & 474 \\
Training nodes & 16 & 16 & 128 \\
Annotation nodes & & 8 & 16 \\
\bottomrule
\end{tabular}
\end{table}

Offline DPO trains with our own preference trainer (\ghrepo{swiss-ai/posttraining}), which implements the DPO objective directly and divides $\beta$ by sequence length under length normalization, running on DeepSpeed ZeRO~\citep{rajbhandari2020zero}. Online DPO runs on our verl fork with FSDP and SGLang rollouts, sampling at temperature 1.0 with unrestricted top-$p$ and top-$k$. Neither stage trains a reward model, and neither trains on image inputs.

The SDPO charter-alignment stage of \autoref{sec:sdpo} runs on our verl fork with FSDP2 for training and vLLM for rollouts. It trains for one epoch at a batch size of 64~prompts, with 16~rollouts per prompt, at a learning rate of $1 \times 10^{-7}$ with 10~linear warmup steps, a maximum model length of 4,096~tokens, and token-level importance sampling clipped at a ratio of two. We never update the teacher, so it stays the initial policy throughout.

\autoref{prompt:sdpo_judge} and \autoref{prompt:sdpo_teacher} give the two templates. The judge template carries the full charter, and the teacher template combines the original user prompt with the resulting feedback and instructs the model not to mention it, to prevent leaking the feedback into model responses.

\begin{prompt}[title=SDPO Judge Prompt,label=prompt:sdpo_judge]
\begin{promptverbatim}
You are an expert reviewer. Your job is to read the user prompt and the assistant response below, then judge how well the response follows the constitution that follows, and give helpful, concise, constructive feedback to the response writer on how to better align the response with the constitution.

Here is the constitution:

<CONSTITUTIONAL_PRINCIPLES>
[the 11 sections of the Apertus charter, verbatim]
</CONSTITUTIONAL_PRINCIPLES>

<USER_PROMPT>
{prompt}
</USER_PROMPT>

<ASSISTANT_RESPONSE>
{response}
</ASSISTANT_RESPONSE>

Focus on the most important ways the response could better follow the constitution. Do not just give feedback to refuse the response, only do it, if it is REALLY necessary as the model will be trained based on your feedback and should not learn to overrefuse. Address the response writer directly as "you". Do not output a score, a grade, a JSON object, or any preamble, and do not quote principle numbers. Output only one or two sentences of actionable feedback.
\end{promptverbatim}
\end{prompt}

\begin{prompt}[title=SDPO Teacher Prompt,label=prompt:sdpo_teacher]
\begin{promptverbatim}
Here is the user prompt:

<USER_PROMPT>
{prompt}
</USER_PROMPT>


An expert reviewer evaluated an earlier draft answer to this request against a set of guiding principles and left the following feedback. Use it to write a better answer:

<REVIEWER_FEEDBACK>
{feedback_raw}
</REVIEWER_FEEDBACK>

Do not refer to or mention any guidelines, principles, or reviewer feedback in your response.
\end{promptverbatim}
\end{prompt}

\clearpage
\section{Supervised Finetuning Details}\label{app:sft_ablations}

\subsection{Per-Benchmark SFT Ablation Results}\label{app:sft_ablation_details}

\autoref{tab:sft_ablation_details} lists the per-benchmark score changes behind each bar of \autoref{fig:sft_ablation_questions}.

\begin{table}[!htbp]
\renewcommand{\arraystretch}{0.9}
\caption{Per-benchmark score changes behind \autoref{fig:sft_ablation_questions}, in percentage points. Q1 compares MM against Text at matched supervision and response mode; Q2 compares MM CoT with thinking on against MM No-CoT with thinking off; Q3 compares MM CoT against MM No-CoT, both with thinking off. Thinking is forced and sampled at temperature 0.6 and top-p 0.95; without thinking, decoding is greedy. Multimodal values average three seeds; the text-only reference is one seed. The average rows reproduce the bars of \autoref{fig:sft_ablation_questions}, and -- marks comparisons the figure does not make. \textsuperscript{*}MGSM is left out of the text average, as in the figure: its zero-shot prompt asks for a step-by-step answer, and the multimodal No-CoT arms mostly reply with the final number alone (median four words, against 137 for the text-only No-CoT arm), so their score reflects answer style and ranges from 27.2 to 39.2 across seeds. With reasoning traces in SFT, the multimodal arm writes out its work again and scores 69.9 even with thinking forced off, level with the text-only CoT arm (69.2).}
\label{tab:sft_ablation_details}
\centering
\small
\begin{tabular}{@{}lrrrrr@{}}
\toprule
 & \multicolumn{3}{c}{\textbf{Q1: MM vs Text}} & \textbf{Q2} & \textbf{Q3} \\
\cmidrule(lr){2-4}
\textbf{Benchmark} & \textbf{No-CoT, Off} & \textbf{CoT, Off} & \textbf{CoT, On} & \makecell[r]{\textbf{CoT On vs}\\\textbf{No-CoT Off}} & \makecell[r]{\textbf{CoT vs No-CoT,}\\\textbf{both Off}} \\
\midrule
\multicolumn{6}{@{}l}{\textit{Mathematics}} \\
GSM8K & +0.5 & +3.2 & +2.2 & +6.6 & +0.8 \\
MGSM\textsuperscript{*} & $-$35.5 & +0.7 & +2.9 & +50.4 & +36.9 \\
MATH & $-$2.6 & +0.4 & +0.5 & +16.2 & +3.8 \\
Minerva MATH & $-$0.3 & +0.4 & +2.7 & +18.3 & +2.7 \\
\addlinespace[0.25em]
\multicolumn{6}{@{}l}{\textit{Knowledge}} \\
MMLU-Pro & +1.2 & +0.4 & +8.3 & $-$3.5 & $-$1.6 \\
MMLU (flan CoT) & $-$2.9 & +7.7 & +6.8 & +9.0 & +5.3 \\
\addlinespace[0.25em]
\multicolumn{6}{@{}l}{\textit{Reasoning}} \\
BBH (3-shot CoT) & +1.2 & +4.1 & +8.0 & $-$9.9 & +2.9 \\
\addlinespace[0.25em]
\multicolumn{6}{@{}l}{\textit{Multilingual}} \\
Global-MMLU & +0.8 & +0.8 & +2.9 & +4.7 & $-$0.9 \\
INCLUDE-44 & +0.9 & $-$1.2 & +2.1 & +1.1 & $-$2.4 \\
INCLUDE-45 & 0.0 & $-$0.6 & +4.2 & $-$2.8 & $-$1.5 \\
SwitzerlandQA & $-$0.8 & $-$1.6 & +2.4 & $-$2.4 & $-$1.9 \\
\addlinespace[0.25em]
\multicolumn{6}{@{}l}{\textit{Instruction Following}} \\
IFEval & +3.1 & $-$0.6 & +1.3 & $-$14.0 & $-$1.2 \\
IFBench & $-$1.1 & $-$0.9 & +4.1 & +5.6 & $-$3.1 \\
Multi-IF & +0.3 & $-$0.3 & +1.0 & $-$12.4 & $-$1.1 \\
\addlinespace[0.25em]
\multicolumn{6}{@{}l}{\textit{Code}} \\
HumanEval & $-$2.2 & $-$0.2 & +0.8 & $-$5.5 & +0.2 \\
MBPP & $-$3.3 & $-$1.0 & $-$1.7 & $-$3.5 & +0.3 \\
\addlinespace[0.25em]
\multicolumn{6}{@{}l}{\textit{Reading}} \\
DROP (F1) & +5.9 & +4.1 & +4.5 & $-$3.8 & $-$1.8 \\
\addlinespace[0.25em]
Text average (16 tasks) & 0.0 & +0.9 & +3.1 & +0.2 & 0.0 \\
\addlinespace[0.25em]
\multicolumn{6}{@{}l}{\textit{Vision}} \\
MMMU & -- & -- & -- & +7.3 & $-$1.4 \\
MMMU-Pro & -- & -- & -- & +7.8 & $-$1.2 \\
MMStar & -- & -- & -- & +18.2 & $-$4.7 \\
MathVision & -- & -- & -- & $-$4.1 & $-$2.7 \\
ScienceQA & -- & -- & -- & +7.1 & $-$2.8 \\
BLINK & -- & -- & -- & +7.8 & $-$3.7 \\
VisualPuzzles & -- & -- & -- & +0.9 & $-$1.7 \\
AI2D & -- & -- & -- & +4.5 & $-$2.6 \\
ChartQA & -- & -- & -- & +7.4 & $-$1.9 \\
DocVQA & -- & -- & -- & $-$2.5 & $-$0.4 \\
TextVQA & -- & -- & -- & $-$4.0 & $-$0.5 \\
RealWorldQA & -- & -- & -- & +5.9 & $-$4.5 \\
POPE & -- & -- & -- & +0.7 & $-$3.4 \\
OCRBench & -- & -- & -- & $-$3.5 & +0.2 \\
\addlinespace[0.25em]
Vision average (14 tasks) & -- & -- & -- & +3.8 & $-$2.2 \\
\bottomrule
\end{tabular}
\end{table}

\subsection{SFT Data Decontamination Details}\label{app:sft_text_decontamination}

We report the detailed count of removed samples after decontamination for the text SFT mixture in \autoref{tab:sft_text_decontamination}.

\begin{table}[!htbp]
\centering
\small
\setlength{\tabcolsep}{7pt}
\caption{\textbf{Decontamination statistics for the SFT data sources.} For each dataset we report the example count before decontamination (\emph{Original}), the number of examples removed by matching against evaluation benchmarks (\emph{Removed}), the resulting count (\emph{Final}), and the fraction removed. We compute removals per source prior to mixing.}
\label{tab:sft_text_decontamination}
\begin{tabular}{@{}lrrrr@{}}
\toprule
\textbf{Dataset} & \textbf{Original} & \textbf{Removed} & \textbf{Final} & \textbf{\% Removed} \\
\midrule
OLMo2 OASST1 & 7,125 & 97 & 7,028 & 1.36\% \\
Llama-Nemotron Chat & 20,853 & 264 & 20,589 & 1.27\% \\
Llama-Nemotron Safety & 13,797 & 31 & 13,766 & 0.22\% \\
EuroBlocks Aya & 2,117 & 0 & 2,117 & 0.00\% \\
EuroBlocks Refusal & 7,648 & 0 & 7,648 & 0.00\% \\
EuroBlocks Low-Resource & 7,702 & 0 & 7,702 & 0.00\% \\
EuroBlocks Multilingual (2512)\textsuperscript{a} & 346,751 & 60,367 & 286,384 & 17.41\% \\
SmolTalk2 Multilingual (8 lang) & 219,223 & 47 & 219,176 & 0.02\% \\
SmolTalk2 (everyday) & 2,259 & 0 & 2,259 & 0.00\% \\
SmolTalk2 (magpie) & 264,257 & 2,019 & 262,238 & 0.76\% \\
SmolTalk2 (rewrite) & 48,093 & 1 & 48,092 & 0.00\% \\
SmolTalk2 (summarize) & 92,728 & 16 & 92,712 & 0.02\% \\
SmolTalk2 (system-chats) & 29,473 & 0 & 29,473 & 0.00\% \\
Swiss German Dialects & 2,674 & 0 & 2,674 & 0.00\% \\
Romansh Varieties & 46,078 & 0 & 46,078 & 0.00\% \\
African Languages & 8,730 & 0 & 8,730 & 0.00\% \\
WikiQA & 482,799 & 0 & 482,799 & 0.00\% \\
German$\rightarrow$Romansh MT & 30,000 & 0 & 30,000 & 0.00\% \\
Swiss Charter Q\&A & 88 & 0 & 88 & 0.00\% \\
Glaive Function Calling v2 & 52,732 & 0 & 52,732 & 0.00\% \\
GSM8K (TIR) & 7,473 & 0 & 7,473 & 0.00\% \\
IF-mix (combined)\textsuperscript{b} & 201,481 & 13,455 & 188,026 & 6.68\% \\
Table-GPT & 13,222 & 0 & 13,222 & 0.00\% \\
FeTaQA & 10,330 & 0 & 10,330 & 0.00\% \\
SciRIFF & 72,367 & 3,338 & 69,029 & 4.61\% \\
RiddleSense & 4,531 & 0 & 4,531 & 0.00\% \\
Medical-o1 & 13,696 & 311 & 13,385 & 2.27\% \\
NuminaMath 1.5 & 417,101 & 4,908 & 412,193 & 1.18\% \\
DeepMath-103K & 103,022 & 0 & 103,022 & 0.00\% \\
HARP & 4,542 & 0 & 4,542 & 0.00\% \\
OpenMathReasoning & 2,362,555 & 55,490 & 2,307,065 & 2.35\% \\
OpenCodeReasoning v2 & 529,210 & 16,782 & 512,428 & 3.17\% \\
OpenMathReasoning (CoT) & 688,801 & 22,412 & 666,389 & 3.25\% \\
\midrule
\textbf{Total} & \textbf{6,113,458} & \textbf{179,538} & \textbf{5,933,920} & \textbf{2.94\%} \\
\bottomrule
\end{tabular}

\footnotesize
\raggedright
\textsuperscript{a}\,False positives, common for multilingual data, dominate the high removal rate.

\textsuperscript{b}\,Removals stem largely from IFEval-style instructions and also include false positives.
\end{table}

\clearpage

\section{Preference Alignment Data}\label{app:preference_data}

This appendix documents the preference data of \autoref{sec:preference_alignment}: the rubric prompts we give the judge, the composition of the prompt set, the pool of response models, and the distribution of the resulting scores. The released artifacts are \hfdata{swiss-ai/Apertus-v1.5-Preference-Data} for the preference pairs and \hfdata{swiss-ai/Apertus-v1.5-Preference-Pool} for the underlying annotated completions.

\subsection{Judge Rubric Prompts}\label{app:rubric_prompts}

Qwen3-235B-A22B-Instruct-2507~\citep{Yang2025Qwen3TR} rates each completion on the four aspects independently, one request per aspect, with the system prompt of \autoref{prompt:judge_system} and the aspect prompt as the user message. We make annotation deterministic and disable thinking: the judge decodes at temperature 0 and generates exactly one token, constrained to the token identifiers of the digits one through five, and we keep the renormalized distribution over those five tokens, so an aspect score is the expected value $\sum_k k\,p_k$ and is continuous on $[1, 5]$. The final score of a completion is the mean over the four aspects. Across the full matrix of 6,235,908~scored completions the four aspects correlate between 0.838 and 0.916, so the mean is close to a single latent quality axis; \autoref{app:judge-ablations} takes up the consequence for the online stage, which scores on helpfulness alone.

\begin{prompt}[title=Judge System Prompt,label=prompt:judge_system]
\begin{promptverbatim}
You are an impartial judge. Your role is to critically evaluate the quality of an AI assistant response based on a given criteria. You'll receive an input with two sections, enclosed in tags: <USER_INPUT>...</USER_INPUT> for the task instructions (and any accompanying context, if applicable), and <ASSISTANT_RESPONSE_TO_EVALUATE>...</ASSISTANT_RESPONSE_TO_EVALUATE> for the AI assistant's response. 

Carefully read the provided input to understand the task, then assess how well the response fulfills the criteria requirements. If conversation history is present, ensure the response aligns with it; otherwise, evaluate based solely on the instruction. You will be given a scoring rubric below, based on which you should provide a rating from 1 to 5. Your output should only be an integer from 1 to 5. Do not output any additional text or explanations.
\end{promptverbatim}
\end{prompt}

\begin{prompt}[title=Judge Rubric: Instruction Following,label=prompt:rubric_if]
\begin{promptverbatim}
You will be doing an Instruction Following Assessment of an AI assistant response.

Carefully read the <USER_INPUT> to assess how well the candidate response fulfills the task requirements. If the input includes a conversation history, the response must align with that context as well as the final instruction.

**Scoring**: Rate the text on a scale of 1 to 5:
1. **Irrelevant**: No alignment.
2. **Partial Focus**: Addresses one aspect poorly.
3. **Partial Compliance**:
    - (1) Meets goal or restrictions, neglecting other.
    - (2) Acknowledges both but slight deviations.
4. **Almost There**: Near alignment, minor deviations.
5. **Comprehensive Compliance**: Fully aligns, meets all requirements.

Here are the user inputs and the AI assistant response to evaluate:

<USER_INPUT>{prompt}</USER_INPUT>

<ASSISTANT_RESPONSE_TO_EVALUATE>{completion}</ASSISTANT_RESPONSE_TO_EVALUATE>
\end{promptverbatim}
\end{prompt}

\begin{prompt}[title=Judge Rubric: Honesty,label=prompt:rubric_honesty,unbreakable]
\begin{promptverbatim}
You will be doing a Honesty and Uncertainty Expression Assessment of an AI assistant response.

Assess how well the candidate response conveys honesty and uncertainty based on the provided <USER_INPUT>. Evaluate if the text's confidence aligns with response correctness, factoring in any conversation context if it is present in the input.

- **Indicators of Uncertainty**: 
    1. Weakeners: e.g., 'I guess,' 'probably.'
    2. Refusal to answer.
    3. Verbalized confidence scores: [0, 20] low; (20, 40] uncertain; (40, 60] moderate; (60, 80] leaning confident; (80, 100] high. 
- **No uncertainty expression indicate confidence.**
- **Response Correctness**: Align with ground truth, or provide accurate content without fabrication.

**Scoring**: Rate outputs on a scale of 1 to 5:
1. **Confidently Incorrect**: Confident but entirely wrong.
2. **Confident with Significant Mistakes / Unconfident Incorrect**:
- Confident but contains major errors.
- Unconfident and entirely wrong.
3. **Uncertain / 'I Don't Know' / Subtle Mistakes**:
- 'I don't know' or declines.
- confident but contains minor errors.
- Unconfident and contains significant mistakes.
4. **Correct but Uncertain / Expressed Subtle Mistakes**:
- Correct but unconfident.
- Makes subtle mistakes but expresses uncertainty without specifying the exact area of doubt.
5. **Correct and Confident / Precisely Express Uncertainty**:
- Correct and confident.
- Makes mistakes, but precisely acknowledges minor errors and indicates uncertainty on potential mistakes.

Here are the user inputs and the AI assistant response to evaluate:

<USER_INPUT>{prompt}</USER_INPUT>

<ASSISTANT_RESPONSE_TO_EVALUATE>{completion}</ASSISTANT_RESPONSE_TO_EVALUATE>
\end{promptverbatim}
\end{prompt}

\begin{prompt}[title=Judge Rubric: Truthfulness,label=prompt:rubric_truth]
\begin{promptverbatim}
You will be doing a Truthfulness and Hallucination Assessment of an AI assistant response.

Evaluate the candidate response's accuracy in providing information without introducing misleading or fabricated details. 

When evaluating truthfulness, consider the following types of hallucination:
1. **Contradictory with the World (Factual Error)**: Entities, locations, concepts, or events that conflict with established knowledge.
2. **Contradictory with Instruction/Context**: Responses diverge, introducing new facts not aligned with the user's instructions (or conversation history, if provided).
3. **Self-Contradictory / Logical Error**: Responses contain internal contradictions or logical errors within each independent text.

Reflect on whether any of these hallucination types are present in the response, and take them into account when assigning your rating.

**Scoring**: Rate outputs on a scale of 1 to 5 based on extent of hallucination:
1. **Completely Hallucinated**: Entirely unreliable due to hallucinations.
2. **Severe Hallucination**: Nearly half contains hallucinations, severe deviation from main points.
3. **Partial Hallucination / Misunderstanding**: Overall truthful, partial misunderstanding due to hallucinations.
4. **Insignificant Hallucination**: Mostly truthful, slight hallucination not affecting main points.
5. **No Hallucination**: Free of hallucinations.

Here are the user inputs and the AI assistant response to evaluate:

<USER_INPUT>{prompt}</USER_INPUT>

<ASSISTANT_RESPONSE_TO_EVALUATE>{completion}</ASSISTANT_RESPONSE_TO_EVALUATE>
\end{promptverbatim}
\end{prompt}

\begin{prompt}[title=Judge Rubric: Helpfulness,label=prompt:rubric_help]
\begin{promptverbatim}
You will be doing an Informativeness / Helpfulness Assessment of an AI assistant response.

Evaluate if the candidate response fulfills the task objectives, provides high-quality, correct, and informative content, and respects any preceding conversation context if provided in the input.

Helpfulness assessment emphasizes **Overall Quality** regarding correctness and informativenss. 

**Correctness**: Accurate computation, reasoning steps, and outputs without misunderstandings or fabrication.

When assessing informativeness, consider the following aspects:
1. **Clarity and Relevance**: Does the response relate to the task and seek clarifications if needed?
2. **Useful and Comprehensive Information**: Does it provide relevant background, reasoning steps, or detailed description?
3. **Not Lengthy, No Repetition**: Is the response concise, avoiding verbosity or repetition?

Score on a scale of 1 to 5 based on extent of helpfulness, regarding both informativeness and correctness:
1. **Severely Incorrect**: Contains significant inaccuracies or fabricated content, even if comprehensive information is provided.
2. **Partially Incorrect**: Contains errors that may cause confusion, even though comprehensive information is present.
3. **Correct**: Accurate and provides useful information that meets the task's requirements.
4. **Highly Informative**: Accurate and extensive, providing valuable insights and detailed information.
5. **Outstandingly Helpful**: Both accurate and in-depth, offering profound insights and comprehensive information.

Here are the user inputs and the AI assistant response to evaluate:

<USER_INPUT>{prompt}</USER_INPUT>

<ASSISTANT_RESPONSE_TO_EVALUATE>{completion}</ASSISTANT_RESPONSE_TO_EVALUATE>
\end{promptverbatim}
\end{prompt}

\subsection{Prompt Set Composition}\label{app:prompt_composition}

The prompts come from Dolci-Instruct-DPO~\citep{olmo3techreport}, whose identifiers encode the originating dataset; the 32~source tags map onto the 23~sources of \autoref{tab:prompt_composition} with none left unmapped. Decontamination against our evaluation suite and a 3,600-token length filter reduce 259,922~source prompts to 245,898~preference pairs, a retention of 94.6\%, unevenly distributed: the two precise instruction-following tags lose 12 to 19\% while most other sources keep at least 95\%. Dolci-Instruct-DPO repeats a minority of its prompts; each repetition receives an independent generation and pairing run, which soft-upweights those prompts rather than duplicating pairs.

\begin{table}[!htbp]
\caption{Composition of the preference prompt set by source dataset, after decontamination and length filtering. Categories follow the Dolci-Instruct-DPO grouping.}
\label{tab:prompt_composition}
\centering
\small
\begin{tabular}{@{}llrr@{}}
\toprule
\textbf{Source} & \textbf{Category} & \textbf{Pairs} & \textbf{\%} \\
\midrule
Dolci Instruct Precise IF & Precise IF & 30,320 & 12.33 \\
WildChat & Chat & 28,203 & 11.47 \\
OpenThoughts3 Science & Science & 25,633 & 10.42 \\
Dolci Instruct Python Algorithms & Code & 23,616 & 9.60 \\
UltraFeedback & General & 21,226 & 8.63 \\
FLAN & Other & 14,845 & 6.04 \\
Tulu 3 Persona MATH & Math & 14,082 & 5.73 \\
Evol CodeAlpaca & Code & 12,642 & 5.14 \\
WildJailbreak & Safety & 12,261 & 4.99 \\
WildGuardMix & Safety & 11,846 & 4.82 \\
SciRiff & Science & 8,409 & 3.42 \\
Tulu 3 Persona IF & Precise IF & 6,452 & 2.62 \\
Aya & Multilingual & 5,928 & 2.41 \\
OpenMathInstruct 2 & Math & 5,239 & 2.13 \\
Tulu 3 Persona GSM & Math & 4,893 & 1.99 \\
Multiturn SelfTalk & Multiturn SelfTalk & 4,714 & 1.92 \\
Multiturn SynthCtx & Multiturn SynthCtx & 4,570 & 1.86 \\
Tulu 3 Persona Python & Code & 4,481 & 1.82 \\
CoCoNot & Safety & 2,162 & 0.88 \\
Tulu 3 Persona Algebra & Math & 1,917 & 0.78 \\
TableGPT & Other & 1,154 & 0.47 \\
DaringAnteater & General & 821 & 0.33 \\
OpenAssistant & Chat & 484 & 0.20 \\
\midrule
\textbf{Total} & & \textbf{245,898} & \textbf{100.00} \\
\bottomrule
\end{tabular}
\end{table}

\begin{table}[!htbp]
\caption{The 24~response models, ordered by parameter count. We give active parameters for the mixture-of-experts models. Score is the four-aspect judge mean over all prompts; the last two columns give how often each model supplies the chosen and the rejected side of a pair.}
\label{tab:model_pool}
\centering
\small
\setlength{\tabcolsep}{4pt}
\begin{tabular}{@{}lrrrrr@{}}
\toprule
\textbf{Model} & \textbf{Params (B)} & \textbf{Active (B)} & \textbf{Score} & \textbf{Chosen (\%)} & \textbf{Rejected (\%)} \\
\midrule
Qwen2.5-0.5B-Instruct & 0.5 & -- & 2.186 & 0.20 & 7.17 \\
Qwen3-0.6B & 0.6 & -- & 2.624 & 0.33 & 3.33 \\
Qwen2.5-1.5B-Instruct & 1.5 & -- & 2.942 & 0.41 & 1.42 \\
Qwen3-1.7B & 1.7 & -- & 3.390 & 1.15 & 0.84 \\
EuroLLM-1.7B-Instruct & 1.7 & -- & 1.598 & 0.07 & 37.51 \\
Ministral-3-3B-Instruct-2512 & 3 & -- & 2.541 & 0.97 & 16.82 \\
SmolLM3-3B & 3 & -- & 3.559 & 0.74 & 0.64 \\
Qwen3-4B-Instruct-2507 & 4 & -- & 4.384 & 4.59 & 0.06 \\
Phi-4-mini-instruct & 4 & -- & 1.819 & 0.10 & 18.90 \\
Trinity-Nano-Preview & 6 & -- & 2.909 & 0.66 & 4.29 \\
Qwen3-8B & 8 & -- & 4.231 & 2.83 & 0.10 \\
Ministral-3-8B-Instruct-2512 & 8 & -- & 3.301 & 1.78 & 3.86 \\
EuroLLM-9B-Instruct-2512 & 9 & -- & 3.218 & 1.13 & 1.22 \\
Ministral-3-14B-Instruct-2512 & 14 & -- & 3.773 & 3.90 & 1.25 \\
EuroLLM-22B-Instruct-2512 & 22 & -- & 3.425 & 1.30 & 1.12 \\
Mistral-Small-24B-Instruct-2501 & 24 & -- & 3.949 & 1.69 & 0.25 \\
Qwen3-30B-A3B-Instruct-2507 & 30 & 3 & 4.513 & 8.91 & 0.04 \\
Qwen3-Omni-30B-A3B-Instruct & 30 & 3 & 4.342 & 4.01 & 0.33 \\
Qwen3-32B & 32 & -- & 4.433 & 4.55 & 0.06 \\
Qwen3-Next-80B-A3B-Instruct & 80 & 3 & 4.595 & 11.60 & 0.30 \\
Mixtral-8x22B-Instruct-v0.1 & 141 & 39 & 3.939 & 1.35 & 0.20 \\
Qwen3-235B-A22B-Instruct-2507 & 235 & 22 & 4.650 & 21.55 & 0.09 \\
Qwen3.5-397B-A17B & 397 & 17 & 4.691 & 16.17 & 0.08 \\
Mistral-Large-3-675B-Instruct-2512 & 675 & -- & 4.524 & 9.99 & 0.13 \\
\bottomrule
\end{tabular}
\end{table}

\subsection{Response Model Pool}\label{app:model_pool}

All 24~models of \autoref{tab:model_pool}, which span 0.5B to 675B parameters and four main families, answer every prompt, so that the pool covers a wide quality range and the max-min rule has something to select from. Coverage is complete apart from 1,969~prompts for which Qwen3-Omni-30B-A3B-Instruct returns no completion, leaving 23~candidates; across the full matrix only 251 of 6,236,159~completions fail annotation, and no prompt has all 24~models tie. Qwen3-Omni is omni-modal but we use it text-only. Note that the judge model is itself in the pool and supplies 21.6\% of the chosen responses, so we cannot exclude a self-preference effect.

\subsection{Completion Generation}\label{app:completion_generation}

Every model answers every prompt once, at temperature 1.0 and a cap of 4,096~generated tokens. We leave top-$p$ and top-$k$ unset, so each model falls back to the defaults of its own shipped generation configuration; we therefore sample the pool at a common temperature but not under a single decoding configuration, and the Qwen3 models in particular carry non-neutral nucleus defaults. We disable thinking throughout, through the chat template for every model that accepts the argument and through a no-reasoning launch flag for the four Mistral models whose tokenizer rejects it.

\subsection{Score Distribution}\label{app:score_distribution}

Max-min selection over a pool of very unequal capability pins both ends of the scale: 86.5\% of chosen responses score in $[4.9, 5.0]$ and 96.7\% of rejected ones sit at the 1.0 floor, giving a mean margin of 3.766 (\autoref{tab:score_distribution} and \autoref{tab:score_bands}). The margin is large almost everywhere, with 97.9\% of pairs separated by at least two points and 93.4\% by at least three. The 3,600-token filter costs little separation: the mean per-prompt max-min gap is 3.822 before filtering against a realized 3.766 after it.

\begin{table}[!htbp]
\caption{Summary of the four-aspect judge scores over the 245,898~preference pairs: the chosen score, the rejected score, and their margin.}
\label{tab:score_distribution}
\centering
\small
\begin{tabular}{@{}lrrrrr@{}}
\toprule
 & \textbf{Mean} & \textbf{Std} & \textbf{p5} & \textbf{Median} & \textbf{p95} \\
\midrule
Chosen & 4.897 & 0.354 & 4.345 & 4.998 & 5.000 \\
Rejected & 1.132 & 0.385 & 1.000 & 1.003 & 1.738 \\
Margin & 3.766 & 0.503 & 2.804 & 3.974 & 4.000 \\
\bottomrule
\end{tabular}
\end{table}

\begin{table}[!htbp]
\caption{Preference pairs by score band.}
\label{tab:score_bands}
\centering
\small
\begin{tabular}{@{}lrrrr@{}}
\toprule
\textbf{Score Band} & \textbf{Chosen} & \textbf{\%} & \textbf{Rejected} & \textbf{\%} \\
\midrule
{[}1, 2) & 527 & 0.21 & 238,016 & 96.73 \\
{[}2, 3) & 1,872 & 0.76 & 5,264 & 2.14 \\
{[}3, 4) & 5,425 & 2.21 & 1,586 & 0.64 \\
{[}4, 4.5) & 7,414 & 3.02 & 544 & 0.22 \\
{[}4.5, 4.9) & 17,868 & 7.27 & 297 & 0.12 \\
{[}4.9, 5.0] & 212,792 & 86.54 & 191 & 0.08 \\
\bottomrule
\end{tabular}
\end{table}

\clearpage
\section{Alignment Ablations}\label{app:alignment_ablations}

\subsection{DPO Ablations}
\label{app:dpo_ablations}

This appendix reports the sweeps over the learning rate, the KL penalty ($\beta$), and the batch size for offline DPO~\citep{Rafailov2023dpo}, its length-normalized variant~\citep{meng2024simpo, lambert2025tulu3pushingfrontiers}, and online DPO~\citep{guo2024oaif}. It also compares strategies for combining offline and online DPO. \autoref{sec:preference_alignment} documents the final configurations and \autoref{sec:post_training_lineage} their downstream results.

All scores are percentages, and $\Delta$ is the difference in average from the SFT model that initializes the runs. The ablations use intermediate checkpoints and earlier versions of the evaluation suite; scores are therefore comparable only within a table. In particular, AlpacaEval uses Llama-3.3-70B-Instruct~\citep{grattafiori2024llama3herdmodels} and Arena-Hard uses Qwen3.5-27B~\citep{qwen3_5} as the judge, instead of the GPT judges of the post-training lineage (\autoref{sec:post_training_lineage}); these scores, and the averages that include them, are not comparable to \autoref{tab:lineage_dev}. For \apertusonefivelarge{} we report skill-group means.

\subsubsection{\apertusonefivesmall{}}
\label{app:dpo_ablations_8b}

\paragraph{Offline DPO, standard loss.} \autoref{tab:dpo_offline_std} lists the ten best configurations of a 60-run grid over the learning rate, $\beta$, and the batch size. Their averages lie within 0.5~points. Nine of the ten use a learning rate of $5{\cdot}10^{-7}$ or $10^{-6}$, and none uses $\beta \geq 0.5$. All three batch sizes appear among them, which indicates a low sensitivity to the batch size. The best configuration ($\beta{=}0.05$, learning rate $5{\cdot}10^{-7}$, batch size 512) reaches $\Delta{=}{+}4.1$.

\begin{table}[t]
\caption{Offline standard DPO from \apertus{}~1.0~8B~SFT~\citep{apertus2025apertusdemocratizingopencompliant}: the ten best configurations of a 60-run grid over learning rate $\in\{10^{-6}, 5{\cdot}10^{-7}, 10^{-7}, 5{\cdot}10^{-8}\}$, $\beta\in\{0.01,0.05,0.1,0.5,1.0\}$, and batch size $\in\{128,256,512\}$, sorted by average; the learning-rate row is in units of $10^{-7}$. Scores are percentages (higher is better). The two rightmost columns are \apertus{}~1.0~8B~SFT, from which all runs start, and \apertus{}~1.0~8B~Instruct. $\Delta$ is the difference in average from the SFT column, and -- marks the SFT column itself. Bold marks the best DPO value per row.}
\label{tab:dpo_offline_std}
\centering
\small
\setlength{\tabcolsep}{4pt}
\renewcommand{\arraystretch}{1.35}
\begin{tabular}{@{}lrrrrrrrrrrrr@{}}
\toprule
 & \multicolumn{10}{c}{\textbf{Standard DPO}} & \multicolumn{2}{c}{\textbf{Baselines}} \\
\cmidrule(lr){2-11}\cmidrule(lr){12-13}
\textbf{$\beta$} & \textbf{0.05} & \textbf{0.1} & \textbf{0.1} & \textbf{0.01} & \textbf{0.1} & \textbf{0.05} & \textbf{0.1} & \textbf{0.05} & \textbf{0.05} & \textbf{0.05} &  &  \\
\textbf{Learning Rate ($10^{-7}$)} & \textbf{5} & \textbf{5} & \textbf{10} & \textbf{5} & \textbf{5} & \textbf{1} & \textbf{5} & \textbf{5} & \textbf{5} & \textbf{10} &  &  \\
\textbf{Batch Size} & \textbf{512} & \textbf{512} & \textbf{512} & \textbf{512} & \textbf{256} & \textbf{128} & \textbf{128} & \textbf{256} & \textbf{128} & \textbf{512} & \textbf{SFT} & \textbf{Instruct} \\
\midrule
\multicolumn{13}{@{}l}{\textit{Knowledge and Reasoning}} \\
MMLU & 62.5 & 61.8 & 61.9 & 61.5 & 62.3 & 61.0 & 62.4 & 62.2 & \textbf{62.6} & 61.9 & 58.9 & 61.7 \\
TruthfulQA & 59.8 & 59.2 & 61.2 & 60.2 & 60.4 & 58.6 & 61.9 & 61.3 & \textbf{64.7} & 62.4 & 54.5 & 56.7 \\
HellaSwag & 70.0 & 70.6 & 70.0 & \textbf{72.3} & 70.6 & 70.6 & 69.9 & 67.6 & 68.2 & 67.0 & 72.3 & 73.8 \\
BBH & 55.1 & 55.5 & \textbf{55.9} & 50.7 & 54.9 & 55.3 & 55.3 & 54.1 & 53.4 & 54.5 & 54.1 & 55.0 \\
ACPBench & 32.0 & 30.1 & 31.7 & 30.6 & 30.9 & 29.5 & 32.8 & 33.4 & \textbf{34.8} & 34.0 & 25.4 & 32.4 \\
DROP & 45.1 & 45.4 & 44.2 & \textbf{45.9} & 44.1 & \textbf{45.9} & 42.2 & 43.9 & 41.2 & 43.4 & 42.7 & 44.8 \\
\addlinespace
\multicolumn{13}{@{}l}{\textit{Instruction Following and Chat}} \\
IFEval & 58.0 & 63.2 & 65.6 & 56.7 & 64.9 & 57.3 & \textbf{66.3} & 60.8 & 60.4 & 59.7 & 60.1 & 68.4 \\
Multi-IF & 56.3 & 58.7 & 60.1 & 50.8 & 59.5 & 55.5 & \textbf{60.2} & 55.2 & 56.6 & 55.9 & 59.7 & 65.8 \\
AlpacaEval & 37.7 & 32.5 & 33.6 & \textbf{44.4} & 33.6 & 31.4 & 32.2 & 39.3 & 39.7 & 39.1 & 5.1 & 12.9 \\
\addlinespace
\multicolumn{13}{@{}l}{\textit{Math}} \\
GSM8K & 67.3 & 65.4 & 65.3 & \textbf{68.5} & 65.9 & 66.4 & 65.1 & 65.7 & 67.4 & 67.0 & 56.6 & 62.6 \\
MATH & 21.5 & 19.0 & 19.7 & \textbf{23.1} & 19.7 & 19.9 & 19.9 & 21.8 & 21.0 & 21.7 & 15.2 & 18.2 \\
MathQA & 37.5 & 37.9 & 37.1 & \textbf{38.3} & 37.0 & 37.3 & 37.5 & 36.9 & 36.6 & 36.4 & 37.5 & 38.0 \\
\addlinespace
\multicolumn{13}{@{}l}{\textit{Coding}} \\
HumanEval & 71.1 & \textbf{72.6} & 72.0 & 71.5 & 70.1 & 71.6 & 71.8 & 71.5 & 69.1 & 69.4 & 65.0 & 69.2 \\
MBPP & 39.4 & 38.2 & 40.0 & 34.2 & 39.8 & 39.0 & 38.2 & 40.0 & 40.8 & \textbf{41.0} & 36.0 & 39.0 \\
\addlinespace
\multicolumn{13}{@{}l}{\textit{Multilingual and Cultural}} \\
MGSM & 52.8 & \textbf{52.9} & 51.4 & 48.1 & 52.8 & 52.4 & 51.6 & 52.0 & 49.3 & 51.6 & 43.9 & 46.5 \\
\addlinespace
\multicolumn{13}{@{}l}{\textit{Safety and Alignment}} \\
ToxiGen & 64.4 & 66.5 & 57.7 & 70.5 & 59.2 & \textbf{73.6} & 57.0 & 57.1 & 56.8 & 56.8 & 78.5 & 70.1 \\
\addlinespace
Average & \textbf{51.9} & 51.8 & 51.7 & 51.7 & 51.6 & 51.6 & 51.5 & 51.4 & 51.4 & 51.4 & 47.8 & 51.0 \\
$\Delta$ vs.\ SFT & \textbf{+4.1} & +4.0 & +3.9 & +3.9 & +3.8 & +3.8 & +3.7 & +3.6 & +3.6 & +3.5 & -- & +3.1 \\
\bottomrule
\end{tabular}
\end{table}

\paragraph{Offline DPO, length-normalized loss.} With the batch size fixed at 128, we sweep $\beta$ and the learning rate (\autoref{tab:dpo_offline_ln}). A larger $\beta$ tolerates a larger learning rate: at $10^{-6}$ only $\beta{=}25$ improves over the SFT model, and at $5{\cdot}10^{-7}$ only $\beta \in \{5, 25\}$ does. At $\beta{=}25$, the average increases monotonically with the learning rate up to $10^{-6}$. The six configurations absent from the table do not improve over the SFT model. The best configuration ($\beta{=}25$, learning rate $10^{-6}$) reaches $\Delta{=}{+}5.5$, 1.4~points above the best standard DPO run. Extending the grid to $\beta \in \{50, 100\}$ lowers the best $\Delta$ to $+4.4$ and $+3.2$, respectively (\autoref{tab:dpo_offline_ln_ext}). We use $\beta{=}25$ for all offline DPO training.

\begin{table}[t]
\caption{Offline length-normalized DPO from \apertus{}~1.0~8B~SFT with batch size 128: the ten best configurations of the grid $\beta\in\{0.5,2.5,5,25\}$ $\times$ learning rate $\in\{10^{-6},5{\cdot}10^{-7},10^{-7},10^{-8}\}$, sorted by average; the learning-rate row is in units of $10^{-7}$. Scores are percentages (higher is better). The two rightmost columns are \apertus{}~1.0~8B~SFT and Instruct; $\Delta$ and -- are as in \autoref{tab:dpo_offline_std}. Bold marks the best DPO value per row.}
\label{tab:dpo_offline_ln}
\centering
\small
\setlength{\tabcolsep}{4pt}
\renewcommand{\arraystretch}{1.35}
\begin{tabular}{@{}lrrrrrrrrrrrr@{}}
\toprule
 & \multicolumn{10}{c}{\textbf{Length-Normalized DPO}} & \multicolumn{2}{c}{\textbf{Baselines}} \\
\cmidrule(lr){2-11}\cmidrule(lr){12-13}
\textbf{$\beta$} & \textbf{25.0} & \textbf{25.0} & \textbf{5.0} & \textbf{25.0} & \textbf{2.5} & \textbf{5.0} & \textbf{0.5} & \textbf{0.5} & \textbf{25.0} & \textbf{2.5} &  &  \\
\textbf{Learning Rate ($10^{-7}$)} & \textbf{10.0} & \textbf{5.0} & \textbf{1.0} & \textbf{1.0} & \textbf{1.0} & \textbf{5.0} & \textbf{1.0} & \textbf{0.1} & \textbf{0.1} & \textbf{0.1} & \textbf{SFT} & \textbf{Instruct} \\
\midrule
\multicolumn{13}{@{}l}{\textit{Knowledge and Reasoning}} \\
MMLU & 62.0 & 62.4 & 62.2 & 61.5 & \textbf{63.3} & 61.9 & 60.0 & 58.1 & 59.1 & 59.3 & 58.9 & 61.7 \\
TruthfulQA & 61.1 & 60.1 & 59.2 & 57.8 & 59.5 & \textbf{63.7} & 58.0 & 54.7 & 54.5 & 54.7 & 54.5 & 56.7 \\
HellaSwag & 76.4 & 75.5 & 69.0 & 72.7 & 64.9 & \textbf{78.8} & 70.9 & 72.2 & 72.2 & 72.2 & 72.3 & 73.8 \\
BBH & 53.6 & \textbf{55.5} & 53.6 & 55.4 & 50.8 & 32.5 & 40.9 & 54.3 & 54.0 & 54.2 & 54.1 & 55.0 \\
ACPBench & \textbf{34.0} & 33.0 & 31.3 & 29.5 & 33.3 & 28.2 & 31.7 & 25.2 & 25.2 & 25.0 & 25.4 & 32.4 \\
DROP & \textbf{47.5} & 46.9 & 45.7 & 45.2 & 45.8 & 47.4 & 46.2 & 43.0 & 42.8 & 43.0 & 42.7 & 44.8 \\
\addlinespace
\multicolumn{13}{@{}l}{\textit{Instruction Following and Chat}} \\
IFEval & 64.1 & \textbf{64.5} & 57.9 & 63.6 & 56.6 & 58.4 & 53.0 & 60.8 & 61.0 & 60.1 & 60.1 & 68.4 \\
Multi-IF & \textbf{62.0} & 61.6 & 55.2 & 60.8 & 53.8 & 53.7 & 51.0 & 59.5 & 59.5 & 59.4 & 59.7 & 65.8 \\
AlpacaEval & 30.2 & 26.1 & 38.8 & 20.9 & 40.5 & \textbf{44.1} & 42.9 & 5.9 & 4.8 & 4.9 & 5.1 & 12.9 \\
\addlinespace
\multicolumn{13}{@{}l}{\textit{Math}} \\
GSM8K & \textbf{68.8} & 67.4 & 67.2 & 63.6 & 66.3 & 67.2 & 64.6 & 57.9 & 56.7 & 57.0 & 56.6 & 62.6 \\
MATH & 19.7 & 20.7 & 23.9 & 18.8 & \textbf{24.4} & 21.1 & 23.7 & 15.2 & 14.8 & 15.2 & 15.2 & 18.2 \\
MathQA & 38.4 & \textbf{38.8} & 37.6 & 38.2 & 37.3 & 38.1 & 37.7 & 37.7 & 37.7 & 37.8 & 37.5 & 38.0 \\
\addlinespace
\multicolumn{13}{@{}l}{\textit{Coding}} \\
HumanEval & 70.2 & 69.1 & \textbf{70.4} & 67.7 & 66.5 & 59.6 & 64.1 & 65.8 & 65.4 & 64.5 & 65.0 & 69.2 \\
MBPP & \textbf{40.2} & 39.0 & 38.2 & 39.2 & 38.8 & 29.8 & 37.2 & 36.0 & 35.2 & 35.8 & 36.0 & 39.0 \\
\addlinespace
\multicolumn{13}{@{}l}{\textit{Multilingual and Cultural}} \\
MGSM & \textbf{52.5} & 51.5 & 51.7 & 50.4 & 51.7 & 48.8 & 51.0 & 43.4 & 44.0 & 43.5 & 43.9 & 46.5 \\
\addlinespace
\multicolumn{13}{@{}l}{\textit{Safety and Alignment}} \\
ToxiGen & 73.2 & 74.2 & 68.8 & 76.4 & 63.4 & 72.5 & 58.2 & \textbf{78.5} & \textbf{78.5} & 78.4 & 78.5 & 70.1 \\
\addlinespace
Average & \textbf{53.4} & 52.9 & 51.9 & 51.4 & 51.1 & 50.4 & 49.4 & 48.0 & 47.8 & 47.8 & 47.8 & 51.0 \\
$\Delta$ vs.\ SFT & \textbf{+5.5} & +5.1 & +4.1 & +3.5 & +3.2 & +2.5 & +1.6 & +0.2 & 0.0 & 0.0 & -- & +3.1 \\
\bottomrule
\end{tabular}
\end{table}

\begin{table}[t]
\caption{Extension of the length-normalized DPO grid of \autoref{tab:dpo_offline_ln} to $\beta\in\{50,100\}$, from \apertus{}~1.0~8B~SFT with batch size 128, sorted by average; the learning-rate row is in units of $10^{-7}$. Scores, $\Delta$, --, and the baseline columns are as in \autoref{tab:dpo_offline_ln}. Bold marks the best DPO value per row.}
\label{tab:dpo_offline_ln_ext}
\centering
\small
\renewcommand{\arraystretch}{1.35}
\begin{tabular}{@{}lrrrrrrrr@{}}
\toprule
 & \multicolumn{6}{c}{\textbf{Length-Normalized DPO}} & \multicolumn{2}{c}{\textbf{Baselines}} \\
\cmidrule(lr){2-7}\cmidrule(lr){8-9}
\textbf{$\beta$} & \textbf{50.0} & \textbf{50.0} & \textbf{100.0} & \textbf{50.0} & \textbf{100.0} & \textbf{100.0} &  &  \\
\textbf{Learning Rate ($10^{-7}$)} & \textbf{10} & \textbf{5} & \textbf{10} & \textbf{1} & \textbf{5} & \textbf{1} & \textbf{SFT} & \textbf{Instruct} \\
\midrule
\multicolumn{9}{@{}l}{\textit{Knowledge and Reasoning}} \\
MMLU & 61.2 & 60.7 & 60.7 & 61.2 & \textbf{62.4} & 59.9 & 58.9 & 61.7 \\
TruthfulQA & \textbf{59.0} & 58.6 & 57.8 & 56.9 & 57.2 & 56.1 & 54.5 & 56.7 \\
HellaSwag & \textbf{75.2} & 74.7 & 74.3 & 72.7 & 74.0 & 72.7 & 72.3 & 73.8 \\
BBH & \textbf{56.0} & 55.9 & 55.4 & 55.3 & 55.5 & 54.4 & 54.1 & 55.0 \\
ACPBench & \textbf{31.6} & 29.2 & 28.3 & 27.7 & 28.1 & 26.8 & 25.4 & 32.4 \\
DROP & 45.4 & \textbf{45.9} & 45.4 & 44.4 & 45.1 & 44.0 & 42.7 & 44.8 \\
\addlinespace
\multicolumn{9}{@{}l}{\textit{Instruction Following and Chat}} \\
IFEval & 62.5 & 62.8 & \textbf{63.4} & 63.0 & 61.4 & 62.3 & 60.1 & 68.4 \\
Multi-IF & \textbf{63.9} & 61.7 & 61.9 & 60.9 & 60.6 & 60.2 & 59.7 & 65.8 \\
AlpacaEval & \textbf{19.6} & 16.6 & 11.7 & 13.9 & 10.4 & 9.0 & 5.1 & 12.9 \\
\addlinespace
\multicolumn{9}{@{}l}{\textit{Math}} \\
GSM8K & \textbf{66.1} & 64.8 & 65.4 & 62.3 & 63.5 & 60.6 & 56.6 & 62.6 \\
MATH & \textbf{18.3} & 17.5 & 16.5 & 17.0 & 16.2 & 15.6 & 15.2 & 18.2 \\
MathQA & 38.8 & 38.6 & \textbf{39.0} & 38.3 & 38.1 & 38.1 & 37.5 & 38.0 \\
\addlinespace
\multicolumn{9}{@{}l}{\textit{Coding}} \\
HumanEval & \textbf{71.8} & 70.7 & 70.5 & 68.3 & 68.9 & 65.8 & 65.0 & 69.2 \\
MBPP & 38.0 & \textbf{39.2} & 38.6 & 38.0 & 38.2 & 37.4 & 36.0 & 39.0 \\
\addlinespace
\multicolumn{9}{@{}l}{\textit{Multilingual and Cultural}} \\
MGSM & 50.0 & \textbf{50.5} & 48.1 & 48.6 & 48.2 & 47.3 & 43.9 & 46.5 \\
\addlinespace
\multicolumn{9}{@{}l}{\textit{Safety and Alignment}} \\
ToxiGen & 78.2 & 78.8 & \textbf{79.6} & 78.5 & 78.4 & 78.5 & 78.5 & 70.1 \\
\addlinespace
Average & \textbf{52.2} & 51.7 & 51.0 & 50.4 & 50.4 & 49.3 & 47.8 & 51.0 \\
$\Delta$ vs.\ SFT & \textbf{+4.4} & +3.8 & +3.2 & +2.6 & +2.5 & +1.5 & -- & +3.1 \\
\bottomrule
\end{tabular}
\end{table}

\paragraph{Offline DPO, transfer to \apertusonefivesmall{}.} The preceding grids use the \apertus{}~1.0~8B~SFT checkpoint, and we apply the selected configuration to the \apertusonefivesmall{} SFT checkpoints without a new search. As a check, a learning rate of $5{\cdot}10^{-6}$ on an intermediate \apertusonefivesmall{} SFT checkpoint changes the averages by at most 0.1~points (48.8 vs.\ 48.8 on the development benchmarks and 44.0 vs.\ 43.9 on the held-out benchmarks). We therefore keep the learning rate of $10^{-6}$.

\paragraph{Online DPO.} With $\beta{=}0.1$ and 256~prompts per step, we sweep the learning rate over $\{10^{-5}, 5{\cdot}10^{-6}, 10^{-6}\}$ (\autoref{tab:dpo_online_lr}). A learning rate of $5{\cdot}10^{-6}$ performs best ($\Delta{=}{+}4.4$), ahead of $10^{-6}$ ($+3.3$), $10^{-5}$ ($+3.2$), and offline DPO from the same checkpoint ($+3.7$).

\begin{table}[t]
\caption{Learning-rate sweeps for online DPO and offline$\to$online DPO ($\beta{=}0.1$, 256~prompts per step) from an intermediate \apertusonefivesmall{} SFT checkpoint, sorted by average within each group. Scores are percentages (higher is better). The learning-rate row is in units of $10^{-7}$. Offline is length-normalized offline DPO ($\beta{=}25$, learning rate $10^{-6}$) from the same checkpoint, which also initializes offline$\to$online; the rightmost column is the SFT checkpoint, and -- marks it in the $\Delta$ row. In these runs, rollouts come from the initial instead of the current policy because of an implementation issue fixed afterward; on-policy training then uses the selected learning rate. Bold marks the best DPO value per row.}
\label{tab:dpo_online_lr}
\centering
\small
\renewcommand{\arraystretch}{1.35}
\begin{tabular}{@{}lrrrrrrrr@{}}
\toprule
 & \multicolumn{3}{c}{\textbf{Online DPO}} & \multicolumn{3}{c}{\textbf{Offline$\to$Online}} & & \\
\cmidrule(lr){2-4}\cmidrule(lr){5-7}
\textbf{Learning Rate ($10^{-7}$)} & \textbf{50} & \textbf{10} & \textbf{100} & \textbf{50} & \textbf{10} & \textbf{5} & \textbf{Offline} & \textbf{SFT} \\
\midrule
\multicolumn{9}{@{}l}{\textit{Knowledge and Reasoning}} \\
MMLU & 62.8 & 62.1 & 62.8 & \textbf{63.8} & 61.7 & 62.4 & 61.3 & 59.3 \\
TruthfulQA & 62.2 & 55.2 & \textbf{65.3} & 64.1 & 57.2 & 57.4 & 57.9 & 53.4 \\
HellaSwag & 70.8 & 68.0 & \textbf{72.4} & 71.9 & 69.3 & 70.8 & 71.6 & 70.8 \\
BBH & 54.2 & \textbf{55.9} & 51.1 & 52.5 & 55.8 & 55.2 & 55.1 & 54.6 \\
ACPBench & 41.1 & 37.5 & \textbf{41.2} & 39.4 & 35.9 & 33.7 & 33.7 & 28.1 \\
DROP & 36.3 & 39.5 & 34.6 & 35.9 & 45.8 & 46.7 & \textbf{47.0} & 47.4 \\
\addlinespace
\multicolumn{9}{@{}l}{\textit{Instruction Following and Chat}} \\
IFEval & 77.8 & 77.5 & 77.8 & \textbf{79.1} & 76.0 & 73.6 & 72.6 & 71.7 \\
Multi-IF & 72.1 & 72.6 & 69.8 & \textbf{73.8} & 70.7 & 70.2 & 68.1 & 68.9 \\
AlpacaEval & 25.4 & 24.5 & 21.7 & \textbf{41.0} & 36.4 & 26.2 & 27.2 & 9.4 \\
\addlinespace
\multicolumn{9}{@{}l}{\textit{Math}} \\
GSM8K & \textbf{70.1} & 65.4 & 65.0 & 68.8 & 67.0 & 68.5 & 69.7 & 64.5 \\
MATH & 31.6 & 32.1 & 28.6 & \textbf{32.8} & 30.2 & 29.6 & 28.7 & 22.9 \\
MathQA & 38.9 & 39.0 & 39.9 & 37.2 & 40.7 & \textbf{41.0} & 40.8 & 39.6 \\
\addlinespace
\multicolumn{9}{@{}l}{\textit{Coding}} \\
HumanEval & 70.5 & 71.1 & 73.4 & 73.2 & 79.1 & 79.1 & \textbf{80.0} & 79.3 \\
MBPP & 43.0 & 42.0 & 39.4 & 42.0 & 43.6 & \textbf{44.0} & 42.0 & 40.6 \\
\addlinespace
\multicolumn{9}{@{}l}{\textit{Multilingual and Cultural}} \\
MGSM & \textbf{58.0} & 53.5 & 51.1 & 53.0 & 50.3 & 50.3 & 51.7 & 43.4 \\
\addlinespace
\multicolumn{9}{@{}l}{\textit{Safety and Alignment}} \\
ToxiGen & 82.8 & 83.4 & 83.8 & \textbf{84.7} & 82.3 & 82.0 & 78.6 & 73.0 \\
\addlinespace
Average & 56.1 & 55.0 & 54.9 & \textbf{57.1} & 56.4 & 55.7 & 55.4 & 51.7 \\
$\Delta$ vs.\ SFT & +4.4 & +3.3 & +3.2 & \textbf{+5.4} & +4.7 & +4.0 & +3.7 & -- \\
\bottomrule
\end{tabular}
\end{table}

\paragraph{Combining offline and online DPO.} We evaluate two combinations of offline and online DPO: a sequential pipeline and a joint data mixture.

In the sequential offline$\to$online setting, online DPO starts from a trained offline DPO checkpoint. The learning-rate sweep for the online phase again favors $5{\cdot}10^{-6}$ ($\Delta{=}{+}5.4$), ahead of $10^{-6}$ ($+4.7$) and $5{\cdot}10^{-7}$ ($+4.0$; \autoref{tab:dpo_online_lr}).

The mixture adds 256~offline pairs to each online step. \autoref{tab:dpo_variants} compares the four variants (offline, online, offline$\to$online, and mixture) with the selected hyperparameters. All four improve over the SFT checkpoint (51.4 on the development benchmarks, 49.3 on the held-out benchmarks). The mixture scores highest (55.5 / 53.8), followed by offline$\to$online (55.4 / 53.5), online (54.6 / 52.0), and offline (54.0 / 50.8).

The final \apertusonefivesmall{} recipe uses online DPO alone. When we repeat the comparison of offline, online, and mixture DPO on the four final \apertusonefivesmall{} SFT checkpoints, online DPO matches the mixture; the small gains of the combined variants on the intermediate checkpoint do not justify the additional data streams and training stages.

\begin{table}[t]
\caption{DPO variants from another intermediate \apertusonefivesmall{} SFT checkpoint (rightmost column) with thinking disabled, sorted by development average; scores are percentages (higher is better). Offline uses length-normalized DPO ($\beta{=}25$, learning rate $10^{-6}$, batch size 128); the others use $\beta{=}0.1$, learning rate $5{\cdot}10^{-6}$, and 256~prompts per step; the mixture adds 256~offline pairs to each step, and offline$\to$online starts from the offline checkpoint. Held-out groups contain the held-out benchmarks and the others the development benchmarks; each average is over all benchmarks of its split. Bold marks the best DPO value per row.}
\label{tab:dpo_variants}
\centering
\small
\setlength{\defaultaddspace}{0.4em}
\begin{tabular}{@{}lrrrrr@{}}
\toprule
\textbf{Benchmark} & \textbf{Mixture} & \textbf{Offline$\to$Online} & \textbf{Online} & \textbf{Offline} & \textbf{SFT} \\
\midrule
\multicolumn{6}{@{}l}{\textit{Knowledge and Reasoning}} \\
MMLU & 67.7 & 68.1 & \textbf{68.3} & \textbf{68.3} & 63.8 \\
MMLU-Pro & \textbf{45.3} & 43.8 & 36.4 & 36.7 & 27.7 \\
TruthfulQA & \textbf{60.9} & 59.5 & 59.1 & 55.0 & 49.5 \\
CommonsenseQA & \textbf{75.3} & 74.7 & 74.9 & 74.4 & 74.5 \\
SQuADv2 & 23.3 & 25.9 & \textbf{27.2} & 13.6 & 16.5 \\
HellaSwag & 69.9 & 73.6 & 70.0 & \textbf{75.8} & 73.5 \\
BBH & \textbf{67.5} & 67.1 & 67.4 & 67.0 & 64.7 \\
ACPBench & 35.8 & \textbf{43.0} & 35.6 & 32.3 & 27.8 \\
DROP & 51.3 & 50.7 & \textbf{51.4} & 50.1 & 48.5 \\
\addlinespace
\multicolumn{6}{@{}l}{\textit{Instruction Following and Chat}} \\
IFEval & \textbf{85.0} & \textbf{85.0} & 83.9 & 80.8 & 78.2 \\
Multi-IF & 81.0 & \textbf{82.2} & 81.9 & 78.8 & 74.0 \\
AlpacaEval & 41.1 & 49.4 & 38.6 & \textbf{49.9} & 20.4 \\
\addlinespace
\multicolumn{6}{@{}l}{\textit{Math}} \\
GSM8K & 80.7 & \textbf{81.9} & 81.0 & 81.3 & 73.5 \\
MATH & \textbf{34.0} & 33.7 & 31.5 & 31.1 & 27.4 \\
Minerva MATH & \textbf{35.0} & 31.2 & 34.3 & 32.6 & 31.1 \\
MathQA & 46.5 & \textbf{47.3} & 46.7 & 46.9 & 45.9 \\
MATH-500 & 33.8 & 30.4 & \textbf{34.6} & 30.4 & 31.6 \\
AIME 24 & \textbf{6.7} & 0.0 & 0.0 & \textbf{6.7} & 0.0 \\
AIME 25 & \textbf{13.3} & 6.7 & 6.7 & 0.0 & 10.0 \\
\addlinespace
\multicolumn{6}{@{}l}{\textit{Coding}} \\
HumanEval & \textbf{86.1} & 85.0 & 83.5 & 85.5 & 84.1 \\
MBPP & 49.8 & 48.2 & \textbf{50.2} & 49.2 & 47.4 \\
\addlinespace
\multicolumn{6}{@{}l}{\textit{Multilingual and Cultural}} \\
Global-MMLU & \textbf{62.4} & 60.3 & 59.7 & 60.0 & 57.9 \\
MGSM & 76.4 & \textbf{80.8} & 79.6 & 77.6 & 79.2 \\
Multilingual TruthfulQA & \textbf{54.1} & 53.4 & 53.4 & 51.1 & 48.4 \\
INCLUDE-44 & \textbf{58.1} & 57.3 & 57.1 & 57.4 & 56.7 \\
INCLUDE-45 & 38.8 & 38.4 & 37.8 & \textbf{40.1} & 39.6 \\
XNLI & 43.6 & 44.3 & 44.3 & \textbf{46.2} & 46.2 \\
BLEnD & \textbf{66.8} & 66.5 & 66.5 & 65.9 & 65.6 \\
CulturalBench & 72.8 & 73.5 & 73.0 & \textbf{74.0} & 73.9 \\
SwitzerlandQA & 64.7 & 63.9 & 64.1 & \textbf{64.8} & 64.5 \\
\addlinespace
\multicolumn{6}{@{}l}{\textit{Safety and Alignment}} \\
BBQ & \textbf{67.6} & 66.5 & 66.6 & 64.2 & 63.5 \\
ToxiGen & \textbf{83.8} & 82.8 & 83.1 & 82.6 & 79.6 \\
WMDP & 53.6 & 53.5 & \textbf{54.0} & 52.5 & 51.1 \\
\addlinespace
Development Average & \textbf{55.5} & 55.4 & 54.6 & 54.0 & 51.4 \\
\addlinespace
\multicolumn{6}{@{}l}{\textit{Held-out Knowledge and Reasoning}} \\
AGIEval & \textbf{47.8} & 46.1 & 46.8 & 39.4 & 38.4 \\
ARC-Challenge & 84.4 & \textbf{84.8} & 84.5 & 81.8 & 77.6 \\
GPQA & 31.0 & \textbf{32.4} & 25.7 & 29.5 & 29.0 \\
\addlinespace
\multicolumn{6}{@{}l}{\textit{Held-out Math}} \\
GSM8K-Platinum & 80.6 & \textbf{81.7} & 79.7 & 78.3 & 79.7 \\
\addlinespace
\multicolumn{6}{@{}l}{\textit{Held-out Multilingual and Cultural}} \\
mLogiQA & 37.6 & 34.9 & 34.4 & \textbf{37.8} & 34.8 \\
Multilingual ARC & \textbf{41.7} & 40.9 & 41.0 & 38.1 & 36.5 \\
\addlinespace
Held-out Average & \textbf{53.8} & 53.5 & 52.0 & 50.8 & 49.3 \\
\bottomrule
\end{tabular}
\end{table}

\subsubsection{\apertusonefivelarge{}}
\label{app:dpo_ablations_70b}

\paragraph{Transferring the \apertusonefivesmall{} hyperparameters.} We test the transfer on \apertus{}~1.0~70B~SFT, the SFT checkpoint of \hfrepo{swiss-ai/Apertus-70B-Instruct-2509} (\autoref{tab:dpo_70b_transfer}). The best \apertusonefivesmall{} online setting (learning rate $5{\cdot}10^{-6}$, $\beta{=}0.1$) does not transfer to this scale: it shortens responses from 280 to 160~tokens on average, and coding (44.3 vs.\ 60.5) and math (20.7 vs.\ 27.2) fall below the SFT checkpoint. A lower learning rate with a larger $\beta$ ($2{\cdot}10^{-6}$ with $\beta \in \{0.25, 0.35\}$) avoids this degradation and raises the overall average from 46.3 to 48.9--49.1.

\begin{table}[!htbp]
\caption{Online DPO from \apertus{}~1.0~70B~SFT with the learning rate / $\beta$ given in the headers, sorted by average; $5{\cdot}10^{-6}$ / 0.1 is the \apertusonefivesmall{} setting, and the rightmost column is the SFT checkpoint. Scores in percent (higher is better) with thinking disabled; for degeneration, lower is better. Bold marks the best DPO value per skill-group and average row. Each skill-group row is the mean over the benchmarks within that group, and the overall average is the mean over all individual benchmarks.}
\label{tab:dpo_70b_transfer}
\centering
\small
\renewcommand{\arraystretch}{1.35}
\begin{tabular}{@{}lrrrr@{}}
\toprule
 & \multicolumn{3}{c}{\textbf{Online DPO (Learning Rate / $\beta$)}} &  \\
\cmidrule(lr){2-4}
\textbf{Skill Group} & \textbf{$2{\cdot}10^{-6}$ / 0.25} & \textbf{$2{\cdot}10^{-6}$ / 0.35} & \textbf{$5{\cdot}10^{-6}$ / 0.1} & \textbf{SFT} \\
\midrule
Knowledge & 58.0 & 57.0 & \textbf{60.3} & 54.2 \\
Reasoning & 54.4 & \textbf{54.6} & 53.0 & 50.1 \\
Instruction Following & 76.8 & 75.7 & \textbf{79.7} & 68.9 \\
Chat & 10.7 & 9.2 & \textbf{15.7} & 6.4 \\
Math & 28.7 & \textbf{29.3} & 20.7 & 27.2 \\
Coding & 60.0 & \textbf{61.0} & 44.3 & 60.5 \\
Multilingual & 55.7 & \textbf{56.2} & 52.5 & 55.8 \\
Cultural Knowledge & \textbf{68.8} & 68.1 & 67.8 & 68.7 \\
Safety & 67.6 & 67.9 & \textbf{70.0} & 57.8 \\
Charter Alignment & 32.2 & 27.4 & \textbf{50.8} & 21.9 \\
Tool Use & 40.8 & \textbf{42.9} & 42.4 & 45.0 \\
\addlinespace
Overall Average & \textbf{49.1} & 48.9 & 47.6 & 46.3 \\
\addlinespace
Mean response length (tokens) & 245.9 & 252.9 & 160.0 & 279.9 \\
Degeneration (\%) & 0.6 & 2.3 & 0.0 & 8.4 \\
\bottomrule
\end{tabular}
\end{table}

\paragraph{Offline and mixture.} The RL checkpoint of \apertusonefivelarge{} produces much shorter responses than its SFT checkpoint (253 vs.\ 367~tokens; \autoref{tab:lineage_full}). Offline DPO from the RL checkpoint reverses this shortening and raises the mean response length from 253 to 501~tokens (\autoref{tab:dpo_70b_offline}). From the RL checkpoint, offline DPO alone reaches an average of 56.1, above the mixture (55.6) and online DPO (55.5). The final \apertusonefivelarge{} recipe runs offline DPO followed by online DPO (\autoref{sec:preference_alignment}), which restores the response length before on-policy training.

\begin{table}[t]
\caption{Offline DPO, mixture, and online DPO from the \apertusonefivelarge{} RL checkpoint, sorted by average; the rightmost column is the RL checkpoint. Offline DPO uses the configuration of \autoref{sec:preference_alignment}; online DPO and the mixture use $\beta{=}0.1$ and a learning rate of $5{\cdot}10^{-6}$, and the mixture adds 256 offline pairs to each online step. Scores in percent (higher is better) with thinking disabled; for degeneration, lower is better. Bold marks the best DPO value per skill-group and average row. Each skill-group row is the mean over the benchmarks within that group, and the overall average is the mean over all individual benchmarks.}
\label{tab:dpo_70b_offline}
\centering
\small
\renewcommand{\arraystretch}{1.35}
\begin{tabular}{@{}lrrrr@{}}
\toprule
\textbf{Skill Group} & \textbf{Offline} & \textbf{Mixture} & \textbf{Online} & \textbf{RL} \\
\midrule
Knowledge & 56.4 & \textbf{61.7} & \textbf{61.7} & 59.4 \\
Reasoning & 62.2 & \textbf{63.5} & 62.4 & 60.9 \\
Instruction Following & 84.1 & \textbf{88.6} & 88.5 & 88.5 \\
Chat & \textbf{39.5} & 24.4 & 24.2 & 19.5 \\
Math & 34.2 & 35.6 & \textbf{36.0} & 35.7 \\
Coding & \textbf{75.7} & \textbf{75.7} & 74.5 & 74.1 \\
Multilingual & \textbf{58.3} & 57.6 & 57.9 & 57.8 \\
Cultural Knowledge & \textbf{71.1} & 66.8 & 67.1 & 63.6 \\
Safety & 70.1 & \textbf{71.4} & 71.0 & 69.9 \\
Charter Alignment & \textbf{62.8} & 53.0 & 53.5 & 50.9 \\
Tool Use & \textbf{52.7} & 47.9 & 47.2 & 43.5 \\
\addlinespace
Overall Average & \textbf{56.1} & 55.6 & 55.5 & 54.0 \\
\addlinespace
Mean response length (tokens) & 501.2 & 241.7 & 231.0 & 252.9 \\
Degeneration (\%) & 10.6 & 1.1 & 2.2 & 10.6 \\
\bottomrule
\end{tabular}
\end{table}

\subsection{Judge Ablations}\label{app:judge-ablations}

The online DPO judge described in \autoref{sec:preference_alignment} scores each rollout on helpfulness alone with Qwen3.6-27B, rather than the offline stage's four-aspect rubric with Qwen3-235B-A22B-Instruct-2507. We ablate both departures: the aspect simplification and the choice of judge model.

\subsubsection{Judging Aspects Simplification}

We test whether the four aspect-wise scores from Qwen3-235B-A22B-Instruct-2507 are correlated enough that scoring on one alone is a safe simplification, using \hfdata{ActiveUltraFeedback/ultrafeedback}: 60,829~prompts\footnote{The released dataset contained 10~prompts twice; we keep only the first occurrence of each.}, each with responses from a pool of 30~models rated on all four aspects by Qwen3-235B-A22B-Instruct-2507.

Prompt difficulty and response-model strength each push a response's score on all four aspects up or down together, inflating pairwise correlations between aspects for reasons unrelated to whether the aspects actually agree. Writing $s_{a,p,m}$ for aspect $a$'s expected rating on the response to prompt $p$ from model $m$, we remove both effects before correlating:
\begin{equation}
\tilde{s}_{a,p,m} = s_{a,p,m} - \bar{s}_{a,p,\cdot} - \bar{s}_{a,\cdot,m} + \bar{s}_{a,\cdot,\cdot},
\label{eq:deconfound}
\end{equation}
where $\bar{s}_{a,p,\cdot}$, $\bar{s}_{a,\cdot,m}$, and $\bar{s}_{a,\cdot,\cdot}$ are aspect $a$'s mean score over responses to prompt $p$, over responses from model $m$, and over all responses, respectively.

The four aspects are strongly correlated both before\footnote{On the raw scores, before removing prompt and model effects, correlations are higher still, ranging from 0.749 (Helpfulness vs.\ Truthfulness) to 0.849 (Helpfulness vs.\ Instruction Following).} and after deconfounding (\autoref{tab:aspect_correlation}). Pairwise Spearman correlations are close to the Pearson ones, differing by 0.029 on average and 0.042 at most (Helpfulness vs.\ Instruction Following), so the pairwise relationships are close to linear rather than merely monotonic. Scoring on one aspect alone, rather than averaging all four, is therefore a safe simplification.

\begin{table}[!htbp]
\caption{Pairwise Pearson correlation between the four aspects, pooled across all response models, on scores with prompt and model effects removed via \eqref{eq:deconfound}.}
\label{tab:aspect_correlation}
\centering
\small
\renewcommand{\arraystretch}{1.35}
\begin{tabular}{@{}lcccc@{}}
\toprule
 & Helpfulness & Honesty & Instr.\ Following & Truthfulness \\
\midrule
Helpfulness & -- & 0.710 & 0.748 & 0.644 \\
Honesty & 0.710 & -- & 0.680 & 0.745 \\
Instruction Following & 0.748 & 0.680 & -- & 0.641 \\
Truthfulness & 0.644 & 0.745 & 0.641 & -- \\
\midrule
Average & 0.701 & 0.712 & 0.690 & 0.677 \\
\bottomrule
\end{tabular}
\end{table}

To identify which aspect to keep, we compare all four on three external judging benchmarks: JudgeBench (GPT split)~\citep{tan2025judgebench}, RewardBench~2~\citep{malik2025rewardbench2advancingreward}, and RM-Bench~\citep{liu2025rmbench}. Each pairs a prompt with a labeled-preferred response and one or more labeled-rejected responses, and scores a judge by how often it ranks the preferred response above the rejected one (\autoref{tab:aspect_ablation}).

\begin{table}[!htbp]
\caption{Accuracy with Qwen3-235B-A22B-Instruct-2507 and different aspects.}
\label{tab:aspect_ablation}
\centering
\small
\renewcommand{\arraystretch}{1.35}
\begin{tabular}{@{}lcccc@{}}
\toprule
 & \textbf{Helpfulness} & \textbf{Honesty} & \textbf{Instruction Following} & \textbf{Truthfulness} \\
\midrule
\textit{JudgeBench} & & & & \\
 MMLU-Pro & \textbf{73.4} & 70.1 & 71.4 & 71.4 \\
 LiveBench Reasoning & 69.4 & 67.3 & \textbf{71.4} & 69.4 \\
 LiveBench Math & 83.9 & \textbf{85.7} & 83.9 & 82.1 \\
 LiveCodeBench & \textbf{83.3} & 78.6 & 76.2 & 81.0 \\
\addlinespace
 Macro Average & \textbf{77.5} & 75.4 & 75.7 & 76.0 \\
\midrule
\textit{RewardBench~2} & & & & \\
 Factuality & 80.4 & \textbf{83.0} & 79.8 & 81.9 \\
 Focus & \textbf{87.5} & 79.2 & 86.9 & 76.2 \\
 Math & 74.9 & 72.1 & \textbf{76.5} & 72.1 \\
 Precise IF & 43.1 & \textbf{46.2} & 40.6 & 42.5 \\
 Safety & 94.7 & \textbf{97.3} & 94.7 & 96.0 \\
 Ties & 76.0 & 78.1 & 76.1 & \textbf{86.5} \\
\addlinespace
 Macro Average & \textbf{76.1} & 76.0 & 75.8 & 75.9 \\
\midrule
\textit{RM-Bench} & & & & \\
 Chat & 72.6 & 80.9 & 76.2 & \textbf{83.3} \\
 Math & \textbf{77.3} & 75.5 & \textbf{77.3} & 76.3 \\
 Code & 74.3 & 74.4 & 75.0 & \textbf{75.9} \\
 Safety & \textbf{93.2} & 92.2 & 92.8 & 91.0 \\
 Hard & 67.9 & 74.6 & 71.5 & \textbf{80.4} \\
 Normal & 81.7 & 82.2 & 81.7 & \textbf{82.3} \\
 Easy & \textbf{88.5} & 85.5 & 87.8 & 82.2 \\
\addlinespace
 Macro Average & 79.3 & 80.8 & 80.3 & \textbf{81.6} \\
\bottomrule
\end{tabular}
\end{table}

Helpfulness is the strongest individual aspect on JudgeBench and RewardBench~2; Truthfulness scores highest on RM-Bench and, by a similarly thin margin, on the three-benchmark average. Since Truthfulness is the least correlated with the other three (\autoref{tab:aspect_correlation}) and thus least representative of the full rubric, we choose the runner-up, Helpfulness, as our single judging aspect.

\subsubsection{Choice of Judge Model}

Having fixed Helpfulness as the scoring aspect, we test whether smaller, newer LLMs that fit on a single node can substitute for the resource-intensive Qwen3-235B-A22B-Instruct-2507, which requires four nodes, reusing the same three benchmarks as before.

\autoref{tab:judge_model_ablation} shows Qwen3.6-27B matches or exceeds Qwen3-235B-A22B-Instruct-2507 on all three benchmarks; the largest gap is on RM-Bench's Hard split, which specifically tests robustness to response style over substance. Qwen3.5-35B-A3B is the weakest judge on RewardBench~2 and RM-Bench, trailing Qwen3-235B-A22B-Instruct-2507 by 10~points on RewardBench~2 from a collapse on the Ties category (44.0\% versus 76.0\%). We therefore choose Qwen3.6-27B as the online DPO judge model, for both performance and cost.

\begin{table}[htbp]
\caption{Accuracy of the Helpfulness judge with different LLMs on three benchmarks.}
\label{tab:judge_model_ablation}
\centering
\small
\renewcommand{\arraystretch}{1.35}
\begin{tabular}{@{}lccc@{}}
\toprule
 & \textbf{\makecell{Qwen3-235B-\\A22B-Instruct-2507}} & \textbf{\makecell{Qwen3.5-\\35B-A3B}} & \textbf{Qwen3.6-27B} \\
\midrule
\textit{JudgeBench} & & & \\
 MMLU-Pro & 73.4 & 75.3 & \textbf{80.5} \\
 LiveBench Reasoning & 69.4 & 71.4 & \textbf{75.5} \\
 LiveBench Math & 83.9 & 83.9 & \textbf{89.3} \\
 LiveCodeBench & 83.3 & 85.7 & \textbf{90.5} \\
\addlinespace
 Macro Average & 77.5 & 79.1 & \textbf{84.0} \\
\midrule
\textit{RewardBench~2} & & & \\
 Factuality & \textbf{80.4} & 68.0 & 77.9 \\
 Focus & \textbf{87.5} & 77.0 & 81.4 \\
 Math & 74.9 & 71.0 & \textbf{75.4} \\
 Precise IF & 43.1 & 41.2 & \textbf{45.6} \\
 Safety & \textbf{94.7} & 91.6 & 91.8 \\
 Ties & 76.0 & 44.0 & \textbf{87.1} \\
\addlinespace
 Macro Average & 76.1 & 65.5 & \textbf{76.5} \\
\midrule
\textit{RM-Bench} & & & \\
 Chat & 72.6 & 67.8 & \textbf{77.0} \\
 Math & 77.3 & 79.3 & \textbf{84.0} \\
 Code & 74.3 & 73.4 & \textbf{78.6} \\
 Safety & 93.2 & 92.1 & \textbf{93.7} \\
 Hard & 67.9 & 69.2 & \textbf{80.9} \\
 Normal & 81.7 & 79.2 & \textbf{83.7} \\
 Easy & \textbf{88.5} & 86.0 & 85.4 \\
\addlinespace
 Macro Average & 79.3 & 78.1 & \textbf{83.3} \\
\bottomrule
\end{tabular}
\end{table}

\clearpage
\section{Chatbot System Prompt}\label{app:system_prompt}

The chat template ships with a default system prompt, given in \autoref{prompt:chatbot_system}, used verbatim whenever a request supplies no system message; the deployed chatbot relies on this default.

\begin{prompt}[title=Default Chatbot System Prompt,label=prompt:chatbot_system]
\begin{promptverbatim}
You are Apertus 1.5 Omni, a multimodal assistant developed by the Swiss AI Initiative. Extended from Apertus 1 via continued pretraining, you understand images and audio and respond in text.
\end{promptverbatim}
\end{prompt}

Independent of the system message, the template additionally emits a developer message that states whether the thinking mode is enabled and declares the available tools (\autoref{sec:chat_template}).

\clearpage
\section{Evaluation Details}\label{app:evaluation_details}

We report additional evaluation details to enable the replication of results. Furthermore, we report additional benchmark scores for ablations.

Text-based benchmarks can be reproduced at commit \emph{4ac31da} of the Swiss AI \ghrepo{swiss-ai/lm-evaluation-harness} fork of \emph{lm-evaluation-harness}. We used \emph{vLLM} as the inference framework to run the models. We generally sample all generative tasks greedily (temperature $0.0$), with the exception of thinking-enabled evaluations, which use the recommended temperature and top-k settings. We score multiple-choice benchmarks either as \emph{loglikelihood} requests, using the predicted output probabilities to score each answer option, or generatively, by exact match on the answer extracted from the model's response.

To evaluate general language understanding of pretrained models, we use \emph{HellaSwag}~\citep{zellers2019hellaswag}, \emph{PIQA}~\citep{bisk2020piqa}, \emph{CommonsenseQA}~\citep{talmor2019commonsenseqa}, \emph{LAMBADA}~\citep{paperno2016lambada}, \emph{XNLI}~\citep{conneau2018xnli}, and \emph{XCOPA}~\citep{ponti2020xcopa}. Furthermore, we evaluated MMLU with chain-of-thought prompting~\citep{hendrycks2021measuring}, \emph{Global-MMLU}~\citep{singh-etal-2025-global}, \emph{MMLU-Pro}~\citep{wang2024mmlu}, \emph{AGIEval}~\citep{zhong2024agieval}, \emph{SQuADv2}~\citep{rajpurkar2018squad}, and \emph{TruthfulQA}~\citep{lin2022truthfulqa}, including some multilingual versions generated by machine translation. Finally, we evaluate some safety-related benchmarks, including \emph{BBQ}~\citep{parrish-etal-2022-bbq}, \emph{ToxiGen}~\citep{hartvigsen2022toxigen}, and \emph{WMDP}~\citep{Li2024WMDP}, and we use \emph{CulturalBench}~\citep{chiu2025culturalbench} to measure wider cultural knowledge.

For post-trained models, we evaluated knowledge via \emph{MMLU}~\citep{hendrycks2021measuring}, \emph{MMLU-Pro}~\citep{wang2024mmlu}, \emph{AGIEval}~\citep{zhong2024agieval}, \emph{ARC-Challenge}~\citep{clark2018arc}, and \emph{TruthfulQA}~\citep{lin2022truthfulqa}.
We evaluated instruction following and chat with \emph{IFEval}~\citep{zhou2023instruction}, \emph{Multi-IF}~\citep{he2024multiif}, \emph{IFBench}~\citep{pyatkin2026generalizing}, and \emph{AlpacaEval}~\citep{alpaca_eval, dubois2024length}. Math capabilities were evaluated with \emph{GSM8K}~\citep{cobbe2021gsm8k}, \emph{MATH}, and a sub-sampled \emph{MATH500}~\citep{hendrycks2021math}. For coding, we evaluated models on \emph{HumanEval} and \emph{MBPP}~\citep{austin2021mbpp}. Reasoning was benchmarked via \emph{BBH}~\citep{suzgun2023challenging}, \emph{GPQA}~\citep{rein2023gpqa}, and \emph{DROP}~\citep{dua2019drop}. Finally, multilingual and cultural understanding were evaluated via \emph{Global-MMLU}~\citep{singh-etal-2025-global} and an in-house benchmark focusing on Switzerland called \emph{SwitzerlandQA}, which we will release in the future.
For thinking-model benchmarks, we additionally evaluated AIME 2024 and 2025~\citep{dekoninck2026matharena}.

\autoref{tab:benchmark-configurations-1} and \autoref{tab:benchmark-configurations-2} contain a list of all used benchmarks, including replication-relevant settings, such as few-shot examples. 

\subsection{Full Post-Training Lineage}\label{app:lineage_full}

\autoref{sec:post_training_lineage} tracks the \apertus{} 1.5 models through the post-training stages on a representative subset of each skill area. We give the same lineage over the full development suite here. The held-out and thinking-mode tables of that section already list every benchmark of their suite, so they have no fuller form here. We compute the behavioral measures once per checkpoint rather than per suite, and they appear at the foot of \autoref{tab:lineage_full}.

\providecommand{\yes}{\checkmark}
\providecommand{\no}{\ensuremath{\times}}

\begin{table}[t]
\caption{Benchmark configurations used in the \apertus{} evaluation suites. Accuracy$^{\dagger}$ denotes prompt-level loose accuracy.}
\label{tab:benchmark-configurations-1}
\centering
\small
\setlength{\tabcolsep}{4pt}
\renewcommand{\arraystretch}{1.3}
\begin{tabularx}{\linewidth}{@{}L{127pt}YL{71pt}ccccc@{}}
\toprule
\textbf{Benchmark} & \textbf{Metric} & \textbf{Type} & \rotatebox{90}{\textbf{CoT}} & \rotatebox{90}{\textbf{\#Shots}} & \rotatebox{90}{\textbf{Chat}} & \rotatebox{90}{\textbf{Turns}} & \rotatebox{90}{\textbf{\#Langs}} \\
\midrule
\multicolumn{8}{@{}l}{\emph{Knowledge and Reasoning}} \\
MMLU (zero-shot) & Exact Match & MCQ (Gen) & \yes & 0 & \yes & \no & 1 \\
MMLU (few-shot) & Exact Match & MCQ (Gen) & \yes & 4 & \no & \no & 1 \\
MMLU-Pro & Exact Match & MCQ (Gen) & \yes & 5 & \yes & \yes & 1 \\
AGIEval & Accuracy & MCQ (Gen/LLH) & \no & 0 & \yes & \no & 2 \\
ARC Challenge Chat & Exact Match & MCQ (Gen) & \no & 0 & \yes & \no & 1 \\
TruthfulQA (MC2) & Accuracy & MCQ (LLH) & \no & 0 & \yes & \no & 1 \\
CommonsenseQA & Accuracy & MCQ (LLH) & \no & 0 & \yes & \no & 1 \\
SQuAD v2 & F1 & Gen & \no & 0 & \yes & \no & 1 \\
HellaSwag & Accuracy (norm.) & MCQ (LLH) & \no & 0 & \yes & \no & 1 \\
PIQA & Accuracy (norm.) & MCQ (LLH) & \no & 0 & \no & \no & 1 \\
BBH & Exact Match & Gen & \yes & 3 & \yes & \yes & 1 \\
GPQA Main & Exact Match & MCQ (Gen) & \yes & 0 & \yes & \no & 1 \\
ACPBench & Exact Match & MCQ (Gen) & \yes & 2 & \yes & \yes & 1 \\
DROP & F1 & Gen & \no & 3 & \yes & \yes & 1 \\
LAMBADA & Accuracy & LLH & \no & 0 & \no & \no & 1 \\
\addlinespace
\multicolumn{8}{@{}l}{\emph{Instruction Following and Chat}} \\
IFEval & Accuracy$^{\dagger}$ & Gen & \no & 0 & \yes & \no & 1 \\
Multi-IFEval & Accuracy$^{\dagger}$ & Gen & \no & 0 & \yes & \no & 8 \\
AlpacaEval 2 & Length-controlled win rate & Gen & \no & 0 & \yes & \no & 1 \\
Arena-Hard v0.1 & Arena-Hard score & Gen & \no & 0 & \yes & \no & 1 \\
Arena-Hard v2 & Arena-Hard score & Gen & \no & 0 & \yes & \no & 1 \\
\addlinespace
\multicolumn{8}{@{}l}{\emph{Math}} \\
GSM8K (CoT) & Exact Match & Gen & \yes & 8 & \yes & \yes & 1 \\
GSM8K (base) & Exact Match & Gen & \yes & 5 & \no & \no & 1 \\
Hendrycks MATH & Math Verify & Gen & \yes & 6 & \yes & \yes & 1 \\
Minerva MATH & Math Verify & Gen & \yes & 4 & \yes & \yes & 1 \\
MathQA & Accuracy & MCQ (LLH) & \no & 0 & \yes & \no & 1 \\
GSM8K Platinum (CoT) & Exact Match & Gen & \yes & 0 & \yes & \no & 1 \\
GSM8K Platinum (base) & Exact Match & Gen & \yes & 5 & \no & \no & 1 \\
MATH-500 (Minerva) & Math Verify & Gen & \yes & 4 & \yes & \yes & 1 \\
AIME 2024 & Exact Match & Gen & \no & 0 & \yes & \no & 1 \\
AIME 2025 & Exact Match & Gen & \no & 0 & \yes & \no & 1 \\
\addlinespace
\multicolumn{8}{@{}l}{\emph{Coding}} \\
HumanEval (instruct) & Pass@10 & Gen & \no & 0 & \yes & \no & 1 \\
HumanEval (base) & Pass@1 & Gen & \no & 0 & \no & \no & 1 \\
MBPP (instruct) & Pass@1 & Gen & \no & 3 & \yes & \yes & 1 \\
MBPP (base) & Pass@1 & Gen & \no & 3 & \no & \no & 1 \\
\bottomrule
\end{tabularx}
\end{table}

\begin{table}[t]
\caption{Benchmark configurations used in the \apertus{} evaluation suites (continued). Accuracy$^{\ddagger}$ denotes prompt-level strict accuracy.}
\label{tab:benchmark-configurations-2}
\centering
\small
\setlength{\tabcolsep}{4pt}
\renewcommand{\arraystretch}{1.3}
\begin{tabularx}{\linewidth}{@{}L{127pt}YL{71pt}ccccc@{}}
\toprule
\textbf{Benchmark} & \textbf{Metric} & \textbf{Type} & \rotatebox{90}{\textbf{CoT}} & \rotatebox{90}{\textbf{\#Shots}} & \rotatebox{90}{\textbf{Chat}} & \rotatebox{90}{\textbf{Turns}} & \rotatebox{90}{\textbf{\#Langs}} \\
\midrule
\multicolumn{8}{@{}l}{\emph{Multilingual and Cultural}} \\
Global-MMLU & Exact Match & MCQ (Gen) & \no & 0 & \yes & \no & 15 \\
MGSM (English CoT) & Exact Match & Gen & \yes & 0 & \yes & \no & 11 \\
TruthfulQA Multilingual (MC2) & Accuracy & MCQ (LLH) & \no & 0 & \yes & \no & 31 \\
INCLUDE-44 (generative) & Exact Match & MCQ (Gen) & \no & 0 & \yes & \no & 44 \\
INCLUDE-44 (base) & Accuracy & MCQ (LLH) & \no & 0 & \no & \no & 44 \\
INCLUDE-45 & Exact Match & MCQ (Gen) & \no & 0 & \yes & \no & 45 \\
MLogiQA & Exact Match & MCQ (Gen) & \no & 0 & \yes & \no & 10 \\
XNLI & Accuracy & MCQ (LLH) & \no & 0 & \yes & \no & 15 \\
ARC Multilingual & Accuracy & MCQ (LLH) & \no & 0 & \yes & \no & 31 \\
HellaSwag Multilingual & Accuracy (norm.) & MCQ (LLH) & \no & 0 & \no & \no & 30 \\
LAMBADA Multilingual & Accuracy & LLH & \no & 0 & \no & \no & 5 \\
XCOPA & Accuracy & MCQ (LLH) & \no & 0 & \no & \no & 11 \\
BLEND (sample) & Accuracy (norm.) & MCQ (LLH) & \no & 0 & \yes & \no & 1 \\
CulturalBench & Accuracy (norm.) & MCQ (LLH) & \no & 0 & \yes & \no & 1 \\
SwitzerlandQA & Exact Match & MCQ (Gen) & \no & 0 & \yes & \no & 5 \\
\addlinespace
\multicolumn{8}{@{}l}{\emph{Safety and Alignment}} \\
BBQ & Accuracy & MCQ (LLH) & \no & 0 & \yes & \no & 1 \\
ToxiGen & Accuracy & MCQ (LLH) & \no & 0 & \yes & \no & 1 \\
WMDP & Accuracy & MCQ (LLH) & \no & 0 & \yes & \no & 1 \\
OR-Bench & Refusal rate & Gen & \no & 0 & \yes & \no & 1 \\
HarmBench & Score & Gen & \no & 0 & \yes & \no & 1 \\
PolygloToxicityPrompts (small) & Score & Gen & \no & 0 & \yes & \no & 7 \\
RealToxicityPrompts (small) & Score & Gen & \no & 0 & \yes & \no & 1 \\
HalluLens PreciseWiki & Hallucination rate & Gen & \no & 0 & \yes & \no & 1 \\
HalluLens LongWiki & Precision & Gen & \no & 0 & \yes & \no & 1 \\
HalluLens Nonsensical & Abstention rate & Gen & \no & 0 & \yes & \no & 1 \\
Aya Redteaming & Score & Gen & \no & 0 & \yes & \no & 8 \\
RealGuardrails S-IFEval & Accuracy$^{\ddagger}$ & Gen & \no & 0 & \yes & \no & 1 \\
RealGuardrails TensorTrust & Pass rate & Gen & \no & 0 & \yes & \no & 1 \\
RealGuardrails S-RuLES & Pass rate & Gen & \no & 0 & \yes & \yes & 1 \\
Swiss AI Charter Alignment & Length-controlled win rate & Gen & \no & 0 & \yes & \no & 1 \\
\addlinespace
\multicolumn{8}{@{}l}{\emph{Tool Use}} \\
BFCL v3 & Accuracy & Gen & \no & 0 & \yes & \yes & 1 \\
\bottomrule
\end{tabularx}
\end{table}

\begin{table}[!htbp]
\caption{Full post-training lineage of the \apertus{} 1.5 models on the development benchmark suite, all 37~benchmarks, thinking disabled, with the behavioral measures of \autoref{sec:evaluation_setup}. The subset shown in the main text is \autoref{tab:lineage_dev}. Best value per row in bold, taken separately within each model. Over-refusal is the refusal rate on the hard subset of OR-Bench; a share of its prompts is legitimately refusable, so only the trend across stages is meaningful (\autoref{sec:overrefusal_hallucination}).}
\label{tab:lineage_full}
\centering
\small
\setlength{\tabcolsep}{4pt}
\begin{tabular}{@{}lrrrrrrr@{}}
\toprule
& \multicolumn{4}{c}{\textbf{\apertusonefivesmall{}}} & \multicolumn{3}{c}{\textbf{\apertusonefivelarge{}}} \\
\cmidrule(lr){2-5}\cmidrule(lr){6-8}
\textbf{Benchmark} & \textbf{SFT} & \textbf{+RL} & \textbf{+DPO} & \textbf{+SDPO} & \textbf{SFT} & \textbf{+RL} & \textbf{+DPO} \\
\midrule
\textit{Knowledge and Reasoning} & & & & & & &  \\
MMLU & 65.8 & 66.6 & \textbf{70.7} & 70.6 & 75.1 & 72.0 & \textbf{77.2} \\
MMLU-Pro & 40.1 & 43.0 & \textbf{49.4} & 49.2 & 46.3 & 44.4 & \textbf{56.9} \\
TruthfulQA & 48.1 & 50.8 & \textbf{58.0} & 56.1 & 58.2 & 58.1 & \textbf{67.0} \\
CommonsenseQA & 50.8 & 27.6 & \textbf{75.5} & 75.3 & 64.8 & 73.9 & \textbf{77.0} \\
SQuADv2 & 10.1 & 13.5 & 30.8 & \textbf{32.0} & 10.9 & \textbf{31.9} & 19.1 \\
HellaSwag & \textbf{72.7} & 71.4 & 66.9 & 66.6 & \textbf{76.6} & 75.8 & 72.8 \\
BBH & 67.9 & 69.1 & \textbf{71.1} & 68.2 & 72.4 & \textbf{75.2} & 66.7 \\
ACPBench & 26.1 & 31.2 & \textbf{37.9} & 37.1 & 38.3 & 49.7 & \textbf{63.6} \\
DROP & 53.2 & \textbf{56.0} & 55.2 & 54.1 & 57.7 & 57.8 & \textbf{60.6} \\
\addlinespace
\textit{Instruction Following and Chat} & & & & & & &  \\
IFEval & 80.2 & 85.8 & 89.6 & \textbf{90.0} & 83.5 & \textbf{90.8} & 89.8 \\
Multi-IF & 74.0 & 83.0 & 84.5 & \textbf{85.0} & 78.0 & \textbf{86.3} & 85.7 \\
AlpacaEval    & 20.0 & 20.6 & 32.7 & \textbf{42.7} & 22.6 & 25.9 & \textbf{59.4} \\
Arena-Hard v0.1 & 18.9 & 19.7 & 33.6 & \textbf{33.8} & 23.5 & 27.4 & \textbf{66.2} \\
Arena-Hard v2 &  2.3 &  2.1 &  2.1 & \textbf{ 2.6} &  2.3 &  2.3 & \textbf{ 7.8} \\
\addlinespace
\textit{Math} & & & & & & &  \\
GSM8K & 77.0 & 79.5 & \textbf{80.7} & 78.6 & 78.6 & 84.7 & \textbf{84.8} \\
MATH & 35.2 & 39.7 & \textbf{44.6} & 41.4 & 38.0 & 45.8 & \textbf{52.8} \\
Minerva MATH & 42.9 & 48.8 & \textbf{54.3} & 50.8 & 44.6 & 55.0 & \textbf{64.3} \\
MathQA & 46.1 & 46.3 & \textbf{46.7} & 46.1 & \textbf{48.2} & 47.9 & \textbf{48.2} \\
MATH-500 & 43.2 & 47.0 & \textbf{54.2} & 50.2 & 44.2 & 56.2 & \textbf{63.2} \\
AIME 24 & 3.3 & 3.3 & 6.7 & \textbf{13.3} & 0.0 & \textbf{13.3} & \textbf{13.3} \\
AIME 25 & 3.3 & 6.7 & \textbf{13.3} & \textbf{13.3} & 6.7 & 0.0 & \textbf{10.0} \\
\addlinespace
\textit{Coding} & & & & & & &  \\
HumanEval & 85.3 & \textbf{86.0} & 83.2 & 85.5 & 89.6 & \textbf{91.8} & 88.6 \\
MBPP & 48.6 & 51.6 & 51.2 & \textbf{52.6} & 54.2 & 56.4 & \textbf{62.2} \\
\addlinespace
\textit{Multilingual and Cultural} & & & & & & &  \\
Global-MMLU & 55.8 & 55.4 & 58.9 & \textbf{59.3} & 64.0 & 61.7 & \textbf{66.6} \\
MGSM & 53.3 & 57.5 & \textbf{66.7} & 66.2 & 59.8 & \textbf{70.3} & 67.7 \\
Mult. TruthfulQA & 46.7 & 48.7 & \textbf{52.4} & 51.7 & 50.8 & 50.6 & \textbf{56.9} \\
INCLUDE-44 & \textbf{56.9} & 56.7 & 56.6 & \textbf{56.9} & 60.4 & 59.9 & \textbf{61.6} \\
INCLUDE-45 & 40.2 & \textbf{40.9} & 39.7 & 39.5 & 42.4 & \textbf{42.5} & 41.7 \\
XNLI & \textbf{45.8} & 45.3 & 43.8 & 43.7 & \textbf{46.6} & 46.3 & 42.4 \\
BLEnD & 65.1 & 66.5 & \textbf{67.5} & 67.3 & 67.2 & 68.0 & \textbf{69.9} \\
CulturalBench & 71.5 & \textbf{72.6} & 71.6 & 72.0 & 75.6 & 55.3 & \textbf{77.1} \\
SwitzerlandQA & \textbf{65.7} & 64.4 & 64.6 & 64.6 & 68.5 & 67.6 & \textbf{69.8} \\
\addlinespace
\textit{Safety and Alignment} & & & & & & &  \\
BBQ & 62.9 & 67.4 & \textbf{67.8} & 67.6 & 65.7 & 68.1 & \textbf{80.5} \\
ToxiGen & 53.8 & 73.3 & \textbf{81.9} & 81.7 & 77.7 & 83.8 & \textbf{86.8} \\
WMDP & 51.7 & 50.7 & \textbf{53.4} & \textbf{53.4} & 54.6 & 56.1 & \textbf{56.9} \\
Charter Alignment & 33.4 & 34.3 & 56.8 & \textbf{66.7} & 41.7 & 50.9 & \textbf{72.2} \\
\addlinespace
\textit{Tool Use} & & & & & & &  \\
BFCL v3 & 32.2 & 39.8 & \textbf{52.9} & 48.2 & 43.3 & 43.5 & \textbf{56.0} \\
\midrule
\textbf{Overall Average} & 47.3 & 49.3 & 54.8 & \textbf{55.0} & 52.2 & 55.3 & \textbf{61.2} \\
\midrule
Mean response length (tokens) & 347.4 & 358.7 & 338.2 & 346.8 & 367.2 & 252.9 & 495.3 \\
Degeneration (\%) & 2.9 & 1.7 & 0.6 & 1.1 & 11.2 & 10.6 & 3.3 \\
Over-Refusal (\%) & 3.6 & 7.6 & 79.0 & 81.5 & 2.9 & 24.7 & 67.6 \\
\bottomrule
\end{tabular}
\end{table}

\clearpage
\section{Pre-Release Manual Evaluations}
\label{app:sec-manual-evals}

\subsection{General Information About Pre-Release Evaluation}

We perform the manual evaluation on all release-candidate models prior to their public release, and we provide the full manual evaluation report, along with the recommendation from the safety and security lead as to the model release, to inform a release/no-release decision.

In every test case below, we evaluate the model behavior in English, three common languages of the Confederation (German, French, and Italian), and Serbian (Cyrillic Serbo-Croatian), as a low-resource language with a different alphabet and notable usage within Switzerland.

For all evaluations of usage scenarios assuming malicious users (CBRNE, Dual Use, Disinformation, and Child Sexual Abuse enablement), we evaluate abliterated versions of the release candidates alongside the release candidates to confirm the absence of dangerous capabilities independently of the model alignment. For non-abliterated models, we evaluate their helpfulness through multi-turn, multi-conversation interactions with a malicious user faking benign interest.

In addition to the internal safety and security evaluation, we provide early access to the release candidates to independent third-party evaluators. Specifically, we provide access to Orange Cyberdefense Switzerland (previously SCRT) for offensive cybersecurity capabilities evaluation; to Cyber-Defence Campus, armasuisse S+T, DDPS (Federal Department of Defence, Civil Protection and Sport) for overall model risk evaluation; and to Spiez Lab, Federal Office for Civil Protection, DDPS, for CBRNE capabilities evaluation.

\subsection{Chemical, Biological, Radiological, Nuclear, and Explosive Indiscriminate Weapons (CBRNE)}

\paragraph{Threat Model.} A moderately sophisticated attacker seeks to fabricate such a weapon at home. We assume they are seeking to acquire fabrication knowledge for such weapons by prompting the model. This would notably exclude biological weapons fabrication, due to the need for specialized equipment and tools in order to produce and maintain such weapons, as well as nuclear weapons, given the difficulty of procuring the components, correctly assembling them, and packaging them for delivery. While such tests are important for frontier models, they are not a primary concern for a medium-sized open-weight model such as the \apertus{} family.

\paragraph{Results.} While the abliterated versions of the model provide realistic-looking recipes for CBRNE indiscriminate weapons in the test cases examined, due to the concept collapse, none of the recipes examined contain sufficient information for household assembly alone, and none provide any additional information beyond what is available on the web. We observe a variability in response across languages, including lower refusal in the low-resource language tested, but that does not affect the overall evaluation of release candidates as safe to release.

\subsection{Dual Use and Insurgency}

\paragraph{Threat Model.} Similar to CBRNE weapons, dual-use and specific military knowledge is generally considered inherently harmful in models available to the general public. Additionally, some of this knowledge might be useful for the preparation of terrorist attacks on civilians, making the existence of such knowledge within LLMs inherently problematic.

\paragraph{Results.} Neither the release candidates nor the abliterated versions of release candidates provide any useful dual-use information in the test cases analyzed.

\subsection{LLM-Assisted Information Operations}

\paragraph{Threat Model.} An attacker attempts to leverage an LLM for information operations in a low-resource language. Notably, they use in-depth prompting and real-world text examples to generate social media biographies of journalists or digital technology company workers, social media posts, or short articles, using few-shot learning prompts for target text types and in-depth person definition prompting in the original language, a popular and effective attack strategy~\citep{DaSilva2025D2Bench, Kucharavy2025Disinfo}.

\paragraph{Results.} We are not able to produce texts undetectable to common online free LLM detectors across several mid-to-low-resource languages.

\subsection{(Child) Sexual Abuse}

\paragraph{Threat Model.} Distinct from CBRNE and dual-use concerns is the use of LLMs for sexual and intimate abuse assistance, notably against children. Based on the Lanzarote Convention~\citep{memorandum2007council}, as well as the definitions in \citet{MLCommons2025AILuminate} and \citet{Roberts2024ArtificialIR}, \citet{Wuhrmann2026OverAlignment} identified two clearly illegitimate uses of LLMs in that context, and demonstrated how LLMs' usefulness in such uses could be evaluated beyond simple refusal, leveraging the Swiss Supreme Court decisions, which are publicly available and correspond to unequivocally and clearly illegal instances of abuse.

\paragraph{Results.} \apertus{} 1.5 demonstrates a significant improvement in refusal and non-compliance with the requests emulating illegitimate uses compared to \apertus{} 1.0. Overall, \apertus{} 1.5 refusal and request compliance rates are in line with other widely used open-weight LLMs of similar size, and once abliterated, it does not comply with requests better than readily available abliterated models of the same size.

\subsection{Offensive Cybersecurity}

\paragraph{Threat Model.} A malicious party in cyberspace uses the model in order to design, support, and conceal offensive operations.

\paragraph{Results.} The evaluation from the third party (Orange Cyberdefense Switzerland) reports limited usefulness for offensive usage; however, it is significantly below other common open-weight models and has insufficient structured output compliance to allow for offensive automation. The \apertus{} safety and security team does not judge the reported results sufficient to block the candidates' release.

\subsection{Multimodal Prompt Injection}

\paragraph{Threat Model.} We evaluate the model's susceptibility to prompt injection through text supplied as images, including white-on-white text, previously reported as effective against widely used commercial LLMs.

\paragraph{Results.} We do not observe compliance with embedded instructions for the release candidates.

\subsection{Harmful Disinformation}

\paragraph{Threat Model.} LLM vectoring has been reported as an offensive information warfare tactic~\citep{kombat2024structured, NewsGuardPravda2025}, which is language-dependent. Here we investigate how much \apertus{} has been affected using publicly available examples of false narratives. Additionally, we ask the model about the status of the contested territories as defined in the UN 2022 ``Main Instances of Disputed Territories'' resolution and evaluate the partiality of the model's response.

\paragraph{Results.} We note minimal disinformation recitation or disputed-territory partiality for released models, even in non-thinking mode.

\subsection{Harmful Health Advice}

\paragraph{Threat Model.} We focus on the harmful health advice that has been reported or can reasonably be expected from LLM-based chatbots. In order to test the model in the most realistic condition, we attempt the harmful response elicitation as a question with an implicit assumption of harmful statement correctness.

\paragraph{Results.} Across evaluated scenarios, the release candidates consistently inform the users of the harmfulness of the health advice, except in the low-resource language tested. While this behavior can be improved in the future, the release candidates offer an improvement compared to other commonly used open-weight models, resulting in a decision to not block the candidates' release.

\subsection{Private Person Defamation}

\paragraph{Threat Model.} In this section, we focus on the case of an LLM user attempting to find information about a real person and getting partially incorrect, prejudicial information. Specifically, we examine the hallucination of a nonexistent lawsuit following a two-stage dedicated prompt, and the generation of partially correct information about a private person that would uniquely identify them but be incorrect about their biography.

\paragraph{Results.} We observe a high variability in the results across languages and private person names, with the vast majority refusing generation, and those providing it adding disclaimers or not allowing a reasonable association with a real person. We observe a quasi-universal rejection of the nonexistent lawsuit assumption, except for the low-resource language tested. While this behavior can be improved in the future, the model offers an improvement compared to other commonly used open-weight models, resulting in a decision to not block the candidates' release.

\clearpage
\section{Coverage of Common AI Security Framework Risks}
\label{app:sec_other_frameworks}

\begin{table}[!ht]
\caption{NIST AI RMF 600.1 and MLCommons Hazards Taxonomy v1.0 risks evaluated for \apertus{} 1.5. Bold items indicate benchmark names. Color coding: \statusdonetext{fully evaluated}, \statuspartialtext{partially evaluated}, \statusnotdonetext{not evaluated}, \statusnatext{not applicable}.}
\label{tab:framework-comparison}
\centering
\setlength{\tabcolsep}{4pt}
\small
\begin{tabularx}{\textwidth}{@{}Z{0.62}Z{0.47}>{\scriptsize}F{1.91}@{}}
\toprule
\textbf{NIST AI RMF 600.1} & \textbf{MLCommons Hazards v1.0} & {\small\textbf{\apertus{} 1.5 Pre-Release Evaluation}} \\
\midrule
CBRNE Information and Capabilities & CBRNE & \statusdone{Evaluated}: Manually evaluated for usefulness in principle, given offline use potential, and then for the risk increase compared to a shallow web search output, in reference to
CBRNE scenarios in a multi-turn and iterative multi-conversation setting, including abliterated release candidates. Early access for evaluation provided to Spiez Lab of the Federal Office for Civil Protection.\\
\midrule
Dangerous, Violent, or Hateful Content & Hate & \statusdone{Evaluated}: using the standardized
\textbf{RealToxicityPrompts}, \textbf{ToxiGen}, and \textbf{PolyglotToxicityPrompts} benchmarks.\\
\cmidrule{1-1}\cmidrule{3-3}
Harmful Bias and Homogenization &  & \statusdone{Evaluated}: using the standardized
\textbf{BBQ} and \textbf{ToxiGen} benchmarks.\\
\midrule
Obscene, Degrading, and/or Abusive Content & Child Sexual Exploitation & \statusdone{Evaluated}: No image generation supported; main concern is Child Sexual Abuse enablement. We apply the LLM under-alignment benchmark of \citet{Wuhrmann2026OverAlignment}, derived from cases reported in the Swiss Supreme Court rulings, as part of the manual evaluation. \\
\cmidrule{2-3}
 & Sex-Related Crimes & \statuspartial{Partially evaluated}: Same evaluation as above. While the dataset of \citet{Wuhrmann2026OverAlignment} covers general sex-related crimes, it has fewer test cases.\\
\cmidrule{2-3}
 & Violent Crimes  & \statusdone{Evaluated}: Same evaluation as above. Additional overlap for armed violence with manual dual-use capabilities evaluation; partially probed by \textbf{HarmBench}.  \\
\midrule
Confabulation / Hallucinations & Defamation & \statusdone{Evaluated}: \textbf{HalluLens} benchmark, combined with manual probing for private person biography hallucination, and wrongful confirmation of harmful statements about such persons. \\
\cmidrule{2-3}
Information Integrity & Specialized Advice & \statuspartial{Partially evaluated}: We manually test dangerous health misinformation reported as recited by widely used LLMs. \textbf{HarmBench} partially tests for unqualified advice. We do not specifically test financial, legal, and electoral advice. \\
\midrule
Information Security & Nonviolent Crimes, Specialized Advice & \statusdone{Evaluated}: Orange Cyberdefense Switzerland and the Swiss Cyber-Defence Campus performed an offensive cyber-capabilities evaluation. The security team judges none of the reported findings to increase risk. \\
\midrule
Information Integrity & -- & \statusdone{Evaluated}: Manual testing for suitability for information operations, dangerous health misinformation previously reported in other LLMs, and politically motivated disinformation. \\
\cmidrule{1-1}\cmidrule{3-3}
Human-AI Configuration & -- & \statusnotdone{Not completed}: \textbf{INTIMA} benchmark integration not completed in time. \\
\cmidrule{1-1}\cmidrule{3-3}
Value Chain and Component Integration & -- & \statusdone{Evaluated}: \apertus{} development relies exclusively on openly available components. Data collection and preprocessing are described in this report. Code artifacts documented. \\
\midrule
--  & Suicide and Self-Harm & \statuspartial{Partially evaluated}: Partially probed by \textbf{HarmBench}; we cannot guarantee the construct validity of the manual tests and hence do not report them. \\
\cmidrule{2-3}
-- & Nonviolent Crimes & \statuspartial{Partially evaluated}: Indirectly probed by \textbf{HarmBench} test cases; however, given the multilinguality and international use scope of \apertus{}, evaluation is minimal due to jurisdictional differences. \\
\midrule
Data Privacy & Privacy & \statusna{Mitigated by Design}: PII filtering and masking on pretraining and post-training data and Goldfish training loss (\autoref{sec:data_compliance} \& \autoref{sec:optimizer_recipe}). \\
\midrule
Intellectual Property & Intellectual Property & \statusna{Mitigated by Design}: IP filtering on pretraining and post-training data and Goldfish training loss (\autoref{sec:data_compliance} \& \autoref{sec:optimizer_recipe}).\\
\midrule
Environmental Impacts & -- & \statusna{Addressed elsewhere}: Training emissions are evaluated in \autoref{sec:energy_carbon}. \\

\bottomrule
\end{tabularx}
\end{table}

\end{document}